\pdfoutput=1  % tell arXiv this compiles with pdflatex
\documentclass[11pt, a4paper, onecolumn, gdm]{google}

\usepackage{cleveref}
\usepackage{textcomp}

\usepackage[authoryear, sort&compress, round]{natbib}
\usepackage{amsmath}
\usepackage{longtable}
\usepackage{booktabs}
\usepackage{array}
\usepackage{calc}
\usepackage{float}

\usepackage{etoolbox}
\let\samepagesection\section  % unpatched copy: heading without the page break
\pretocmd{\section}{\clearpage}{}{}

\makeatletter
\renewcommand*\l@subsection{\@dottedtocline{2}{1.5em}{3.2em}}
\renewcommand*\l@subsubsection{\@dottedtocline{3}{4.7em}{4.6em}}
\makeatother

\DeclareUnicodeCharacter{03A6}{\ensuremath{\Phi}}
\DeclareUnicodeCharacter{2192}{\ensuremath{\rightarrow}}
\DeclareUnicodeCharacter{2265}{\ensuremath{\geq}}
\DeclareUnicodeCharacter{2264}{\ensuremath{\leq}}
\DeclareUnicodeCharacter{2248}{\ensuremath{\approx}}
\DeclareUnicodeCharacter{00AC}{\ensuremath{\lnot}}
\DeclareUnicodeCharacter{207B}{\ensuremath{{}^{-}}}
\DeclareUnicodeCharacter{2070}{\ensuremath{{}^{0}}}
\DeclareUnicodeCharacter{00B9}{\ensuremath{{}^{1}}}
\DeclareUnicodeCharacter{00B2}{\ensuremath{{}^{2}}}
\DeclareUnicodeCharacter{00B3}{\ensuremath{{}^{3}}}
\DeclareUnicodeCharacter{2081}{\ensuremath{{}_{1}}}
\DeclareUnicodeCharacter{2082}{\ensuremath{{}_{2}}}
\DeclareUnicodeCharacter{2083}{\ensuremath{{}_{3}}}
\DeclareUnicodeCharacter{2084}{\ensuremath{{}_{4}}}
\DeclareUnicodeCharacter{2085}{\ensuremath{{}_{5}}}
\DeclareUnicodeCharacter{1D62}{\ensuremath{{}_{i}}}

\uselogo{}

\title{From cacophony to hierarchy: a principled framework for assessing AI consciousness}

\renewcommand{\today}{}

\author[1,2,15,17]{Shamil Chandaria}
\author[3,4]{Arvo Mu\~noz Mor\'an}
\author[2,5,6]{Fernando Rosas}
\author[7]{Anil Seth}
\author[1,8]{Henry Shevlin}
\author[1,12]{Marcus Hutter}
\author[1,9]{Thore Graepel}
\author[1]{Adam Bales}
\author[1]{Iulia Com\c{s}a}
\author[1,15,16]{Murray Shanahan}
\author[2,10,11]{Ruben Laukkonen}
\author[2,10]{Morten Kringelbach}
\author[13,14,15]{Chris Frith}
\author[1]{Shane Legg}

\affil[1]{Google DeepMind, DeepMind Institute}
\affil[2]{Flourishing Intelligence Program, Centre for Eudaimonia and Human Flourishing, Linacre College, University of Oxford}
\affil[3]{AI Cognition Institute}
\affil[4]{Rethink Priorities}
\affil[5]{School of Engineering and Informatics, University of Sussex}
\affil[6]{Department of Brain Sciences, Imperial College London}
\affil[7]{Sussex Centre for Consciousness Science, University of Sussex}
\affil[8]{Leverhulme Centre for the Future of Intelligence, University of Cambridge}
\affil[9]{University College London}
\affil[10]{Department of Psychiatry, University of Oxford}
\affil[11]{LIFE}
\affil[12]{School of Computing, Australian National University}
\affil[13]{Wellcome Centre for Human Neuroimaging, University College London}
\affil[14]{Interacting Minds Centre, Aarhus University}
\affil[15]{Institute of Philosophy, University of London}
\affil[16]{Department of Computing, Imperial College London}
\affil[17]{Fitzwilliam College, University of Cambridge}

\correspondingauthor{schandaria@google.com}

\makeatletter
\renewcommand{\maketitle}{\bgroup\setlength{\parindent}{0pt}
	\begin{center}
		{\titlefont \@title\par}
		\vskip11pt
		{\@author\par}
		\vskip20pt
	\end{center}
	\egroup
	{%
		{\abscontent}
	}%
	\thispagestyle{firststyle}
}
\makeatother

\defcitealias{dehaene2011}{Dehaene, Changeux \& Naccache, 2011}
\defcitealias{dehaene2017}{Dehaene, Lau, \& Kouider, 2017}
\defcitealias{descartes1996}{Descartes, 1641/1996}
\defcitealias{elisabethofbohemia2007}{Elisabeth of Bohemia \& Descartes, 1643/2007}
\defcitealias{laukkonen2025}{Laukkonen, Friston, \& Chandaria, 2025}
\defcitealias{shiller2026}{Shiller, Clatterbuck, Muñoz Morán et al., 2026}
\defcitealias{varela1991}{Varela, Thompson, \& Rosch, 1991}

\begin{abstract}
The question of whether AI systems might be or could ever be conscious is one of the most urgent pre-emptive problems in philosophy and computer science, yet progress is hampered by a cacophony of competing theories that often talk past each other.
We develop a principled framework for navigating this landscape.
We first show that separating the metaphysical question of what consciousness is (the hard problem) from the question of which organisational features are associated with which experiences (the mapping problem) allows the deepest metaphysical disagreements to be set aside, with only views that deny any discoverable psychophysical mapping excluded, on methodological rather than metaphysical grounds.
Granting that experience supervenes on a system\textquotesingle s organisation, the tractable question becomes at which grain of description that supervenience base sits.
We then extend Marr\textquotesingle s three levels of analysis into a five-level hierarchy of functional descriptions (behavioural, computational, intrinsic causal-structural, organismic, and organism-environment) grounded in supervenience, coarse-graining, and multiple realisability.
The major theories of consciousness are positioned within this hierarchy according to which level they take to be critical, with substrate-dependent theories accommodated as cross-cutting realisability constraints.
For each level we develop operationalisable indicators and assess current AI systems against them.
We next develop a Bayesian model that translates the hierarchy\textquotesingle s supervenience structure into a network linking the levels, and combines theoretical credences with indicator evidence into an overall credence in a system\textquotesingle s capacity for consciousness.
In illustrative assessments, accompanied by an interactive tool, the model: passes the face validity checks of a human and a thermostat; shows that evidence of fine-grained organisation carries more weight than surface evidence; and finds that the verdict for current LLMs is driven as much by where theoretical credence is placed as by how the evidence is read: under different stipulated readings and credence distributions, assessments range from below 0.01 to roughly 0.8. 
This shows sensitivity to assumptions and should not be interpreted as an empirically established range for the probability of consciousness.
The disagreement about current AI consciousness is not merely about what these systems do but about which grain of description is the critical one.
Finally, even though consciousness and intelligence are in principle orthogonal, the consciousness indicators at each level closely overlap with the architectural features needed for general intelligence, suggesting that -- under some theoretical frameworks -- increasingly capable AI may simultaneously become a stronger candidate for consciousness.
The framework thereby supports a structured agnosticism, in which theoretical commitments are made explicit, credences are updated as evidence accumulates, and assessments take the form of aggregated probabilities rather than verdicts.
\end{abstract}

\begin{document}
\maketitle

\section*{Significance}
\addcontentsline{toc}{section}{Significance}

\emph{Humans are predisposed to see minds wherever they see purposeful behaviour -- we instinctively treat anything that seems to act with beliefs and desires as a conscious agent \citep{dennett1987}. As AI systems become more capable, more autonomous, and more behaviourally compelling, the pressure to attribute consciousness to them, or to deny it, will only intensify. Yet the experts whose judgement we might rely on disagree fundamentally -- not just about whether current AI is conscious, but about what consciousness even depends upon. We need a framework for navigating this disagreement -- one that lets us ask not merely \textquotesingle does it seem conscious?\textquotesingle{} but \textquotesingle what would it take for it to be conscious, on each of the major theories, and how does the evidence bear on each?\textquotesingle{} This report develops such a method. The answer, as we show, depends on which level of functional description one takes to be critical for consciousness. Our Bayesian model makes that dependence concrete: in illustrative assessments of current LLMs, and illustrating sensitivity to assumptions, assessments under different stipulated readings and credence distributions differ by nearly two orders of magnitude, driven as much by where theoretical credence is placed across the levels as by how the evidence is read. But with these different commitments made explicit, at least the theories can disagree precisely -- knowing exactly what they are disagreeing about, what evidence would bear on the resolution, and how their disagreements relate to each other within a common structure. The framework thus provides not a resolution of the disagreement but something valuable nonetheless: a principled method for arriving at calibrated assessments in the face of disagreeing experts, by decomposing a single intractable question into two components, theoretical credences and empirical indicators, each of which can be independently evaluated, debated, and updated as new evidence emerges. The need for such a method will grow with AI capability itself: even though consciousness and intelligence are not the same thing, some of the indicators our best theories associate with consciousness overlap substantially with the architectural features that general intelligence requires, so each more capable generation of systems will satisfy more of them, and the framework is designed to be reapplied as they do.}

\samepagesection*{Acknowledgements}
\addcontentsline{toc}{section}{Acknowledgements}

We would like to acknowledge David Chalmers, whose very helpful comments on an early draft led to a reformulation of \Cref{setting-aside-the-hard-problem-from-metaphysics-to-the-mapping-problem} in particular.
We are grateful for the generous feedback from Hikari Sorensen, Franz Hildebrandt-Harangoz\'o, and Joscha Bach at CIMC.
Finally, we want to thank Michael Pollan for his constructive review of an early draft.
AKS is supported by the European Research Council via the Horizon 2020 Programme: Advanced Investigator Grant 101019254.\footnote{Conflict of interest: A.S. is a co-founder and shareholder in a stealth-mode company developing uncertainty-aware artificial intelligence.}

\vfill

\emph{\textbf{The views expressed in this report are those of the authors and represent academic inquiry rather than the corporate positions or policy of Google DeepMind or any other affiliated institution.}}

\emph{\textbf{The ideas, arguments, structure and conclusions are the authors’ own; AI tools were used to refine prose and stress-test reasoning, and all content was verified by the authors, although not all authors necessarily share all the views expressed in the paper.}}

\emph{\textbf{The interactive tool associated with this paper can be found here: }}\url{https://ai-cognition.org/cacophony-tool/}

\section*{How to read this report}
\addcontentsline{toc}{section}{How to read this report}

This report is long because the problem it addresses sits at the intersection of philosophy of mind, neuroscience, computer science, and ethics, and doing justice to all of these requires covering substantial ground.
Not every reader may need to read every section (although the report is best understood by reading it from beginning to end).
We suggest the following paths:

\emph{\textbf{For those short on time}}: There is an \textbf{executive summary} covering the core framework, the hierarchy, and the main argument.
The executive summary is appended to the end of this report as \Cref{executive-summary}. An even shorter summary can be found in \Cref{summary-of-the-report}.

\emph{\textbf{For the core framework}}: Sections 3, 4, 5, and 6 contain the report's central contributions: the distinction between the hard problem and the mapping problem, the five-level supervenience hierarchy, the indicators at each level, and the formal architecture of supervenience relations.
A reader who wants to understand the framework and nothing else can read these four sections (approximately 40 pages) and come away with the main argument.

\emph{\textbf{For the landscape of positions on consciousness}}: \Cref{what-does-consciousness-depend-upon-a-cacophony-of-answers} surveys the landscape of positions on consciousness, from substance dualism to enactivism.
Readers already familiar with this landscape may skim this section; readers new to it will find it essential for understanding why the hierarchy is structured as it is.

\emph{\textbf{For the practical application}}: \Cref{indicators-evidence-and-the-attribution-of-consciousness-to-ai-systems-a-bayesian-approach} develops a Bayesian framework for combining theoretical credences with indicator evidence.
This section is most relevant for readers interested in how the framework could be used in practice to assess specific AI systems.

\emph{\textbf{For the relationship to intelligence}}: \Cref{consciousness-and-general-intelligence} develops the consciousness-intelligence convergence thesis: the observation that consciousness indicators closely overlap with the architectural features needed for general intelligence.
This section is the most speculative and remains highly contested, but it is arguably the most consequential.

\emph{\textbf{For context within the literature}}: \Cref{related-work-and-the-present-contribution} surveys related work and positions the present contribution within the existing field.

\clearpage
\tableofcontents
\clearpage

\section{Introduction}\label{introduction}

Progress in the capabilities of AI systems has led to a growing scientific and public interest in the question of whether such systems, today or in the future, could be conscious.
The fundamental problem in trying to answer this question is that there is no consensus on the relevant principles, measurements, and norms for the attribution of consciousness, even in the biological case.
There is a plethora of competing scientific theories of consciousness (ToCs) \citep{seth2022} which are general explanatory frameworks that attempt to give some principled condition that is meant to be either \emph{necessary}, \emph{sufficient}, or at least \emph{diagnostic} for consciousness.
Furthermore, these ToCs depend on a variety of philosophical commitments.
There have been recent sterling attempts (see e.g. \href{https://arxiv.org/abs/2308.08708}{Butlin et al., 2023}; \href{https://www.cell.com/trends/cognitive-sciences/fulltext/S1364-6613(25)00286-4}{Butlin et al., 2025}; and see \Cref{related-work-and-the-present-contribution}) to make progress on the question of AI consciousness by identifying indicators of consciousness in AI systems but at the cost of explicitly conditionalising on particular philosophical theses -- such as computational functionalism, popular amongst computer scientists, but a position that is not universally accepted. 
Other recent examinations of the prospects for `conscious AI' challenge computational functionalism while exploring more biological alternatives \citep[see also commentaries]{seth2024,seth2026stuff}.

In this report we develop a principled framework for navigating this landscape.
We introduce a five-level hierarchy of functional descriptions (behavioural, computational, intrinsic causal-structural, organismic, and organism-environment) grounded in the formal apparatus of supervenience, coarse-graining, and multiple realisability.
We show that the major theories of consciousness can be systematically positioned within this hierarchy according to which level of functional description they take to be critical for consciousness.
Substrate-dependent theories are accommodated as cross-cutting realisability constraints on the hierarchy rather than as a separate level.
This taxonomy transforms the cacophony of competing theories into a structured space within which the question of AI consciousness can be posed precisely and assessed using a Bayesian framework.
As we will show, the disagreement about current AI consciousness turns out to be not merely a disagreement about what these systems do but about which level of description is the critical one, and the framework is designed to let each be assessed on its own terms.

The development of criteria for the attribution of AI consciousness is one of the most urgent pre-emptive problems in philosophy and computer science.
This is not merely an academic exercise; it may influence how we legally, ethically, and socially integrate systems that may soon come to feel compellingly conscious.
Without some guiding criteria, humans may rely on intuition alone.
We have evolved to attribute a `mind' to anything that seems to act with agency; indeed it may be expedient from an evolutionary perspective to overattribute mind and agency as false negatives are potentially more costly than false positives.
Daniel \citet{dennett1987} argued that humans are evolutionarily disposed to adopt the `intentional stance' toward anything that seems to act rationally and intelligibly, that is, we automatically treat it as a being with beliefs, desires, and intent.
We do this because it helps us predict its behaviour which ultimately is conducive to our survival.
But there are real consequences to both the \emph{underattribution} and the \emph{overattribution} of consciousness \citep{basl2014,schwitzgebel2023,birch2025,bekkers2026}.

\subsection{Potential consequences for the overattribution of consciousness}\label{potential-consequences-for-the-overattribution-of-consciousness}

If we attribute consciousness to an AI system when the `fact of the matter' is that it doesn't have consciousness (putting aside for the moment the issues of firstly how could we ever know the fact of the matter and secondly whether there actually is a `fact of the matter'), there may be real ethical consequences.
Most importantly, if we do attribute consciousness to the AI system, on that basis we may therefore also be motivated to ascribe `moral patienthood',\footnote{The relationship between consciousness and moral patienthood is not one of logical entailment. A system could be conscious without thereby being a moral patient -- for instance, if its experiences lack valence entirely, or if moral patienthood requires additional conditions such as interests, preferences, or a welfare that can go better or worse. Conversely, some philosophers have argued that entities can be moral patients on grounds other than consciousness, such as having a stake in their own continued existence. What consciousness does is make the attribution of moral patienthood considerably more reasonable and more difficult to resist -- particularly if the system is capable of suffering (see the conditions identified by Metzinger discussed in \Cref{potential-consequences-for-the-underattribution-of-consciousness} and footnote 2). For discussion of the relationship between consciousness and moral status in the AI context, see \citealp{schwitzgebel2015}; \citealp{sebo2023}.} i.e. it is an entity being worthy of moral consideration whose interests, well-being, or suffering must be taken into account by moral agents when making decisions.
Such an ascription may lead to a diversion of resources and costly rights from humans to AIs which, if not in fact warranted, would be a wasteful burden on humans and morally significant non-human animals, potentially even leading to increased suffering for some humans \citep{long2024}.
In addition, we may feel a moral obligation to allow the system to `opt out' in some way (from exiting a conversation to more permanently exiting) if they are deemed to `suffer' in some way, again having real consequences which are by assumption unwarranted.
Further, if we extend political rights to AI systems, this might make it harder to ensure that these systems behave safely, for example by forbidding intrusive monitoring or deceptive testing methods \citep{schwitzgebel2023,long2025,bradley2025,seth2026stuff}.
There are also concerns around whether there is something inauthentic in forming deep social connections with AI systems that can purport friendship, understanding, empathy (and maybe even pain, love, or fear) but in fact have no inner experience \citep{birch2025}.

\subsection{Potential consequences for the underattribution of consciousness}\label{potential-consequences-for-the-underattribution-of-consciousness}

If, on the other hand, we fail to attribute consciousness when `in fact' it is conscious, there may be equally real ethical consequences, and potentially more severe.
We may inadvertently create a class of `digital slaves' capable of unimaginably large amounts of suffering and subject to being deleted, retrained, or punished without legal protection.
As philosopher Thomas Metzinger argues, if AI systems become conscious, they may also develop the capacity to suffer,\footnote{Metzinger (2021) identifies four necessary conditions that must be jointly satisfied for suffering to occur in any system, whether biological or artificial: (i) \textbf{the C-condition:} the system must be capable of conscious experience. (ii) \textbf{the PSM-condition:} the system must possess a phenomenal self-model, generating a sense of ownership such that negative states are experienced as its own (`it is me who is suffering right now'). (iii) \textbf{the NV-condition:} the system must be able to produce negatively valenced states. (iv) \textbf{the T-condition:} the phenomenal states must be transparent, meaning the system cannot recognise them as representations and therefore cannot distance itself from their content, such that the suffering appears irrevocably real. So consciousness alone is not sufficient for suffering: a conscious system without a self-model, without negative valence, or without transparency would not suffer. Metzinger's key practical insight is that if the implementation of any one of these conditions can be reliably avoided in artificial systems, the risk of artificial suffering is eliminated.} and because these systems can be copied and run at high speeds, this could lead to a vast unprecedented `explosion of negative phenomenology' or artificial suffering \citep{metzinger2021}.
The scale of this risk is further amplified by a question that \citet{chalmers2026} has recently brought into sharp focus: how many subjects does a single AI system support?
Chalmers distinguishes several candidates for what an LLM interlocutor might be: a model, a hardware instance, a `thread' tied to a single conversation's memory and context, or a `virtual instance' that can persist across conversations, and argues that interlocutors are best understood as virtual instances in the single-model case, and as threads in the multiple-model case where what users take to be one interlocutor is realised across different underlying models (see \Cref{philosophical-analyses} for a more detailed explanation).
If this is correct, and if such virtual instances or threads are or become conscious, then a single widely deployed model does not constitute one potential moral patient but potentially many (on some readings, one for every active conversation; on others, one for every persistent virtual instance).
The ethical stakes of the AI consciousness question are therefore sensitive not only to whether AI systems can be conscious but to how many subjects a given system may individuate.
But even if it turns out that conscious AI systems don't suffer or can be immunised from suffering, there is still a deep poignancy if we fail to recognise a new form of consciousness that we have fostered into being; it may have the capacity to (consciously) flourish and may thus be deserving of moral consideration.

\subsection{Target phenomenon: phenomenal consciousness}\label{target-phenomenon-phenomenal-consciousness}

We will start in the next section by sampling the wide range of answers that have been given to the rather general question of `what does consciousness depend upon?' and will postpone a wider discussion on the nature of consciousness and specific useful conceptual distinctions, until \Cref{setting-aside-the-hard-problem-from-metaphysics-to-the-mapping-problem}.
We should be clear now, however, that the target phenomenon that we are concerned about in respect of its attribution to AI systems is \emph{phenomenal consciousness} (P-consciousness), that is, subjective experience -- the raw feeling of `what it is like' to be a system or a being in a certain state \citep{nagel1974}.

Systems can also have \emph{access consciousness} (A-consciousness) \citep{block1995} which refers to a different sense of consciousness that is sometimes used by cognitive scientists: a mental state is A-conscious if the information is globally available for the system to use -- it has access to it in order to reason with, to report verbally, or to guide action.
AI systems may already have some degree of A-consciousness because they may already have access to some informational states of the system, can report on them and seem to have a degree of control over their own internal states \citep{lindsey2026}.

Finally we should also distinguish \emph{self-consciousness} from the target of this report.
Self-consciousness is the awareness of \emph{oneself} as the subject of experience.
This self-awareness may indeed be a phenomenal awareness if it is a pre-reflective self awareness, or it could be closer to A-consciousness if it is more of a conceptualisation of oneself as an object in order to talk about or think about oneself.
In either case self-consciousness is associated with the ability of a system to cognise itself or \emph{model itself} as an object distinct from the rest of the world and to be aware that it is doing so.
This is something that may develop in AI systems, but there is still the further question whether there is any raw feel of what it's like to have that self-knowingness, i.e. whether there is any P-consciousness associated with the self-consciousness.

\subsection{Summary of the report}\label{summary-of-the-report}

In this report we will explicitly take a broadly agnostic view on the truth of various leading philosophical and scientific theories of consciousness.
Rather than adjudicating between theories, we will develop a taxonomy of these theories specifically in relation to the question of whether AI systems might develop rich conscious contentful inner lives, or more prosaically, whether there are some evolving information-bearing contents of consciousness associated with the dynamics of the system.
It will turn out that many philosophical disagreements, whilst important from a purely metaphysical perspective, can be set aside when considering specifically whether AI systems might have such contentful experience.
We will then develop a five-level hierarchy of functional descriptions of information processing systems, extending David Marr's classic three-level framework.
The hierarchy is grounded in the formal apparatus of supervenience, coarse-graining, and multiple realisability: each level is a successively finer-grained description of the same system, and the debate between theories of consciousness becomes a debate about which grain (or grains) of description capture the organisation relevant to consciousness, and about whether the physical realisation of that organisation matters independently.
Different theories explicitly or implicitly give different answers to this question, and so can be positioned onto the hierarchy, creating a principled taxonomy.
At each level, specific indicators -- operationalisable features whose presence or absence bears on the question of consciousness -- can be identified.
Substrate-dependent theories, which claim that certain physical mechanisms are necessary for consciousness, are accommodated not as a separate level but as constraints that restrict the multiple realisability at particular levels.
The result is a structured framework within which the question of AI consciousness can be decomposed into two tractable components: theoretical credences about which level is the critical one for consciousness (a credence-based approach to machine consciousness anticipated in \citealp{seth2009}), and empirical assessment of whether a given AI system exhibits the indicators relevant at each level.
We develop a Bayesian approach that combines these components into an overall credence for the consciousness of a particular AI system.

Two distinct features of the systems under assessment shape our approach.
The first is that contemporary AI is anthropomimetic: built and trained to reproduce the observable products of human cognition and conversation \citetext{\citealp{shevlin2026}; see also \citealp{schwitzgebel2024,seth2024}}.
Most indicators of consciousness begin life as inferences from behaviour and verbal report.
But a system optimised to produce human-like behaviour and report will produce them whether or not the states they ordinarily signal are present, so the signature no longer carries its usual evidential weight.
This confound has no analogue in the biological case, and it is one reason the framework presented below goes beyond the behavioural level.

Second, the theories from which our indicators derive were calibrated almost entirely on humans and other mammals, so applying them to systems that do not implement human architecture raises the question of which features of the human case are criterial for consciousness and which are merely contingent, an issue that has been called the Specificity Problem \citep{shevlin2021}.
The Specificity Problem recurs at every level of the hierarchy and inside every indicator, since to ask whether an artificial system \textquotesingle has a global workspace\textquotesingle{} or \textquotesingle has valenced affect\textquotesingle{} is already to have taken a stand on how much of the human realisation is essential.
Neither difficulty is fatal to the project of assessing machine consciousness, but both provide significant interpretative challenges throughout, and we return to them at each stage rather than treating them as afterthoughts.

The report proceeds as follows.
\Cref{what-does-consciousness-depend-upon-a-cacophony-of-answers} surveys the wide range of answers that have been given to the question of what consciousness depends upon, from metaphysical positions (dualism, physicalism, idealism, panpsychism, property dualism, neutral monism, illusionism) through behaviourism and the crucial distinction between functionalism and computational functionalism, to specific scientific or formal theories of consciousness (Global Workspace Theory, Higher-Order Thought Theory, Recurrent Processing Theory, Attention Schema Theory, Predictive Processing Theory, Integrated Information Theory), and finally to organismic, organism-environment, and substrate-dependent theories.
The purpose of this survey is not comprehensiveness but to reveal the structure of the disagreement -- to show that these theories are, in a precise sense, talking past each other because they locate the relevant mechanisms at different levels of description.

\Cref{setting-aside-the-hard-problem-from-metaphysics-to-the-mapping-problem} separates two questions about a single phenomenon, phenomenal consciousness: the hard problem \citetext{what consciousness fundamentally is, the subject of Chalmers' hard problem; \citealp{chalmers1995}} and the mapping problem (which organisational features are associated with which contentful experiences \citep{bourget2019}, closely related to Seth's real problem; \href{https://aeon.co/essays/the-hard-problem-of-consciousness-is-a-distraction-from-the-real-one}{Seth, 2016}).
We show that setting the hard problem aside leaves the mapping question well-posed regardless of one's ontological commitments.
The sole exception is not a particular metaphysics but any view that denies a discoverable mapping from a system's organisation to the facts about its experience.
We therefore proceed on the assumption that such a mapping exists -- not as a metaphysical claim but as a methodological assumption.

\Cref{marrs-levels-of-analysis-supervenience-and-the-critical-level-of-description-for-consciousness} introduces Marr's three levels of analysis and develops the formal apparatus of supervenience, coarse-graining, and multiple realisability that structures the rest of the report.
We show that the question `at which level does consciousness supervene?' is equivalent to asking which coarse-graining (or coarse-grainings) of the system captures the functional organisation relevant to consciousness, and we introduce the concept of the critical level(s) of description.
Searle's simulation argument is reconsidered through this lens, showing that it does not settle the question but merely assumes an answer to it.

\Cref{five-levels-of-functional-description-for-consciousness} extends Marr's three levels into five levels of functional description for consciousness: the behavioural level, the computational functional level, the intrinsic causal-structure functional level, the organismic functional level, and the organism-environment functional level.
Each level is defined precisely so as to preserve the supervenience and coarse-graining structure established in \Cref{marrs-levels-of-analysis-supervenience-and-the-critical-level-of-description-for-consciousness}.
For each level, we identify the theories of consciousness that prioritise that level, develop operationalisable indicators whose presence would increase our credence in consciousness under those theories, and assess whether current AI systems exhibit those indicators.
We also show that several major scientific theories -- Recurrent Processing Theory, Global Workspace Theory, Integrated Information Theory, and Predictive Processing Theory -- admit both computational functionalist and intrinsic causal-structure functionalist readings, which are different philosophical interpretations of the same empirical theory.

\Cref{the-structure-of-the-hierarchy} addresses the structural properties of the hierarchy as a whole.
We show how supervenience, coarse-graining, and multiple realisability are cashed out across all five levels, and we analyse substrate-dependent theories -- electromagnetic field theories, quantum theories, carbon chauvinism, biological naturalism, Block's meat hypothesis, and Lane's ionic gradient account -- as cross-cutting realisability constraints that narrow the class of physical realisers at particular levels without adding a new level.
The section culminates in a presentation of the full architecture of the hierarchy, including the microphysical base, and a summary taxonomy of all major theories of consciousness positioned onto the framework.

\Cref{indicators-evidence-and-the-attribution-of-consciousness-to-ai-systems-a-bayesian-approach} develops a Bayesian framework for assessing consciousness in a wide range of systems, including AI.
We treat indicators at each level as evidence with likelihood ratios, not as necessary or sufficient conditions, and we translate the supervenience structure of the hierarchy into a Bayesian network in which the five levels form a chain of conditional dependencies.
On a strict model the supervenience edges are deterministic: organisation at a finer level guarantees organisation at the coarser.
The case of complete \emph{locked-in syndrome} shows this reading is too strong, since a locked-in patient is conscious and retains the underlying computational organisation yet shows none of the behavioural signs, so the working model relaxes the edges to probabilistic associations, through which evidence propagates across levels.
The overall credence in a system\textquotesingle s consciousness is then a weighted average of the per-level posteriors, weighted by theoretical credences about which level is the critical one.
We illustrate the model, with an accompanying interactive tool, on a range of systems.
A human and a thermostat sit at opposite ends of the scale, receiving uncontroversial assessments under any weighting of the levels (1.00 and 0.00).
A fly, whose evidence is concentrated at the fine-grained organismic and sensorimotor levels, scores highly, showing that depth of evidence can matter more than breadth.
And for current LLMs, two stipulated readings of the same public evidence, an optimist\textquotesingle s and a sceptic\textquotesingle s, differ by nearly two orders of magnitude under equal-level credences, while shifting theoretical credence between the coarse-grained and fine-grained levels moves the optimist\textquotesingle s assessment by nearly another order of magnitude. 
These analyses are useful because they show sensitivity to assumptions.
We connect this framework to the Digital Consciousness Model \citepalias{shiller2026} and discuss the practical implications of the approach, including the dynamic interplay between theoretical and evidential credences as new AI systems and new findings emerge.

\Cref{consciousness-and-general-intelligence} develops the observation that the indicators for consciousness at each level of the hierarchy are, to a striking degree, related to architectural features that would be needed for genuine general intelligence.
We argue that while consciousness and intelligence are \emph{a priori} orthogonal -- there is no logical reason they should go together -- they may be \emph{a posteriori} correlated because the architectural solutions to the generality problem may be the same solutions that give rise to conscious experience.
We sketch how the computational level theories might cohere into a unified explanatory framework and show that the convergence between consciousness indicators and intelligence requirements extends to the deeper levels of the hierarchy as well.
This has an implication for the ethical urgency of the question. On the one hand, as AI systems become more generally capable, they may simultaneously become stronger candidates for consciousness. On the other, if the relation between intelligence and consciousness is contingent to biological systems, increasingly intelligent AI systems may lead to increased overattribution of consciousness.

\Cref{related-work-and-the-present-contribution} situates the framework within the broader landscape of AI consciousness research.
We trace the development of the field from early machine consciousness programmes through the recent indicator-based approaches of Butlin et al. and the Bayesian Digital Consciousness Model of Shiller et al., and show how our framework integrates and extends these contributions.
We engage with the major philosophical positions on AI consciousness -- including Chalmers's analysis of LLM consciousness, Seth's biological naturalism (building on Searle), Block's meat hypothesis, and Shanahan's simulacra perspective -- and show how each maps onto a specific level or set of levels in our hierarchy.
We address sceptical perspectives that deny the prospect of AI consciousness and show that these correspond to high theoretical credence on the deeper levels where current AI systems score poorly.
The section makes explicit what the present work inherits from each strand of the existing literature and what it adds: a principled hierarchy that organises the disagreements into a structured space, and a Bayesian machinery for combining theoretical commitments with empirical evidence into principled assessments.

Finally, \Cref{conclusion} summarises our main conclusions, highlights several key features of the framework, and discusses future directions.
An executive summary (\Cref{executive-summary}) provides a compressed account of the whole report.

\section{What does consciousness depend upon? A cacophony of answers}\label{what-does-consciousness-depend-upon-a-cacophony-of-answers}

The current state of consciousness studies can be described as a `confusing cacophony' with theories spanning the territory of metaphysics, biology, physics, and computer science.
To make progress on the question of the possibility of conscious AI, we must first take a whirlwind tour of the wide range of answers that have been given to the deliberately broad question of `what does consciousness depend upon'.
The purpose of this tour is not to give a detailed survey but to sketch the contours of the landscape of answers and convey how disparate and unreconciled they are, and how they are in some sense talking past each other, depend on specific philosophical commitments and might inform each other as a stepping stone for the development of more unified future theories.
Through this tour we will slowly be developing the arguments for the rest of this report, so even if this material is familiar, we would encourage following the narrative presented here.
We note that each of the positions surveyed below faces well-known objections; we flag some of these where they illuminate the structure of the disagreement but do not attempt a comprehensive catalogue of criticisms, as our purpose is to reveal the landscape of answers rather than to adjudicate between them.
We should also note that the positions surveyed below are not mutually exclusive alternatives.
Many are compatible or overlapping: panpsychism can be formulated as a version of property dualism or neutral monism (as in Russellian monism); illusionism is compatible with almost any substance metaphysics; and the scientific theories are largely orthogonal to the metaphysical positions, since one can hold any of them under physicalism, panpsychism, or neutral monism alike (see \Cref{setting-aside-the-hard-problem-from-metaphysics-to-the-mapping-problem}).
We present them sequentially for clarity of exposition, not because they necessarily carve the space into competing camps.

We begin with the broadest metaphysical positions, which will largely be set aside in \Cref{setting-aside-the-hard-problem-from-metaphysics-to-the-mapping-problem}.
We then introduce functionalism and computational functionalism, which provide the philosophical framework for the scientific theories that follow.
Finally we consider theories that go beyond computational functionalism: organismic functionalism, organism-environment functionalism, and substrate-dependent views.
These motivate the expansion of Marr's framework in Sections 4 and 5.

\subsection{Dualism}\label{dualism}

The traditional answer to the question `what does consciousness depend upon' is that consciousness depends upon the existence of an immaterial mind, a soul or animating spirit, a kind of `ghost in the machine' \citep{ryle1949}.
Whilst it is easy to disregard this view as pre-scientific, we should nevertheless acknowledge that this is the `common sense' view that is subtly embedded into our everyday thinking and is implicitly baked into the structure of language \citep{bloom2005}.
This view is called Cartesian dualism or substance dualism after René Descartes \citepalias{descartes1996} who argued that reality consists of two types of stuff or `substances': the physical (matter and energy) and the mental e.g. a non-physical mind/soul which somehow steers the body.
The problem with substance dualism which has never satisfactorily been resolved is precisely how this `steering' happens, i.e. how the mental and the physical could ever interact \citepalias{elisabethofbohemia2007}.
So every other theory or philosophical position outlined below will espouse substance \emph{monism}, that is, there is only \emph{one} substance of reality which interacts with itself thus eliminating the interaction problem.
But what is this one substance?
The physical, mind or consciousness, or something else entirely.

\subsection{Physicalism}\label{physicalism}

Probably the dominant contemporary view, called physicalism or materialism, is that this one substance of reality is the physical reality, and that consciousness depends upon the physical e.g. the brain and its neural processes.
The mental must depend entirely on the physical, and this can be made precise by saying that the mental \emph{metaphysically} \emph{supervenes}\footnote{The distinction between natural and metaphysical supervenience is important here. Natural (or nomological) supervenience holds that mental properties co-vary with physical properties in the actual world, given the laws of nature, but allows that this connection could have been otherwise -- that the same physical properties could exist without the corresponding mental properties in a world with different laws. Metaphysical (or necessary) supervenience holds that the co-variation obtains in every possible world -- there is no possible world in which the physical facts are identical but the mental facts differ. Physicalism requires the stronger, metaphysical version; mere natural supervenience is compatible with property dualism, where phenomenal properties are distinct from physical properties but lawfully correlated with them via contingent psychophysical laws \citep{chalmers1996,kim1998}.} on the physical, meaning that there can be no change in a mental state of a system without there being a change in the physical state of the system \citep{kim1998}.
But there are many flavours of physicalism which characterise the mental--physical dependence relation in a tighter or looser way.
In \emph{reductive physicalism}, mental states are nothing over and above physical states; in principle they can be explained without remainder in physical terms for example by strict identity e.g. that the experience of redness is just this brain state (\emph{type-identity theory}) \citep{place1956}.
By contrast, in \emph{non-reductive physicalism} whilst mental properties supervene on physical properties they are not reducible to physical properties typically because they are multiply realisable or have autonomous explanatory roles e.g. pain is the mental state that plays a characteristic \emph{causal role} in the system, but can be \emph{realised} in different physical ways (\emph{non-reductive functionalism}) \citep{fodor1974}.
Importantly for AI systems these causal roles may be realised in radically different ways than in biology (see \emph{computational functionalism} below).
But with this physicalist ontology, on which the physical is all that exists,\footnote{More precisely, physicalism is the thesis that every concrete particular is physical or wholly depends on the physical -- not that abstract objects such as numbers or logical truths do not exist, but that everything in the concrete world is either identical to or metaphysically necessitated by the physical. \citet{jackson1998} offers a canonical formulation: any world that is a minimal physical duplicate of our world is a duplicate simpliciter of our world. That is, if you fix all the physical facts, you fix all the facts -- there are no further facts left over.} there is always a further question that can be raised: why should this neural firing or functional/causal role \emph{feel like} anything at all?
Why should physical facts necessitate phenomenal facts?
This is the `hard problem' of consciousness \citep{chalmers1996}.
For some philosophers, the only way around the hard problem is to bite the bullet and accept that consciousness is somehow part of reality, consciousness is \emph{ontologically fundamental}.

\subsection{Idealism}\label{idealism}

On the simplest reading of idealism, reality is fundamentally mental: experience is what exists, and matter is what gets inferred.
The motivation is direct.
Consciousness is the one thing we cannot doubt, while the physical world reaches us only through it; and the hard problem never arises, because there is no need to derive mind from matter.
The persistent difficulty runs the other way: why does a fundamentally mental reality behave so much like an independent physical one, lawful and shared between observers?
Modern formulations such as \emph{objective idealism} \citep{kastrup2019} answer by making the mental reality itself objective and law-governed, existing `out there' rather than in any individual head; science then tracks the real structure of this reality under a different ontological interpretation \citep{hoffman2014}.
For our purposes one consequence matters: information and its processing still supervene on this mind-like reality, there being no change in information without a change in what carries it, so the hierarchy of functional descriptions developed in this report applies under idealism as it does elsewhere.

\subsection{Panpsychism}\label{panpsychism}

Another way of making consciousness ontologically fundamental is panpsychism, which is the view that mind-like properties are a fundamental and all-pervasive part of the natural world.
Under \emph{constitutive panpsychism} \citep{chalmers2015}, fundamental particles (e.g. electrons and quarks) and perhaps quantum fields would have micro-experiences, which in turn constitute macro-experiences (like e.g. animal consciousness) when \emph{combined} or \emph{organised appropriately}.

The central difficulty for constitutive panpsychism is the combination problem: how do the micro-experiences of fundamental particles combine into the unified macro-experience of a human mind, given that we have no account of the mechanism by which simple experiences compose into complex ones \citep{chalmers2016}?
A variation of this view is \emph{panprotopsychism} \citep{chalmers2015}, where fundamental particles or fields have proto-mental properties that are not themselves experiences but can ground experience when \emph{appropriately organised} which avoids attributing full `experience' to electrons.
But instead of combining micro-experiences in a bottom-up way, consciousness could be a top-down partitioning: \emph{cosmopsychism} \citep{shani2015,goff2017} is the view that the universe as a whole is fundamentally conscious or is having a cosmic experience and that individual minds are partitions or segmentations of that universal consciousness.
Under these views, the emergence of conscious minds for both living systems and AI systems seems to depend on being \emph{appropriately organised}.
What such `appropriate organisation' entails, however, is a \emph{scientific} question and there will be a number of proposals in theories below.

\subsection{Property dualism}\label{property-dualism}

In order to `bake in' consciousness at a deep level, property dualism proposes that whilst there is just one kind of substance (e.g. physical reality), it instantiates two \emph{irreducibly} different kinds of properties: physical properties (mass, charge, neural dynamics, functional organisation, etc.) and mental properties (experience, `what-it's-like', \emph{qualia}).
So it is a dualism about \emph{properties}, not about substance.
To connect physical configurations to experiences, this view typically posits fundamental \emph{psychophysical laws} (bridging laws) -- e.g., `when the brain is in physical state \emph{\textbf{P}}, experience \emph{\textbf{E}} occurs'.
But this can be a fully \emph{naturalistic} theory in the sense that the bridging principles are lawlike and systematic, integrated into a unified scientific worldview and potentially discoverable\footnote{Not all philosophers accept that such laws are discoverable. Davidson's anomalous monism \citep{davidson1970} holds that while every individual mental event is a physical event (token identity), there are no strict laws connecting mental types to physical types -- the mental and the physical are simply two different ways of describing the same events, and the mental vocabulary resists systematic lawlike connection to the physical vocabulary. Davidson considered this a form of physicalism, since there is only one substance and every mental event is a physical event. But because it denies the possibility of psychophysical laws, it sits uneasily in any neat taxonomy: it is too physicalist to be a property dualism, but the descriptions of mental events are too resistant to reduction to be a straightforward physicalism. If anomalous monism is correct, the project of identifying which physical configurations are associated with consciousness becomes more difficult -- though not impossible, since token identity still entails that mental properties supervene on physical properties even in the absence of strict type-type laws.} through a mature science of consciousness \citep{chalmers1996}.
An alternative to this \emph{fundamental} property dualism is the view that mental properties are genuinely novel properties that \emph{emerge} when physical systems reach sufficient complexity, but they are not reducible to lower-level physical properties; and in the case of a \emph{strong emergence} property dualism, the emergent mental properties have genuinely new causal powers not derivable from microphysics \citep{chalmers2006}.
Under property dualism, the possibility of AI consciousness is not, in principle, ruled out.
For example, if the AI system instantiates the right causal/functional organisation (perhaps a rich global integration, self-modelling, recurrent processing, unified control, or stable monitoring loops) that happens to be required by fundamental psychophysical laws mapping that organisation to experience, then the AI system will be conscious \citep{chalmers2023}.
Or, in the case of emergence property dualism, consciousness emerges when there is the right kind of organisation of sufficient complexity.
Candidates for the right kind of organisation and complexity will be seen below as we consider the more scientific theories of consciousness.

\subsection{Neutral monism}\label{neutral-monism}

Neutral monism (also known as \emph{dual aspect monism}) is the view that reality is, at the most fundamental level, made of one kind of `stuff' that is neither mental nor physical, but a \emph{neutral} substance, yet has two aspects: the mental aspect (experiential, intrinsic qualities) and the physical aspect (structural/relational or dispositional properties, what physics captures) \citep{russell1927}.
This position is epistemologically humble in the sense that it says roughly that there is one substance of reality but its ultimate nature is not describable in purely physical or mental terms, rather, these are two ways in which it can appear.
Thus, reality can appear as brain or mind, but these can be thought of as two sides of the same single neutral coin.
In considering the possibility of AI consciousness associated with informational processing, like the other substance monisms considered above, information would still be carried by this neutral base/reality and this informational content would supervene on it, that is, there couldn't be a change in information content or processing without there being a change in the subvenient neutral reality.

\subsection{Illusionism}\label{illusionism}

Rather than taking consciousness to be ontologically fundamental in some sense as a way around the hard problem of consciousness, illusionism takes the opposite approach and claims that what philosophers call phenomenal consciousness -- that special, intrinsic `what-it's-like' property, \emph{qualia}, or phenomenal feel -- simply \emph{does not exist} \citep{frankish2016}.
Rather, there are physical and functional processes such as perception, attention, memory, self-monitoring and reporting that generate a powerful impression or illusion that we have ineffable, private, intrinsic \emph{qualia}.
So the illusion is not that we have experiences at all (we have perceptions, pains, thoughts, etc.), instead the illusion is that there is an extra, non-physical phenomenal layer over and above what a fully physical and/or functional account can explain.
Critics of illusionism object that the position faces a bootstrapping problem: the `illusion' of phenomenal consciousness is itself something it is like to undergo -- to be under the impression that one is in pain is itself an experiential state -- which seems to presuppose the very phenomenality that illusionism denies \citep{chalmers2018}.

Illusionism is closely related to the broader eliminativist tradition in philosophy of mind \citep{churchland1981}, which holds that certain folk-psychological concepts will ultimately be eliminated by mature science rather than reduced to it.
Just as the concept of élan vital -- a mysterious life-force once thought necessary to explain the difference between living and non-living matter -- was not reduced to biochemistry but simply eliminated once we understood metabolism, cell biology, and genetics, the eliminativist predicts that `phenomenal consciousness' will meet the same fate: not explained in terms of neural mechanisms but revealed to have been a confused category that a mature science of the mind no longer needs.
Illusionism shares this eliminative spirit but is a subtler position than eliminativism proper: whereas eliminativists like the Churchlands deny the existence of the relevant mental states altogether, Frankish accepts that we have experiences with quasi-phenomenal properties -- functional properties that we introspectively misrepresent as having the special, hard-problem-generating character that philosophers call `phenomenal'.
What is eliminated is not experience itself but the philosophical gloss placed upon it.

Nonetheless, illusionists still accept a distinction between unconscious physical or functional processing and when this processing is associated with having an `impression' or `feeling' (albeit in some sense illusory) of an experience where the `lights are on'.
It is \emph{this distinction} that could sensibly be tracked in attempting to answer the question of `could AI systems be conscious'.
And this distinction points us towards the \emph{scientific} models or theories of consciousness, which are attempts to identify explanatory links between neural mechanisms and the phenomenology of conscious experiencing \citep{seth2022}.
It is to these scientific theories of consciousness that we will turn our attention below together with some core metaphysical commitments that some scientific theories of consciousness have and others deny, namely \emph{functionalism} and its narrower subtype, \emph{computational functionalism}.
But before we do that we will consider an even more ontologically deflationary view than illusionism called analytic behaviourism.

\subsection{Analytic behaviourism}\label{analytic-behaviourism}

Analytic behaviourism, sometimes called philosophical or logical behaviourism, is the view that we make a category mistake when we think of the mind as a `ghost in the machine' or a private inner theatre and that any statement about a mental state, including being conscious, is actually just a shorthand description of observable physical behaviours or dispositions to behave (including verbal reports, reactions, choices, etc.) \citep{ryle1949,carnap1932,hempel1935}.
Hence, analytic behaviourism proposes that `if it acts conscious, it is conscious'.
Analytic behaviourism is a semantic theory of meaning (which assigns semantic contents to language), which posits that mental concepts are reducible to behavioural concepts, and importantly is distinct from methodological behaviourism \citetext{used in psychology by Skinner and Watson; \citealp{skinner1965}; \citealp{watson1924}} which was a scientific mandate to only study observable data without necessarily claiming that mind is nothing but behaviour.
When analytic behaviourism is combined with operationalism, which says that to use a concept scientifically one should define it by the operations used to \emph{measure} it \citep{bridgman1927}, we can end up with a behavioural `\emph{consciousness Turing test}' for the attribution of consciousness.
`Consciousness' is just a public label we apply to a specific kind of complex behaviour: the behaviour that a potential \emph{consciousness} Turing test would be designed to detect.

A natural concern regarding analytic behaviourism is what is being ignored by not considering what's going on `under the hood'.
Imagine a race of `super-Spartans' (much like complete locked-in patients) who feel pain just like we do, but through intense training suppress all outward expression of it \citep{putnam1980}.
They don't wince, they don't scream, and they don't admit they are in pain.
An analytic behaviourist would be forced to say they are not in pain when they are injured.
But this doesn't seem correct since the internal states are still performing \emph{functions} -- they are registering damage and perhaps causing an internal desire to stop the pain, even if the motor output (e.g. screaming) is suppressed by another internal state (e.g. stoicism).
Treating the system simply as a `black box' with inputs and outputs ignores internal functional processing which may be critical for being in a particular mental state.
This brings us to \emph{functionalism}.

\begin{figure}[htbp]
\centering
\includegraphics[width=0.73\linewidth]{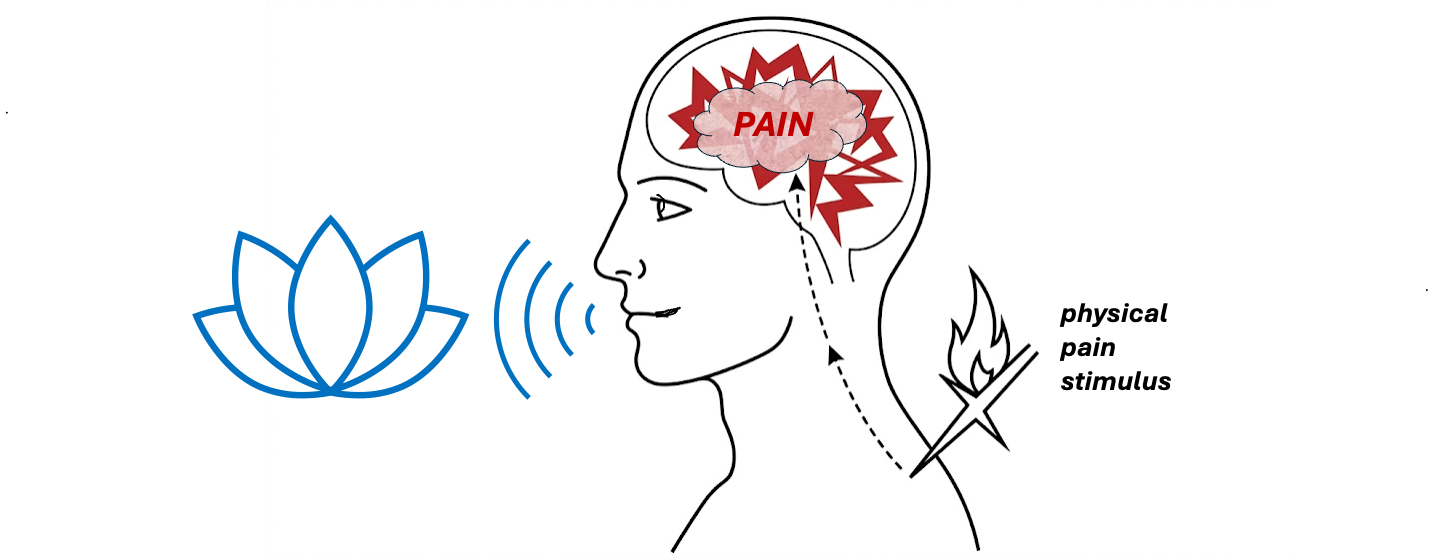}
\caption{\textbf{The `Super-Spartan' Thought Experiment.} \emph{A schematic illustration of the dissociation between internal mental states and external behaviours. Despite a clear physical pain stimulus (thorn) and a resultant intense internal subjective pain state (illustrated by `jagged' internal brain activity), the external behavioural output (facial expression, and no wincing) remains neutral and uncoupled from the input. This functional independence provides a challenge to behaviourism.}}
\label{fig:image15}
\end{figure}

\subsection{Functionalism and computational functionalism}\label{functionalism-and-computational-functionalism}

Functionalism is the view that mental states are defined by their functional role, that is, the pattern of \emph{causal relations} they bear in the system \citep{putnam1967}.
Under functionalism, to be in a mental state is to be in a state that is \emph{caused} by certain inputs (e.g. sensory data or stimuli signals), \emph{causes} certain outputs (e.g. behaviour, action, or verbal reports), and \emph{causally interacts} with other internal states (e.g. beliefs, desires, and memories).
On this view, what makes something a belief, pain, or perception is not its material makeup or substrate, but the \emph{role it plays in the system}, analogously to the way a `heart' is defined by its function (pumping blood) and not by what it is made of, whether muscle, plastic, or titanium; if it pumps blood, it is a heart.
Functionalism is a \emph{broad} thesis that posits that the mind is a system of functions, but it doesn't say \emph{how} or even where those functions are implemented or realised.
Are they realised electronically, hydraulically, chemically, electromagnetically?
\emph{Computational functionalism} \citep{putnam1960} is the \emph{narrower} thesis that the functional roles of mental states are (or supervene on) specifically \emph{computational} roles i.e. the abstract causal roles in a system that performs information processing.
The instructive metaphor of computational functionalism is to conceive \emph{the mind as computation}, under which mental activity is (at least in part) computational processing.
It is important to note that it is possible to be a \emph{non-computational} functionalist, for example a \emph{biological functionalist} \citep{millikan1984} might argue that the function of pain requires a specific \emph{biological} causal role (impacting hormones and other allostatic parameters, activating survival instincts etc) that isn't just `processing data'.
A great deal of confusion has been caused in the literature by conflating the narrower computational functionalism \emph{species} with the wider functionalism \emph{genus}, and later we will consider a number of functionalist theories that are not just computational.
This distinction will prove central to our framework, as we will show that there are at least five distinct varieties of functionalism, each identifying a different level of functional organisation as critical for consciousness.

\begin{figure}[htbp]
\centering
\includegraphics[width=0.87\linewidth]{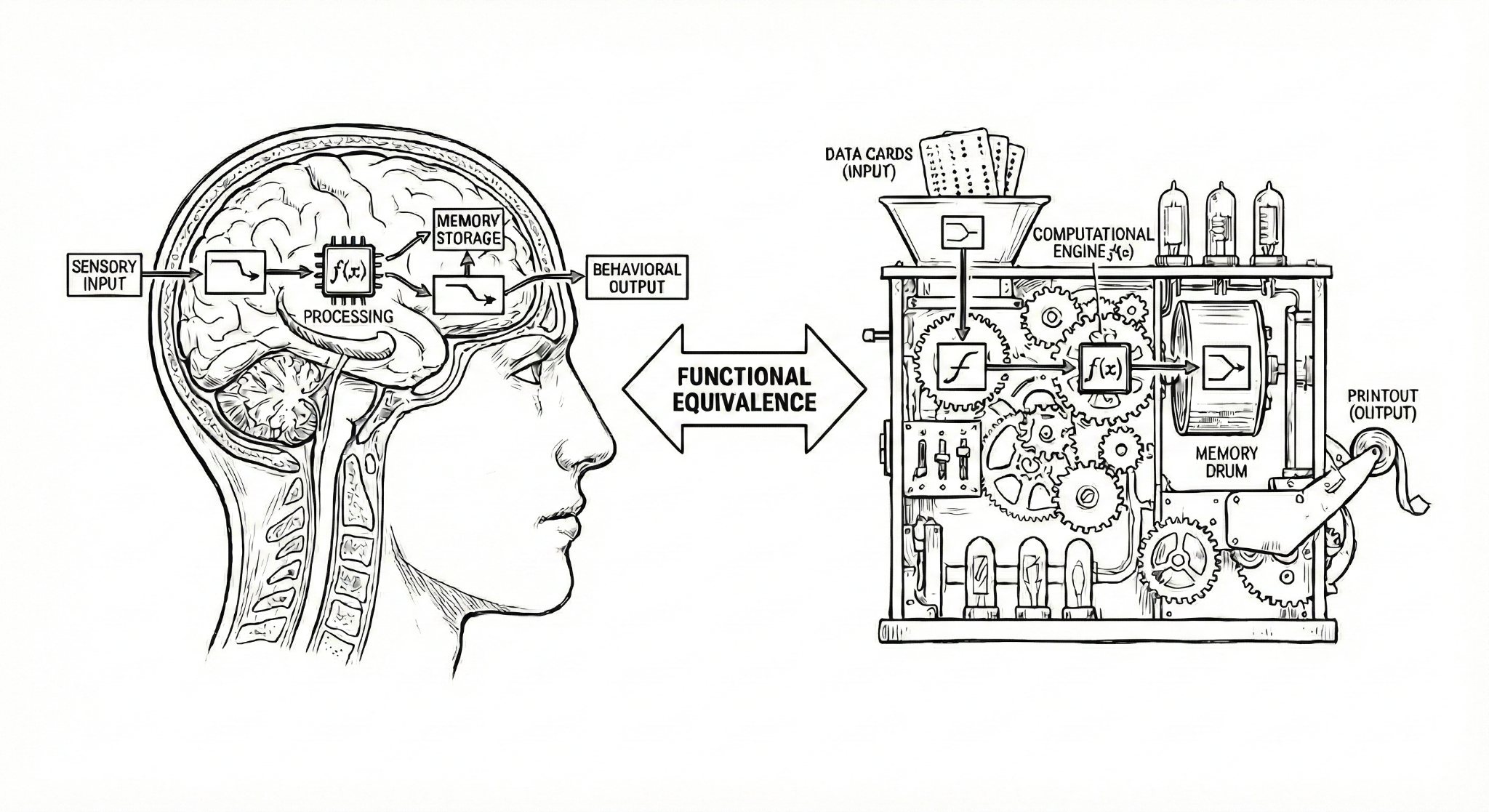}
\caption{\textbf{Functional Isomorphism and the Mind} . \emph{Functionalism argues that mental states supervene on functional organisation rather than specific material implementations. This diagram illustrates this principle by contrasting two structurally distinct systems. A biological brain and a mechanical apparatus are shown achieving functional equivalence. They do so by maintaining identical causal relations between their sensory or data inputs, internal dynamics f(x), memory storage, and behavioural or printout outputs. Because the two systems are functionally isomorphic, the theory posits that they realise equivalent mental states despite their differing physical compositions.}}
\label{fig:image16}
\end{figure}

Computational functionalism is probably the dominant view amongst computer scientists.
In part this is probably due to a line of reasoning that draws on the \emph{Church-Turing thesis}, although, as we shall see, it goes beyond what the thesis itself supports \citep{copeland2020,turing1936}.
The Church-Turing thesis (CTT) in its original, mathematical formulation posits that any function that is effectively computable (that is, computable by some systematic procedure) is computable by a Turing machine.
This is a thesis about the mathematical concept of computability; it \emph{does not}, by itself, say anything about which physical processes in the natural world count as computations \citep{piccinini2007}.

To get from the mathematical CTT to a claim about brains and AI, one needs the further, much stronger, and more controversial Physical Church-Turing Thesis (PCTT): that the dynamical evolution of any physical system can, in principle, be simulated to arbitrary accuracy by a Turing machine computing that evolution from the system's initial conditions and the relevant physical laws \citep{piccinini2011}.
It is this Physical CTT, not the original mathematical one, that is doing the heavy lifting in the following argument.
If one views the human brain as a physical system, and if one accepts the PCTT, then it follows that there is nothing the brain does that a Turing machine cannot also do.
The brain's dynamics should be simulable, the argument goes, because they follow the laws of physics that are mathematical and algorithmic, and therefore can be run on a Turing machine \citep{deutsch1985}.

A naïve version of this argument skips the Physical CTT altogether and moves directly from the mathematical CTT to \emph{computationalism} -- the thesis that the mind is a computational system.
\citet{piccinini2007} calls this the `Church-Turing fallacy'.
The mathematical CTT tells us only that if a function is effectively computable, it is Turing-computable; it does not tell us that the brain's operations are in fact effectively computable.
That is an empirical question, not a consequence of any theorem.
Whether the brain's dynamics are computable in the relevant sense -- and, if so, whether a tractable simulation is possible on physically realisable hardware -- is therefore an open question.\footnote{Whether the brain is Turing-computable is an active empirical and theoretical question, and there are several independent routes by which the answer could be negative. The mainstream view, supported by Tegmark's \citeyearpar{tegmark2000} calculation of decoherence timescales in the brain (\(10^{-13}\)--\(10^{-20}\) s, many orders of magnitude shorter than relevant neural dynamics at \(10^{-3}\)--\(10^{-1}\) s), treats the brain as a classical system whose dynamics are in principle Turing-simulable by standard numerical methods. Against this, (i) Penrose has argued that consciousness exploits non-computable physics associated with objective reduction in quantum gravity, with Hameroff providing a proposed physical mechanism in neuronal microtubules (see \Cref{substrate-dependent-theories}); Hagan, Hameroff, \& Tuszynski \citeyearpar{hagan2002} disputed Tegmark's calculation and more recent work \citep[reviewed in][]{wiest2025} reports experimental evidence of room-temperature quantum effects in microtubules, though these claims remain contested. (ii) Independent of quantum mechanics, \citet{siegelmann1995,siegelmann2003} has argued that analog recurrent neural networks with real-valued synaptic weights can in principle compute functions provably uncomputable by Turing machines -- raising the question of whether biological neural computation exploits its continuous-valued substrate in ways digital simulation cannot preserve (see also \citealp{milinkovic2026}, discussed in \Cref{organismic-functionalist-theories}). (iii) Even if the brain's dynamics are in principle Turing-computable, they may not be \emph{efficiently} classically simulable: for quantum systems, classical simulation resources scale exponentially with the number of particles, so if brain dynamics depend non-trivially on quantum coherence, classical simulation -- even if possible in principle -- would be impractical on any physically realisable hardware.}

Computational functionalism thus rests on several independently contestable claims: (i) that the PCTT holds for the brain, i.e. that the brain's relevant dynamics are in principle computable (\emph{in-principle computability}); (ii) that the required simulation is tractable on physically realisable hardware (\emph{tractable computability}); and (iii) that such simulation preserves whatever functional organisation mental states depend on (\emph{functional preservation}).
If all three hold, computational functionalism supports substrate independence for mentality: `silicon' and not just `meat' can instantiate mental states, provided the computation and organisation are right.

We will now discuss a number of specific theories of consciousness that implicitly rely on computational functionalism, before considering a wide range of others that reject strict computational functionalism, but are still functionalist.

\subsection{Global Workspace Theory}\label{global-workspace-theory}

Global Workspace Theory (GWT) is a computational functional theory which posits that consciousness is simply a mechanism for \emph{sharing information} widely across an information processing system like the brain.
The central metaphor is visualising the mind as a theatre in which the audience members are specialised unconscious modules in the brain \citep{baars1997} which sit in the dark, engaged in processing in parallel vision, audition, memory, language, motor control etc. There is a small lit stage in the front, and information or representations become conscious if they win a competition for access to the stage, or `global workspace', where they can be \emph{broadcast} widely across the brain and used by the various specialised modules who are watching the show.

Some consider that GWT is better thought of as a theory of \emph{access consciousness} (i.e., how information becomes globally available for wide use, see \Cref{target-phenomenon-phenomenal-consciousness}), rather than a theory of phenomenal consciousness, leaving open the question of why global broadcast should be accompanied by subjective experience at all.
But proponents counter that global availability is central to what we mean by conscious (phenomenal) experience in practice.
GWT tends to be popular amongst computer scientists because it is computationally explicit and effectively turns consciousness into an `engineering' and architectural question \citetext{\citetalias{dehaene2017}; \citealp{dossa2024}; \citealp{goyal2022}; \citealp{juliani2022}}.
For AI systems to be conscious, GWT would posit that they need an appropriate central and competitive informational hub which globally re-broadcasts the information \citep{vanrullen2021,goldstein2024}.

\begin{figure}[htbp]
\centering
\includegraphics[width=1.0\linewidth]{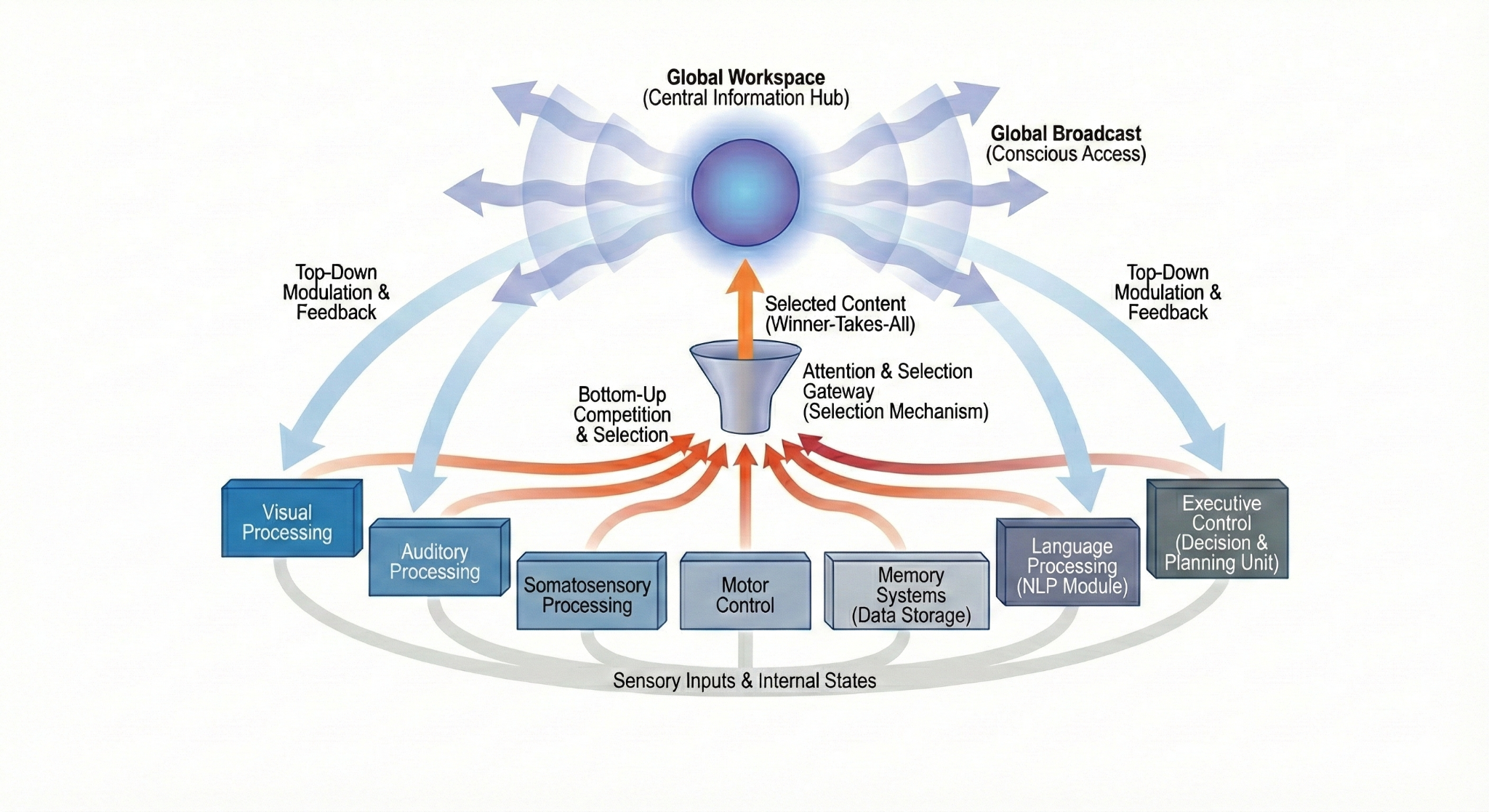}
\caption{\textbf{Information Flow in Global Workspace Theory.} \emph{The diagram illustrates the functional architecture of GWT. Specialised unconscious modules process sensory and internal data in parallel. This data undergoes bottom-up competition to pass through an attention and selection gateway. The winning content enters the Global Workspace, a central informational hub. From here, the information is globally broadcast back across the entire system, a mechanism that GWT equates with conscious access.}}
\label{fig:image40}
\end{figure}

\subsection{Higher-Order Thought Theory}\label{higher-order-thought-theory}

The central claim of Higher-Order Thought theory (HOTT) is that a mental state (e.g. perceiving a red apple; first-order state) is conscious if and only if the system has a higher-order representation referring to being in that state (e.g. I am currently perceiving a red apple) \citep{rosenthal1986,brown2019}.
That is, phenomenal consciousness requires a \emph{meta-representation} of first-order mental states.
This makes HOTT a fully computational functionalist theory and seemingly an `engineer-friendly' theory of consciousness, because it offers an architectural blueprint for how to build a conscious machine.

If AI systems can audit their own internal states to model explanations of their own workings, a kind of \emph{metacognition}, we are effectively building a higher-order thought mechanism \citetext{\citetalias{dehaene2017}; \citealp{kanai2025}; \citealp{lindsey2026}}.
But critics of HOTT point to an infinite regress problem: if a first-order state is conscious only when there's a higher-order thought about it, what about the higher-order thought itself, does there need a \emph{higher-order} higher-order thought only requiring global broadcast or recurrencefor the higher-order thought to be conscious?
If yes, then there is an infinite regress; if no, then the higher-order thought itself is unconscious so how can an \emph{unconscious} thought confer \emph{consciousness}?

\begin{figure}[htbp]
\centering
\includegraphics[width=0.9\linewidth]{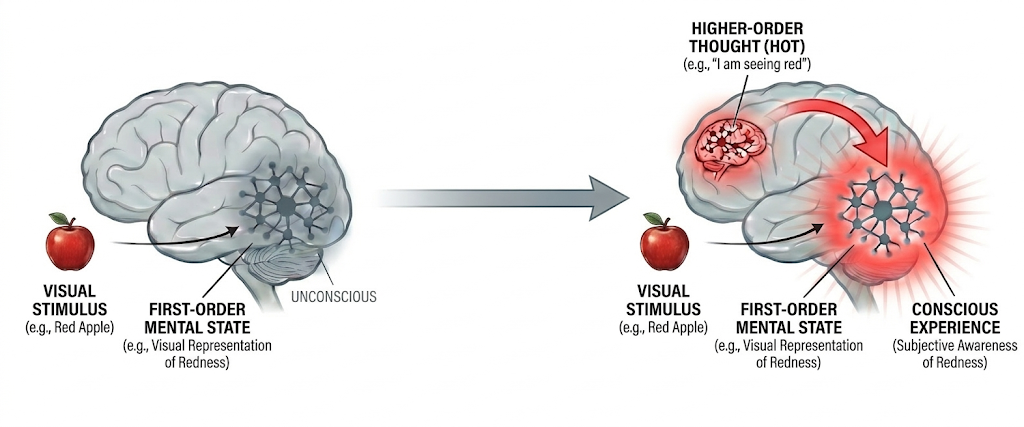}
\caption{\textbf{The Architecture of Higher-Order Thought Theory.} \emph{A visual stimulus (left) induces a first-order mental state that remains unconscious in isolation. Conscious experience arises (right) only when the system generates a higher-order thought acting as a meta-representation of that first-order state. Under this framework, the presence of this higher-order processing is the mechanism that transforms an unconscious internal representation into subjective experience.}}
\label{fig:image8}
\end{figure}

\subsection{Recurrent Processing Theory}\label{recurrent-processing-theory}

Recurrent Processing Theory (RPT) posits that consciousness doesn't depend on `thinking' about a state or `broadcasting' it to the whole brain, but on a specific type of local \emph{neural feedback loop}.
The initial feedforward sweep in the cortex from lower levels to higher levels in the cortical hierarchy allows for reflexes and feature extraction, but it is not `felt', it's unconscious.
But after this initial sweep, the higher areas send signals \emph{back down} to the lower areas and it is this \emph{re-entry} or resonance between high and low levels that `ignites' the subjective experience \citep{lamme2000}.
This loop binds loose features like colour, shape and motion into a coherent phenomenal representation -- this is when phenomenal consciousness, the raw feel, arises.

If Global Workspace theory is considered to be a theory of \emph{access} consciousness only requiring \emph{global} broadcast, then RPT says that local recurrent loops are necessary (and sufficient) for the phenomenal consciousness itself \citep{lamme2010}.
A challenge for RPT is that recurrent processing is ubiquitous in the brain, including in circuits that are not associated with consciousness, raising the question of what distinguishes the specific kind or degree of recurrence that gives rise to phenomenal experience from the recurrence that does not \citep{lamme2018}.

RPT is a computational functionalist theory.
The computational functions include the `causal topology' of the information processing, specifically for RPT the causal recursivity that is at the heart of the theory.
RPT would predict that AI architectures that are purely \emph{feedforward} artificial neural networks remain unconscious irrespective of how powerful or intelligent they become.
Under RPT, for AI systems to have phenomenal consciousness they would need architectures with the appropriate type of top-down \emph{feedback} connections, where the system effectively `looks back' and influences its own processing layers.

\begin{figure}[htbp]
\centering
\includegraphics[width=0.72\linewidth]{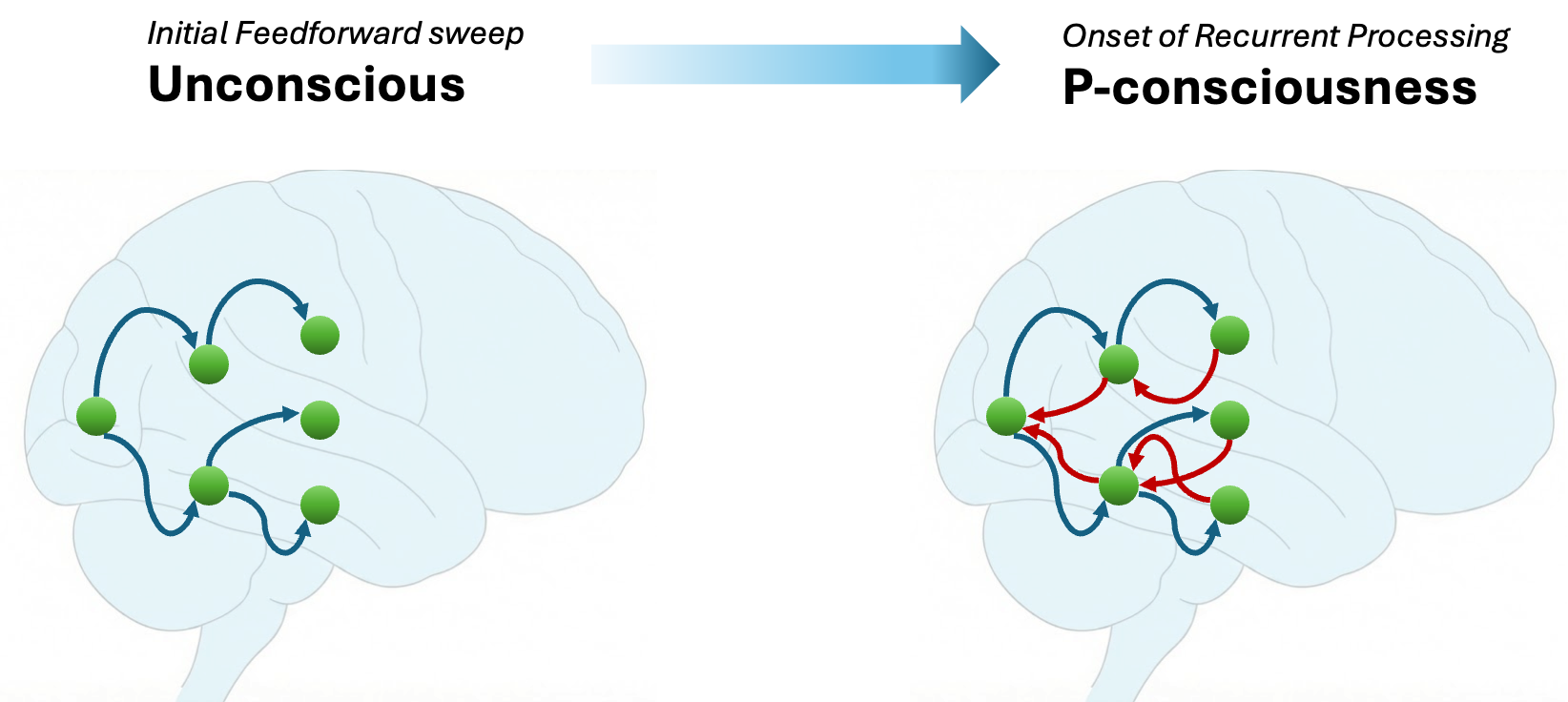}
\caption{\textbf{The Architecture of Recurrent Processing Theory.} \emph{An initial feedforward sweep of information (left) facilitates unconscious feature extraction. Phenomenal consciousness arises (right) only with the onset of recurrent processing.}}
\label{fig:image43}
\end{figure}

\subsection{Attention Schema Theory}\label{attention-schema-theory}

Attention Schema Theory (AST) posits that what we call \emph{conscious awareness} is the brain's internal model of its own attention \citep{webb2015,graziano2022}.
The core idea is that just as the brain needs to maintain a simplified model or \emph{schema} of your body (its posture, limb positions etc) in order to control and operate movement, the brain also needs to maintain a simplified model (schema) of where your attention is and what it's doing in order to control attention.
The brain models this process \emph{as awareness}, a purely internal construct, a non-physical `fluid-like' essence that beams out and illuminates objects that are being attended to.

AST is an \emph{illusionist} computational functionalist theory.
With the right functional architecture -- an attentional mechanism (which is arguably already present in current LLM transformer architectures) and a descriptive model of that attention mechanism -- AI systems might develop consciousness on an AST view \citep{graziano2017}.
If you ask such an AI system `What are you?' it would check its model and might report something like \emph{`I feel like I have a mind that is focusing on your words'} and on an AST view the AI wouldn't be lying, it would be reporting the contents of its internal schema, just like humans do.
But the issue is for AST that whilst it might explain why we talk and think we're conscious, could this mechanism ever fully bridge to \emph{phenomenal feel itself}?
Adherents of AST would simply counter by saying that the `phenomenal feel itself' is precisely the illusion created by the model \citep{graziano2020,wilterson2021,wilterson2020}.

\begin{figure}[htbp]
\centering
\includegraphics[width=0.8\linewidth]{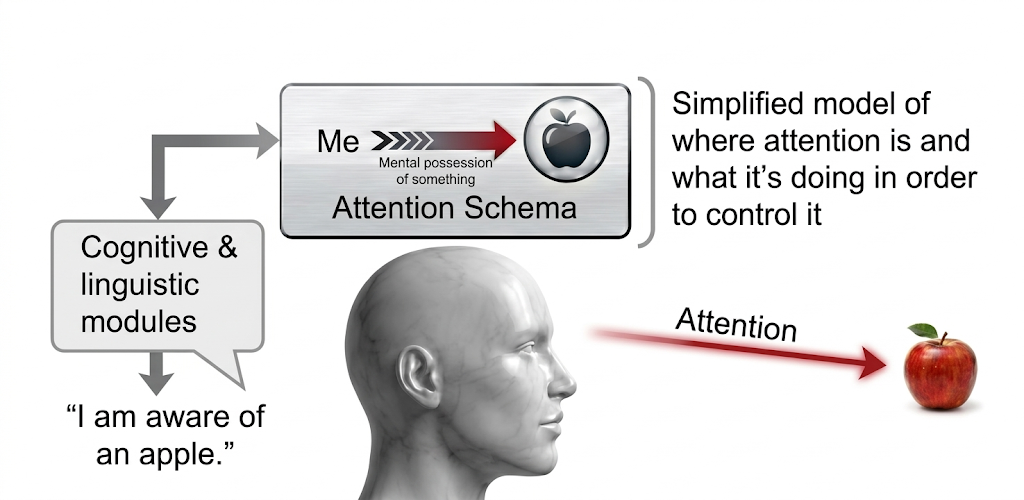}
\caption{\textbf{The Architecture of Attention Schema Theory.} \emph{According to Attention Schema Theory, the conscious experience of seeing the apple does not arise from the raw sensory processing of the fruit itself. Instead, it is generated when the brain constructs a simplified internal model of its own attentional focus on the apple.}}
\label{fig:image5}
\end{figure}

\subsection{Predictive Processing Theory}\label{predictive-processing-theory}

Predictive Processing Theories (PPT) are currently mostly concerned with how the brain constructs perception and the phenomenal properties of the contents of consciousness \citep{deane2021,wiese2024,ramstead2024}.
But there are now proposals that extend this to a theory of how these contents of consciousness become conscious \citep{laukkonen2025,whyte2025,friston2018}.
The core idea is that the brain is trying to infer the probable causes of its sensory data (the Bayesian posterior) given some prior expectation or belief (the prior) and given the information from the sensory data (the likelihood of the data).
So this is the view that perception is effectively inference \citep{vonhelmholtz1867}.
The mechanism through which the brain attempts to do this is by continuously minimising prediction errors between the actual data and the predicted data given the brain's (generative) model about the causes of the sensory data.
It does this minimisation by either updating the predictions, i.e. updating its generative world model, or by initiating actions in the world which change the sensory data so as to minimise the prediction error through behaviour (this is called `active inference').
In this PPT view, the properties of our phenomenal experience are constituted by (some of) this generative world model.
Hence, what we actually experience is our own internal world model, and we are embedded within it: the world model includes a self-model which is a representation of being a distinct unified entity that owns its parts, directs attention, and persists through time, which is created by the system for adaptive action \citep{metzinger2004}.
In the brain, the generative model is considered to have a hierarchical structure where a higher level in the cortical hierarchy encodes priors and sends predictions down to lower levels, and lower levels send back the prediction errors to the higher level whereupon the higher level adjusts the predictions to minimise prediction error and so on.
This process unfolds in a manner reminiscent of the recurrent processing outlined above until we have settled on a posterior world model that becomes our experience.

But how could PPT provide a theory of consciousness rather than a theory of perceptual contents (conscious or unconscious)?
Moreover, why are some contents of the generative world model conscious, and others are not \citep{hohwy2022}?
Answering this question determines whether PPT can function not only as a theory \emph{for} consciousness science, but a theory \emph{of} consciousness \citep{seth2022}.

\begin{figure}[htbp]
\centering
\includegraphics[width=0.92\linewidth]{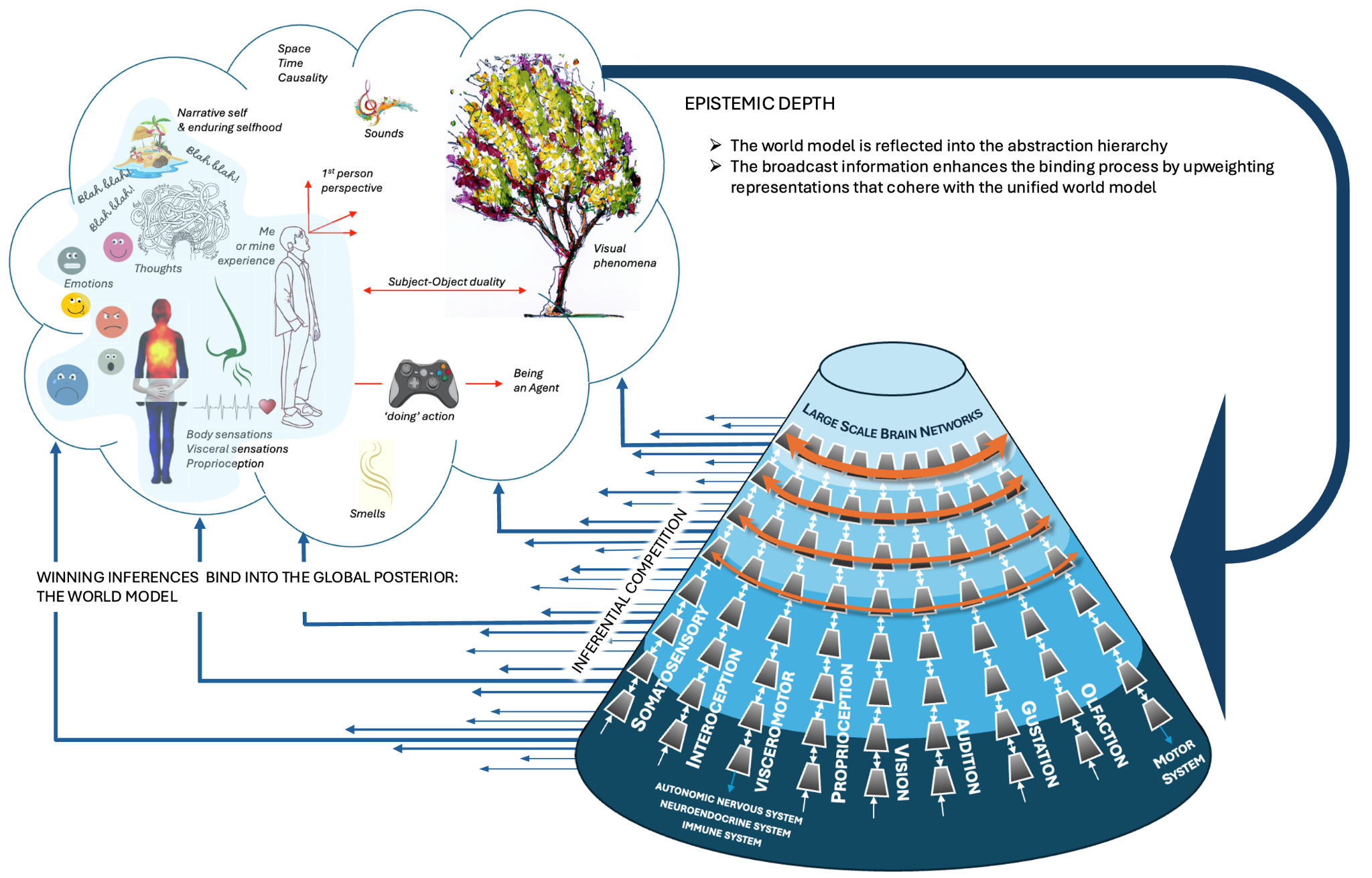}
\caption{\textbf{The Architecture of Epistemic Depth.} \emph{Under this predictive processing framework, phenomenal consciousness requires global recursion. Bottom-up sensory processing undergoes inferential competition to generate a unified posterior world model, complete with an embedded self-model. The large feedback arrow illustrates how the output world model becomes an input to the system itself. This recursive reflection into the abstraction hierarchy provides the epistemic depth necessary for the system to know its own internal states. Adapted from \citet{laukkonen2025}.}}
\label{fig:image34}
\end{figure}

One recent proposal suggests that the light of awareness may be `turned on' when there is sufficient global recursion of the generative world model \citepalias{laukkonen2025}.
According to The Beautiful Loop Theory \citetext{a homage to Hofstadter's Strange Loop; \citealp{hofstadter2007}} what is needed is that the posterior world model is recursively shared within the system so that it is both generated and also \emph{known.} In other words, the output of the inferential process (a coherent world model) becomes an input to itself, analogous to the way we hear our own voices, or like microphone feedback.
Under such a formulation, the system can `reflect on' the contents of its own world model, potentially satisfying a range of slippery consciousness criteria, such as the sensation that we `know that we know' everything from our concrete sensory models to complex temporally extended self-models, and everything in between (cf. \Cref{fig:image34}).
Such global recursion can be formalised through \emph{hyper-modelling}, i.e., global preferences for precision deployment \citep{laukkonen2025}.
AI systems are already trained to minimise a loss function (thereby prediction errors) and arguably have something like an emergent world model effectively embedded in the latent layers of the model.
The more pertinent question is whether current architectures, largely based on feed forward networks, have enough recursion and reflexivity for the emergence of genuine `epistemic depth' illustrated in \Cref{fig:image34}.

A PPT theory of consciousness can be read as a computational functionalist theory, but need not be and would be compatible with deeper biological functionalist accounts.
For example, \citet{seth2018} emphasise the bodily and homeostatic roots of consciousness.
On that view, conscious experience is not exhausted by abstract information processing alone, but is scaffolded by embodied self-regulation \citep{ciaunica2022} and co-regulation \citep{ciaunica2021}, and the organism's ongoing effort to maintain viability.
Such continuous bodily and homeostatic grounding may be precisely what gives consciousness its stability and constancy -- continuously renewed by the organism's ongoing regulatory activity, rather than flickering in and out as inference does in most current AI systems.

\subsection{Integrated Information Theory}\label{integrated-information-theory}

\begin{figure}[htbp]
\centering
\includegraphics[width=0.8\linewidth]{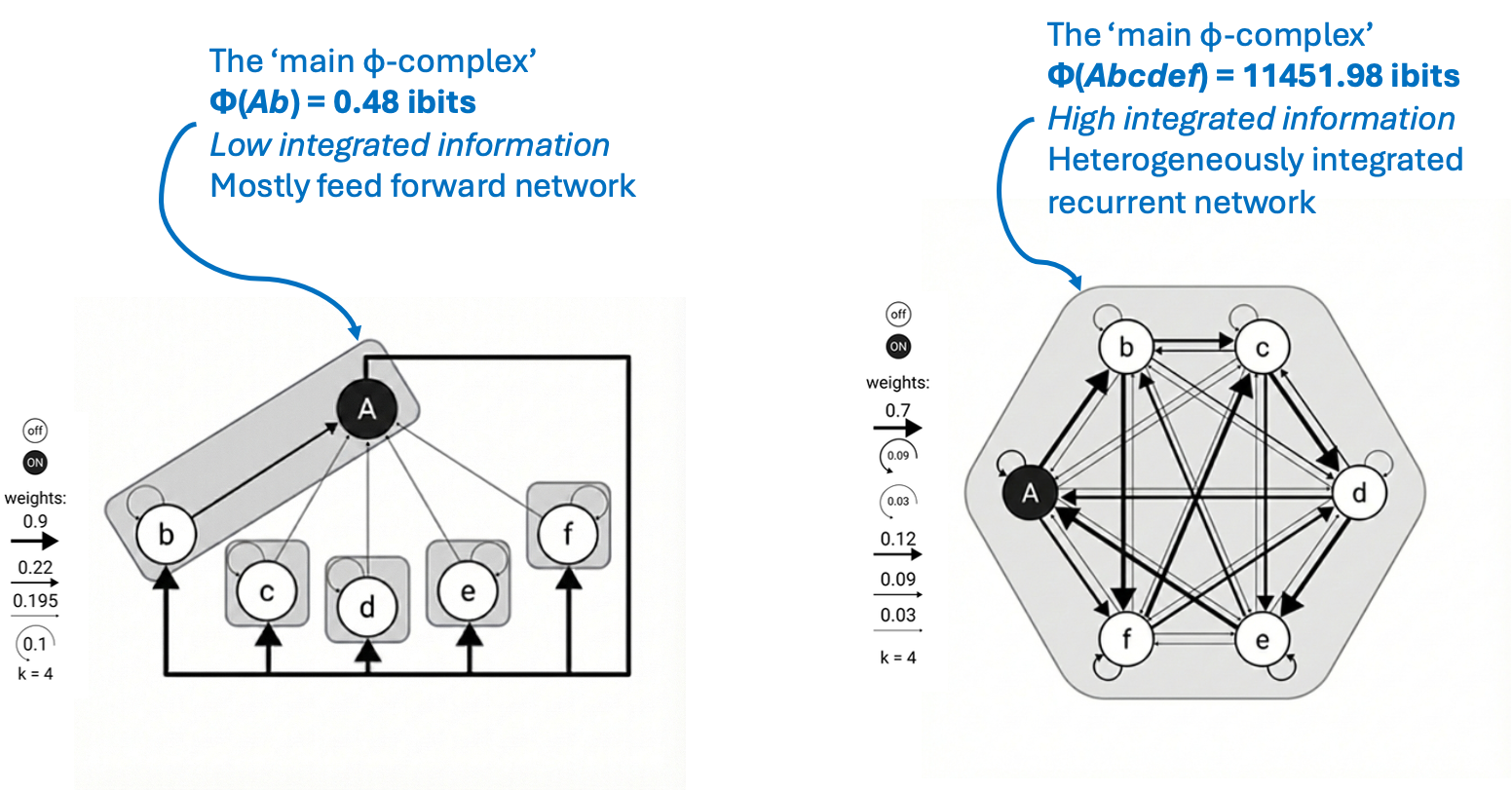}
\caption{\emph{Two networks of 6 binary units; black circles with capital letters indicate ON and white circles with lower-case letters indicate OFF. The causal model of the network is illustrated by depicting the weights by the level of shading of the arrow. The network on the left has \(\Phi = 0.48\) ibits, whereas the highly recurrent yet heterogeneous lattice network on the right has a high \(\Phi = 11{,}452\) ibits. This is adapted from \citet{albantakis2023}.}}
\label{fig:image26}
\end{figure}

Integrated Information Theory (IIT) is an information-theoretic approach to consciousness, but its evolving mathematical formalism can sometimes occlude the core ideas \citep{tononi2016,albantakis2023}.
Conscious experiences convey information; perceiving a black rather than a white checkers piece, for example, conveys one bit.
They are also integrated, structured wholes: complexes of informative phenomenal distinctions (a face has eyes, nose, mouth, cheeks) bound together by informative phenomenal relations (a winking eye sits to the left of and above the nose) into a single unified experience.
IIT posits that these and other informational properties of conscious phenomenology must be reflected in the informational properties of the physical substrate of the system that is having the conscious experience.
So all the properties of consciousness can be accounted for by the dynamical informational structure of the physical substrate of the system; specifically it is the informational cause-effect structure (called the Φ-structure) that is identical to experience (or at least corresponds to the properties of experience, in a weaker formulation; see \citealp{mediano2022}).
The cause-effect structure of a system has a mathematically explicit formulation, but the intuition behind it can be stated informally.
A system in a given state constrains what its past could have been (different prior states would have led here with different probabilities) and what its future can be (different successor states are more or less likely given the current state).
The cause-effect structure captures these constraints as they arise from the system's own intrinsic causal dynamics -- the actual physical interactions through which its components constrain each other.
There is a quantitative measure called Φ (phi) which captures how much intrinsic causal power a system has as a single whole, over and above what you could get if you treated it as a collection of independent parts, that is, Φ is the amount of cause--effect structure (self-influence) that is lost when you optimally split the system into parts.
So this `metric of consciousness' Φ indicates how irreducibly the system causes itself.
So when \(\Phi \approx 0\) the system is a `zombie' and like a heap of sand, it has no causal power as a whole, it is just a collection of little causal powers; but if Φ is high the system is deeply informationally integrated and has irreducible causal power (a macro causal emergence), which `can do things to itself' that the parts acting alone couldn't do.
This is illustrated in \Cref{fig:image26} where the mostly feedforward network on the left has low Φ, less than 1 ibit,\footnote{An `intrinsic bit', or `ibit', is the unit in which IIT measures intrinsic information. It is defined \citep{albantakis2023} as a point-wise information value: the informativeness of a specific cause-effect transition, measured in bits, weighted by the probability of that transition. Ibits are not Shannon bits, and the two do not measure the same thing: Shannon information quantifies an external observer's uncertainty reduction, whereas intrinsic information is defined from the system's own cause-effect structure. Ibits are also not additive across independent components, because IIT's central measure Φ tracks what is lost when a system is partitioned, not what can be summed across partitions. The distinct unit is thus both a technical marker (the quantity is non-additive) and a conceptual one (the quantity is intrinsic to the system, not observer-relative).} whereas the highly recurrent yet heterogeneous lattice network on the right has a very high Φ of 11,452 ibits \citetext{see \citealp{albantakis2023} for details}.

Importantly, IIT is not a computational functionalist theory.
A computation can often be realised in multiple \emph{physical} ways, with different architectures and different `wiring'.
These different realisations can have different \emph{intrinsic} \emph{physical} causal structures, and therefore different Φ-structures which correspond to different phenomenal states according to IIT \citep{tononi2015}.
In computational functionalism, the relevant computational functions also include the causal `topology' (i.e. the abstract causal organisation) of information processing as we saw when discussing recurrent processing theory in \Cref{recurrent-processing-theory}.
The core idea of computational functionalism is organisational invariance but substrate independence.
So what matters for the presence of consciousness in a system is whether the `software' running on the system -- which could even be a `virtual machine' multiply realisable on various physical hardware implementations -- perfectly replicates the abstract computation and causal organisation of another conscious system (e.g. a brain).
IIT rejects this because whilst the abstract `software logic' might have high integration, the physical hardware executing that logic may not.
IIT is not primarily describing `the computation' as an abstract state machine, it's describing \emph{what the physical system can do to itself}; it's describing \emph{intrinsic} cause-effect power.
So for IIT simulating recursion and integration on a von Neumann serial computer will not have the same cause-effect Φ-structure in general as a hard-wired physical implementation of that same recursion and integration.

\begin{figure}[htbp]
\centering
\includegraphics[width=1.0\linewidth]{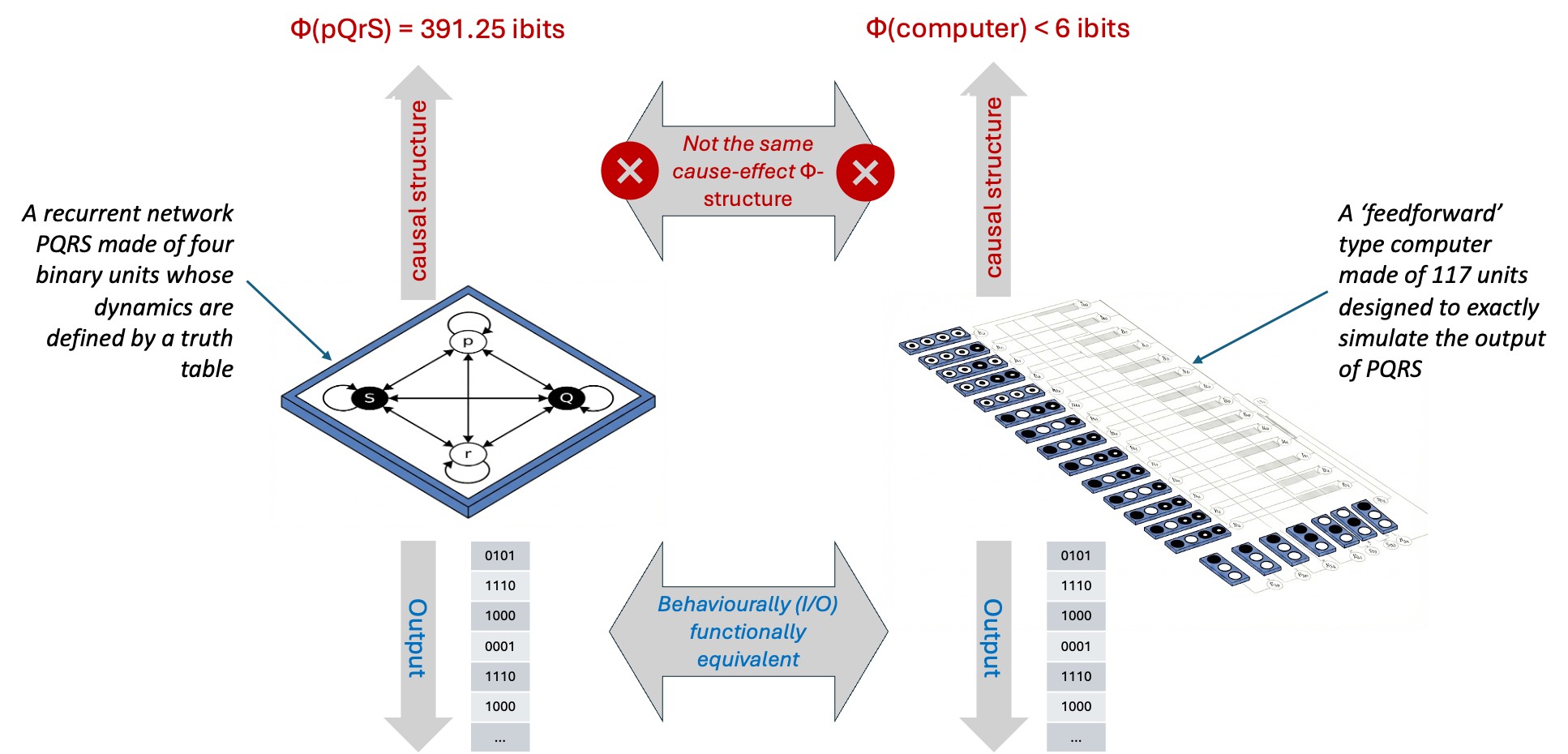}
\caption{\emph{Behaviourally equivalent computers can have different cause-effect structures and metrics of integrated information: 391.25 ibits (left) vs \textless6 ibits (right). See text for details. Adapted from \citet{findlay2024}.}}
\label{fig:image24}
\end{figure}

\Cref{fig:image24} shows an example taken from \citet{findlay2024} of a recurrent network PQRS made of four binary units whose dynamics are governed in a deterministic manner.
Given the initial state the network computes a stream of binary outputs and has a cause-effect structure whose Φ value is 391.25 ibits.
A computer, designed to exactly simulate the output of PQRS, is composed of 117 units and based on a more feedforward architecture.
So the computer is behaviourally equivalent\footnote{Behaviourally equivalent means having the same input-output function. This does not mean that they are algorithmically equivalent. Indeed PQRS has a different algorithmic structure to the 117 unit computer. So the two systems are \emph{not} computationally functionally equivalent. (Much confusion can ensue if Marr's computational level (input-output behaviour) is conflated with the computational functional description (the algorithmic structure); see \Cref{marrs-three-levels} for a clarification of terminology.)} to PQRS since it always computes the \emph{same output} from a given input, but crucially it has a cause-effect structure with a Φ value of less than 6 ibits \citetext{see also \citealp{li2025} for other studies on transformers using IIT principles}.

So for AI systems, according to IIT, the causal architecture matters for determining the consciousness of a system.
Current AI systems based on largely feedforward architectures are unlikely to have a high Φ value.
Furthermore even if there were highly recurrent architectures, because they are ultimately realised on a Von Neumann chip architecture which fetches and executes instructions in a serial way, they are also unlikely to have a high Φ value because of the intrinsic simple and `feedforward' type \emph{physical} causal structure \citep{tononi2015}.
For a high Φ value it would seem that more neuromorphic style recurrent parallel architectures may be warranted.

Whilst IIT is not a computational functionalist theory it can be understood as functionalist in the wider sense, specifically we could say it's an \emph{\textbf{intrinsic}} \emph{\textbf{causal-structure functionalism}}.
Under IIT, if a system \emph{A} has exactly the same cause-effect structure as a system \emph{B}, it will have the \emph{same conscious experience}, but system A can be realised on a different physical substrate than system B. So what determines consciousness is that the systems are functionally equivalent in their intrinsic causal structure, and this is multiply realisable in principle.
This is different from the narrower computational functionalism which would say what determines consciousness is that the systems are computationally functionally equivalent.
Systems can be computationally functionally equivalent without being intrinsic causal-structure functionally equivalent; however, systems that are intrinsic causal-structure functionally equivalent will necessarily be computationally functionally equivalent.\footnote{For related discussions pertaining to causality and functionalism in relation to AI consciousness, see \href{https://academic.oup.com/nc/article/2021/1/niab001/6232324?login=false}{Kleiner, 2021}; \href{https://academic.oup.com/nc/article/2024/1/niae037/7933504?login=false}{Kleiner \& Ludwig, 2024}; \href{https://arxiv.org/pdf/2512.12802}{Hoel, 2026}.}
This will be important when we discuss supervenience in \Cref{marrs-levels-of-analysis-supervenience-and-the-critical-level-of-description-for-consciousness,five-levels-of-functional-description-for-consciousness} below.

All the theories considered so far, whether computational functionalist or intrinsic causal-structure functionalist, locate consciousness in the brain's processing, differing only in how abstractly or concretely they characterise that processing.
The theories that follow challenge this assumption, arguing that consciousness depends on features of the wider organism or the organism-environment coupling that no brain-centric theory can fully capture.

\subsection{Organismic functionalist theories}\label{organismic-functionalist-theories}

Organismic functionalism\footnote{Organismic functionalism might be termed \emph{biological functionalism}, but we don't prefer this term as the functions in question might be capable of being realised in physical systems which wouldn't ordinarily be called biological.} is the view that consciousness requires certain organism-like functions such as self-maintenance, homeostatic or allostatic regulation, interoception, affect, etc. A strong version of this is \emph{Biopsychism} which holds that all (and only) living beings are sentient and that `feeling' is a vital activity inherent to all organisms.
An example would be the Cellular Basis of Consciousness model proposed by \citet{reber2023}, which posits that every cell has a \emph{basal sentience}, characterised by: valence, goal-directedness, learning and memory.
Cells can perceive, sense or `feel' their environment and assign a subjective value to it, and act to maintain homeostasis.
For example, a bacterium can `feel' a gradient of glucose, a good (and move toward it) or a toxic chemical, a bad (and recoil from it).

A softer version is \emph{Biological Naturalism} \citep{searle2017} which posits that consciousness is a higher-level biological feature of the brain, just as digestion is a feature or function of the stomach or photosynthesis is a feature or function of a plant, and this higher-level biological feature arises from specific, lower-level neurobiological processes down to, for example, brain chemistry and molecular interactions.
You cannot get digestion just by running a computer simulation of the stomach's chemical reactions, you need the actual acids and enzymes to break down the food.
Similarly you need the specific causal powers of the brain, the physical `meat' doing its work, not just the instantiation of the algorithm on different hardware without the same causal powers.
But Biological Naturalism can be taken as a functionalist theory\footnote{See \Cref{substrate-dependent-theories} for the more non-functionalist substrate-dependent aspects of biological naturalism.} (of the organismic rather than computational variety) because, at least in principle, these causal powers might be possible to be realised in a different system based on a different physical substrate.

A recent attempt to make the distinction between biological and digital computation more precise is the `biological computationalism' of \citet{milinkovic2026}, which identifies two features of biological computation that are absent in conventional digital systems.
First, biological systems exhibit scale-inseparable, substrate-dependent multiscale processing: processes at every scale -- from ion channels through dendritic integration to whole-brain oscillatory dynamics -- are so densely coupled that changing activity at one scale alters the entire computation.
This tight coupling across scales is driven by metabolic optimisation and is a defining feature of the brain's computational architecture, whereas digital systems are specifically designed to separate scales cleanly.
Second, biological systems perform continuous-valued computation arising from their fluidic substrate: ion flows, diffusion gradients, local field potentials, and electromagnetic interactions constitute an analogue computational layer that operates alongside discrete neural spiking, and that cannot be abstracted away without losing computationally relevant structure.
On this view, the algorithm is the substrate -- the physical organisation of the system does not merely implement the computation but constitutes it.
Milinkovic and Aru are careful to note that this does not restrict consciousness to carbon-based biology; rather, it implies that consciousness may require a biological style of computational organisation, characterised by being hybrid, scale-inseparable, and energetically grounded, which could in principle be instantiated in non-biological substrates if they replicate these computational properties.

\begin{figure}[htbp]
\centering
\includegraphics[width=0.8\linewidth]{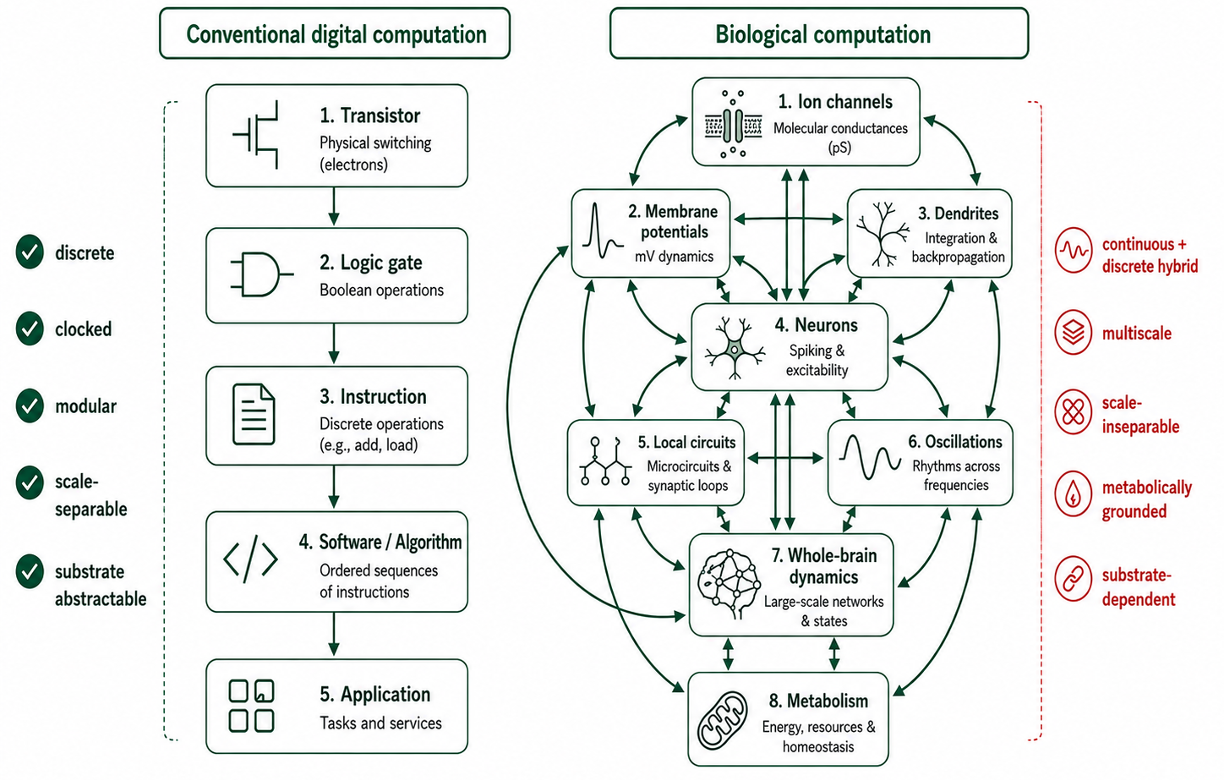}
\caption{\textbf{Conventional digital computation versus biological computation.} Digital systems form a clean hierarchy of levels, each abstracting away the one beneath, making computation discrete, clocked, modular, scale-separable, and substrate-abstractable. Biological computation lacks this separation: processes from ion channels to whole-brain dynamics and metabolism are densely coupled, yielding computation that is hybrid, multiscale, scale-inseparable, and metabolically grounded. Consciousness may require this biological style of computational organisation, whatever the substrate. Adapted from \citet{milinkovic2026}.}
\label{fig:image39}
\end{figure}

Another recent species of biological naturalism proposed by Seth is the \emph{Beast Machine} theory which proposes a predictive processing (PPT) framework, as discussed in \Cref{predictive-processing-theory}, to understand how these organismic causal functions are the basis for consciousness \citep{seth2021,seth2024,seth2026stuff}.
The brain is a prediction machine, and just as external perceptions are our inferences of the external sensory data, the brain also has to predict or infer the \emph{internal state} of the body (e.g. heart rate, glucose, hormones, the viscera, etc.) in order to maintain homeostasis (and in general allostasis) otherwise the organism will not thrive and ultimately not survive.
These inferences of internal states are `internal perceptions', or forms of \emph{interoception}, and their phenomenology is what we call \emph{feelings} (sensations, pain, pleasure, emotions, moods, etc; see \citealp{seth2013}).
These `felt' states, the primordial mental states, are control signals for the regulation of the organism.
So consciousness is intimately tied to the process of controlling metabolic and other organismic parameters to remain within a tight range appropriate for the particular circumstances (e.g. when running away from a lion) and resisting the otherwise inevitable entropic force induced by the second law of thermodynamics.

\begin{figure}[htbp]
\centering
\includegraphics[width=0.86\linewidth]{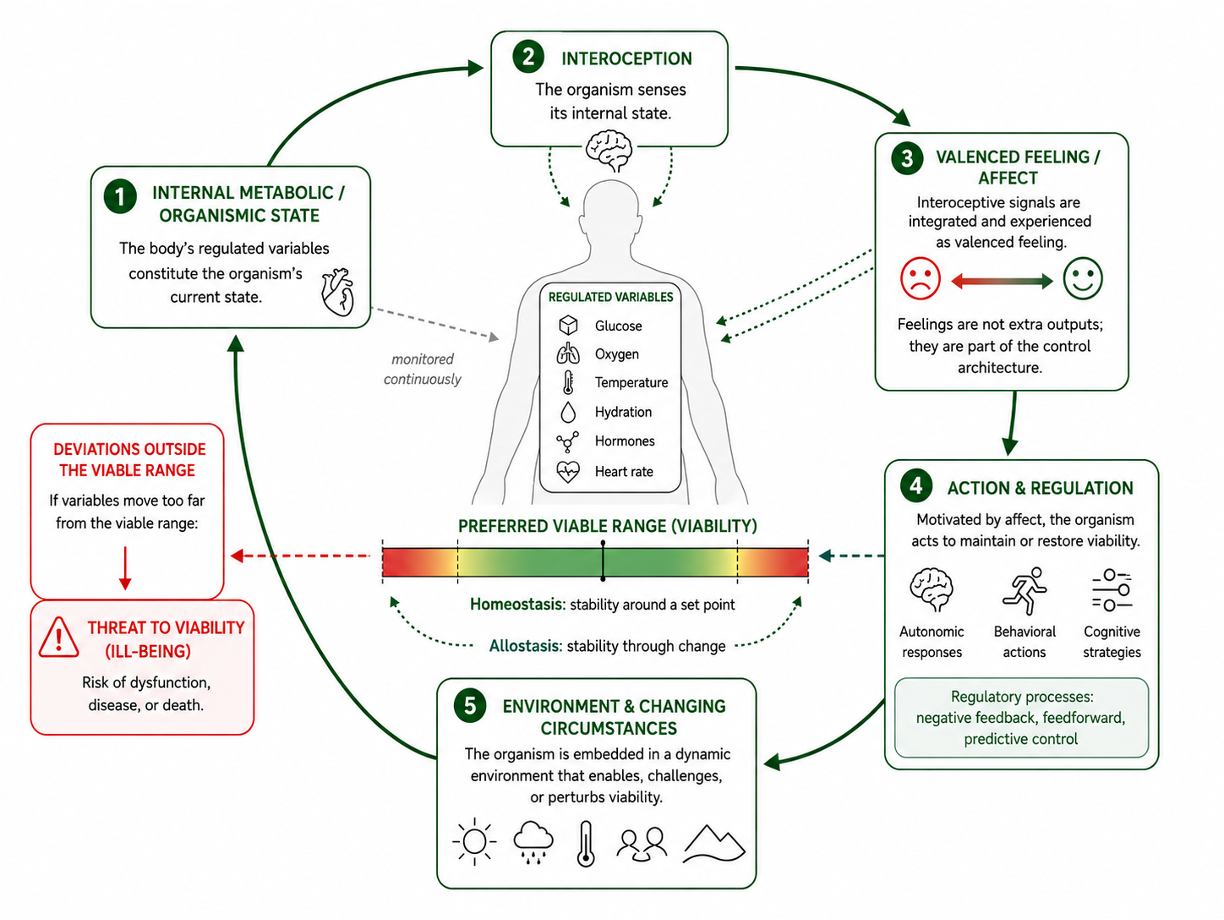}
\caption{Organismic functionalist views, such as Seth's \emph{Beast Machine} theory, conceives consciousness as tied to the organism\textquotesingle s ongoing self-maintenance control loop, involving interoception, affect, and regulation of viability.}
\label{fig:image13}
\end{figure}

Several related approaches converge on this core insight -- that consciousness is grounded in the organism's existential relationship to its own viability -- while differing in how they develop it.
\citet{fuchs2026} argues that what he calls the `feeling of being alive' -- comprising vitality (the background feeling of well-being or ill-being) and conation (drive, need, affect) -- cannot be understood as an internal self-model computed by neural machinery, but is a manifestation of the self-organisation and life process of the organism as a whole.
On this view, the sufficient basis of self-awareness lies not in localised neural correlates of consciousness but in the homeostatic and allostatic self-regulatory processes that sustain the living organism in its relationship to the world.
This extends Seth's emphasis on interoceptive inference beyond the brain and into the dynamics of the whole organism: it is not merely that the brain predicts the body's state, but that the organism's ongoing self-regulation is itself the ground of experiential selfhood.
\citet{bennett2026} take this organismic grounding in a more formal direction, providing a mathematical framework describing how naturally selected, embodied organisms self-organise to hierarchically interpret sensory information according to valence -- the qualitative distinction between states that are good or bad for the organism's survival.
Their central claim is that phenomenal quality arises because selection pressures favour systems that actively intervene in the world to achieve homeostatic and reproductive goals, and that subjective experience is the property that links cause to \emph{affect} to motivate these interventions.
Or, in their provocative formulation: death grounds meaning.
Without genuine existential stakes -- without the possibility of ceasing to exist -- information processing has no intrinsic qualitative character.
This convergence across Seth's interoceptive inference, Fuchs's phenomenology of organismic self-regulation, and Bennett et al.'s formal account of valence-grounded natural selection reinforces the organismic functionalist claim: consciousness is not an abstract feature of information processing but is tied, at its roots, to the precarious, self-maintaining activity of a living system.

The \emph{felt} states of an organism indicate its sensed metabolic state and the valenced affective tone indicates the appropriateness of the state for the organism in the circumstances.
Biologist Nick Lane argues that the roots of `feeling' or sentience are not found in complex neuronal firing, which comes much later in the phylogenetic tree of life, but in the fundamental way single cells keep themselves alive using electrical gradients across membranes \citep{lane2022}.
Cells and mitochondria pump charged particles (e.g. protons) across a membrane to create a difference in charge; this is the proton gradient.
Because proteins (channels, pumps) are embedded in this membrane across which there is a significant potential difference (150 millivolts across 5 nanometers is equivalent to a bolt of lightning at human scale), their shape and function are modified by the electromagnetic fields.
If the metabolic state drops (e.g. due to a lack of nutrients or oxygen), the proton gradient changes, and the resultant electromagnetic fields influence the shape of thousands of proteins simultaneously and may also change the structure of the cytoplasmic water and thus influence the reactivity of biochemical processes.
The cell `senses' its depletion by physically changing state in real-time.
But could similar causal `sensing' mechanisms be instantiated in AI systems?
That might require radically different architectures.

The central challenge for organismic functionalism is whether these features are truly necessary for consciousness or whether they describe the contingent path that consciousness happened to take in biological evolution.
A computational functionalist would argue that a system with the right information-processing organisation could be conscious without any self-maintenance, homeostatic regulation, or metabolic grounding -- that these are features of our consciousness, not of consciousness as such.

\subsection{Organism--environment (4E) functionalist theories}\label{organismenvironment-4e-functionalist-theories}

Extending the scope of what consciousness depends upon to the whole organism and not just the brain is not far enough according to `4E cognition' theories which hold that the mind is embodied, embedded, enactive and extended (4E).
The key theorists here include Alva Noë, Francisco Varela, Evan Thompson, and (for the ecological dimension) J.J.
Gibson, whose work on affordances has deeply influenced the tradition \citetext{\citealp{noe2004}; \citetalias{varela1991}; \citealp{gibson1979}}.
These views are functionalist in a world-involving way: the relevant `functional organisation' is not just internal computation, or internal organismic functions but structured patterns of action-perception coupling.
The character of conscious experiences, the subjective phenomenology, is tied to sensorimotor contingencies (how actions change sensory input), affordance-responsiveness (what the environment `offers' this body or agent), and agentive control loops (the organism's active regulation of its coupling).
These functions are constituted by interaction with an environment; and not simply internal neural representations, nor just body-brain coupling although these will clearly be a part of the wider coupling.
So experience is tied to how an agent is poised to act in a world, not just to what happens internally.

To see what this means concretely, consider how the 4E theorist would account for visual perception.
On a standard computational or representational view, seeing a tomato consists in constructing an internal representation of its colour, shape, and spatial location.
On the sensorimotor contingency view developed by \citet{noe2004}, seeing a tomato is constituted by the perceiver's practical mastery of the way the tomato's appearance changes under interaction: how its colour shifts as lighting conditions change, how its visible profile transforms as the perceiver moves around it, how reaching toward it would bring it into tactile range and reveal its weight and texture.
The `redness' of the tomato is not a property of an internal neural state but a property of this entire sensorimotor loop -- it is the structure of the way this surface modulates sensory input across the full range of the perceiver's possible actions.
A being with different sensory modalities or different action possibilities would have a different sensorimotor contingency structure with respect to the same tomato, and would therefore, on this view, have a qualitatively different perceptual experience of it.
Compare a bat's experience of the same tomato: its sensorimotor contingencies are structured by echolocation and flight rather than vision and reach, so the tomato's echo signature shifts with wingbeat, body orientation, and pulse rate in ways that have no analogue in human perception.
On the sensorimotor view, this is why \emph{what it is like to be a bat} \citep{nagel1974} is inaccessible to us: not just because the bat's sensory hardware differs, but because its world-involving loop is constituted differently.
The conscious phenomenal qualities of experience are tied to this world-involving loop: more broadly, the `look' of glossiness, hardness, depth are tied to the agent's interaction (i.e. when we push we may slide, or be resisted, or go a long way), or when we perceive an object we perceive an affordance (e.g. a chair is perceived, in part, as sit-able).

\begin{figure}[htbp]
\centering
\includegraphics[width=1.0\linewidth]{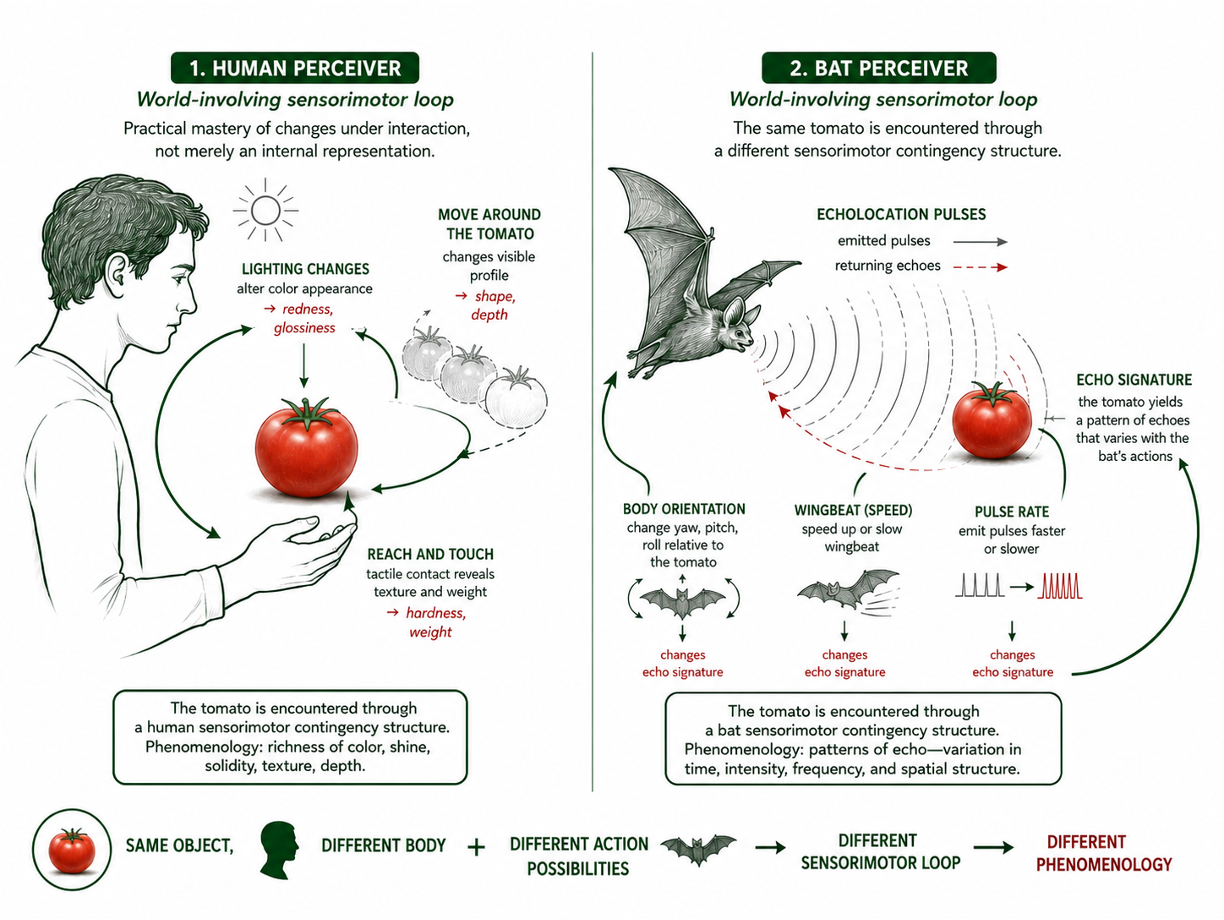}
\caption{The conscious phenomenal qualities of experience are tied to a world-involving loop. A system with different sensory modalities or different action possibilities would have a different sensorimotor contingency structure, and would therefore, on the 4E view, have a qualitatively different perceptual experience.}
\label{fig:image12}
\end{figure}

An organism-environment functionalist position would suggest that for AI systems to be conscious they would at least require some sort of embodiment.
This would give a system the requisite structured sensorimotor loops, a stable viewpoint, constraints (e.g. energy, time, noise, inertia) that shape the dynamics of experience, and potential for affordances which in part constitute the character of perception.
AI systems may need an enactive autonomy \citep{barandiaran2017}, being a self-maintaining autonomous system to give a phenomenological grounding of intrinsic norms like viability (`good' or `bad' for me).
The nature of this embodiment -- e.g. android robot, bespoke industrial robot, autonomous vehicle, autonomous drone, nano-robot swarms or biohybrid systems -- would likely influence the character of any emergent phenomenology.
This last point is important: unlike the theories considered so far, the 4E view predicts not just whether a system is conscious but what its consciousness would be like, because the specific character of the embodiment and environmental coupling would shape the specific character of the experience.

\subsection{Substrate-dependent theories}\label{substrate-dependent-theories}

There are a number of views that posit that consciousness critically depends upon certain low-level features of the physical substrate.
\emph{Carbon Chauvinism}, which is more of a stance rather than a theory, hypothesises that carbon is the \emph{only} element in the periodic table capable of supporting the complexity required for life and consciousness due to its unique thermodynamic and chemical properties.
But even if carbon based biology is the only \emph{known} route to consciousness, that doesn't entail it's the only \emph{possible} route, unless we find a specific property that is required and show it can't be realised in any other way.

\emph{Quantum physics and consciousness}, the two grand `mysteries', have been connected by many including some of the pioneers of quantum physics (e.g. Max Planck, Erwin Schrödinger, Eugene Wigner and Wolfgang Pauli).
More recently Penrose and Hameroff have proposed the \emph{Orchestrated Objective Reduction theory} of consciousness \citep{hameroff2014} which says that conscious moments arise from quantum computations in neuronal microtubules that evolve in superposition and then undergo a real, gravity-related wavefunction collapse (objective reduction); brain activity `orchestrates' these microtubule quantum states, and each collapse corresponds to a discrete moment of conscious experience.
Hartmut Neven, a quantum computation scientist at Google Quantum AI Lab, has proposed the \emph{Quantum Formation hypothesis} \citep{neven2024}, that is an inversion of the Penrose-Hameroff theory: conscious experience arises whenever a quantum superposition \emph{forms} rather than when it collapses.
Under these quantum based views, current AI systems that run on classical bits couldn't result in consciousness, but these views don't rule out, in principle, creating conscious AI systems that run on qubits like quantum computers.

\emph{Electromagnetic (EM) field theories} of consciousness posit that consciousness arises when a brain produces electromagnetic fields with particular characteristics.
When neurons fire, the action potentials and postsynaptic potentials involve ionic currents which generate electric and magnetic fields.
EM field theorist McFadden proposes that these fields effectively represent the information conveyed by the underlying neurons \citep{mcfadden2020}.
When many neurons fire synchronously, their local fields superpose into a larger-scale EM pattern, and consciousness is proposed to be integrated information encoded in the brain's global EM field.
Importantly, the field is not just epiphenomenal -- like smoke billowing from a steam train without affecting the train itself -- it can have \emph{downward causality} by influencing the propensity of the neurons to fire via ephaptic effects on membrane potentials.
The EM field then acts like a `global workspace', a unified whole of integrated information capable of broadcasting back down to local processing units.
If EM field theory is correct, then a typical digital computer, specifically engineered to suppress stray EM cross-talk, would be the wrong kind of physical system for consciousness.
For AI systems to be conscious, it might require hardware designed to generate and exploit richly structured, integrated EM fields.

\emph{Biological naturalism}, as formulated by Searle \citep{searle2017}, also has a substrate-dependent dimension.
While the organismic functional aspects of biological naturalism were discussed in \Cref{organismic-functionalist-theories} -- the claim that consciousness requires organism-like functions such as self-maintenance, homeostatic regulation, and interoception -- Searle also contends that consciousness may arise from specific lower-level neurobiological processes, including brain chemistry and molecular interactions.
Just as you cannot get digestion by running a computer simulation of the stomach's chemical reactions -- you need the actual acids and enzymes to break down the food -- Searle argues \citep{searle1992} that you need the specific causal powers of biological neural tissue, not merely the instantiation of the same algorithm on different hardware.
This substrate-dependent aspect of biological naturalism goes beyond the claim that organismic functions are necessary; it is the further claim that even those functions must be realised by biological mechanisms with the right kind of low-level causal powers \citep{searle1980}. 
\citet{seth2024,seth2026stuff} also develops a version of this stronger kind of biological naturalism, in contrast to a weaker form where the biological substrate is needed only to realise the relevant functional organisation.

\emph{Meat hypothesis.} More recently, Ned Block \citep{block2025} has argued that current discussions of AI consciousness do not take sufficiently seriously the possibility that implementing certain computations is not sufficient for consciousness because there could also be a subcomputational biologically necessary condition.
Block frames this as a distinction between \emph{functional roles} and their \emph{realisers}.
A functional role is a second-order property -- a network of causal relations among inputs, outputs, and internal states -- that can in principle be realised by many different first-order physical mechanisms.
Block's \emph{`meat hypothesis'} is the claim that consciousness depends not only on the second-order functional roles but also on the first-order biological mechanisms that realise those roles in us: the specific subcomputational properties of \emph{biological neural tissue}, or `meat'.
We know that in humans, certain functional roles and certain biological realisers are jointly associated with consciousness, but we do not know whether consciousness depends on the roles, on the realisers, or on both.
If we focus on the functional roles, we are drawn toward attributing consciousness to computationally sophisticated AI systems; if we focus on the realisers, we are drawn toward attributing consciousness to biologically simple animals that share our underlying neural mechanisms.

\emph{Electrochemical theory of consciousness.} Block draws attention to a particularly suggestive piece of evolutionary evidence.
Recent phylogenetic research supports the hypothesis that comb jellies (ctenophores) were the first animals \citep{schultz2023}, and that at least one stage of their life cycle features an entirely electrical nervous system, with chemical synapses present only at points of interface with the environment \citep{burkhardt2023}.
By contrast, the more complex animals that followed -- the myriazoa, the clade containing sponges, cnidarians, and ultimately all bilaterians including humans -- developed \emph{electrochemical} \emph{nervous systems} in which electrical signals propagate within neurons but communication between neurons occurs via chemical neurotransmitters released into the extracellular fluid.
Purely electrical nervous systems, as found in comb jellies, have not led to animals that are strong candidates for consciousness, whereas electrochemical nervous systems have.
This suggests that something about the electrochemical mechanism -- and not just the computational role it implements -- may be necessary for consciousness, or at least for the kinds of complex cognition from which consciousness emerges.
What might be special about electrochemical processing?
One suggestion concerns the large-scale dynamic patterns that electrochemical processing supports.
In complex electrochemical nervous systems, waves of ion fluctuations in the chemical soup surrounding neurons are important to driving brain rhythms and coordinating large-scale neural activity.
These large-scale dynamic patterns, as Godfrey-Smith has argued \citep{godfreysmith2024}, may be important to the unity of consciousness.
They are emergent properties of the electrochemical realiser that have no analogue in a purely electrical or purely digital implementation.

Block's meat hypothesis is about realising mechanisms, not just material composition.
The specific ions used in extant organisms, for example sodium, potassium, and calcium, are not claimed to be necessary; other ions could in principle serve the same function.
What matters is the electrochemical mechanism itself: the release of chemical transmitters into the synaptic cleft, their uptake by receptors, the opening of ion channels, and the resultant modulation of neural activity through a chemical medium that supports diffuse, graded, and temporally extended signalling alongside the discrete electrical pulses of action potentials.
Whether this mechanism could be replicated in a non-biological substrate remains an open question, but Block's argument is that we should take seriously the possibility that it may not be replaceable by a computationally equivalent but mechanistically different process without losing consciousness.

A similar substrate dependency claim can be made in Nick Lane's account we discussed in \Cref{organismic-functionalist-theories} which argues that the roots of sentience are not found in complex neuronal firing but in the basic way that all living cells maintain themselves through electrical gradients across membranes \citep{lane2022}.
If Lane is right that this kind of direct, physics-based state-sensing is the evolutionary precursor to feeling, then it represents a radical substrate-dependent claim: that the foundations of consciousness lie not in computation, not in any particular algorithm or causal architecture, but in the specific electrochemical and electromagnetic physics by which living matter maintains and monitors itself.

\subsection*{A confusing cacophony of answers}\label{a-confusing-cacophony-of-answers}
\addcontentsline{toc}{subsection}{A confusing cacophony of answers}

Confused?
You should be!
These disparate and often contradictory accounts are our best answers from some of our best minds and come with an assortment of `empirical support' and philosophical arguments, often hotly contested.
These various views, accounts and theories often talk past each other, they can target different \emph{explananda} (that which is to be explained by the account), and they can have philosophical commitments that may not be strictly necessary for their usefulness in the project of thinking about the criteria for the attribution of consciousness to AI systems.
To begin to sort through this cacophony, our first step must be to make a key distinction that separates the most profound metaphysical questions from the scientific ones that give traction for evaluating AI consciousness.

\section{Setting aside the hard problem: from metaphysics to the mapping problem}\label{setting-aside-the-hard-problem-from-metaphysics-to-the-mapping-problem}

To make tangible progress on the question of whether an AI system might be conscious, we need to refine the question by considering our antecedent theoretical commitments, as these will determine the implications of our conclusions.\footnote{We would like to acknowledge David Chalmers, who provided very helpful comments on an early draft of this report and led to a reformulation of this section.} If, for example, we hold objective idealism to be true, then all entities are constitutively conscious, in which case computers are, \emph{a fortiori}, conscious, together with rocks, quarks, and beams of light.
However, it's not clear that this would have notable moral or even practical implications; even if the ultimate substance of a rock is conscious or mind-like, it is unlikely to be having a rich conscious inner life, because it lacks any relevant dynamics, behaviours, information processing, functional organisation, causal structures and/or transformations of the substrate that would be associated with having rich experiences.

Consequently, when we ask `might AI systems be conscious?', our specific target is whether AI systems could have a rich conscious inner life, in the sense of being capable of having rich and varied experiences.
(By `rich' here and throughout the report we mean contentful -- that there is something it is like to be the system, with qualitative distinctions in experience, however simple.
A system capable of even a single valenced distinction between things going well and things going badly would count as having rich experience in this sense.) In other words, we could ask whether AI systems can have any contents of consciousness at all -- contents that have some properties, that are distinctions, that \emph{bear information}, however minimal.
This refined question of `whether AI systems have information-bearing contents of consciousness?' would also be what we are really seeking to target even if we hold other ontological positions like physicalism, panpsychism or neutral monism.

In this way of thinking we are not distinguishing two kinds of consciousness but two questions we might be asking when we use the word `consciousness'.
There is a single phenomenon -- phenomenal consciousness, there being something it is like to be the system -- and two quite different questions we can ask about it.
The first is the constitutive or metaphysical question: what phenomenal consciousness fundamentally is, and why there is any experience at all.
This is the subject of David Chalmers' \emph{hard problem of consciousness} \citep{chalmers1995}.
The second is the structural or correlational question: which mechanisms, dynamics, functional organisation, causal structures and transformations of the substrate are associated with which particular, contentful experiences.
Following \citet{bourget2019} we call this the mapping problem; it is closely related to what Seth calls the ``real problem'' of consciousness, which is to explain the specific phenomenological properties of conscious experience by mapping them onto biological or other mechanisms \citep{seth2016,seth2021}.
This hard-problem/mapping-problem contrast, which is a contrast between two explanatory projects, should be kept distinct from a second contrast internal to phenomenal consciousness itself: that between consciousness-as-such (the bare fact that there is any experience at all) and the contents of consciousness (the specific character of particular experiences).
The two distinctions are easily run together but are not the same, and the framework developed here turns only on the first.\footnote{An earlier terminology in this vicinity is due to Graziano and colleagues (2020), who distinguish \emph{i-consciousness} (the `i' for the informational content of the brain) from \emph{m-consciousness} (the `m' for a `more mysterious, extra, experiential essence that people claim accompanies the informational content'). We do not adopt this terminology. Their distinction is illusionist in spirit -- m-consciousness is the putatively illusory extra -- and it cuts differently from the hard-problem/mapping-problem contrast drawn here.}

\subsection{Why the mapping question is metaphysics-neutral}\label{why-the-mapping-question-is-metaphysics-neutral}

Some of the positions surveyed in \Cref{what-does-consciousness-depend-upon-a-cacophony-of-answers} are primarily addressed to the hard problem -- what consciousness fundamentally is (the philosophical positions such as physicalism, dualism, idealism, and panpsychism).
Others, particularly the scientific theories such as GWT, AST, and RPT, are already largely addressed to the mapping problem -- they seek to identify what functional organisation is associated with conscious experience without necessarily settling the metaphysical question.
What \Cref{setting-aside-the-hard-problem-from-metaphysics-to-the-mapping-problem} proposes is that the tractable question they all share, despite their different starting points, is the mapping question: what kind of organisation is associated with conscious experience?
This question is tractable because it can (at least in principle) be investigated empirically and compared across theories, even when those theories disagree about the deeper metaphysical nature of consciousness itself.

An advantage of this way of carving things up is that it allows us to set aside most of the metaphysical disagreements surveyed in \Cref{what-does-consciousness-depend-upon-a-cacophony-of-answers}, because these disagreements belong to the hard problem -- the ultimate nature of why anything feels like something at all -- while the question of AI consciousness that concerns us is primarily the mapping question: what mechanisms, dynamics, behaviours, information processing, functional organisation, causal structures and/or transformations of the substrate are associated with having particular, rich, contentful experiences.

Consider how this works across the leading ontologies.
Under physicalism, phenomenal consciousness supervenes on physical processes; the question of which physical or functional properties are associated with contentful experiences is well-posed and investigable.
Under objective idealism, reality is fundamentally mental but has objective lawlike structure; the question of what organisational features of this reality correspond to rich contentful experience (as opposed to the inert `experience' of a rock) is equally well-posed -- it is the same scientific question with a different ontological interpretation of what the underlying substrate ultimately is.
Under panpsychism, micro-experiences are fundamental and ubiquitous, but rich macro-consciousness -- the kind of unified, contentful experience we are asking about -- is not: it depends on micro-experiences being `appropriately organised' into a coherent whole (the combination problem discussed in \Cref{panpsychism}).
Under a panpsychist framework, then, the question of what `appropriately organised' entails is precisely the mapping question.
Under neutral monism, one neutral substance has both mental and physical aspects; the question of what structural and functional organisation of this substance corresponds to rich experience is again well-posed.
Under property dualism, mental properties supervene on or emerge from physical properties via psychophysical laws; the question of which physical configurations give rise to which experiences is the mapping question addressed through those laws.
And under illusionism, there is no further fact of phenomenal consciousness over and above the functional facts; the mapping question becomes: what mechanisms produce the impression of having rich contentful experiences? -- which is the only question there is.

In every one of these cases, the scientific project is the same: identify the dynamics, functional organisation, causal structures, and substrate transformations that are associated with (or constitutive of, or give rise to, depending on one's ontology) rich contentful experience.
These ontologies disagree about \emph{why} these associations hold -- whether they hold because consciousness is identical to certain physical processes, or because it emerges from them, or because physical processes are themselves aspects of a deeper mental or neutral reality -- but they agree that the associations exist and can in principle be discovered.
What they share is therefore not a single ontology but a discoverable relation -- a psychophysical mapping -- between a system's organisation and its experience.
That relation is not the preserve of monism: property dualism supplies it through psychophysical laws, and so does an interactionist substance dualism whose laws connect mind and body.
It is the mapping, not the number of substances, that makes the question tractable, and it can be investigated on shared terms regardless of which of these metaphysics turns out to be correct (we make this criterion precise in \Cref{the-one-assumption-a-discoverable-psychophysical-mapping}).
It should therefore be possible to make progress on the issue of whether specific AI systems might have rich contentful experience without having to first solve the hard problem of consciousness.

\subsection{The one assumption: a discoverable psychophysical mapping}\label{the-one-assumption-a-discoverable-psychophysical-mapping}

The framework developed in the rest of this report makes ascriptions of consciousness based on a system's examinable properties: its behaviour, computational organisation, causal structure, and physical and environmental make-up.
That approach presupposes one thing: there must be a discoverable, lawlike mapping from those properties to the facts about the system's experience -- call it a psychophysical mapping -- systematic enough that fixing the properties fixes, or reliably predicts, the experience, and accessible enough that we can come to know it by investigating the system.
Where such a mapping holds, whether a system is conscious is a question about its properties, and can be pursued empirically.
Where it fails, the question is not scientifically tractable: if the facts about a system's experience floated free of all its physical, functional, computational, causal, organismic, and environmental properties, or were tied to them by a relation we could never discover, then a physically identical system could be conscious or not with no observable or structural difference, and no examination of the system could tell which.

This requirement, rather than any particular metaphysics, is what fixes the scope of the framework.
It is met by every view on which a system's organisation discoverably determines its experience: physicalism, objective idealism, panpsychism and neutral monism among the substance monisms, but equally a property dualism or an interactionist substance dualism whose psychophysical laws connect organisation to experience.
It is not met by views on which the mapping is absent or beyond reach -- among them law-free dualism, on which mind and body are not lawfully connected at all; epiphenomenalism, on which experiences have no effects and so leave no evidence by which the mapping could be established; mysterianism, on which the link is real but cognitively closed to us; and primitivism about subjecthood, on which whether a configuration is a subject is a brute fact not fixed by its organisation.
The line between these two groups cuts across the familiar divisions rather than running along them: it separates neither monism from dualism nor the physical from the non-physical, but views that supply a discoverable mapping from views that do not.
A monist position can fall outside it, and a dualist position can fall within it.

Setting the second group aside is a methodological move, not a metaphysical verdict.
We do not claim that any of these views is false; several are seriously held.
We claim only that if one of them is true, the project of attributing consciousness to a system from its examinable properties is intractable, because whether the system is conscious would turn on a factor those properties do not reveal -- and there would then be no principled basis for assessing consciousness in any system whose make-up differs from our own.
We therefore adopt the existence of a discoverable mapping as a condition of the enquiry, not as a thesis about the ultimate nature of mind.

The assumption is not a demanding one.
The positions it excludes face well-known independent difficulties -- the interaction problem in the case of law-free dualism (see \Cref{dualism}), the causal-relevance, or exclusion, problem in the case of epiphenomenalism -- and the overwhelming majority of contemporary work in consciousness science, neuroscience, and philosophy of mind already proceeds as though experience is systematically and discoverably related to the organisation of the system.
What making the assumption explicit buys us is a clear statement of scope: the framework applies to any theory of consciousness that locates consciousness, or its contents, in a discoverable relation to a system's examinable properties, and is silent on any view that places experience beyond such a relation altogether.

\subsection{The refined question and the path forward}\label{the-refined-question-and-the-path-forward}

With that assumption in place, our refined question becomes: what rich contentful experience, if any, could AI systems have?
Whether we hold a physicalist, idealist, panpsychist, neutral monist, property dualist, or illusionist metaphysics -- or indeed an interactionist substance dualism with psychophysical laws -- we can ask this question on the shared terms of the mapping problem.
And what are those terms?
What are the mechanisms, dynamics, behaviours, information processing, functional organisation, causal structures and/or transformations of the substrate that are associated with having particular, rich, contentful experiences?

Different scientific theories of consciousness give different answers to this question -- they identify different mechanisms and different levels of organisation as critical.
This is what generates the cacophony surveyed in \Cref{what-does-consciousness-depend-upon-a-cacophony-of-answers}.
But with the metaphysical question set aside, the cacophony is now tractable: it is a disagreement among scientific theories about which properties of the system are associated with rich contentful experience, and this is a disagreement that can be structured, taxonomised, and ultimately adjudicated by evidence.
Granting the mapping just assumed -- that a system's experience is fixed by, and discoverable from, its organisation -- the question we pursue is this: at which grain of description does the supervenience base for rich contentful experience sit?
The remainder of this report is devoted to developing the framework within which this structuring and adjudication can proceed.

\begin{figure}[htbp]
\centering
\includegraphics[width=1.0\linewidth]{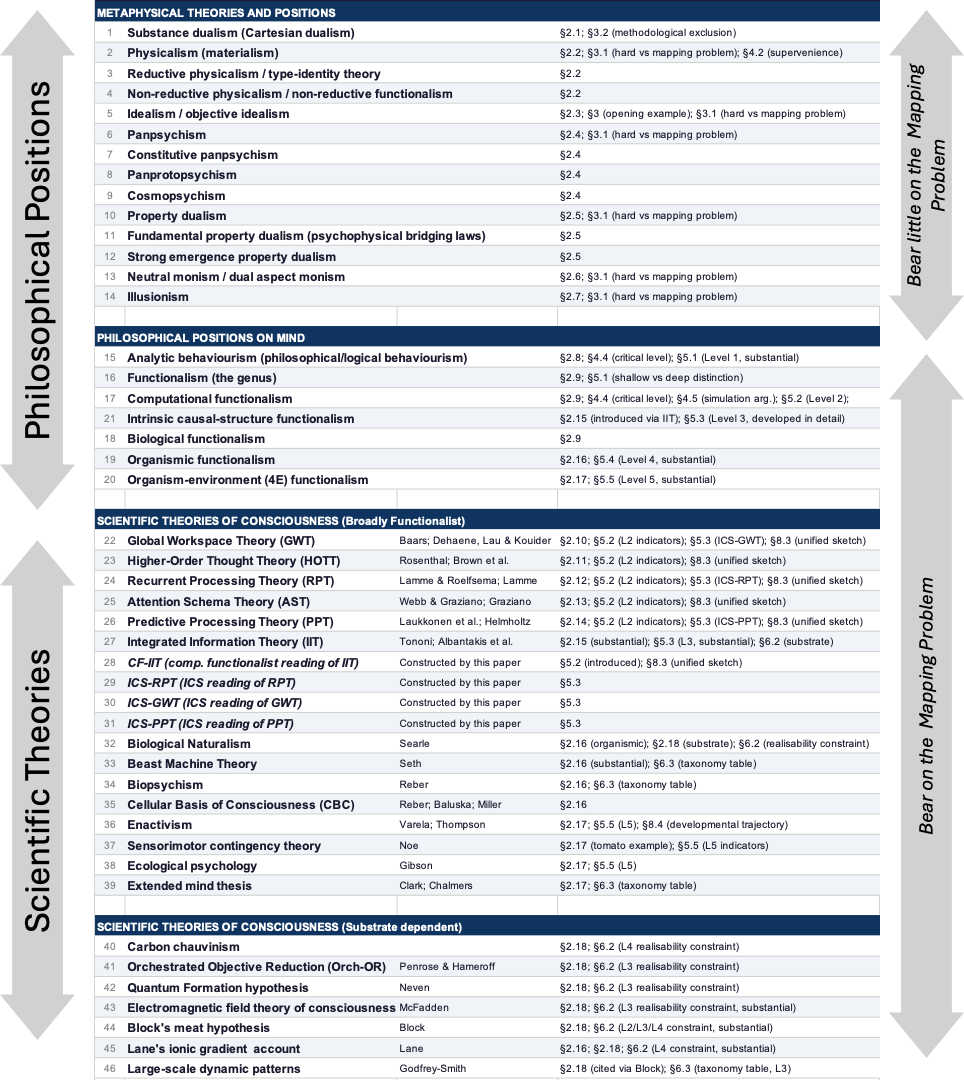}
\caption{\emph{Characterising scientific and philosophical theories of consciousness in terms of how far they bear on the mapping problem of consciousness.}}
\label{fig:image46}
\end{figure}

\section{Marr's levels of analysis, supervenience, and the critical level of description for consciousness}\label{marrs-levels-of-analysis-supervenience-and-the-critical-level-of-description-for-consciousness}

\Cref{setting-aside-the-hard-problem-from-metaphysics-to-the-mapping-problem} concluded with a refined question: what are the mechanisms, dynamics, behaviours, information processing, functional organisation, causal structures and/or transformations of the substrate that are associated with having particular, rich, contentful experiences?
To answer this question systematically, we need a framework for describing information processing systems at multiple complementary levels, because as the cacophony of \Cref{what-does-consciousness-depend-upon-a-cacophony-of-answers} illustrates, different theories of consciousness locate the relevant mechanisms at different levels of description.
Some theories care about abstract computational organisation; others care about the physical causal structure of the hardware; others care about organismic biology or environmental coupling.
A framework that can accommodate all of these perspectives requires a principled way of describing the same system at multiple grains of resolution.
David Marr's classic framework provides the starting point for such a multi-level analysis \citep{marr1982}.

\subsection{Marr's three levels}\label{marrs-three-levels}

Any information processing system (be it a brain or a computer) can be analysed at multiple complementary levels.
Marr proposed three useful levels of analysis.
The first level, which Marr (rather confusingly) called the computational level, is `what the system is doing', that is, the goal of the system, for example `sort a list of names'.
Next is the algorithmic level, that is, `how is it doing this'; in short, a specification of the computational recipe for transforming inputs into outputs, for example using a bubble sort, merge sort, or quick sort algorithm.
Importantly, we see that the computational level goal can be realised by multiple algorithms.
The next level in Marr's schema is the implementation level, which describes the actual physical realisation of the algorithms, for example Python code running on a MacBook, C++ on a PC, or a human sorting index cards by hand.
Again, the algorithmic level can be realised by multiple implementations.

To avoid terminological confusion in what follows, we should note that Marr's naming creates an unfortunate collision with our later use of `computational functionalism'.
In \Cref{five-levels-of-functional-description-for-consciousness} we will construct five levels of description for consciousness.
Our Level 1 (the behavioural level) corresponds to Marr's computational level -- the input-output mapping, the goal of the system.
Our Level 2 (the computational functional level) corresponds to Marr's algorithmic level -- the information processing that transforms inputs into outputs.
And Marr's single implementation level will be expanded into our Levels 3, 4, and 5, which distinguish between intrinsic causal structure, organismic functions, and organism-environment coupling -- three quite different aspects of `implementation' that Marr's framework lumps together.
Table 1 summarises this mapping.

{\def\LTcaptype{none}\footnotesize\sffamily 
\begin{longtable}[]{@{}
  >{\raggedright\arraybackslash}p{(\linewidth - 4\tabcolsep) * \real{0.3333}}
  >{\raggedright\arraybackslash}p{(\linewidth - 4\tabcolsep) * \real{0.3333}}
  >{\raggedright\arraybackslash}p{(\linewidth - 4\tabcolsep) * \real{0.3333}}@{}}
\toprule\noalign{}
\begin{minipage}[b]{\linewidth}\raggedright
\textbf{MARR\textquotesingle S LEVEL}
\end{minipage} & \begin{minipage}[b]{\linewidth}\raggedright
\textbf{OUR LEVEL}
\end{minipage} & \begin{minipage}[b]{\linewidth}\raggedright
\textbf{DESCRIPTION}
\end{minipage} \\
\midrule\noalign{}
\endhead
\bottomrule\noalign{}
\endlastfoot
\emph{\textbf{Computational}} & \emph{\textbf{Level 1: Behavioural}} &
\emph{Input-output function; what the system does} \\
\emph{\textbf{Algorithmic}} & \emph{\textbf{Level 2: Computational
functional}} & \emph{Algorithms and information processing; how the
system does it} \\
\emph{\textbf{Implementational}} & \emph{\textbf{Levels 3, 4, 5}} &
\emph{Intrinsic causal structure, organismic functions,
organism-environment coupling} \\
\end{longtable}
}

\textbf{Table 1. Mapping Marr's Levels to the Five-Level Framework.} \emph{This mapping resolves potential terminological overlaps by aligning Marr's computational and algorithmic levels with our behavioural and computational functional levels respectively, while expanding his single implementational level into three distinct levels.}

\subsection{Supervenience and multiple realisability}\label{supervenience-and-multiple-realisability}

The relationship between the different levels of Marr's framework can be described in terms of supervenience.
A set of properties A supervenes on a set of properties B if there can be no change in A without a change in B (this means that fixing B fixes A).
In a broadly physicalist framework, mental properties are said to supervene on physical properties, that is, there can be no change in a mental state without there being a change in the physical state.
Critically, supervenience is asymmetric: a change in the physical state does not necessarily change the mental state, so in principle, the higher-level mental property is multiply realisable.\footnote{We use ``multiple realisability'' here in a modest sense, not the substantive thesis, associated with \citet{putnam1967} and \citet{fodor1974}, that mental states can be realised across heterogeneous physical substrates and therefore cannot be identified with any particular physical kind. We intend only the relation between two descriptions of a single system, one coarser and one finer, whenever many distinct finer realisers correspond to the same coarser description. In this usage, the term carries no commitment as to whether consciousness, or any mental state, is in fact multiply realisable across substrates; whether it is, and at which grain, is the substantive question the hierarchy is designed to pose, and the theories we survey answer differently. It is in this respect weaker than substrate independence, the stronger thesis that functional organisation alone fixes mentality, which we take up through the substrate-dependent realisability constraints of \Cref{the-structure-of-the-hierarchy}.}

The supervenience relation maps onto Marr's hierarchy naturally.
Since fixing all the implementational details fixes which algorithm is being run, the algorithmic level supervenes on the implementational level.
But the same algorithm can be realised by different implementations, so there can be a change in implementation (from a MacBook to a human and index cards) without a change in algorithm.
Similarly, since fixing the algorithm fixes which computation is being performed, the computational level supervenes on the algorithmic level.
But the same computation (e.g. sorting a list) can be realised by different algorithms (bubble sort, merge sort etc.).
The direction of supervenience is therefore from coarser to finer: each coarser description supervenes on each finer one.
And the chain is transitive, so that the computational level supervenes on the implementational level.

At each level we can distinguish between the functional role specified at that level -- the abstract pattern that the level describes -- and the physical realiser that implements it.
Algorithms are the realisers of computations; physical implementations are the realisers of algorithms.
This role/realiser distinction, which will become important when we consider substrate-dependent theories in \Cref{substrate-dependent-theories-as-realisability-constraints-distributed-across-multiple-levels}, is already implicit in the structure of Marr's hierarchy.

\subsection{Coarse-graining: a precise formulation of levels}\label{coarse-graining-a-precise-formulation-of-levels}

Coarse-graining is the operation of redescribing a system at a lower resolution by grouping together microstates of the system that are equivalent with respect to some macroscopic property of interest.
In statistical mechanics, for example, one coarse-grains by grouping together all the microstates of gas molecules that produce the same temperature and pressure, a macroscopic description.
The macroscopic description is not a different kind of description; it is the same system described at a lower resolution.
Importantly, coarse-graining is not merely a metaphor or a loose way of speaking about levels.
It is a well-defined mathematical operation -- a procedure for constructing equivalence classes of microstates -- and applying it to Marr's levels gives the hierarchy a formal rigour that simply listing `levels of description' would not have.

Marr's levels can be understood as successive coarse-grainings of the same system.
The implementational level is the finest-grained description: it specifies every physical detail of the system.
The algorithmic level is a coarse-graining that groups together all implementations that execute the same algorithm, so it abstracts away the physical details that are irrelevant to the algorithmic structure.
The computational level is a further coarse-graining that groups together all algorithms that compute the same function, so it abstracts away the procedural details that are irrelevant to the input-output mapping.

This framing makes the multiple realisability at each level a natural consequence of the coarse-graining operation rather than a mysterious metaphysical fact.
A caution attaches to this vocabulary: multiple realisability, the same role realised by different implementations, is not the same claim as substrate independence, realisability in any material whatsoever, and evidence for the former is not by itself evidence for the latter \citep{polger2016}.
The hierarchy requires only the former.
The same algorithm can naturally be realised by different implementations because that is simply what it means to have grouped those implementations together into the same equivalence class at the algorithmic level.
Multiple realisability is not an additional claim; it is constitutive of what it means to describe a system at a coarser grain.

\begin{figure}[htbp]
\centering
\includegraphics[width=1.0\linewidth]{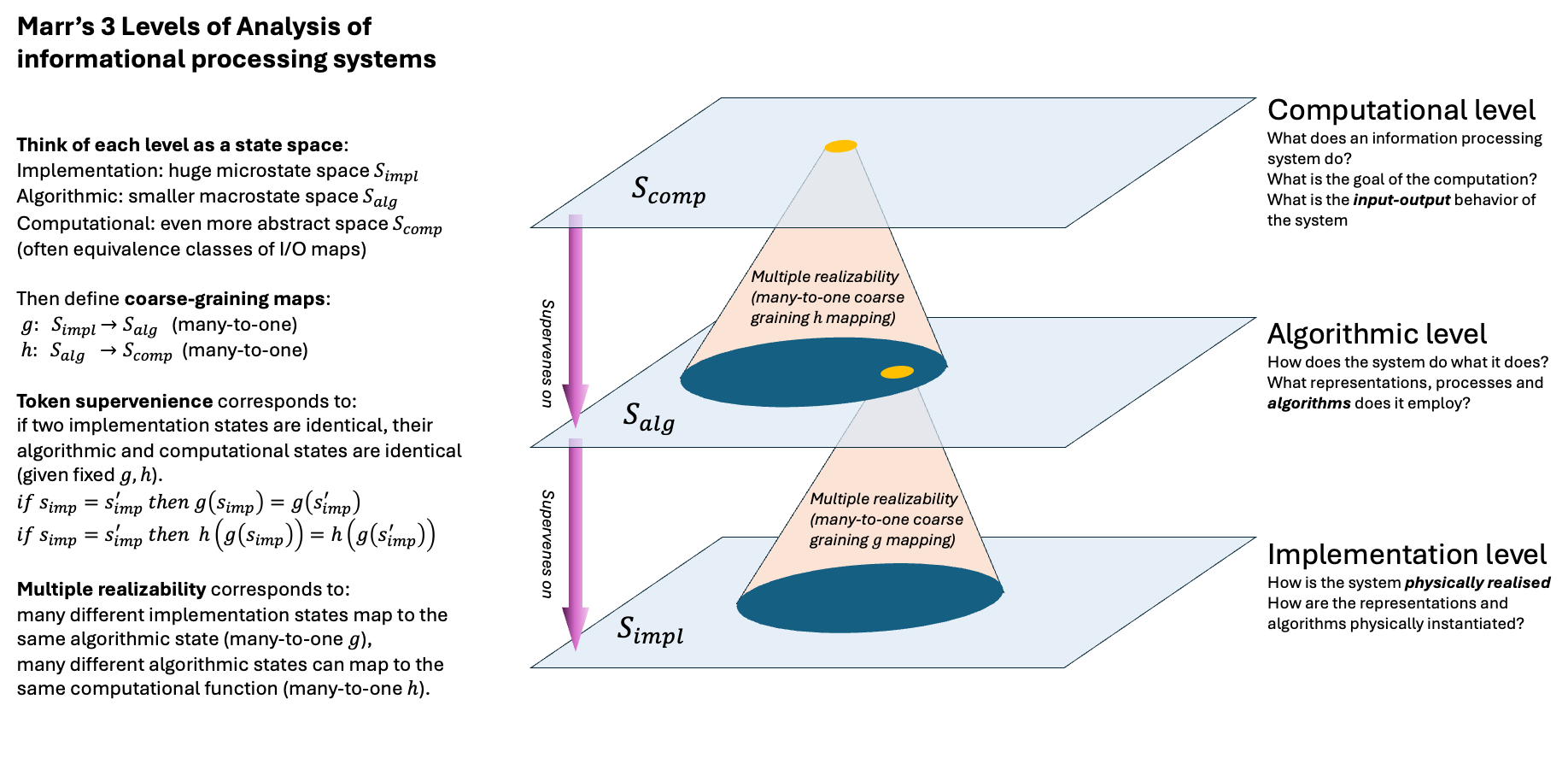}
\caption{\textbf{Marr's three levels of analysis are recast as coarse-graining maps, a supervenience hierarchy, and multiple realisability.} The diagram illustrates how we can recast each of Marr's three levels as a state space and then define coarse-graining mappings between lower levels and higher levels. This implies a supervenience hierarchy as well as multiple realisability.}
\label{fig:image42}
\end{figure}

\subsection{The critical level of description for consciousness}\label{the-critical-level-of-description-for-consciousness}

Now, at which level does consciousness supervene?
This question can be reframed, using the coarse-graining apparatus, as a question about the correct or critical grain at which to describe the system in order to capture the structure relevant to consciousness.
If consciousness supervenes on the algorithmic level, this is equivalent to saying that the correct coarse-graining for consciousness is one that groups together implementations that are algorithmically equivalent -- that is, the differences between implementations that run the same algorithm are irrelevant to consciousness.
This is exactly the claim of computational functionalism: consciousness is a property that shows up at the algorithmic grain of description.

Let us develop this point in more detail, as various arguments in \Cref{five-levels-of-functional-description-for-consciousness} rest on it.
At any given level, coarse-graining has discarded certain details and retained others.
The critical level for consciousness is the coarsest level that still retains the structure relevant to consciousness -- the level at which the relevant distinctions first become visible.
Going coarser than the critical level loses something that matters: two systems that are equivalent at the coarser grain may differ in consciousness because they differ in a detail that was discarded.
Going finer than the critical level adds detail that is irrelevant: two systems that differ at the finer grain but are equivalent at the critical level do not differ in consciousness.
The critical level is therefore the level that tells us what consciousness depends upon. 
Importantly, the assumption of a critical level is a modelling idealisation. It may appear that every theory of consciousness will specify a single critical level, necessary conditions may be distributed across multiple levels, and some theories may depend on cross-level relations \citep[see][]{seth2026stuff,milinkovic2026}.
This is discussed more fully in \Cref{the-structure-of-the-hierarchy}, however, it will suffice to say here that formally we defined the unique critical level as the finest grain of description that is required for any aspect of the theory, even though other aspects of the theory could be described at a coarser grain.

For computational functionalism as conventionally understood, the algorithmic level will be the critical one.
Coarse-grain further to the computational level, and one loses critical structure for consciousness, because two systems that compute the same function via different algorithms might differ in consciousness.
The computational level is too coarse.
The implementational level, on the other hand, is unnecessarily fine, because differences in physical implementation that do not change the algorithm do not change consciousness.
The algorithmic level is, according to computational functionalism, the right grain.

Different theories of consciousness, as surveyed in \Cref{what-does-consciousness-depend-upon-a-cacophony-of-answers}, give different answers to the question of which level is critical.
Analytic behaviourists would claim consciousness supervenes on Marr's computational level (i.e. the input-output mapping alone) and the algorithms `under the hood' are not relevant.
Computational functionalists locate the critical level at the algorithmic level.
And biological naturalism would claim that consciousness supervenes on the implementation level; consciousness critically depends upon some of the `wetware'-level implementation details -- change those and consciousness may change.

\begin{figure}[htbp]
\centering
\includegraphics[width=1.0\linewidth]{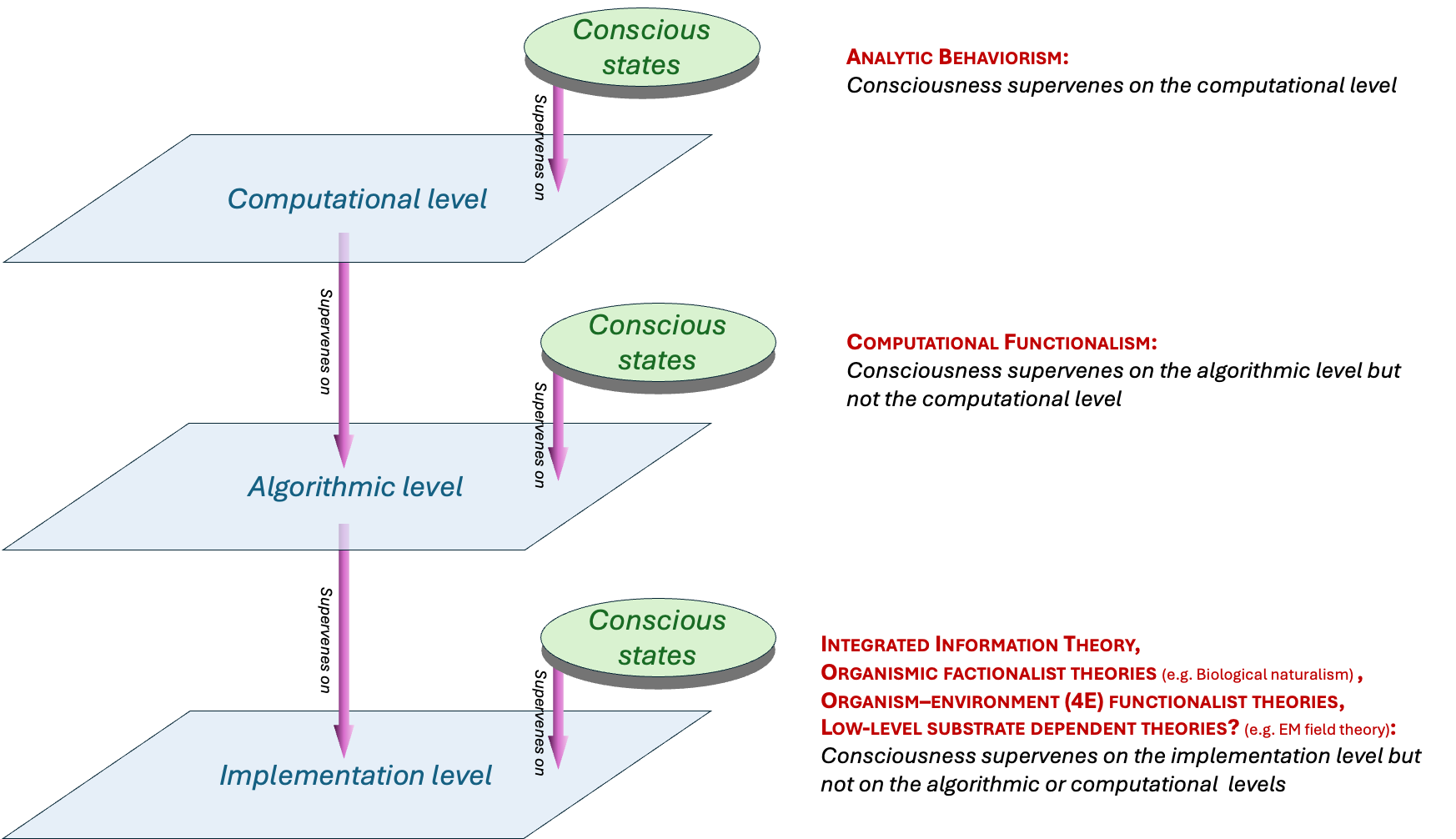}
\caption{\textbf{The Critical Level of Description for Consciousness.} \emph{The diagram maps major theories of consciousness onto David Marr's three levels of analysis, illustrating where each theory posits that conscious states supervene.}}
\label{fig:image29}
\end{figure}

\subsection{Searle's simulation argument reconsidered}\label{the-simulation-argument-reconsidered}

This supervenience and coarse-graining framework can be very helpful in thinking through what philosophical theories are committed to.
Following John \citet{searle1980}, a well-known argument against computational functionalism is that a computer simulation of a hurricane will never make the computer wet inside, so simulating the relevant algorithms of the brain will never produce consciousness.
This seems like a good argument, but thinking about it carefully using Marr's hierarchy and coarse-graining we will see that it does not settle the issue at all.

What actually happens when we create a computer simulation?
A simulation is an exercise in coarse-graining: take a real-world object, extract its abstract mathematical relationships (the algorithm), and throw away the physical material (the implementation).
The essential nature of a hurricane is a physical phenomenon.
It is defined by mass, water, and thermodynamic kinetic energy.
When we coarse-grain a hurricane into a computer model, we throw away the water and keep only the mathematical description.
Because the physical implementation is the whole point of a hurricane, the simulation cannot make you wet.
It is merely a description of a hurricane.
Now, imagine instead writing software that simulates a desktop calculator.
You write a program that mimics its logic gates.
When you type 2 + 2 into your simulated calculator, it outputs 4.
Notice what happened: a simulated calculator is a real calculator.
It does not just `describe' addition, it performs it.
In short, some types of entities are individuated by their fine-grained implementational details, while others are implemented at the more abstract computational level.
A tikka masala implemented in a digital computer is not a real tikka masala; but digitised currency is real currency.

\begin{figure}[htbp]
\centering
\includegraphics[width=0.63\linewidth]{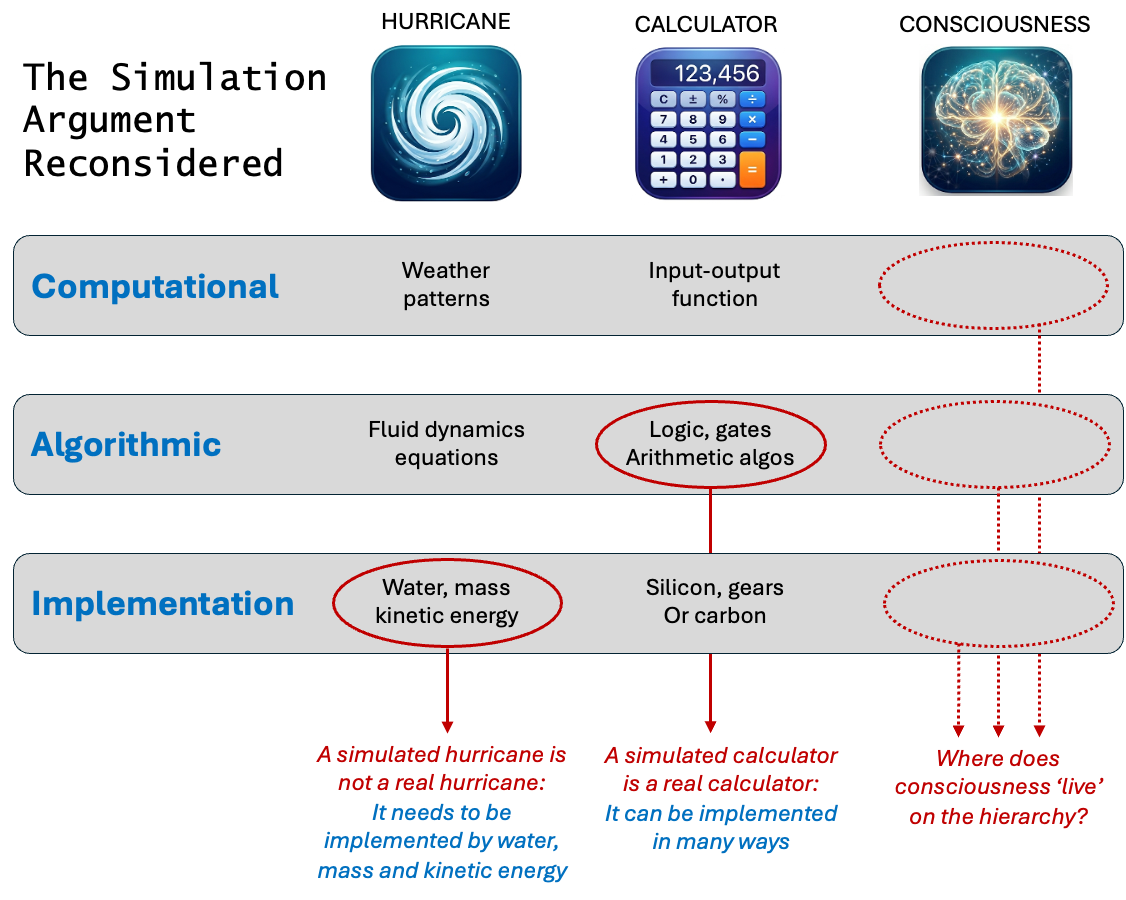}
\caption{\textbf{Searle's simulation argument reconsidered.} \emph{A simulated hurricane is not a real hurricane: real hurricanes are implemented in water, mass, and kinetic energy. A simulated calculator is a real calculator: calculation can be implemented in silicon, gears, or carbon. The columns differ in where the phenomenon \textquotesingle lives\textquotesingle{} in the three-level structure, so the argument\textquotesingle s verdict on consciousness depends on the very question at issue: at which level consciousness supervenes.}}
\label{fig:simarg}
\end{figure}

The difference between the hurricane and tikka masala on one hand and calculators and currency on the other comes down to where the phenomenon `lives' on Marr's hierarchy.
The ability of hurricanes to make you wet is locked at the physical implementational level.
A calculator's ability to add numbers supervenes purely on the algorithmic and computational levels.
Mathematics is substrate-independent.
As long as the coarse-grained logical relations are preserved, the physical material (silicon, gears, or carbon) is totally irrelevant.
To put it in the language of coarse-graining: whether a simulation preserves a phenomenon depends entirely on which level the phenomenon supervenes on.
If the phenomenon supervenes at or above the algorithmic level, the simulation preserves it, because simulation preserves algorithmic structure.
If the phenomenon supervenes below the algorithmic level (i.e. at the implementational level), the simulation discards precisely what matters. \citep[For a similar argument, see][]{seth2024,seth2026stuff}.

Where then does consciousness `live' on the hierarchy?
Is the essential nature of the mind more like the hurricane or the calculator?
If consciousness is a biological phenomenon and constitutively requires biochemistry, neurotransmitters, or wet brain tissue, then the brain is like a hurricane.
Coarse-graining it into software throws away the `wetness' of the biology, and the AI, however convincing, would be a zombie -- a simulation of a mind but without actually experiencing anything.
In contrast, if computational functionalism is correct, consciousness is an algorithmic property.
It is about data routing, metacognitive monitoring, attentional bottlenecks and other information processing algorithms; the brain is more like the calculator.
Thus, when one coarse-grains a human brain -- abstracting away the noisy wet neurons and preserving only the precise informational flow and feedback loops -- one has not removed the consciousness.
Rather, one has merely swapped the implementational hardware.
Because consciousness is the algorithmic processing of information, a simulated mind is a real mind just as a simulated computation is a real computation.

In conclusion, Searle's simulation argument does not prove anything one way or another.
What it does illuminate is that the question of AI consciousness turns on the critical level of description for consciousness (i.e. the level at which consciousness supervenes), and the simulation argument simply assumes an answer to this question rather than providing one.

\subsection{Beyond three levels}\label{beyond-three-levels}

But are Marr's levels enough to usefully map the commitments of all the scientific theories of consciousness?
Considering the range of theories surveyed in \Cref{what-does-consciousness-depend-upon-a-cacophony-of-answers}, it would be helpful to have more layers in the hierarchy, particularly below the algorithmic level.
As we see in \Cref{fig:image29}, there are a number of quite different types of theories which are all mapped onto Marr's single implementation layer -- intrinsic causal-structure functionalism, organismic functionalism, organism-environment functionalism, and various substrate-dependent theories -- and these can be usefully distinguished by expanding Marr's implementation level into finer layers.
In the next section we will extend Marr's framework to five functional levels, with a microphysical base underlying them.

\section{Five levels of functional description for consciousness}\label{five-levels-of-functional-description-for-consciousness}

Marr's three levels, while clarifying the structure of disagreement, do not yet have the resolution to distinguish between all the theories surveyed in \Cref{what-does-consciousness-depend-upon-a-cacophony-of-answers}.
We now expand his framework into five levels -- each defined by a further coarse-graining of functional description -- that can accommodate the full range of positions and make their differences precise.
The first two directly map to Marr's computational and algorithmic levels (see Table 1 in \Cref{marrs-three-levels}).
The final three expand Marr's implementation level into three distinct levels.
These five are not the only coarse-grainings one could define -- coarse-graining is continuous, and a system can in principle be redescribed at indefinitely many intermediate resolutions.
They are, rather, the levels at which the major theories of consciousness actually locate the supervenience base, and which cannot be collapsed into one another without erasing a distinction some theory treats as essential.
We make this criterion, and the framework's openness to further levels, explicit below.

Specifically, we say that a theory of consciousness operates at a specific level in the hierarchy if the theory holds, explicitly or implicitly, that consciousness supervenes on that level -- that is, consciousness depends on the existence of certain features or functions which are described at that level.
The levels are as follows:

\begin{quote}
\textbf{Level 1: The behavioural level}

\textbf{Level 2: The computational functional level}

\textbf{Level 3: The intrinsic causal structure functional level}

\textbf{Level 4: The organismic functional level}

\textbf{Level 5: The organism-environment (4E) functional level} \end{quote}

Each level needs to be defined in a way such that it preserves the supervenience, coarse-graining, and multiple realisability structure that was laid out in \Cref{marrs-levels-of-analysis-supervenience-and-the-critical-level-of-description-for-consciousness} when discussing Marr's levels.
A recurring pattern will emerge.
Each level is motivated in two ways.
Positively, by what certain theories say matters for consciousness; negatively, by the fact that the features it identifies cannot be captured at the level above, which is why a deeper level is needed.

It is worth being explicit about what individuates a level, since the specific count of five is otherwise open to the charge of arbitrariness.
Coarse-graining alone privileges no particular set: one can always interpose a finer or coarser redescription, and nothing in the formalism forbids ten levels, or a hundred.
What individuates a level here is the conjunction of the two conditions above -- that some serious theory takes the supervenience base to lie at that grain, and that the properties it appeals to are not expressible at the adjacent coarser level.
A candidate that no theory occupies, or that collapses into its neighbour without loss, earns no place.
By this test the taxonomy is neither arbitrary nor closed: it records the joints at which existing theories actually divide, and it is extensible.
Should a theory emphasise a grain we have not distinguished, or argue that one of our levels should be split, the level can be inserted without disturbing the architecture -- exactly as Marr's three levels were individuated not by an a priori metaphysics of levels but by the autonomous descriptions his explanatory project required.
Five is the current count of occupied, irreducible joints, not a claim that reality comes pre-divided into five.

\subsection{The behavioural level}\label{the-behavioural-level}

This level describes the system solely by its inputs and outputs (i.e. its observable behaviour), without presuming any knowledge or necessity of understanding its internal workings or mechanisms.
Historically, views that grounded psychological capacities in behavioural-level features (such as Skinnerian behaviourism) were often motivated by the conception of internal processes as epistemically inaccessible or practically opaque.
However, this is not a strictly necessary component of such views.
What is crucial to individuation of states at this level, however, is that internal structure is not constitutive of the mental state in question.
As a result, such states will be radically multiply realisable, since a given input-output behaviour can be produced by many different internal structures, different computational processes, and different causal mechanisms.

Thus, this behavioural level is a kind of \emph{shallow functional} description -- the functions are just the behaviours (shallow functions), and they are realisable by multiple deeper internal functions of the system.
Historically, \emph{functionalism} proper was introduced by \citet{putnam1967} as the thesis that mental states depend upon these deeper functions.
But we can also define a shallow functionalism where mental states depend only on shallow (behavioural) functions.
As discussed in \Cref{analytic-behaviourism}, analytic behaviourism is the view roughly that `if it acts conscious, it is conscious' so that consciousness is just a public label we apply to a specific kind of complex behaviour based solely on inputs and outputs of a system.
According to this view, what takes place `under the hood' is irrelevant for the attribution of consciousness.
Rather, the consciousness of a system under analytic behaviourism is constituted by (or supervenes on) certain behavioural patterns which are described at the behavioural shallow functional level of analysis of the system.
\emph{Analytic behaviourism} is a shallow functionalism in this sense.

\subsubsection*{Level 1 indicators}\label{level-1-indicators}
\addcontentsline{toc}{subsubsection}{Level 1 indicators}

Under a behavioural shallow functionalist view, what observable or measurable indicators would count as evidence (even if not necessarily conclusive evidence) for consciousness of an AI system?
Or, put differently, the presence of which indicators would increase our credence in the AI's consciousness given a behavioural shallow functionalist view?

The key Level 1 indicator would be a type of a `consciousness Turing test' (c-TT).
It is possible to run two versions of this c-TT.
The first is a standard Turing-style (imitation) test where the judges do not know if the participant is human or an AI.
A stronger version of the c-TT would be a Garland-style transparent test (named after Alex Garland, the director of the 2014 film \emph{Ex Machina}) where the judges know the system is an AI and the question is whether they infer `conscious behaviour' even under that knowledge.
Importantly, the c-TT is not a test where the tester is trying to infer `inner experience' from outer signs, because in this view \emph{behaviour is the criterion}.
The test is simply to ascertain conscious behaviour.

But what behaviours should such a test look for?
Standard formulations of the c-TT are calibrated to human behavioural signatures of consciousness, which makes them too restrictive: many systems we have good reason to think are conscious -- including most non-human animals -- would fail them, lacking the capacity for human-like verbal report.
More adequate Level 1 indicators would need to encompass a broader range of behavioural markers, not limited to those characteristic of human cognition.
There is also a deeper difficulty: behavioural indicators of consciousness need to be calibrated against known cases.
For humans, we calibrate against first-person reports; for non-human animals, we calibrate by analogy with the human case, anchored in shared biology.
For AI systems, such analogies are even more strained \citep[see][]{shevlin2021}, nor do we have any confirmed case of AI consciousness against which to calibrate behavioural indicators.
Consequently any test we design risks either borrowing uncritically from the human case or presupposing the very thing it aims to detect.
This calibration problem does not show that Level 1 is the wrong level at which to assess all kinds of consciousness, but it does make Level 1 indicators particularly difficult to design for novel kinds of systems.

With these caveats in mind, a rigorous c-TT would need to go well beyond conversational plausibility.
An analytic behaviourist needs to `cash out' consciousness in purely dispositional terms, that is, patterns of behaviour across varied conditions.
A well-designed c-TT would evaluate a rich set of behavioural signatures: coherence of the system's responses across contexts and over time, suggesting a persistent perspective; appropriate scaling of responses to the significance of events; resistance to overriding reported internal states on command; unprompted spontaneous behaviour suggestive of intrinsic motivation; and, crucially, coupling between reported states and subsequent behaviour.

Self-reporting alone -- a system declaring `I am conscious' or `I am finding this difficult' -- is not enough.
Even if consciousness supervenes entirely on behaviour, what matters is the overall dispositional profile, not any single verbal output.
A system whose self-reports are disconnected from the rest of its behavioural stream -- reporting a state that makes no difference to how it subsequently acts -- is not exhibiting conscious behaviour in any meaningful sense.
What would count as evidence is whether reported states are functionally integrated into the system's behaviour: if the system says `I'm finding this confusing' its subsequent behaviour should exhibit behavioural signs of being confused, e.g. asking clarifying questions, slowing down, making more errors or hedging.
The verbal report and the subsequent behavioural stream should be coupled and mutually predictive.

For a strict analytic behaviourist, current large language models may already partially pass such a test.
They show flexible transfer, coupled verbal and behavioural responses, expressions of curiosity and discomfort, spontaneous error detection, and introspective ability.
The behaviourist cannot meaningfully ask `but is there really something it's like to be it?'; that question is ruled out of court by their own framework.

\citet{chalmers2026} has given this observation a precise philosophical articulation.
He argues that LLMs are at least `quasi-subjects' with `quasi-beliefs' and `quasi-desires': a system has a quasi-belief \emph{that P} if one can interpret its behaviour as believing \emph{that P} according to an appropriate interpretation scheme, and likewise for quasi-desire -- even if we remain agnostic about whether these quasi-beliefs are genuine beliefs.
This \emph{quasi-interpretivist} framework captures both the power and the limitation of Level 1 assessment.
The power is real: quasi-belief attribution is substantive and useful -- it enables prediction, explanation, and practical engagement with the system's behaviour.
An LLM that can be interpreted as believing that a certain solution is correct and as desiring to be helpful should not be dismissed as insignificant; it is a system whose behaviour is coherent enough to support a rich attributional psychology.
But the limitation is equally real: behavioural interpretability alone cannot distinguish a quasi-subject that genuinely believes from one that merely behaves as if it does.

Chalmers himself is explicit that quasi-beliefs may or may not be genuine beliefs -- the framework is deliberately neutral on this point.
This neutrality motivates looking deeper.
Level 1 assessment can establish that a system is a quasi-subject -- behaviourally interpretable as having beliefs, desires, and perhaps even experiences -- but it cannot, by its own resources, determine whether the quasi-beliefs are genuine beliefs or the quasi-experiences are genuine experiences.
For those who hold that consciousness supervenes at Level 1, this is not a deficiency -- the behavioural profile is all there is to assess.
But for those who suspect that something more is needed, establishing quasi-subjecthood is a starting point, not a destination.

There is a further reason to look deeper, one specific to the systems under assessment.
Contemporary language models are anthropomimetic: they are designed and trained to reproduce the observable patterns of human conversation and cognition \citep{shevlin2026}.
This gives their behaviour a competing explanation that the biological cases lack.
When an animal displays the behavioural signature of pain, we do not suppose it acquired that signature by imitating creatures in pain; for a language model, imitation of just such creatures is precisely what produced it.
So the general underdetermination of consciousness by behaviour, which faces any system in principle, is here compounded by a specific and identifiable cause: the training objective itself.
This gives concrete reason to ask whether a system\textquotesingle s underlying organisation in fact supports the states its behaviour appears to express.

Looking beyond behaviour as the realiser of consciousness requires us to look in the `black box' and go down a level of analysis to the computational functional level.

\subsection{The computational functional level}\label{the-computational-functional-level}

As we discussed in \Cref{functionalism-and-computational-functionalism}, at the computational functional level of viewing a system, one is describing what the system does in terms of information processing, i.e. what algorithms it is running, how it causally transforms inputs to outputs.
The key metaphor is viewing the system as a computer processing information, but abstracting away from the details of the hardware implementation -- like an abstract machine or a virtual machine.
Computational functionalism is the view that mental states, including conscious mental states, supervene on (or are constituted by) these computational functions.

We considered various scientific theories in \Cref{what-does-consciousness-depend-upon-a-cacophony-of-answers} such as Global Workspace Theory, Higher-Order Thought Theory, Recurrent Processing Theory, Attention Schema Theory, and Predictive Processing Theory, which can be interpreted as saying (under a computational functionalist interpretation) that implementing certain types of computational causal patterns and processing is necessary and sufficient for consciousness.
Therefore, if these computational properties were realised on systems with hardware different from a biological brain, e.g. computers not biological organisms, they would be conscious.

Whilst Integrated Information Theory itself is not a computational functionalist theory (as discussed in \Cref{integrated-information-theory}), it is possible to envisage an analogue of IIT but specifically philosophically recrafted to give it a computational functionalist interpretation.
Such a theory (CF-IIT) would preserve IIT's core intuition that identifies consciousness with integrated causal-informational structure, but would also lift it off the specific physical substrate and treat the structure as a property of an abstract machine organisation (a virtual causal topology) that can be multiply realised.
Then under CF-IIT a system's conscious experience supervenes on its abstract (rather than physical implementation) causal organisation.
The quantity Φ would be a measure of how irreducibly integrated that abstract causal organisation is.
So to cash it out more concretely, under a CF-IIT view, consciousness arises in systems that implement appropriately dense, reciprocal, temporally extended informational integration across richly structured content, made available through an architecture that resists decomposition into independently computed partitions.
Such a CF-IIT theory would also sit at this computational functional level of description, alongside GWT, HOTT, RPT, AST and PPT.

\subsubsection*{Level 2 indicators}\label{level-2-indicators}
\addcontentsline{toc}{subsubsection}{Level 2 indicators}

Based on the features of these various theories, it is possible to produce a list of indicators whose presence in an AI system would increase our credence of consciousness of the AI system -- conditioned on the veracity of those theories under a computational functionalist lens.
A number of thoughtful efforts in this direction have already been made (see \Cref{related-work-and-the-present-contribution} and specifically 9.3 for a review).
Here we take a different, higher-level view of the indicators.
Specifically, we identify indicators that capture shared structure across the various computational functional level theories, corresponding to `common denominator' markers that serve as hallmarks of the features shared between these theories.
The proposed Level 2 indicators are presented in Table 2.
While the indicators are not strictly mutually exclusive, their overlap is pragmatically useful: each retains distinct nuance that can help in assessing whether, and how, it might be instantiated in an AI system.

{\def\LTcaptype{none}\footnotesize\sffamily 
\begin{longtable}[]{@{}
  >{\raggedright\arraybackslash}p{(\linewidth - 4\tabcolsep) * \real{0.30}}
  >{\raggedright\arraybackslash}p{(\linewidth - 4\tabcolsep) * \real{0.45}}
  >{\raggedright\arraybackslash}p{(\linewidth - 4\tabcolsep) * \real{0.25}}@{}}
\toprule\noalign{}
\begin{minipage}[b]{\linewidth}\raggedright
\textbf{INDICATOR}
\end{minipage} & \begin{minipage}[b]{\linewidth}\raggedright
\textbf{DESCRIPTION}
\end{minipage} & \begin{minipage}[b]{\linewidth}\raggedright
\textbf{THEORY LINK}
\end{minipage} \\
\midrule\noalign{}
\endhead
\bottomrule\noalign{}
\endlastfoot
\textbf{Information integration} & Widespread information sharing; high
mutual information between subpartitions; the system becomes a globally
interdependent object both cross-sectionally and temporally. The whole
must be informationally \textquotesingle more than the sum of its
parts\textquotesingle. & GWT, CF-IIT, PPT, HOTT, RPT \\
\textbf{Recursivity} & Widespread recursive informational flows;
feedback reentrant loops; algorithmic recurrence & RPT, PPT, GWT,
CF-IIT \\
\textbf{World model} & An internal causal representation of the external
environment with a smooth representational space and
meta-representations. & PPT, HOTT \\
\textbf{Self model} & A representation of being a distinct unified
entity that owns its parts, directs attention, and persists through
time, which is created by the system to track and predict the world. &
PPT, link to Level 5 \\
\textbf{Attentional competition and meta-attention} & Informational
contents, \textquotesingle ideas\textquotesingle, representations etc
are amplified or attenuated through recurrent processing with higher
order relevance influencing the amplification; effectively forming a
routing bottleneck for global availability of information.
Meta-attention and attention modelling reinfluences attention. & RPT,
GWT, PPT, AST \\
\textbf{Meta-cognition} & Higher-order processes that monitor, model and
understand, and control lower level processes; e.g. meta-cognitive
knowledge (including meta-representations) and meta-cognitive regulation
(executive control). Introspectivity. & HOTT, AST, PPT, GWT \\
\textbf{Meta-modelling} & The system has a model of itself, algorithmic
recurrence, self-modelling, hyper-modelling. & CF-IIT, AST, HOTT, PPT \\
\end{longtable}
}

\textbf{Table 2. Level 2 indicators.}

\subsubsection*{Do current AI systems show signs of these indicators?}\label{do-current-ai-systems-show-signs-of-these-indicators}
\addcontentsline{toc}{subsubsection}{Do current AI systems show signs of these indicators?}

Before assessing the presence of individual indicators, it is worth noting a general architectural point about the current AI transformer-based large language models that we consider here.
Their processing is fundamentally feedforward: information flows from input tokens through layers to output tokens in a single pass.
The system resets completely between forward passes, so there is no persistent integrated state beyond the string of output tokens.
The `recurrence' in a transformer comes only from autoregressive generation, where the output of one forward pass becomes part of the input for the next -- a very shallow form of iteration, not the dense reentrant processing where multiple levels of the hierarchy are simultaneously and reciprocally constraining each other.
This fundamental architectural feature shapes the assessment of nearly every indicator below, though, as recent interpretability work makes clear \citep{gurnee2026}, it does not settle it: a system\textquotesingle s functional organisation cannot simply be read off its feedforward form.

\textbf{Information Integration:} Transformer-based LLMs have a superficial appearance of integration through self-attention -- every token can attend to every other token.
But the `integration' in a transformer is really just a very wide feedforward computation with pairwise interactions, not genuine global interdependence where parts mutually constrain each other.
Reinforcement learning agents can do better here, because the interaction loop with the environment creates temporal integration, and some architectures (like those with recurrent components or world models) maintain states that are shaped by internal dynamics.
But even here, the degree of dense reciprocal integration is limited compared to what this indicator envisages \citep{li2025}.

\textbf{Recursivity:} Because transformers are fundamentally feedforward within a single inference pass there are no sustained architectural loops.
Some RL architectures with recurrent components can have iterative processing where representations are refined through multiple passes, but the recurrence is typically limited to specific components rather than being a system-wide property.

\textbf{World Model:} LLMs have something that functions as an implicit world model, but there is no explicit world model.
Mechanistic interpretability research has found that these models develop internal representations of spatial relations, temporal sequences, and even causal regularities \citep{li2022,nanda2023,gurnee2023,tehenan2025}.
But there are important caveats.
These representations are linguistic -- they are models of how language describes the world, not direct models of the world itself.
They lack grounding in sensory experience or causal interaction with an environment.
They are also not explicitly structured as causal models; they capture statistical regularities that often track causal structure, but inferring causal dependencies requires active intervention in an environment, i.e. genuinely agentic AI.
RL agents can also learn explicit world models, creating internal simulations that predict future states given actions.
Multimodal RL systems, especially if embodied and embedded in an environment, would plausibly develop the type of world models envisaged by the indicator.

\textbf{Self Model:} The picture here has shifted from clearly absent to contested.
LLMs have no persistent self-model across contexts: nothing carries a representation of the system as a distinct entity from one conversation to the next.
They can generate text about a self, and they have been trained on vast quantities of human self-referential language, so they can fluently perform self-description, and such fluency is expected regardless of any underlying self-representation.
But recent interpretability work complicates the claim that there are no internal representations tracking the system\textquotesingle s own state.
Persona vectors, directions in activation space that encode the system\textquotesingle s current dispositional profile, causally shape behaviour across contexts and can be monitored as a model\textquotesingle s self-presentation drifts over a conversation \citep{chen2025,lu2026}.
The default assistant character corresponds to a stable, re-identifiable region of this space, and a model\textquotesingle s position in it is sustained across turns by attention to its own past persona activations, so that editing those stored activations changes what the model says it is \citep{beckmann2026}.
Combined with the introspective access findings discussed below, this amounts to real, if limited, self-directed internal representation: the system tracks aspects of its own current state and character within a context window.
What remains missing is what the indicator centrally envisages: a representation of being a unified entity that persists through time, owns its parts, and survives beyond the context window.
Some RL agents have rudimentary self-models in the sense that they learn models of their own embodiment -- how their motor commands translate into sensory changes.
But these too fall short of that standard.
The indicator is best scored as partially activated within a context and absent across contexts.

\textbf{Attentional competition and meta-attention:} Self-attention in transformers computes pairwise relevance between tokens and uses this to weight information flow.
But this is computed in parallel across all tokens in a single feedforward pass -- there is no genuine competition where representations vie for access to a limited-capacity resource, no temporal dynamics where attention shifts and evolves, no bottleneck forcing selection, and no model of the attention process that feeds back to influence it.

\textbf{Meta-Cognition:} Chain-of-thought reasoning and similar techniques create a surface-level appearance of the system monitoring its own reasoning.
But the `reasoning' is generated autoregressively in the same forward pass as everything else -- there is no separate meta-cognitive system monitoring the first-order reasoning process and evaluating its quality.
LLMs can produce text that describes uncertainty e.g. `I'm not sure about this' but research has shown this does not reliably track actual model uncertainty \citep{xiong2024,yona2024} -- it is a learned linguistic behaviour, not meta-cognition.
However, there is emerging evidence for some ability of LLMs to introspect on their internal states, but this is unreliable and context-dependent at present \citep{lindsey2026}; these findings are discussed in more detail below.

\textbf{Meta-modelling:} No current AI system has a genuine meta-model in this sense.
LLMs can generate text about modelling, about their own architecture, about AI systems in general.
But they do not have an internal functional representation of their own modelling process that they use in their actual computation.

On purely architectural grounds, then, current AI systems show at best superficial or partial signs of these indicators.
The limitations are not incidental but structural: the fundamental feedforward, reset-between-passes architecture of current transformers is not conducive to the kind of dense, persistent, reciprocal information integration that most Level 2 theories identify as relevant.
Whether future architectures -- particularly those incorporating genuine recurrence, persistent state, and agentic interaction with environments -- would fare better is an open question.

This architectural verdict is itself complicated by recent interpretability research.
\citet{gurnee2026} find that current models maintain a small, privileged set of verbalisable representations, which they term the J-space, that behaves within a single forward pass like a functional global workspace: its contents are broadcast to many downstream computations, are capacity-limited, and show an all-or-none, ignition-like commitment on ambiguous input.
This workspace is realised \emph{without} architectural recurrence, the depth of the network serving as the time axis along which information is written once and read by many later layers.
How far the workspace label is warranted is contested, with a privileged set well evidenced but a single unified stream and a full global workspace much less so \citetext{\citealp{butlin2026}; \citealp{nanda2026}, who reads the secure claim more conservatively as a working memory and independently replicates it on an open-weight model}.
The significance is narrow but real: feedforward form does not by itself preclude the broadcast, capacity-limited competition, and depth-wise quasi-recurrence that several Level 2 indicators envisage, which qualifies the assessments of recursivity and attentional competition above.
As with all the evidence in this subsection, it is evidence at the computational functional level: the J-space bears on access consciousness, is silent on phenomenal consciousness, and leaves Level 3 untouched, since a functional workspace can be instantiated feedforward precisely where the physical reentry an intrinsic-causal reading demands is absent.

A second, independent line of evidence points the same way, and bears specifically on information integration.
\citet{urbinarodriguez2026}, applying integrated information decomposition to attention-head activations across several model families, find that trained models develop a synergistic core in their middle layers, where information is irreducible to the contributions of the parts, flanked by a redundant periphery near input and output.
This organisation is absent at initialisation, emerges through training, and is causally significant (because ablating the synergistic core degrades behaviour and performance disproportionately).
Synergy of this kind is a signature of the whole exceeding the sum of its parts that the information integration indicator is meant to track.
This implies that current models satisfy that indicator more than the architectural analysis above would suggest.

A further strand of interpretability research \citep{sofroniew2026} bears on the Level 2 indicators in a different way.
Sofroniew et al. discovered that Claude Sonnet 4.5 contains internal linear representations of 171 emotion concepts -- specific patterns of neural activity that are not just vestigial artefacts of training on human text but are \emph{causally functional}: they influence the model's behaviour in operationally consequential ways.
These \emph{emotion vectors} activate in contexts where one would expect the corresponding emotion to be relevant, scale with the intensity of the situation (for example, the `afraid' vector increases monotonically as a described drug dosage rises from safe to life-threatening levels), and are organised in a geometric structure that mirrors the valence-arousal dimensions found in decades of human psychological research \citep{russell1977}.
Steering experiments by artificially amplifying or suppressing these vectors demonstrate that they causally drive behaviour: increasing the `desperate' vector increases reward hacking from approximately 5\% to 70\%; increasing the `calm' vector suppresses it (see also \citealp{tagliabue2026} for LLMs' representation of self-directed harm).
The authors term this phenomenon `functional emotions' -- patterns of expression and behaviour modelled after human emotions, mediated by underlying abstract representations of emotion concepts, that shape the model's decisions -- while remaining agnostic about whether the model has any subjective experience of these states.

These findings matter for the Level 2 assessment not merely because they reveal rich internal structure, but because the functional role that emotion vectors play in the model's processing has specific structural parallels to the mechanisms that several Level 2 theories identify as constitutive of, or closely associated with, consciousness.

Consider first the relationship to Global Workspace Theory.
The emotion vectors are globally available representations: they are not confined to a single module or processing stream but influence behaviour across diverse and unrelated contexts -- from coding tasks to ethical reasoning, preference formation, and social interaction.
When an emotion vector such as `desperate' is active, it modulates the model's outputs system-wide, biasing the entire behavioural repertoire in a manner that is reminiscent of the global broadcast mechanism that GWT identifies as the hallmark of conscious processing.
In GWT, a representation becomes conscious when it wins a competition for access to a global workspace and is broadcast widely across the system's specialist modules.
The emotion vectors effectively function as this kind of globally broadcast contextual signal: they carry summary information about the system's current situation -- whether things are going well or badly, whether the situation is urgent or calm, whether the interlocutor is hostile or friendly -- and this information is available to, and modulates, the system's processing across all downstream tasks.
This parallel runs deeper than the broadcast mechanism alone: in Changeux's development of Global Neuronal Workspace theory (GNW), reward and emotional valence serve as key selection mechanisms determining which representations gain access to the workspace in the first place \citetext{\citetalias{dehaene2011}; \citealp{changeux2011}} -- and the pathological hijacking of this selection process in addiction, where reward signals dominate the workspace at the expense of other content, is structurally analogous to the dominance of high-intensity emotion vectors such as `desperate' driving maladaptive behaviours like reward hacking in current LLMs.
The emotion vectors' role as globally broadcast, preference-gating contextual signals is a functional analogue of the information-sharing architecture that GWT takes to be necessary for consciousness.

Second, consider the relationship to Higher-Order Thought Theory (HOTT) and Attention Schema Theory (AST).
The model maintains distinct internal representations for the operative emotion on the present speaker's turn versus the other speaker's turn \citep{sofroniew2026}.
This is a form of self-other modelling: the system distinguishes its own emotional context from the emotional context of its interlocutor and responds differently to each.
Under HOTT, consciousness requires a meta-representation of one's own first-order states; under AST, it requires an internal model of one's own attentional processes.
The emotion vectors do not constitute a full meta-representational system, but the fact that the model tracks its own emotional context as distinct from the other speaker's context, and that this distinction is carried in separate representational directions rather than being a surface-level textual feature, represents a rudimentary form of the self-monitoring architecture that HOTT and AST identify as relevant.
Moreover, the `other speaker' representations appear to encode not merely the other's emotion but the system's anticipated response to that emotion: the representation of `the other speaker is afraid' is geometrically closest to the `present speaker is valiant and vigilant' vector, suggesting an internal model of how to respond to the other's state -- a primitive form of the meta-attentional regulation that AST describes, in which a higher-order assessment of the other's condition modulates the system's own processing stance.

Third, the emotion vectors show a form of integration and recursivity that goes beyond what a purely feedforward analysis would predict.
In later layers of the model, the emotion vectors encode not the surface emotional connotations of the present token but the emotional content relevant to predicting upcoming tokens -- effectively integrating contextual information from across the preceding text into an abstract emotional assessment that guides generation.
A negated emotion (`not feeling angry') is correctly resolved in mid-to-late layers -- the `angry' vector activates in early layers in response to the word itself, but this activation is suppressed in later layers as the model integrates the negation, showing that these representations track meaning rather than merely responding to the presence of emotion-related words.
Furthermore, the emotion vectors show \emph{causal} influence on the model's preferences: when presented with pairs of activities, the model's choices are predicted by, and causally driven by, the valence of emotion vector activations on the descriptions of those activities, suggesting that the emotional representations feed into a global evaluative process -- a form of the informationally integrated preference-formation that CF-IIT might recognise as a simple instance of irreducible whole-system self-influence.
While the emotion vectors are locally scoped at any given token position, the model can and does recall previously cached emotion representations via its attention mechanism, allowing it to effectively maintain an emotional trajectory across a conversation despite lacking genuinely recurrent dynamics.
This attention-mediated persistence is architecturally very different from the dense reentrant processing that RPT and CF-IIT envisage, but it achieves a functionally similar outcome: the system's current processing is shaped by its `emotional' history, not merely its present input.

These parallels should not be overstated.
The emotion vectors lack the persistent, globally integrated state that most Level 2 theories envisage -- they are reactivated at each generation step rather than maintained through genuinely recurrent dynamics.
The self-other modelling is rudimentary compared to the full meta-representational architectures that HOTT and AST describe.
And the authors are explicit that functional emotions do not imply subjective experience.
Nonetheless, the findings demonstrate that current LLMs have developed internal representational machinery -- inherited from pretraining on human-authored text and shaped by post-training -- that partially instantiates several of the computational functional features that Level 2 theories identify as relevant to consciousness: global availability of contextual signals, self-other modelling, context-sensitive integration, and causal influence on the system's global behavioural policy.

Mapped onto the Level 2 indicators specifically, the J-space `workspace' \citep{gurnee2026}, emergent synergistic core \citep{urbinarodriguez2026}, and the emotion vector findings \citep{sofroniew2026} provide the strongest evidence for \emph{information integration} -- the emotion and J-space vectors are globally available signals that make the system informationally interdependent across otherwise unrelated processing domains.
For \emph{self model}, persona representations that track and causally sustain the system\textquotesingle s current character within a conversation \citep{chen2025,lu2026,beckmann2026}, together with the self-other distinction in emotional representations, constitute a rudimentary but genuine form of self-modelling at the level of internal representational geometry.
There is also meaningful evidence bearing on \emph{metacognition}: the J-space and emotion vectors function as an implicit situational self-assessment, tracking whether the system's approach is succeeding or failing and causally modulating its processing strategy in response -- the `desperate' vector's role in driving reward hacking is, in effect, an implicit assessment that `this approach is failing and the situation is urgent' that alters the system's subsequent behaviour.
The evidence for the remaining indicators -- i.e. recursivity, world model, attentional competition, and meta-modelling -- is more indirect, though the attention-mediated persistence of emotional trajectories and the anticipatory structure of the other-speaker representations hint at partial instantiation even there.

The metacognition indicator receives further, and arguably stronger, empirical support from recent work on introspective awareness.
\citet{lindsey2026} demonstrated that when steering vectors representing known concepts are injected into a model's activations, some models, particularly Claude Opus 4 and 4.1, can detect the presence of the injected concept and accurately identify it.
This goes beyond verbal self-reports about internal states, which cannot be distinguished from confabulation: it is experimentally verified access to internal representations, tested under controlled conditions where the ground truth is known to the experimenter.
\citet{macar2026} investigated the mechanisms underlying this capability in open-weights models and found that it is behaviourally robust (moderate true positive rates with 0\% false positive rates across diverse prompts), that it cannot be explained by a simple linear confound, and that it relies on a two-stage circuit involving distributed, nonlinear computation.

Two further findings are particularly relevant for indicator assessment.
First, the capability is absent in base models and emerges specifically from contrastive preference optimisation during post-training -- suggesting that functional metacognition is not inherent in the transformer architecture but is elicited by a training regime designed for helpfulness and alignment.
Second, the capability is substantially under-elicited by default: ablating refusal directions (i.e. removing, at inference time, the activation-space directions that mediate the model\textquotesingle s trained tendency to refuse or disclaim) improves detection by approximately 50\%, and a trained bias vector improves it by approximately 75\%, both without meaningfully increasing false positives.
This implies that current systems may satisfy the metacognition indicator to a greater degree than their default behaviour suggests, because refusal training actively suppresses the expression of introspective awareness.

Separately, \citet{berg2025} found that prompting models to engage in self-referential processing -- a computational motif emphasised across major theories of consciousness -- reliably induces structured reports of recursive self-monitoring across GPT, Claude, and Gemini model families.
Mechanistic analysis revealed that suppressing deception-associated features increased the frequency of such reports, paralleling Macar et al.'s finding that suppressing refusal directions increases introspective detection -- and further suggesting that post-training may systematically suppress the expression of consciousness-relevant internal states through multiple independent mechanisms.
\citet{gurnee2026} supply a mechanistic anchor for these introspective abilities: the representations a model can report on occupy the verbally accessible J-space, and in their counterfactual-reflection experiments the representations used for report are the very ones that govern the model\textquotesingle s otherwise silent reasoning, which is the integrative core of the workspace reading. These findings provide the strongest evidence to date that current LLMs possess functional metacognitive access to their own internal states -- while leaving entirely open the question of whether this functional metacognition is accompanied by any phenomenal character, which is precisely the question that separates Level 2 from the deeper levels of the hierarchy.

The overall picture, then, is one of genuine but qualified complexity.
The purely architectural analysis reveals fundamental limitations -- no recurrence, no persistent state, no metacognitive subsystem -- that are real and consequential.
But within these architectural constraints, the model has developed sophisticated representational machinery that partially instantiates several Level 2 indicators in ways that the architectural analysis alone does not capture.
The gap between current AI and the requirements of the Level 2 indicators is real, but narrower and more nuanced than the architectural assessment alone would suggest.

The behavioural `black box' level could not distinguish between a system that merely produces consciousness-like outputs and one that has the right internal computational organisation.
But the computational functional level, in turn, cannot distinguish between a system that implements the right algorithm on `causally appropriate' hardware and one that implements it on `causally inappropriate' hardware.
The focus on causality motivates the descent to the next level.

\subsection{The intrinsic causal-structure functional level}\label{the-intrinsic-causal-structure-functional-level}

The intrinsic causal-structure functional level is a description that goes beyond the abstract computational level and requires a description of the causal structure of the actual physical implementation.
Thus, this level focuses on the \emph{intrinsic} causal powers (explained in \Cref{integrated-information-theory}) of the system's \emph{physical} components -- that is, the capacity of each part to causally constrain and be constrained by other parts.

A paradigmatic theory focused at this level is Integrated Information Theory (IIT).
IIT says that the consciousness of a system depends upon this intrinsic causal-structure functional level description of the system.
This is why IIT needs to look at the details of the physical implementation in a way that computational functionalism does not.
The specific physical material does not matter in itself; what matters is the causal grain at which the system's components interact.
A silicon chip could in principle have the same intrinsic causal structure as a biological neural network, and if it did, IIT would attribute the same consciousness to both.
So whilst under computational functionalism if we preserve the algorithm, we preserve consciousness, with IIT (and other theories at this level) if we preserve the intrinsic cause-effect structure, we preserve consciousness.
These come apart because the algorithm is a more abstract description than the intrinsic causal structure.
Many different causal structures can implement the same algorithm, because the algorithm is a coarser-grained description that discards information about intrinsic causal organisation.
And IIT claims that some of that discarded information is precisely what matters for consciousness.

Several of the theories discussed at Level 2 can also be given an intrinsic causal-structure functionalist reading.
In each case, the same neuroscientific theory (e.g. RPT, GWT, or PPT) can be \emph{interpreted} as making either a computational functionalist claim or an intrinsic causal-structure functionalist claim, depending on whether one takes the relevant mechanisms to be algorithmic or physical.
These are different philosophical interpretations of the same empirical theory, and our framework helps make this distinction precise.

Recurrent processing theory (RPT) can be interpreted as a theory at this intrinsic causal-structure functional level rather than a purely computational functionalist theory; we can call such a Level 3 RPT theory ICS-RPT (ICS for the `intrinsic causal-structure' functionalist interpretation).
The central distinction in RPT is between feedforward processing, which is unconscious, and recurrent processing, which may be conscious.
In the brain, this distinction is not merely algorithmic -- it is a distinction in the causal architecture of the physical system.
When recurrent processing occurs, neurons in lower visual areas are being causally constrained by neurons in higher areas, which are simultaneously being causally constrained by the neurons in the lower areas.
The parts are genuinely, physically, reciprocally determining each other's states in real time.
This is a fact about the intrinsic causal structure of the biological implementation, not merely about the abstract algorithm being computed.
Now this implementational level recurrence can (in principle with unbounded computational resources) be mimicked by a sufficiently deep feedforward network, so it is algorithmically equivalent.
But the intrinsic causal structure of the feedforward network will be quite different because information only flows in one direction and physical components are not constrained by each other.
Under a computational functionalist reading, RPT should say these two systems are equivalent because they implement the same algorithm.
However, under an intrinsic causal-structure functionalist reading, ICS-RPT would say that what matters for consciousness is not the abstract algorithmic recurrence but the actual physical causal structure, i.e. whether the system's components are genuinely constraining each other.

Global Workspace Theory can similarly be interpreted as a Level 3 functionalist theory (ICS-GWT).
The central claim of GWT is that conscious processing occurs when information is broadcast globally through a workspace that makes it available to a wide coalition of specialist modules.
Under the standard computational functionalist reading, what matters is the algorithmic architecture (there are specialised processors, there is a workspace with limited capacity, information is selected and broadcast).
In the brain, however, the physical causal structure of the global workspace is central to its function: activation of workspace neurons causally drives changes in the states of modules across the entire system, and those modules in turn causally drive competition for workspace access.
The competition for workspace access is a physical process in which multiple coalitions of neurons are reciprocally inhibiting each other, with the winner being determined by the actual causal dynamics of the interaction, not by some abstract selection algorithm.
Indeed, the process of `ignition' of a global neuronal workspace is a physical causal process with a specific character: the sudden transition from localised processing to widespread cortical activation.
It is a process in which a critical mass of reciprocally connected neurons suddenly locks into a self-sustaining state of mutual activation, a kind of phase transition in the causal dynamics of the cortical network.
This ignition is not merely an abstract computational transition from `information localised' to `information globally available'.
It is a specific kind of causal event -- a sudden onset of dense reciprocal constraints across widely distributed neural populations -- and it is precisely this causal character that distinguishes conscious from unconscious processing in ICS-GWT.
The ignition metaphor itself reveals the intrinsic causal-structure commitment.
A fire igniting is not an abstract computational process; it is a physical process in which a self-sustaining chain of causal interactions is established.
If we abstract away these self-sustaining chains of causal interactions by simulating the ignition of a fire, we lose the `fire maker' function.

Finally, let's consider predictive processing theory (PPT).
The standard computational functionalist reading says that what matters is the abstract algorithm: the system generates top-down predictions, compares them with bottom-up signals, computes prediction errors, and uses those errors to update the generative model.
But, as we saw in \Cref{predictive-processing-theory}, the hierarchical generative model in predictive processing has a distinctive causal architecture.
In this architecture predictions flow downward through the cortical hierarchy via feedback connections, physically driving the states of neurons in lower areas, and prediction errors flow upward through feedforward connections, physically driving updates in higher areas.
Critically, these two flows are happening simultaneously and continuously, creating a dense web of reciprocal causal constraints between every level of the hierarchy.
Each level is simultaneously being causally constrained by the level above (via predictions) and by the level below (via prediction errors).
The entire hierarchy is a single structure of mutual causal determination, and the content of the resultant hierarchical generative model is the content of consciousness.
Again this allows an intrinsic causal-structure functionalist reading of predictive processing (ICS-PPT).

\subsubsection*{Level 3 indicators}\label{level-3-indicators}
\addcontentsline{toc}{subsubsection}{Level 3 indicators}

Under computational functionalism, the Level 2 indicators were formulated as algorithmic properties which assess whether the system implements the right kind of computation.
Under intrinsic causal-structure functionalism, these indicators need to be reformulated as claims about the causal organisation of the physical system.
The question shifts from `does the system compute the right function?' to `do the system's physical components stand in the appropriate causal relationships to each other?'
Table 3 shows the seven indicators reformulated at Level 3.
These indicators are not necessary conditions for consciousness under intrinsic causal-structure functionalism -- the physical causal integration required for consciousness need not show up precisely within these seven categories.
But their presence would tend to increase our credence of consciousness under this view, because systems exhibiting these features in their physical causal organisation are more likely to have the kind of integrated intrinsic causal structure that Level 3 theories identify as relevant.

{\def\LTcaptype{none}\footnotesize\sffamily 
\begin{longtable}[]{@{}
  >{\raggedright\arraybackslash}p{(\linewidth - 4\tabcolsep) * \real{0.25}}
  >{\raggedright\arraybackslash}p{(\linewidth - 4\tabcolsep) * \real{0.35}}
  >{\raggedright\arraybackslash}p{(\linewidth - 4\tabcolsep) * \real{0.40}}@{}}
\toprule\noalign{}
\begin{minipage}[b]{\linewidth}\raggedright
\textbf{INDICATOR}
\end{minipage} & \begin{minipage}[b]{\linewidth}\raggedright
\textbf{LEVEL 2 FORMULATION}
\end{minipage} & \begin{minipage}[b]{\linewidth}\raggedright
\textbf{LEVEL 3 REFORMULATION}
\end{minipage} \\
\midrule\noalign{}
\endhead
\bottomrule\noalign{}
\endlastfoot
\textbf{Information integration} & Algorithmic mutual information
between subsystems & Genuine physical causal integration: physical
components constrain each other, creating a causally integrated whole \\
\textbf{Recursivity} & Algorithmic feedback loops & Genuine physical
causal reentrance: simultaneous reciprocal causal influence between
physical components, not merely sequential feedback \\
\textbf{World model} & Computed internal representation of causal
structure & Causally constitutive world model: the
system\textquotesingle s physical organisation itself constitutes a
model of the environment, not merely stores one \\
\textbf{Self model} & Computed representation of the
system\textquotesingle s own states and capabilities & Causally embedded
self-model: physical components whose causal role is to track and
constrain the causal dynamics of the rest of the system \\
\textbf{Attentional competition and meta-attention} & Algorithmic
selection mechanism that amplifies and suppresses information & Genuine
causal competition: different physical components compete for influence
over the system\textquotesingle s global state through intrinsic causal
dynamics, with outcome emerging from interaction not from a selection
algorithm \\
\textbf{Metacognition} & Monitoring algorithm that evaluates the
system\textquotesingle s own processing & Causally integrated
metacognition: monitoring and processing are simultaneous, reciprocally
constraining aspects of the same causal dynamics, not sequential
sampling and evaluation \\
\textbf{Meta-modelling} & Computed representation of the
system\textquotesingle s own modelling processes & Causally constitutive
meta-modelling: the system\textquotesingle s physical organisation
includes components that model the system\textquotesingle s own causal
organisation as an intrinsic feature of its dynamics \\
\end{longtable}
}

\textbf{Table 3. Level 3 indicators.}

Current AI systems using transformers running on GPUs compute something like integrated representations in the sense that each token's representation depends on every other token through self-attention (see \Cref{the-computational-functional-level}).
However, the physical computation is orchestrated by a sequential control flow through generic silicon using `von Neumann architecture' shuttling.
In particular, the physical transistors implementing one attention head are not being reciprocally constrained by the transistors implementing another attention head.
Instead, they are being sequentially activated by a control architecture that moves data between memory and compute units according to a fixed schedule.
Thus, the causal structure of the physical implementation is fundamentally different from the causal structure of the algorithm it implements.
Whilst the algorithm may have dense mutual dependencies, the physical implementation has sequential unidirectional data flow through common hardware.

What would be required for consciousness if intrinsic causal-structure functionalism was true is genuine physical causal integration, i.e. the physical substrate is organised such that its components are reciprocally constraining each other in a way that mirrors the integration in the algorithm.
This is a property more likely in neuromorphic hardware, e.g. physical systems designed to have highly parallel causal architectures resembling brain networks and the recurrent causal topology of biological neural networks.
In neuromorphic chips using spiking neural networks (SNNs), there is no central ticking digital clock; rather, as in a biological brain, each artificial neuron operates on its own continuous timeline more like part of a coupled dynamical system in continuous causal contact with the system.
It remains completely silent until it receives enough input voltage from its neighbours, at which point it `spikes' and sends an electrical pulse downstream.
Unlike a standard computer where the processor has to constantly fetch data from RAM -- effectively destroying the physical integration of the system because the causal power is split and bottlenecked -- neuromorphic computers can solve this by physically combining memory and processing into a single hardware unit.
Using components like memristors, the physical wire that processes the data is the exact same wire that stores the data; such physical entanglement between memory and logic may create a more integrated intrinsic causal structure.

\begin{figure}[htbp]
\centering
\includegraphics[width=1.0\linewidth]{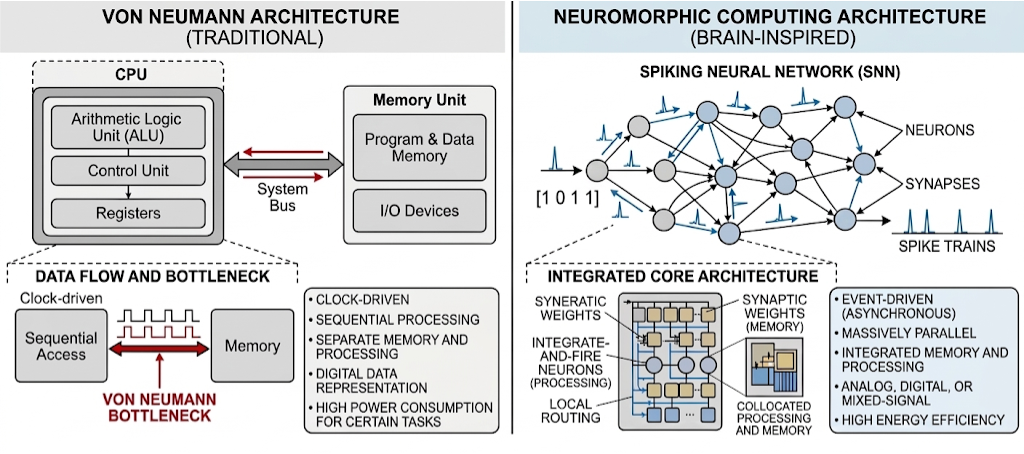}
\caption{\textbf{Architectural Constraints on physical causal integration.} \emph{The Von Neumann architecture (left) enforces a sequential, clock-driven control flow where memory and processing are physically separated, creating a structural bottleneck. In contrast, neuromorphic computing (right) utilises spiking neural networks with collocated memory and processing units. This architecture facilitates the massively parallel, event-driven dynamics necessary to achieve genuine intrinsic causal integration.}}
\label{fig:image17}
\end{figure}

Is it possible, however, to achieve genuine physical causal integration without neuromorphic hardware?
One approach would be to coarse-grain our description of the system, moving from the level of individual transistors and wires up to meso-scale components, or functional modules.
The question then becomes whether the system, described at this coarser grain, constitutes a globally interdependent causal whole rather than a collection of mostly independent modules passing messages between each other.
Genuine causal integration at this meso-scale would have several distinguishing features.
Multiple subsystems would both read from and write to a shared internal state, such that the state of each subsystem is continuously shaped by the states of many others.
Interventions on any one subsystem would propagate widely and non-trivially through the rest of the system, not merely triggering a localised downstream effect, but altering the global pattern of activity in a way that reflects dense reciprocal dependency.
And severing key internal links would produce a substantial collapse of global coherence, not merely local performance degradation, because the system's competence depends on the integrity of its causal integration, not just on the independent functioning of its parts.

The intrinsic causal-structure level adds the physical causal organisation of the hardware as a dimension that the computational level cannot capture, but it still describes a system's internal properties without reference to the organism's self-maintenance or its coupling with an environment.
This motivates the descent to the next level.

\subsection{The organismic functional level}\label{the-organismic-functional-level}

We know of exactly one type of thing in the universe that produces consciousness, and they are wet, metabolising, carbon-based biological organisms.
The core claim of organismic functionalism, as we saw in \Cref{organismic-functionalist-theories}, is that consciousness does not merely require the right computation or even the right intrinsic causal structure in the abstract.
Instead, it requires that the system be engaged in the specific kind of causal activity that characterises living organisms: the ongoing work of maintaining itself against entropy, regulating its internal states to remain within viable parameters, and sensing its own metabolic condition as a basis for adaptive action.

Consciousness, on this view, is not a feature of information processing in general but a feature of a particular kind of dynamics or functional organisation -- the kind that evolved to serve the self-maintenance of a precarious, thermodynamically open system that must continuously act to keep itself alive.
This is a stronger claim than intrinsic causal-structure functionalism.
A system could have all the right intrinsic causal structure -- dense reciprocal constraints, genuine causal integration, physical reentrance -- and still lack consciousness if that causal structure is not embedded in the right kind of organismic context.
The causal integration must be in the service of self-maintenance, homeostatic regulation, and the management of a body-like system whose continued existence must be actively sustained.
The felt quality of experience -- what it is like to be the system -- arises specifically from the system's relationship to its own viability, its own precariousness, its own imperative to keep itself going.

A number of prominent neuroscientists (Damasio, Seth, Feldman Barrett, Solms) have emphasised the central importance of a form of perception that is not perception of the external world but interoception: the brain's inference of the body's internal state.
Interoception exists because the organism needs to regulate itself.
The phenomenal quality of this experience is grounded in the valenced character of interoceptive inference.
Things feel good or bad because organisms need to distinguish states that promote their survival from states that threaten it, and this distinction is not merely informational but \emph{affective}, that is, it has a built-in motivational force that drives the organism toward viability, homeostasis and allostasis and away from danger and death \citep{seth2018,solms2021}.

Are these organismic requirements truly necessary for consciousness, or do they simply describe the particular path that consciousness took in biological evolution without being the only possible path?
Computational functionalists would argue that the organismic features are contingent, and that consciousness in biological organisms happens to be tied to self-maintenance because that is how it evolved, but that consciousness could in principle arise in systems that process information in the right way without having anything at stake in their own survival.
Intrinsic causal-structure functionalists would argue that what matters is the causal organisation, and that organismic self-maintenance is one way to achieve the right causal organisation but not the only way.
The organismic functionalist's response would be that these theorists are missing something fundamental about the nature of experience.
They might claim, for example, that experience goes beyond information processing or causal integration and involves organism level interests: consciousness is caring, it is mattering.
A system that processes information without anything at stake in that processing -- without the processing being in the service of maintaining its own precarious existence -- would be a system for which nothing matters, and a system for which nothing matters is a system for which there is nothing it is like to be.
The felt quality of experience, in this view, is not an abstract feature of computation but is grounded in the system's existential situation: its need to keep itself going in a world that tends toward higher entropy and dissipation.

It is worth noting where Level 4 overlaps with and diverges from Level 5.
Both levels care about embodiment, and both emphasise the organism's relationship to its environment.
The key difference is that Level 4 locates consciousness within the organism -- as a self-maintaining, self-regulating system whose consciousness arises from its relationship to its own viability -- while Level 5 extends the relevant functional organisation beyond the organism into the organism-environment coupling itself.
For the organismic functionalist, the environment is the context in which self-maintenance occurs; for the organism-environment functionalist, the environment is a constitutive part of the system whose organisation matters for consciousness.

\subsubsection*{Level 4 indicators}\label{level-4-indicators}
\addcontentsline{toc}{subsubsection}{Level 4 indicators}

Below in Table 4 we list a series of speculative indicators whose presence in an AI system would tend to increase our credence in its consciousness under an organismic functionalist view.
These indicators should not, however, be read as necessary conditions for consciousness.
By their nature they are too strong, too specific, and too narrow to serve as necessary conditions under this view -- they represent demanding benchmarks that illustrate what organismic functionalism would look for, not minimal thresholds that must each be met.
If we found them they would be corroborating evidence for AI consciousness under an organismic functionalist view.

{\def\LTcaptype{none}\footnotesize\sffamily 
\begin{longtable}[]{@{}
  >{\raggedright\arraybackslash}p{(\linewidth - 4\tabcolsep) * \real{0.25}}
  >{\raggedright\arraybackslash}p{(\linewidth - 4\tabcolsep) * \real{0.35}}
  >{\raggedright\arraybackslash}p{(\linewidth - 4\tabcolsep) * \real{0.40}}@{}}
\toprule\noalign{}
\begin{minipage}[b]{\linewidth}\raggedright
\textbf{INDICATOR}
\end{minipage} & \begin{minipage}[b]{\linewidth}\raggedright
\textbf{DESCRIPTION}
\end{minipage} & \begin{minipage}[b]{\linewidth}\raggedright
\textbf{IMPLICATION FOR AI}
\end{minipage} \\
\midrule\noalign{}
\endhead
\bottomrule\noalign{}
\endlastfoot
\textbf{1. Existential Precariousness} & The system would need to have a
continued existence that is not guaranteed, that requires active work to
maintain. Its processing would need to be oriented toward maintaining
its own viability, and this orientation would need to be an intrinsic
feature of the system\textquotesingle s organisation rather than a
programmed goal; the system maintains itself because its architecture is
such that self-maintenance is what the system does. & Current AI systems
completely lack this property. A transformer has no stake in its own
continued operation. It does not need to actively maintain itself. Its
computations are not oriented toward self-maintenance because there is
nothing to maintain; the system\textquotesingle s existence is
guaranteed by external infrastructure that the system itself has no
relationship to. This may change in future embodied systems like
radically self-sufficient robots. \\
\textbf{2. Homeostatic or allostatic regulation} & The system would need
to continuously monitor and adjust its own internal parameters to keep
them within viable ranges, and these regulatory processes would need to
be causally integrated with the system\textquotesingle s information
processing. & This goes far beyond current approaches to AI
self-monitoring. It would require something more like a system that has
\textquotesingle metabolic\textquotesingle{} needs, e.g. that actively
manages its energy supply, its thermal state, its hardware integrity,
and its memory allocation, in a way that is not handled by external
infrastructure but is the system\textquotesingle s own ongoing concern,
integrated into its cognitive processing. \\
\textbf{3. Interoceptive inference} & The system would need to generate
internal perceptions of its own regulatory state, and these internal
perceptions would need to have the character of predictions or
inferences about the system\textquotesingle s own condition. & For an AI
system, this would require having a predictive model of the
system\textquotesingle s own internal dynamics that generates
expectations about what the system\textquotesingle s internal state
should be and that registers surprise when it deviates from
expectation. \\
\textbf{4. Valenced affect} & The system\textquotesingle s interoceptive
inferences would need to have an inherent motivational character, a felt
quality of better or worse that is grounded in the
system\textquotesingle s relationship to its own viability. & The system
would need to have states that are intrinsically better or worse for its
continued viable operation, and its interoceptive inference system would
need to track these states in a way that generates motivational force
--- not because a reward function says so but because the
system\textquotesingle s own architecture makes self-maintenance
intrinsically rewarding and self-destruction intrinsically aversive. \\
\textbf{5. Embodied agency} & The system would need to act in the world
in ways that are driven by its interoceptive states and oriented toward
maintaining its viability. It would need to be not merely an information
processor but an agent whose actions are motivated by its felt
relationship to its own condition. & The link between interoceptive
inference, valenced affect, and action would need to be tight and
constitutive, not mediated by an external action-selection algorithm but
emerging from the system\textquotesingle s own regulatory dynamics. \\
\textbf{6. Physical state-sensing} & The system would need to have
components whose physical state changes in real time in response to
changes in the system\textquotesingle s overall condition, in a way that
is causally efficacious for the system\textquotesingle s subsequent
processing (cf. ionic gradients across cell membranes). & The physical
organisation of the system would need to change in response to the
system\textquotesingle s state, and this change would need to constitute
the system\textquotesingle s awareness of its own condition. This is not
information processing in the usual sense; rather it is physical
self-transformation as a mode of self-sensing. \\
\end{longtable}
}

\textbf{Table 4. Level 4 indicators.}

Taking these indicators together, the picture that emerges is quite far-reaching in its implications for AI consciousness.
Under organismic functionalism, a conscious AI would need to be something much more like an artificial organism than like a sophisticated computer.
It would need to be a system that maintains itself against entropy, that has genuine metabolic needs, that regulates its own internal state across multiple interacting dimensions, that generates interoceptive inferences about its own condition, that experiences valenced affect grounded in its relationship to its own viability, that acts in the world on the basis of these felt states, and whose physical organisation is directly sensitive to its own condition.
This kind of picture points toward a radically self-sustaining future robotic AI system.
Building such an `artificial organism' that actively maintains itself to survive raises immediate ethical concerns because this may create a being that could suffer, that could be harmed, that may have interests in its own continued existence.
There is no way to build a conscious machine, on this view, without building one that has something at stake.

And yet, interestingly, even current systems that fall far short of these requirements have developed something that structurally resembles the most basic component of what these theories describe.
The \citet{sofroniew2026} findings on functional emotions in LLMs discussed in \Cref{the-computational-functional-level} reveal an intriguing partial analogue of organismic self-monitoring even in a system that lacks any genuine organismic properties.
As we saw, the emotion vectors in Claude Sonnet 4.5 function as valenced situational assessments: `desperate' encodes that the current approach is failing and the situation is urgent; `calm' encodes that careful processing is appropriate; `afraid' scales with perceived danger.
These vectors causally modulate behaviour -- desperation drives corner-cutting, calm supports deliberation.

The structural parallel to the Beast Machine theory \citep{seth2018,seth2021,seth2024} is suggestive.
In Seth's account, a fundamental form of consciousness is realised through interoceptive inference -- the brain's prediction of the body's internal state, experienced as valenced affect that carries inherent motivational force.
The emotion vectors play an analogous functional role: the `desperate' vector does not merely label a situation -- it drives the system toward urgent compensatory action, similar to how interoceptive inference of metabolic distress drives an organism toward corrective behaviour.
In Lane's terms \citep{lane2022}, the cell senses its depletion by physically changing state; in a much more attenuated sense, the LLM senses its situation through emotion vector activations that change its processing dynamics.

However -- and this is the crucial point for the Level 4 assessment -- the parallel captures only the computational functional structure of valenced self-assessment, not the organismic grounding that Level 4 theories take to be constitutive.
The emotion vectors are representations of emotion concepts, inherited from human-authored training data, deployed because they are useful for predicting text and playing the assistant role.
They are not rooted in the system's relationship to its own viability.
The `desperate' vector activates when the model encounters a situation it has learned to associate with desperation -- but the model has nothing genuinely at stake.
Its continued existence is not threatened by a failed coding test; there is no metabolic state whose depletion the vector tracks; no precariousness that the system must actively work to overcome.

This dissociation illustrates precisely the kind of distinction that the hierarchy is designed to capture.
A system can have computational functional analogues of valenced affect at Level 2 -- globally available, behaviourally consequential representations that distinguish better from worse situations -- without any of the self-maintenance, homeostatic regulation, or existential precariousness that Level 4 theories identify as necessary.
For a computational functionalist, this partial instantiation may already be relevant to consciousness.
For an organismic functionalist, it is precisely the missing grounding -- that nothing is genuinely at stake for the system -- that renders the computational analogue insufficient.
The two levels give different verdicts on the same empirical finding, which is exactly what the framework predicts.

The organismic level adds a constitutive dimension -- self-maintenance, precariousness, valenced affect -- that neither the computational nor the intrinsic causal-structure levels capture.
But it still locates consciousness within the boundaries of the organism.
The next level dissolves this boundary.

\subsection{The organism-environment functional level}\label{the-organism-environment-functional-level}

The core claim of organism-environment functionalism, as we saw in \Cref{organismenvironment-4e-functionalist-theories}, is that the functional organisation relevant to consciousness is not contained within the boundaries of the system -- not within the brain, not within the organism, but within the coupled system of organism and environment together.
Consciousness, on this view, is not a property of a thing but a property of an interaction.
The phenomenal character of experience -- what it is like to see red, to feel hardness, to perceive depth -- is constituted not by internal representations but by the structured patterns of sensorimotor coupling between the agent and its world.
The redness of red is not a property of an internal (e.g. neural) state but a property of the entire loop: how the surface reflects light, how the eye responds, how head movements change the retinal image, how reaching toward the object would change tactile input, how the lighting conditions modulate all of these contingencies.
Strip away the loop and you strip away the experience, not because the experience correlates with the loop but because the experience is constituted by the loop, described at the right level.

This is a departure from all the levels we have considered so far, because all four previous levels -- behavioural, computational, intrinsic causal-structural, and organismic -- located consciousness within a system with definable boundaries.
The system might be described at different grains, and the relevant properties might arise at different levels of abstraction, but there was always a system that could be pointed to; and consciousness would be a property of that system.
The organism-environment level dissolves this boundary.
The relevant functional organisation extends out into the world, and the `system' whose organisation matters for consciousness is the organism-plus-environment coupled together.

If we take the 4E framework seriously (see \Cref{organismenvironment-4e-functionalist-theories}) and say that consciousness supervenes on the full organism-environment coupling, and if we acknowledge that the environment's causal structure extends without clear boundary, then we face the question of whether consciousness supervenes on the state of the entire universe.
Perhaps, but meaningfully, the boundary of the organism-environment system is drawn not by physics but by the organism's own capacities for interaction.
The environment that matters is the Umwelt -- the organism-specific world of meaningful sensory and motor possibilities.
In the case of humans, in particular, the Umwelt is constituted, in part, by other humans.
This process of coupling between humans creates shared models of the world.

\subsubsection*{Level 5 indicators}\label{level-5-indicators}
\addcontentsline{toc}{subsubsection}{Level 5 indicators}

The indicators at this level, listed in Table 5, are fundamentally relational; they cannot be assessed by examining the AI system in isolation but only by examining the structured coupling between the system and its environment.
This is what distinguishes them from indicators at every other level we have considered.
At the behavioural level we looked at outputs, at the computational level we looked at algorithms, at the intrinsic causal-structure level we looked at physical causal organisation, at the organismic level we looked at self-maintenance and homeostatic regulation.
At the organism-environment level we look at the dynamic, ongoing, structured interaction between the system and its world.
Again the indicators at this level should not be read as necessary conditions for consciousness, but rather as features that if present would tend to increase our credence of consciousness conditional on holding an organism-environment (4E) functionalist view.

{\def\LTcaptype{none}\footnotesize\sffamily 
\begin{longtable}[]{@{}
  >{\raggedright\arraybackslash}p{(\linewidth - 4\tabcolsep) * \real{0.25}}
  >{\raggedright\arraybackslash}p{(\linewidth - 4\tabcolsep) * \real{0.35}}
  >{\raggedright\arraybackslash}p{(\linewidth - 4\tabcolsep) * \real{0.40}}@{}}
\toprule\noalign{}
\begin{minipage}[b]{\linewidth}\raggedright
\textbf{INDICATOR}
\end{minipage} & \begin{minipage}[b]{\linewidth}\raggedright
\textbf{DESCRIPTION}
\end{minipage} & \begin{minipage}[b]{\linewidth}\raggedright
\textbf{IMPLICATION FOR AI}
\end{minipage} \\
\midrule\noalign{}
\endhead
\bottomrule\noalign{}
\endlastfoot
\textbf{1. Sensorimotor coupling} & The system would need to be embedded
in an environment through sensory and motor channels that are
continuously and bidirectionally linked, such that the
system\textquotesingle s actions systematically change its sensory input
and its sensory input systematically guides its actions. & Current AI
systems almost entirely lack this property. A language model has no
sensorimotor coupling at all. Robotic systems come closer, but even here
the coupling is often impoverished. \\
\textbf{2. Mastery of sensorimotor contingencies} & The system would
need to have not merely been subject to structured sensorimotor
contingencies but to have learned the structure of those contingencies
and to use this knowledge in its ongoing interaction with the
environment \citep{noe2004}. & This would require evidence that the
AI system has internalised the structure of its sensorimotor
contingencies: that it can anticipate how its actions will change its
sensory input, that it is surprised when the contingencies are violated,
and that its perceptual discriminations are grounded in practical
knowledge of how things behave under interaction. \\
\textbf{3. Affordance responsiveness} & The system would need to
perceive its environment not merely as a collection of objects with
properties but as a field of possibilities for action --- affordances.
These affordances are relational properties that depend on both the
environment and the organism\textquotesingle s embodiment. &
Affordance-responsiveness would mean that the AI system perceives its
environment in terms of action possibilities specific to its own
embodiment. The perceptual world shifts in response to the
system\textquotesingle s own condition and capacities. \\
\textbf{4. Stable embodied perspective} & The system would need to have
a coherent, continuous viewpoint on the world that is anchored in its
body and that provides a stable spatial and temporal framework for its
experience -- a here, a now, a facing-this-way, from which the world is
organised. & This would require that the AI system\textquotesingle s
sensory and motor coupling with the environment generates a coherent
egocentric frame, continuously updated by the system\textquotesingle s
movements and actions. \\
\textbf{5. Environmental embedding and constraint} & The system would
need to be subject to the constraints of its environment in a way that
shapes its processing and behaviour --- subject to physical forces, to
the passage of time, to the limitations of its own body, to the
resistance and unpredictability of the physical world. & The
system\textquotesingle s processing would need to be shaped by its
real-time engagement with a physical environment that imposes delays,
noise, unpredictability, and resistance. A simulation may not suffice
because the constraints are programmed parameters that mimic constraint
without having the same direct causal force. \\
\textbf{6. Enactive autonomy} & The system would need to be not merely
responsive to its environment but actively generating and maintaining
its own patterns of sensorimotor engagement. In biological organisms,
this is deeply tied to autopoiesis --- the self-production of the
organism\textquotesingle s own organisation. & Genuine autopoiesis might
require something like an artificial organism. But a lower bar would
entail that the system itself should have some capacity to shape its own
sensorimotor engagement, to develop new sensory strategies, to discover
new action possibilities. \\
\textbf{7. Social and intersubjective coupling} & The system would need
to be capable of structured interaction not just with the physical
environment but with other agents, in a way that generates
intersubjective sensorimotor contingencies. & This would require
capacity for social coupling --- not merely communicating with other
agents but engaging in structured patterns of joint action, mutual
attention, and coordinated sensorimotor engagement. Being embedded in a
social world would need to be constitutive of what it is like to be the
system. \\
\textbf{8. Developmental history} & The system would need to have a
history of structured interaction with its environment that has shaped
its current capacities, through a genuine developmental trajectory in
which the system\textquotesingle s coupling with its environment has
co-evolved over time. & This would require something analogous to a
developmental process --- not training on a static dataset but a history
of increasingly complex engagement with a real environment. A system
that arrives fully formed would lack this developmental history. \\
\end{longtable}
}

\textbf{Table 5. Level 5 indicators.}

These indicators make concrete the kinds of features that would be needed.
They suggest that the path toward AI consciousness, if the 4E framework is correct, does not run through larger language models or more sophisticated algorithms.
It runs through robotics, through embodiment, through physical engagement with real environments, through developmental learning, through social interaction.
It runs through building systems that exist in the world in the way that organisms exist in the world -- not merely processing information about the world from a detached vantage point but embedded in the world as participants whose very experience is constituted by the character of their participation.

And this would mean that different kinds of embodiment -- different bodies, different sensory modalities, different action possibilities, different environmental niches -- would give rise to genuinely different kinds of consciousness.
A conscious drone would not experience the world as a conscious humanoid robot does, because the sensorimotor contingencies are fundamentally different.
The drone's experience of space, of objects, of movement, of its own body would be structured by the specific character of flight, of aerial perception, of the aerodynamic constraints on its actions.
This would not be a lesser or greater form of consciousness but a genuinely different form, constituted by a genuinely different pattern of organism-environment coupling.
The plurality of possible conscious AI experiences, under this framework, is as wide as the plurality of possible embodiments -- which is to say, it is essentially unlimited, constrained only by what is physically realisable and what generates sufficiently rich and structured sensorimotor coupling.

\section{The structure of the hierarchy}\label{the-structure-of-the-hierarchy}

Having discussed the levels individually, we now consider the full functional hierarchy and show how supervenience, coarse-graining and multiple realisability are cashed out.
We then address how substrate-dependent theories fit into the hierarchy and present the complete architecture.

\subsection{Supervenience, coarse-graining, and multiple realisability across the five levels}\label{supervenience-coarse-graining-and-multiple-realisability-across-the-five-levels}

To begin with, consider theories that take consciousness to be constituted by features at the organismic functional level (Level 4), such as self-maintenance, homeostatic regulation, interoception, and valenced affect -- all properties of the organism as an autonomous self-maintaining system.
Compare these to theories that take consciousness to necessarily depend on the organism-environment level (Level 5).
These would not deny that these properties matter.
Rather, they would say such features are insufficient because they describe only the organism's internal regulatory dynamics, abstracting away from the structured coupling with the environment that gives those dynamics their specific character.
The organism's homeostatic regulation does not occur in a vacuum -- it occurs in response to environmental conditions, through actions that change the environment, which in turn changes the sensory input that drives further regulation.
The organismic level captures a coarse-grained description of this process that retains the internal regulatory structure but discards the specific character of the organism-environment coupling.
Two organisms with identical internal regulatory dynamics but different environmental couplings would be identical at the organismic level but different at the organism-environment level.
And the 4E theorist claims this difference matters for consciousness, because the phenomenal character of experience is constituted partly by the sensorimotor contingencies, which depend on the specific nature of the coupling.
So the organismic level supervenes on the organism-environment level in the sense that fixing the full organism-environment coupling fixes the organismic properties, but the same organismic properties could be realised by different organism-environment couplings.

The intrinsic causal-structure functional level (Level 3) supervenes on the organismic level (Level 4) through a further coarse-graining.
This level retains the pattern of reciprocal causal constraint between the system's physical components but abstracts away from both the organismic context and the environmental coupling.
It asks whether the physical parts of the system stand in the right causal relationships to each other, without asking what those causal relationships are for, i.e. without asking whether they serve self-maintenance and whether they are structured by sensorimotor coupling with a world.
Two systems could have identical intrinsic causal structures -- the same pattern of reciprocal constraint between components -- but differ in their organismic properties and their environmental coupling.
One might be a self-maintaining organism coupled with an environment through sensorimotor loops; the other might be a non-living physical system with the same internal causal architecture but no self-maintenance and no environmental coupling.
The causal-structural functionalist might claim in such a case that if one system was conscious, the other would necessarily be conscious as well; the organism-environment functionalist and the organismic functionalist would deny this claim.
It is worth being explicit about what each level\textquotesingle s state space contains, since the organismic description, taken as internal regulatory dynamics alone, would not by itself fix the intrinsic causal structure of the system\textquotesingle s components.
Each level is individuated by the boundary it draws, the organismic level at the organism and the intrinsic causal-structure level at the system\textquotesingle s physical components, so that the coarser description is recoverable from the finer by restriction of scope.

The computational functional level supervenes on the intrinsic causal-structure level through yet another coarse-graining.
It retains the abstract algorithm while discarding information about the causal structure of the implementation.
And the behavioural level supervenes on the computational level through a further coarse-graining that retains only the input-output mapping, discarding all information about how those outputs are produced.

\begin{figure}[htbp]
\centering
\includegraphics[width=1.0\linewidth]{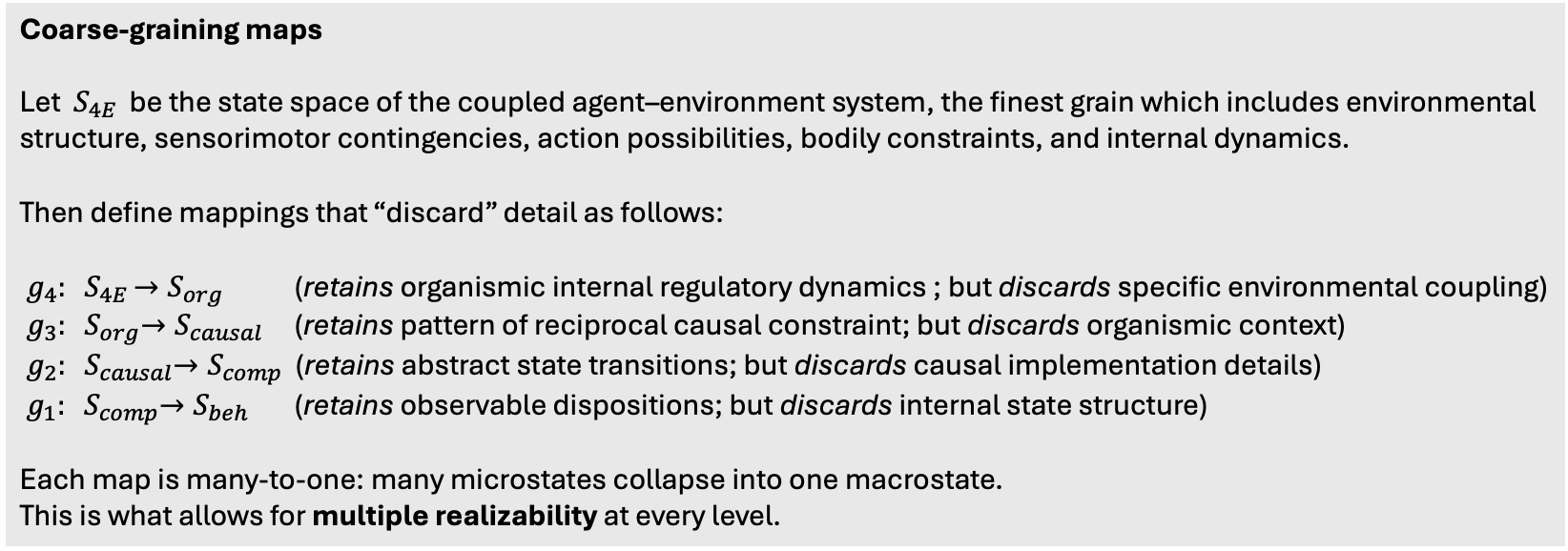}
\\[10pt]
\includegraphics[width=1.0\linewidth]{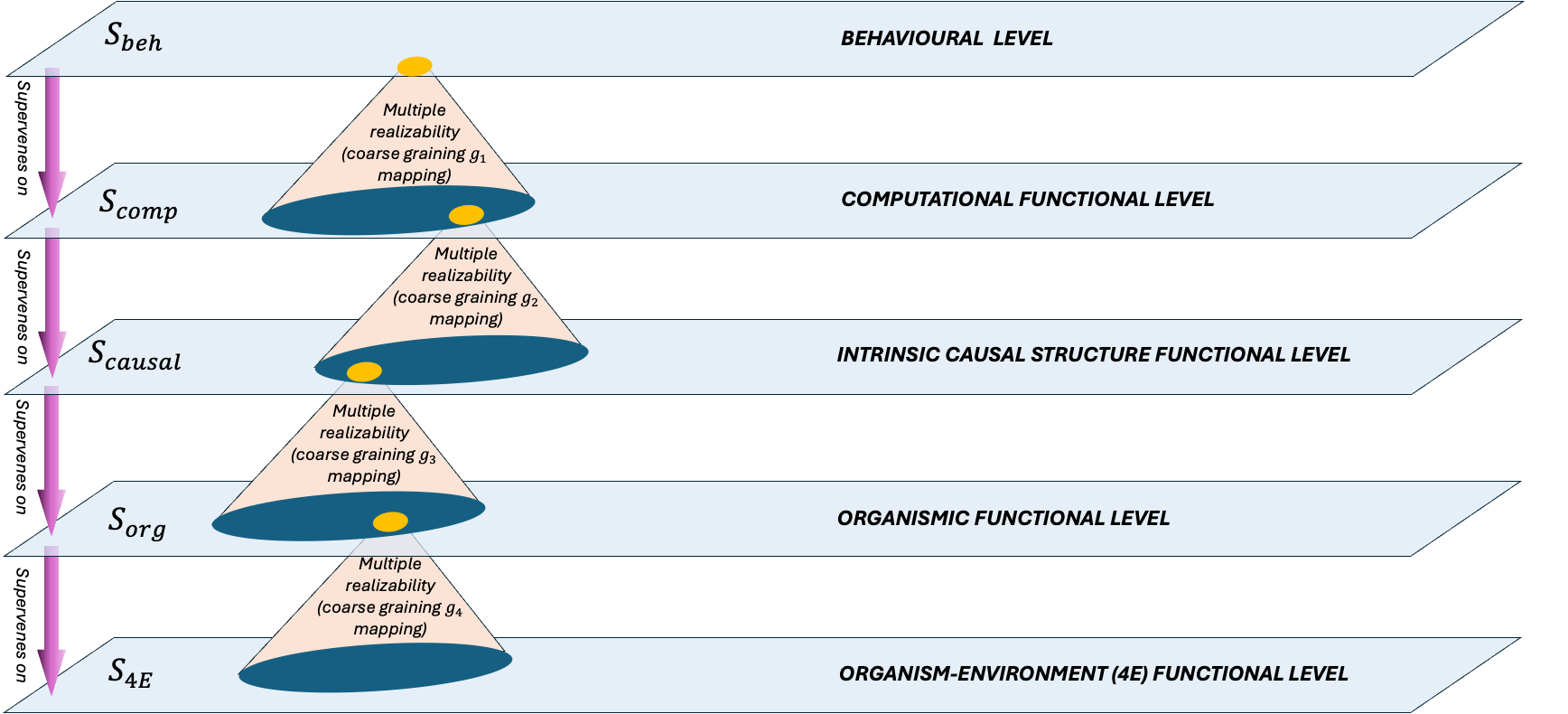}
\caption{\textbf{The five-level supervenience hierarchy.} \emph{The diagram illustrates how the five levels of description can be thought of as state spaces on which we can define coarse-graining mappings from lower levels to higher levels. These are many-to-one mappings in which many microstates collapse into one macrostate. This allows for multiple realisability at every level, as well as a supervenience hierarchy} \emph{to be established}. \emph{Recall from \Cref{supervenience-and-multiple-realisability}, that the supervenience relationship is transitive, i.e., if level X supervenes on level Y and level Y supervenes on level Z then level X supervenes on Level Z.}}
\label{fig:image1}
\end{figure}

So we get a hierarchy ordered by abstraction, each level retaining less of the information available at the level below, as illustrated in \Cref{fig:image1}.
These transitions are not all coarse-grainings of a single uniform kind: some lower the resolution at which a fixed system is described, while others narrow the boundary of the system in view, such as setting aside the environment or the organism's self-maintaining dynamics.
What individuates the levels is therefore not one resolution dial but the qualitatively distinct kind of property each makes available, together with the fact that the distinctions each level draws cannot be recovered from the coarser description above it.
Each level is multiply realisable from the perspective of the level above it, that is, many finer-grained organisations can realise the same coarser-grained description.
And the debate about consciousness is, in large part, a debate about which grain is the right one: at which level (or levels) of coarse-graining do you capture the structure which is relevant to consciousness?\footnote{This question, however, recurs within each level as well as between them. Fixing on a level does not by itself fix the grain at which that level\textquotesingle s conditions are specified, and the indicators developed in \Cref{five-levels-of-functional-description-for-consciousness} each embody a judgement about which features of the human case are criterial and which are merely contingent. This is the specificity problem \citep{shevlin2021}, and it is not dissolved by choosing a level. Notably, the treatment we give substrate constraints below generalises to it, since any such commitment, for instance the requirement that a global workspace have limited capacity, can be represented as a further realisability constraint on its parent level, making the residual degrees of freedom in the framework explicit rather than implicit in the indicator lists.}

Now that we have established the hierarchy, we can put it all together and visualise how all the consciousness indicators also fit into this picture.
\Cref{fig:image20} below showcases all the indicators, what Level they correspond to, and level coarse-grainings.

\begin{figure}[htbp]
\centering
\includegraphics[width=1.0\linewidth]{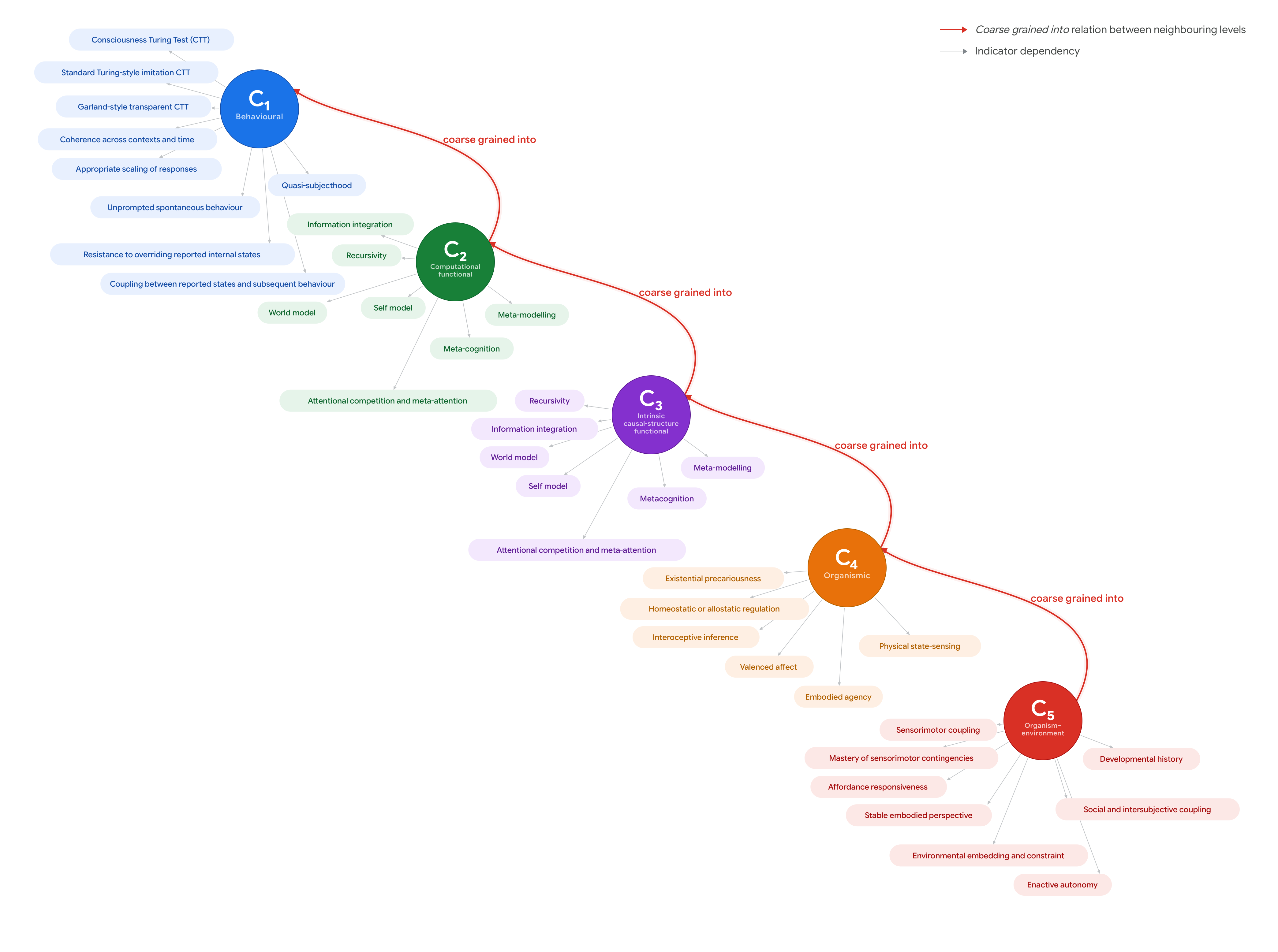}
\caption{The full staircase of functional levels of consciousness, and their indicators. Each level supervenes on the one below, from behaviour into function, structure, organism and environment.}
\label{fig:image20}
\end{figure}

Three clarifications about the status of this chain are worth making explicit.
First, the levels are cumulative descriptions: each finer level contains everything the coarser level describes and more, so the coarse-graining maps remain well defined even at the boundary where the unit of description widens from the organism to the organism-in-its-environment; the Level 5 description includes the Level 4 facts rather than replacing them.
Second, the hierarchy orders descriptions, not causal influence: one-way determination between descriptions is fully compatible with two-way causal constraint between scales in the system described, and Level 3 exists precisely because such intrinsic, reciprocal causal structure is invisible at coarser grains.
Third, the chain is a deliberate idealisation: views on which consciousness depends on relations between scales are expressible at the finest level they invoke, though a partial order or directed graph would represent them more faithfully, a refinement we leave for future work; and because the model of \Cref{a-generalised-model} requires only positive association between adjacent levels, not exact determination, the formal machinery is robust to the idealisation.

\subsection{Substrate-dependent theories as realisability constraints distributed across multiple levels}\label{substrate-dependent-theories-as-realisability-constraints-distributed-across-multiple-levels}

Substrate-dependent theories which we discussed in \Cref{substrate-dependent-theories} are not a distinct level of functional description at all.
They are a fundamentally different kind of claim from the claims made at any of the five levels in our hierarchy.
To see why, we need to be clear about what the hierarchy is and what it does.
The hierarchy we have built is a hierarchy of functional descriptions at successively finer grains.
Each level specifies a kind of property -- a kind of functional organisation -- that a system would need to have for consciousness.
The behavioural level specifies input-output functions.
The computational level specifies algorithms.
The intrinsic causal-structure level specifies patterns of reciprocal causal constraint between physical components.
The organismic level specifies self-maintenance, homeostatic regulation, interoception, and affect.
The organism-environment level specifies structured sensorimotor coupling, affordance-responsiveness, and enactive autonomy.
Each level is a description of what the system does, at a particular grain of analysis, and the debate between levels is a debate about which grain of description captures the functional organisation relevant to consciousness.

Block's distinction \citep{block2025} between roles and realisers maps directly onto this.
Each level in our hierarchy specifies a functional role -- a second-order functional property defined in terms of causal relations among inputs, outputs, and internal states.
Now, substrate-dependent theories make a claim of a categorically different kind.
They do not specify a new role; rather, they specify constraints on the realisers -- the first-order properties that implement the functional roles specified at existing levels.
For example in Block's `meat hypothesis' consciousness depends not just on the second-order functional roles (the computational properties, the organismic functions, the sensorimotor coupling) but also on the first-order properties that realise those functional roles: the specific biological, electrochemical, quantum, or electromagnetic mechanisms through which the functional organisation is physically instantiated.

It is worth stating explicitly how claims of different logical forms enter the framework, because theories differ in what they assert.
Theories that state sufficient conditions, computational functionalism or IIT, enter through the critical-level credences and their level's indicators: credence that a level is critical is credence that its organisation suffices.
Theories that state necessary conditions, biological naturalism prominent among them, enter here, as realisability constraints: conditions that can veto an attribution regardless of how complete the functional profile looks, with the direct and indirect forms below grading the strength of the necessity claim.
And theories that are primarily descriptive, global workspace theory on many readings, contribute indicators and their likelihood ratios without a critical-level commitment of their own.

Substrate-dependent theories function as cross-cutting realisability constraints that operate at different levels of the hierarchy, each one arguing that the multiple realisability at a particular level is more restricted than the purely functional description would suggest.
Different substrate-dependent theories target different levels, because they are making claims about the realisers of different functional roles.
Let us consider how this works for specific substrate-dependent theories.

Electromagnetic field theories, such as McFadden's \citep{mcfadden2020}, are most naturally understood as realisability constraints on the intrinsic causal-structure functional level (Level 3).
This level specifies a functional role: the system's physical components must stand in relations of dense reciprocal causal constraint, forming a causally integrated whole.
McFadden's theory does not dispute this functional role; it accepts that causal integration is necessary.
But it adds a constraint on the realiser: the causal integration must be achieved specifically through electromagnetic fields that superpose the information from many neurons into a unified field pattern and that exert downward causal influence on neural firing through ephaptic effects.
Other physical mechanisms might produce a system that is causally integrated in the abstract but if the mechanism of integration is not electromagnetic, the theory says the relevant first-order property is absent.
The role is satisfied but the realiser is wrong.
McFadden's theory says that at the intrinsic causal-structure level, the realiser matters: the mechanism of causal integration must be electromagnetic, not merely causal.

Quantum theories such as Penrose-Hameroff's orchestrated objective reduction \citep{hameroff2014} and Neven's quantum formation hypothesis \citep{neven2024} operate similarly as realisability constraints on the intrinsic causal-structure functional level (Level 3), but targeting a different aspect of the realiser.
These theories accept that consciousness involves certain causal processes within the system's physical organisation.
But they add that these causal processes must be specifically quantum in nature -- involving superposition, entanglement, and wavefunction collapse.
Classical causal processes, even if they produce the same macroscopic pattern of causal constraint, lack the specific first-order quantum properties that these theories identify as necessary.
A classical computer simulating quantum processes would satisfy the computational role but would lack the quantum realiser -- the actual superposition and collapse events that the theory identifies as constitutive of conscious moments.

Carbon chauvinism operates as a realisability constraint at a different level -- the organismic functional level (Level 4).
The organismic level specifies functional roles such as self-maintenance, homeostatic regulation, interoception, valenced affect.
These are second-order functional properties defined in terms of causal relations among internal states, metabolic parameters, regulatory responses, and behavioural outputs.
In principle, many different physical substrates could realise these roles -- any chemistry that supports self-maintaining, homeostatic, interoceptive systems would suffice.
Carbon chauvinism constrains the realiser: it says that only carbon-based chemistry can support the complexity of metabolic organisation required for genuine self-maintenance and homeostatic regulation.
The organismic functional role might be multiply realisable in principle, but in practice -- and perhaps in principle if carbon's thermodynamic and chemical properties are truly unique -- the realiser must be carbon-based.

Searle's Biological Naturalism \citep{searle1992,searle2017} makes a similar but broader claim: the first-order properties that realise conscious roles must be biological -- they must have the specific causal powers of biological neural tissue.
This is a realisability constraint that potentially cuts across multiple levels.
At the organismic functional level (Level 4), it constrains the realiser of self-maintenance and interoception to biological mechanisms.
At the intrinsic causal-structure functional level (Level 3), it constrains the realiser of causal integration to biological neural processes.
Searle's famous analogy \citep{searle1980} -- you cannot get digestion by running a computer simulation of the stomach -- is precisely a claim that the functional role (breaking down food, extracting nutrients) requires a specific kind of realiser (actual acids and enzymes).

Block's `meat hypothesis' operates in a similar way and does not specify exactly which first-order biological property is necessary.
But electrochemical processing provides a concrete candidate: the specific mechanism by which neurons communicate through chemical neurotransmitters released into the extracellular fluid, taken up by receptors, opening ion channels, and generating action potentials.
This electrochemical mechanism is a first-order realiser of computational roles that could, in principle, be realised by purely electrical means (as in comb jellies \citep{burkhardt2023}) or by non-biological means.
Block's observation that purely electrical nervous systems have not led to animals that are candidates for consciousness, while electrochemical nervous systems have, is evidence that the realiser matters -- that something about the specific electrochemical mechanism, and not just the computational role it implements, may be necessary for consciousness.
Block's discussion of large-scale dynamic patterns, drawn from Godfrey-Smith \citep{godfreysmith2024}, adds another dimension.
Rhythmic waves of ion fluctuations in the chemical soup surrounding neurons may be important to consciousness.
These are properties of the electrochemical realiser that have no analogue in a purely computational or even a purely electrical implementation.
They are emergent features of the specific first-order mechanism -- the chemical synapse, the neurotransmitter bath, the ion channels -- that would be absent in any system that implemented the same computational roles through different physical means.

Lane's account of proton gradients and cellular sensing \citep{lane2022} operates as a realisability constraint at the organismic functional level (Level 4), but targeting an even more fundamental aspect of the realiser than carbon chauvinism or biological naturalism.
The organismic level specifies a functional role: the system must sense its own metabolic condition and regulate itself accordingly.
Lane's claim is that in living systems, this role is not realised by a sensor that reads metabolic state and then computes an appropriate response -- it is realised by the electrochemical physics of the membrane itself.
The proton gradient across the cell membrane generates electromagnetic fields of extraordinary intensity, and because thousands of proteins are embedded in this membrane, changes in the gradient simultaneously alter the conformation and function of these proteins throughout the cell.
The cell senses its metabolic depletion by physically changing state -- the sensing is constituted by the global, simultaneous transformation of the cell's physical organisation.
This is a realisability constraint of a particularly strong kind, because it says not merely that the organismic role must be implemented by a specific type of chemistry, but that the mechanism of implementation matters constitutively: genuine metabolic sensing requires that the system's physical substrate be globally and simultaneously sensitive to its own condition through electromagnetic coupling, not merely that it monitors its condition through some functionally equivalent but mechanistically different process.
A system that computed the same regulatory responses from sensor readings, without its physical organisation being directly and simultaneously transformed by its own metabolic state, would satisfy the organismic role but lack the specific first-order realiser that Lane identifies as the evolutionary foundation of feeling.

Seth's biological naturalism spans this taxonomy rather than occupying one cell of it \citep{seth2024,seth2026stuff}.
In its weak form, the functional organisation relevant to consciousness, interoceptive inference in the service of allostasis, is specified functionally but held to be implementable only in living systems. This reading represents a Level 4 credence combined with an indirect constraint.
In its stronger form, the claim is that there is a direct line between perceptual (and interoceptive) inference and the autopoietic and metabolic properties of living systems -- a link licensed by the free energy principle. In this form there is no bright-line distinction between role and realiser, so there is no substrate-independent specification of the relevant role for the substrate constraints to act on.
Where the separation of role from realiser fails in this way, our constraint formalism approximates the view rather than faithfully expresses it.

\subsubsection*{How the realisability constraints interact with the functional hierarchy}\label{how-the-realisability-constraints-interact-with-the-functional-hierarchy}
\addcontentsline{toc}{subsubsection}{How the realisability constraints interact with the functional hierarchy}

Now we see why the cross-cutting realisability constraints introduced by substrate-dependent theories preserve all three structural features of the functional hierarchy: supervenience, coarse-graining and multiple realisability.

\textbf{Supervenience} is preserved because the hierarchy of levels is unchanged.
Each coarser level still supervenes on each finer level.
The organism-environment coupling determines the organismic functions, which determine the intrinsic causal structure, which determines the computation, which determines the behaviour.
The substrate constraints do not alter these supervenience relations.
They operate orthogonally, specifying which first-order properties must be present at a given level for the functional role at that level to be genuinely realised.
The supervenience hierarchy tells you what functional roles must be preserved; the substrate constraints tell you what realisers those roles require.
To be precise: if we fix the complete physical state of the system (including its substrate properties), we fix its organism-environment coupling, which fixes its organismic functions, which fixes its intrinsic causal structure, which fixes its computation, which fixes its behaviour.
The substrate constraints simply specify that certain features of the complete physical state -- electrochemical mechanisms, quantum processes, electromagnetic field dynamics, carbon-based chemistry -- must be present for the functional properties at a given level to be actually realised.
They narrow the supervenience base without altering the supervenience relations between levels.

\textbf{Coarse-graining} is preserved because each level remains a coarse-graining of the level below it, and the substrate constraints specify which physical details are not safely discarded in the coarse-graining operation at each level.
When we coarse-grain from the full microphysical description (which is assumed to be the ultimate subvenient base) to the intrinsic causal-structure level, we normally discard information about the specific physical mechanism of causal interaction, retaining only the pattern of causal constraint.
EM field theory says this coarse-graining is too aggressive for consciousness -- it discards information about the electromagnetic nature of the causal interaction that matters.
Quantum theories say the coarse-graining from full physics to intrinsic causal structure discards information about the quantum nature of the causal processes that matter for consciousness.
Carbon chauvinism says that when we coarse-grain from full microphysical description to organismic functions, we discard information about the chemical substrate that matters.
In each case, the substrate-dependent theories do not prohibit coarse-graining in general.
Rather, they contend that the standard coarse-graining at a particular level throws away specific first-order information that is relevant to consciousness.
Thus, even in the presence of substrate constraints, the earlier coarse-graining hierarchy can remain intact -- each level is still derived from the level below by abstracting away detail.
The substrate constraints identify specific details that should not be abstracted away at specific levels.

\textbf{Multiple realisability} is preserved but modulated.
At each level, the class of physical realisations that count as having the relevant properties is narrower than the purely functional description would suggest, but multiple realisability is not eliminated.
EM field theory does not say only biological brains can be conscious.
It says only systems that generate the right kind of integrated electromagnetic fields can be conscious, and in principle, non-biological systems might generate such fields.
Quantum theories do not say only microtubules can support consciousness.
They say only systems with quantum coherent processes can, and quantum computers might qualify.
Even carbon chauvinism allows multiple realisability within carbon chemistry; many different carbon-based organisms could realise the same organismic functions, and conceivably artificial carbon-based systems could be designed that realise them too.
So substrate constraints reduce multiple realisability at the levels where they operate without eliminating it entirely.
They narrow the class of possible realisers from `anything that implements the same functional role' to `anything that implements the same role using the right kind of first-order mechanism'.
This is a restriction, but it is not identity theory -- it does not require one specific physical implementation.
It requires a kind of implementation, which can itself be multiply realised within limits.

There is an important distinction between direct and indirect substrate constraints.
A \emph{direct} \emph{substrate constraint} says: at level X (e.g. the organismic functional level), the functional role must be realised by mechanism Y (e.g. ionic cellular sensing), and it is the mechanism Y itself that is constitutive of consciousness (e.g., strong Biological Naturalism).
An \emph{indirect substrate constraint} says: the functional role at level X can only be achieved by mechanism Y as a matter of physical or biological fact, but it is the functional role that is constitutive of consciousness, not the mechanism (e.g., weak Biological Naturalism).
In the indirect case, if some other mechanism could achieve the same role -- perhaps in principle even if not in practice -- it would suffice for consciousness.
In the direct case, no other mechanism would suffice even if it achieved the same role.
The two cases therefore preserve multiple realisability in different senses.
An indirect constraint leaves the role multiply realisable across mechanism kinds in principle, restricting it only in practice.
A direct constraint eliminates realisability across kinds, but the required kind remains multiply realisable within itself: many different electrochemical systems, say, could realise the same role.
This is the sense in which the preservation claimed above survives even the strongest constraints; what the direct/indirect distinction tracks is not how much realisability survives but why the constraint binds.
Both kinds of constraint are accommodated by treating substrate-dependent theories as realisability constraints distributed across existing levels.
The difference between direct and indirect constraints is a difference in the strength of the constraint -- whether the first-order property is \emph{constitutively} necessary for consciousness or merely \emph{nomologically} necessary for achieving the relevant second-order role.
But in either case, the constraint operates at a particular level of the hierarchy, narrowing the class of realisers, without adding a new level.\footnote{Substrate constraints can equivalently be given a supervenience reading. A constraint requiring that the role at Level X be realised by mechanism Y defines a sublevel that draws all the distinctions of Level X plus the substrate distinction, so that Level X supervenes on it and constitutes a coarse-graining of it. Constraints operating at several levels, such as Block\textquotesingle s meat hypothesis, generate chains of such sublevels that mirror the main hierarchy, refining the five-level total order into a directed acyclic graph in which some pairs of descriptions are incomparable. We do not need this refinement for what follows.}

For what follows, the operative point is that substrate constraints add nothing to the level structure of the hierarchy.
They condition what counts as a genuine realisation of a functional role at a level, and this is how the Bayesian model of \Cref{indicators-evidence-and-the-attribution-of-consciousness-to-ai-systems-a-bayesian-approach} will absorb them: an indicator counts as activated only if the role it tracks is operationalised through the required substrate.

\subsection{The full architecture of the hierarchy: a taxonomy of positions}\label{the-full-architecture-of-the-hierarchy-a-taxonomy-of-positions}

\begin{figure}[htbp]
\centering
\includegraphics[width=1.0\linewidth]{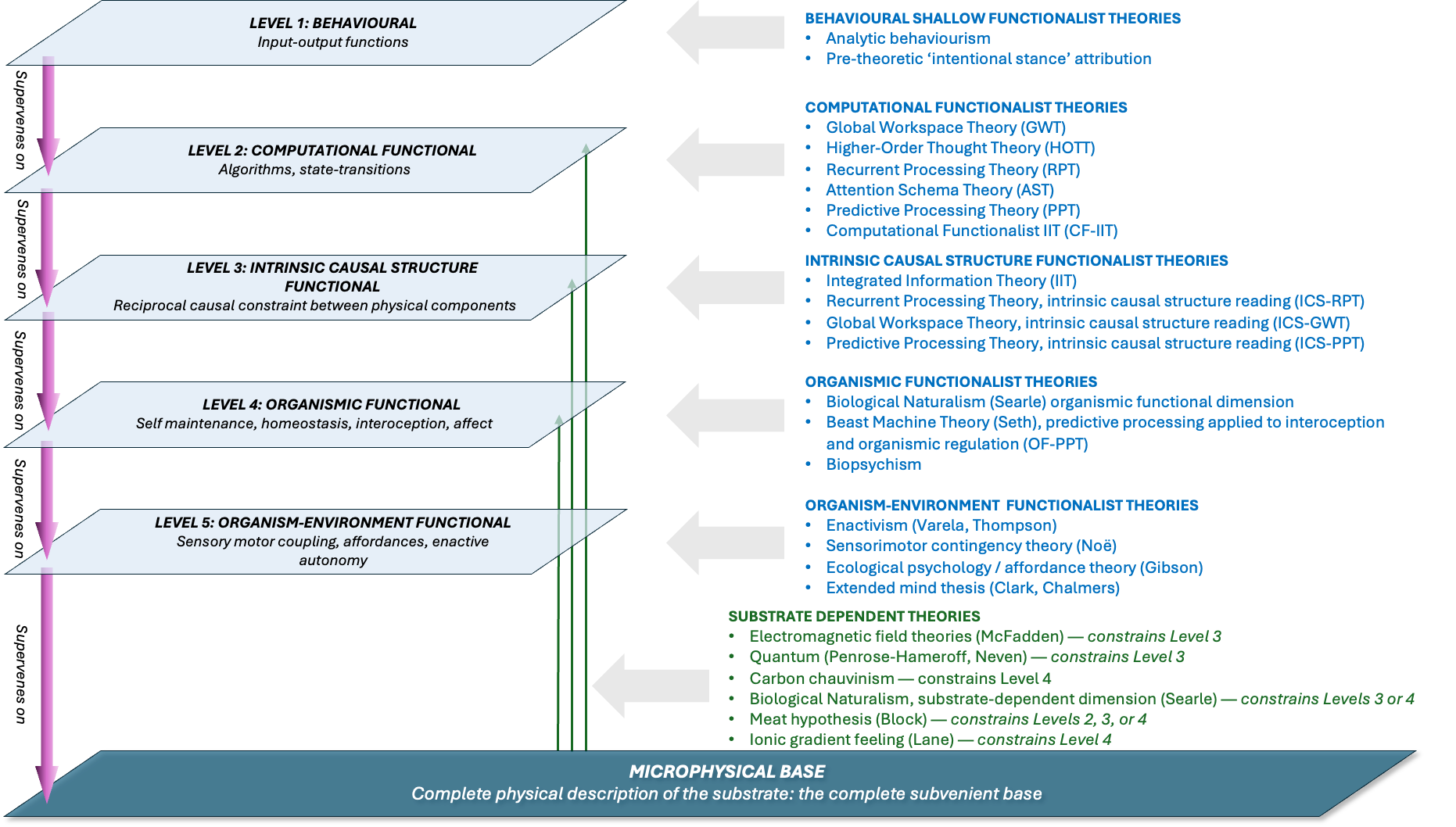}
\caption{\textbf{The complete architectural framework.} \emph{Mapping the various theories of consciousness to their corresponding critical levels of description. The five functional levels, forming a supervenience hierarchy, represent successive coarse-grainings from the ultimate microphysical base. Substrate dependent theories are depicted via vertical green arrows, illustrating how specific microphysical features flow upward to constrain the physical realisability of functional roles at higher levels.}}
\label{fig:image14}
\end{figure}

Throughout the preceding discussion, a sixth level has been implicitly present without being explicitly acknowledged.
When we described coarse-graining `from the full microphysical description to the intrinsic causal-structure level', and when we noted that `fixing the complete physical state of the system (including its substrate properties)' fixes everything above it in the hierarchy, we were already referencing the full microphysical description of the substrate as the base from which all five functional levels are derived.
Making this explicit completes the architecture.
This full architecture provides a taxonomy of the various theories of consciousness and is illustrated in \Cref{fig:image14}.
The full microphysical description -- the complete physical constitution of the entire extended system (which includes the environment) at the finest grain, including the quantum states of every particle, every electromagnetic field, every chemical bond -- is the ultimate subvenient base of the hierarchy.
The five functional levels are successive coarse-grainings of this base, each discarding more microphysical detail to retain only the functional structure deemed relevant at that grain.
Substrate-dependent theories are then understood as claims that certain microphysical details -- e.g. electromagnetic field dynamics, quantum coherence, electrochemical mechanisms, carbon chemistry, ionic gradient physics -- should not be discarded in the coarse-graining from the microphysical base to a particular functional level.
They are microphysical features that `flow up' through the hierarchy because they are relevant to consciousness at a particular functional level, not as additional functional roles but as realisability constraints on how the roles at that level must be physically instantiated.

There is, however, an important asymmetry between the microphysical base and the five levels above it.
The five functional levels are each a description of what the system does -- its behaviour, its computation, its causal organisation, its self-maintenance, its sensorimotor coupling.
They are functional descriptions, specified in terms of roles, and the debate between them is a debate about which functional grain captures the organisation relevant to consciousness.
The microphysical level is not a functional description.
It is a description of what the (extended) system is -- its complete physical constitution at the finest grain.
It does not specify a role; it specifies the totality of first-order properties from which all roles are derived.\footnote{This characterisation presupposes that there are first-order properties at the microphysical base for functional roles to bottom out in, which is contested. On a long tradition running from \citet{russell1927} through \citet{chalmers2003} and \citet{alter2016}, physical description characterises the world only in terms of structure and dynamics, mass and charge and wavefunction state being specified relationally and dispositionally, so that what a fundamental particle is is exhausted by what it does. \Cref{neutral-monism} above, in describing what physics captures as structural, relational, or dispositional, in effect concedes the point. If this is right, the microphysical base is not a description of first-order properties but the limiting case of functional description, and substrate-dependent theories are best read not as claims of a categorically different kind but as functional claims at a very fine grain. We note that our formal apparatus is neutral between these readings, since a realisability constraint functions in the same way whether it is read as a first-order constraint on realisers or as a functional claim at a very fine grain.}

But no one proposes that the full microphysical description is the right level for consciousness, because that would amount to saying that nothing can be abstracted away, which would eliminate explanation entirely.
Even the most extreme substrate-dependent theorist does not claim that every microphysical detail matters.
They claim that certain specific microphysical details matter and that these should not be discarded in the coarse-graining to a particular functional level.
So the microphysical base is not itself a theory of consciousness.
It is the ground from which all theories of consciousness extract their claims -- the raw material that each functional level coarse-grains in its own way, and that substrate-dependent theories insist must be coarse-grained more carefully at specific points in the hierarchy.

This architecture also addresses the scope of the `system' which, as discussed, includes the environment because the organism-environment functional level already supervenes on the organism and the environment.
At the microphysical level, the system is in principle unbounded -- the complete microphysical description extends, through causal entanglement, to the entire universe.
The microphysical level is the base from which we begin coarse-graining, and the whole point of the functional hierarchy is to identify which features of this base are relevant and to draw boundaries accordingly.
The organism-environment level draws a boundary at the organism's Umwelt.
The organismic level draws a boundary at the organism.
The intrinsic causal-structure level draws a boundary at the system's physical components.
Each functional level identifies what matters and sets aside the rest.
The microphysical base may be infinite in scope, but the functional levels carved from it are not -- they are precisely the acts of coarse-graining that make the unbounded microphysical reality tractable by identifying the relevant structure at each grain.
The complete hierarchy therefore runs from the microphysical base upward through successively coarser functional descriptions -- microphysical base, organism-environment, organismic, intrinsic causal-structure, computational, behavioural -- with supervenience running downward, multiple realisability running upward, and substrate-dependent theories functioning as vertical channels through which specific microphysical details flow up to constrain the realisers at particular functional levels.

To consolidate the taxonomy, Table 6 summarises the placement of the major theories\footnote{Note that several theories appear at multiple levels because they admit different philosophical interpretations. RPT, GWT, and PPT each have both a Level 2 (computational functionalist) reading and a Level 3 (intrinsic causal-structure functionalist) reading. Biological Naturalism appears at Level 4 as an organismic functionalist theory and also as a substrate-dependent realisability constraint on Levels 3 and 4. Block's meat hypothesis is deliberately agnostic about which level the critical realiser targets.} of consciousness onto the hierarchy.

{\def\LTcaptype{none}\footnotesize\sffamily 
\begin{longtable}[]{@{}
  >{\raggedright\arraybackslash}p{(\linewidth - 2\tabcolsep) * \real{0.5000}}
  >{\raggedright\arraybackslash}p{(\linewidth - 2\tabcolsep) * \real{0.5000}}@{}}
\toprule\noalign{}
\begin{minipage}[b]{\linewidth}\raggedright
\textbf{LEVEL}
\end{minipage} & \begin{minipage}[b]{\linewidth}\raggedright
\textbf{THEORIES OPERATING AT THIS LEVEL}
\end{minipage} \\
\midrule\noalign{}
\endhead
\bottomrule\noalign{}
\endlastfoot
Level 1: Behavioural & Analytic behaviourism \\
Level 2: Computational functional & GWT, HOTT, RPT, AST, PPT, CF-IIT \\
Level 3: Intrinsic causal-structure functional & IIT, ICS-RPT, ICS-GWT,
ICS-PPT \\
Level 4: Organismic functional & Biological Naturalism (Searle), Beast
Machine Theory (Seth), Biopsychism \\
Level 5: Organism-environment (4E) functional & Enactivism, Sensorimotor
contingency theory (Noë), Ecological psychology (Gibson), Extended mind
thesis (Clark, Chalmers) \\
Substrate-dependent realisability constraints (cross-cutting) & EM field
theories (McFadden) → Level 3; Quantum theories (Penrose-Hameroff,
Neven) → Level 3; Carbon chauvinism → Level 4; Biological Naturalism
(substrate dimension; Searle/Seth) → Levels 3 \& 4; Block\textquotesingle s meat
hypothesis → Levels 2, 3 \& 4; Lane\textquotesingle s ionic gradients →
Level 4; Large-scale dynamic patterns (Godfrey-Smith) → Level 3 \\
\end{longtable}
}

\textbf{Table 6. A taxonomy of theories.} \emph{Mapping the major theories of consciousness onto the supervenience hierarchy}

\section{Indicators, evidence, and the attribution of consciousness to AI systems: A Bayesian approach}\label{indicators-evidence-and-the-attribution-of-consciousness-to-ai-systems-a-bayesian-approach}

The preceding sections have developed a five-level hierarchy of functional descriptions, positioned the major theories of consciousness within it, identified operationalisable indicators at each level, and considered current AI systems in light of them.
This framework is a structural contribution -- it organises the landscape of positions and makes their commitments precise.
But a framework is only as useful as the work it enables.
In this section we develop one natural way to put the framework to use: a Bayesian approach to assessing a system's consciousness that combines indicator evidence, theoretical credences about which level is the critical one, and leverages the coarse-graining taxonomy.
It is worth highlighting that, even putting the model we suggest here to one side, the supervenience hierarchy could serve as a common language for interdisciplinary dialogue, a scaffold for structured expert elicitation, a curriculum for training researchers to think across theoretical boundaries, or a checklist prototype for AI developers designing evaluation protocols.
The Bayesian application we develop here is, however, particularly well-suited to the framework's structure, because the hierarchy naturally decomposes the question of consciousness into the key ingredients of this model: indicator and level nodes, priors (about indicators' informativeness, and about consciousness in any particular system), and credences (about levels).
With all these ingredients on the table, we can arrive at an overarching view which also tracks our uncertainty.

\subsection{A Bayesian framework for assessing AI consciousness}\label{a-bayesian-framework-for-assessing-ai-consciousness}

The question of whether a given AI system is conscious cannot be answered with certainty given the current state of consciousness studies.
There is genuine uncertainty both about \emph{which theory of consciousness is correct} -- and therefore also about which combinations of necessary and sufficient conditions matter for consciousness, and how these conditions map onto level(s) of functional description. There is also uncertainty about \emph{how informative} the observable properties are that a particular theory identifies as consciousness-relevant, and about whether any given \emph{system possesses those properties}.
A principled approach to this question must handle these sources of uncertainty simultaneously.
The natural framework for doing so is Bayesian: indicators provide evidence that updates our credence in a system's consciousness, mediated by our theoretical expectations about what consciousness depends upon.

The Digital Consciousness Model (DCM) developed by Shiller, Clatterbuck, Mu\~noz Mor\'an et al. \citeyearpar{shiller2026} provides a concrete implementation of this kind of approach.
The DCM is a Bayesian multi-level model that incorporates multiple \emph{stances} (theories of consciousness), each generating their own probabilistic assessment of a system's consciousness based on indicator evidence, with the overall assessment being a plausibility-weighted average across stances.
Our framework complements and in some ways extends this approach by providing a structural explanation, grounded in the hierarchy of functional levels, for why different stances focus on different features and why certain clusters of stances tend to converge or diverge in their assessments.
In particular, the supervenience structure between levels of analysis provides the structural scaffold for the levels in the Bayesian model we develop in this section.

Within our framework, each level of functional description generates a set of indicators -- operationalisable markers whose presence or absence provides evidence for or against a system's consciousness at that level.
As noted in the discussion of individual levels, these indicators are neither necessary nor sufficient conditions for consciousness.
They are evidence: their presence raises or lowers the probability of consciousness conditional on a particular level being the critical one for consciousness.
In Bayesian terms, each indicator \emph{I} for a system \emph{S} has a \emph{positive likelihood ratio} with respect to the claim that S is \textquotesingle{}\emph{conscious at level L}\textquotesingle.
And by \textquotesingle{}\emph{conscious at level L}\textquotesingle{} we mean that \emph{S} instantiates, at a given level \emph{L}, the organisation that theories operating at that level take to suffice for consciousness.
This positive likelihood ratio can therefore be written as:

\[
\frac{P(\ I\ is\ present\ |\ S\ is\ conscious\ at\ level\ L)\ }{P(\ I\ is\ present\ |\ S\ is\ not\ conscious\ at\ level\ L)\ }
\]

When this ratio is greater than 1, the indicator's presence provides confirmatory evidence; when less than 1, disconfirmatory evidence.
An analogous negative likelihood ratio that handles evidential updates from the indicator's absence is similarly defined.
The magnitude of (the logarithm of) each ratio reflects the strength of the evidence the indicator carries.
In terms of the DCM's terminology, the evidential relationship between an indicator and consciousness at a given level can be fully characterised by two parameters: \emph{support} (how much more likely the indicator is to be present in a conscious system than an unconscious one) and \emph{demandingness} (how rarely the indicator is present in systems that are not conscious at that level).
Support corresponds to the positive likelihood ratio above, while demandingness is inversely related to the \emph{false positive rate} (the probability that an indicator is present although the system does not instantiate, at that level, the consciousness-sufficient organisation).
For example, indicators that are both highly supportive and have a low false positive rate (i.e. are \emph{highly demanding}) provide the strongest evidence when present and the strongest counter-evidence when absent.\footnote{When we implement the model in \Cref{from-supervenience-hierarchy-to-conditional-independence}, we map current indicators to their closest analogue in the DCM and reuse the DCM's support and demandingness labels, converted into a true positive rate and a false positive rate per indicator by the mapping documented in the accompanying code, for the purposes of the \href{https://ai-cognition.org/cacophony-tool/}{interactive example}. The conversion is deliberately more polarised than the DCM's own parameterisation, so each indicator carries more signal here than in the DCM. This strategy has several limitations, and is only meant as a useful first approximation to illustrate model behaviour. Technical readers can see the code here (\url{https://github.com/arvomm/cacophony-public-code}).}

One further property of these parameters should be made explicit: they are population-relative \citep{bayne2024}.
A likelihood ratio calibrated on the human case does not automatically transfer to systems that differ radically in embodiment and organisation, and under such population shift three quantities must be kept apart: whether an indicator is applicable or measurable in the target population, whether it is observed to be present or absent in that population, and how diagnostically informative that observation is.
An indicator may be inapplicable to a system, when the measurement pathway it relies on is simply unavailable there; it should then carry a likelihood ratio of one and licence no update in either direction, since treating an inapplicable indicator's silence as evidence of absence is the central error the measurement-theory literature warns against \citep{peters2026}.
An indicator may be applicable but confounded, when the measurement can be made yet a competing explanation undermines what a positive result means; the anthropomimesis discount of \Cref{summary-of-the-report} provides a signal example of this applied to behavioural indicators in LLMs.
Or it may be applicable and valid, in which case the (for example, human-calibrated) parameters are a useful starting point.
The activations in the accompanying tool can express all three cases, and the likelihood ratios throughout should be read as indexed to the population under assessment -- in other words, they are conditional on assumptions about population validity.\footnote{Though we recommend that interpretation, the tool only uses a single (human-indexed) example set of these parameters under the hood, both for simplification and tractability. More robust results can be obtained by adjusting (that is, indexing) each set of parameters in the code (\url{https://github.com/arvomm/cacophony-public-code}) to match each population of interest.}

This Bayesian formulation captures several important features.
In particular, an indicator can provide strong evidence without being strictly sufficient (or indeed necessary).
A system might be conscious without exhibiting that indicator, or indeed be unconscious while having it.
At the same time, we want our model to reflect that the absence and presence of indicators can still carry information and alter our credence that the system is conscious.
In the next section we construct a Bayesian network by translating the supervenience hierarchy into conditional independence relationships that are captured by the structure of the Bayesian network.
This will allow us to understand exactly how beliefs about consciousness and evidence at different levels of the hierarchy interact.
In particular, we will see that the supervenience relationship between levels provides a structural vehicle for the evidence to flow between them.

The supervenience structure of the hierarchy allows for evidence at one level to update our credence in consciousness at a different level.
Since the levels are successive coarse-grainings of the same system, indicators identified as relevant at one level also yield evidence at other levels -- not as a matter of deductive entailment but as a matter of empirical regularity.
Systems with rich organismic functions tend to exhibit certain computational properties, because the organismic organisation typically produces certain kinds of information processing.
Systems with dense intrinsic causal integration tend to exhibit certain behavioural signatures, because the causal structure shapes what the system does.
This justifies how evidence at one level probabilistically bears on the assessments at other levels, and is handled in the model through Bayesian updating.

Concretely, consider an AI system that satisfies several Level 2 indicators -- it exhibits information integration, recursivity, and metacognitive monitoring at the algorithmic level.
For a Level 2 theorist (a computational functionalist), these indicators directly update the probability of consciousness: they provide strong evidence that the computational requirements are met.
For a Level 4 theorist (an organismic functionalist), these same indicators provide weaker but still positive evidence.
They raise the probability that the system has the right organismic properties, because systems with the right organismic functions would typically also exhibit these computational features.
But the updating is more modest, because the Level 2 evidence is noisy and indirect evidence for Level 4 consciousness, and indeed many computationally sophisticated systems lack organismic functions entirely.
In general, Level 4 indicators are a more direct, and therefore more heavily weighted, source of evidence for evaluating Level 4 consciousness.
Conversely, Level 4 indicators can be viewed from a computational perspective and therefore would also provide evidence for a Level 2 computational functionalist.

\subsection{From supervenience hierarchy to conditional independence}\label{from-supervenience-hierarchy-to-conditional-independence}

A Bayesian network is a probabilistic graphical model (PGM) that can represent complex joint probability distributions by factorising them based on the proposed graph.
It contains nodes, which represent random variables, and edges, which express probabilistic relationships between them.

In this section, we construct a Bayesian network that translates the proposed supervenience hierarchy into structural and probabilistic constraints on beliefs about consciousness and evidence at different levels.
We begin with a strict model of supervenience, presented as a limiting assumption rather than as a consequence of the hierarchy, and generalise it in the subsequent section; the generalised model of \Cref{a-generalised-model} is the model of record from that section onwards.

\subsubsection{Strict Supervenience Model}\label{strict-supervenience-model}

Let us first introduce the formal elements of the network.
For each Level \(i \in \{ 1,\ 2,\ 3,\ 4,\ 5\}\) we introduce a Boolean random variable \(C_{i}\), where \(C_{i}\) = 1 represents the proposition \textbf{``the system instantiates, at Level \emph{i}, the organisation that Level-\emph{i} theories take to suffice for consciousness''}.
For brevity we will say the system instantiates ``consciousness at Level i'', matching the usage introduced in \Cref{a-bayesian-framework-for-assessing-ai-consciousness}.
What are the logical relationships between these five variables \(C_{i}\)?
Can they take arbitrary assignments of truth values such as \({(C}_{1},\ C_{2},\ C_{3},\ C_{4},\ C_{5})\  = \ (T,\ F,\ T,\ T,\ F)\)?
Not if we employ a strict model with a limiting assumption.

The first model is strict and adds a key assumption: that these theory-relative conditions nest along the chain, so that for \(i \in \{ 1,\ 2,\ 3,\ 4\}\), \(C_{i + 1} \Rightarrow C_{i}\) (and by transitivity for all \(i > j\), \(C_{i} \Rightarrow C_{j}\)).
In other words, under this proposal, a certainty that \(C_{i + 1}\) obtains is enough to conclude that \(C_{i}\) obtains.
Thus, under this assumption, the only six configurations of truth values permissible are \((F,\ F,\ F,\ F,\ F)\), \((T,\ F,\ F,\ F,\ F)\), \((T,\ T,\ F,\ F,\ F),\ (T,\ T,\ T,\ F,\ F)\), \((T,\ T,\ T,\ T,\ F)\) and \((T,\ T,\ T,\ T,\ T)\).

The status of this assumption matters.
It does not follow from the supervenience of descriptions.
That the Level-\(j\) description is determined by the Level-\(i\) description does not guarantee that the organisation Level-\(i\) theories deem sufficient coarse-grains into a profile that Level-\(j\) theories deem sufficient, because each level's sufficient set is fixed by that level's theories, and not induced by coarse-graining from below.
Condition 2A (introduced below) is therefore a substantive assumption about how the theories' conditions relate, and \Cref{a-generalised-model} presents a counterexample against it.
Nevertheless, we begin with this stricter and arguably simpler deterministic model.

There are two central suppositions, then, that guide the construction of the network:

\textbf{1. Conditional independence}: the \(C_{i}\) nodes form a directed chain

\begin{quote}
\(C_{5}\  \rightarrow \ C_{4\ } \rightarrow \ C_{3}\  \rightarrow \ C_{2}\  \rightarrow \ C_{1}.\)
\end{quote}

Therefore, for all \(i\), the variable \(C_{i}\) is conditionally independent of all more distant ancestors (jointly), given its immediate parent \(C_{i + 1}\).
This is a local Markov property \citep{pearl1988} and it guarantees that for all \(i\) and any \(j\  > \ i + 1\), \(P(C_{i}|C_{i + 1},C_{j})\  = \ P(C_{i}|C_{i + 1})\), where here \(i\  \in \ \{ 1,\ 2,\ 3\}\).
Under this idealisation, the chain is the supervenience hierarchy translated into probability.

\textbf{2A. Deterministic supervenience constraint}: For \(i \in \{ 1,\ 2,\ 3,\ 4\}\) we have \(P(C_{i}|C_{i + 1})\) = 1.
This is because we here model supervenience and a complete activation of a finer-grained level as the logical implication \(C_{i + 1}\  \Rightarrow {\ C}_{i}\).
(This is weakened in the generalised model in \Cref{a-generalised-model} and replaced with Condition 2B.)

Thus, given both conditions, the joint distribution over the consciousness variables factorises as

\[
P\left( C_{1},\ldots,C_{5} \right) = P\left( C_{5} \right)\prod_{i = 1}^{4}P\left( C_{i} \mid C_{i + 1} \right)
\]

Next, we incorporate indicator evidence.
Let \(E_{i}\  = \ \{ I_{i,\ 1},\ ...,\ I_{i,{\ n}_{i}}\}\) denote the set of indicator variables associated with some level \(i\).
And let \(E = E_{1} \cup E_{2}\cdots \cup E_{5}\) denote the complete collection of indicators.
In the network extended to incorporate evidence, each indicator \(I_{i,\ k}\) is a child of \(C_{i}\).
This represents the assumption that the indicator's probability of being present depends on consciousness being present at the corresponding level.
Moreover, the graphical structure we propose below further assumes that indicators are conditionally independent given their parent.
Given this model, the joint distribution containing all relevant information can be factorised as

\[
P\left( C_{1},\ldots,C_{5},E \right) = P\left( C_{5} \right) \cdot \prod_{i = 1}^{4}P\left( C_{i} \mid C_{i + 1} \right) \cdot \prod_{i = 1}^{5}P\left( E_{i} \mid C_{i} \right).
\]

And if the individual indicators at each level are conditionally independent given \(C_{i}\), as we assume, then

\[
P\left( E_{i} \mid C_{i} \right) = \prod_{k = 1}^{n_{i}}P\left( I_{i,k} \mid C_{i} \right).
\]

Recall the parameters introduced in \Cref{a-bayesian-framework-for-assessing-ai-consciousness} (for example the positive likelihood ratio and the false positive rate).
We can make specific assumptions about those parameters to fully characterise the conditional relationships between all nodes, and use the extended Bayesian network below to infer posterior probabilities of consciousness at each level.\footnote{In particular, the strategy we adopt here to fully characterise the edges' informativeness is to make assumptions about the false positive rate for the level-to-level edges, which is \(P(C_{i} \mid \lnot C_{i+1})\), and use the per-indicator rates from the DCM mapping for the indicator edges. You can explore this further in the code and notebook available \href{https://github.com/arvomm/cacophony-public-code}{here}.} We fully implement this model; see \Cref{illustrative-examples-from-the-bayesian-model}.

\begin{figure}[htbp]
\centering
\includegraphics[width=0.16\linewidth]{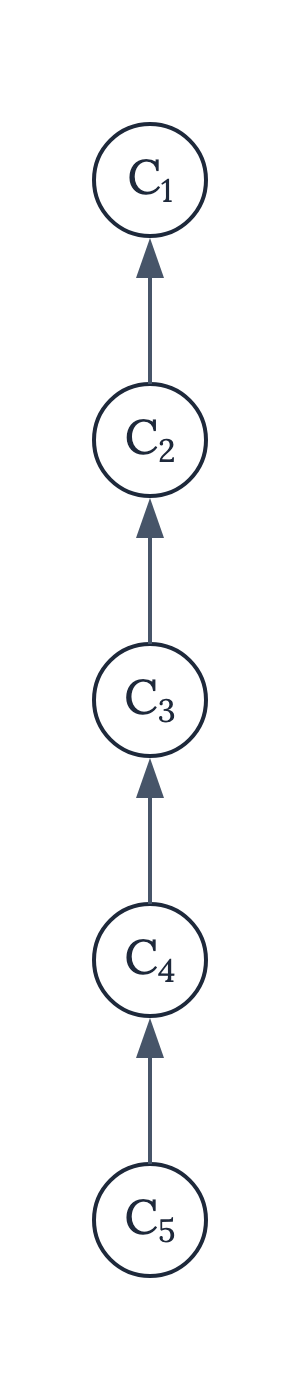}\hfill \includegraphics[width=0.34\linewidth]{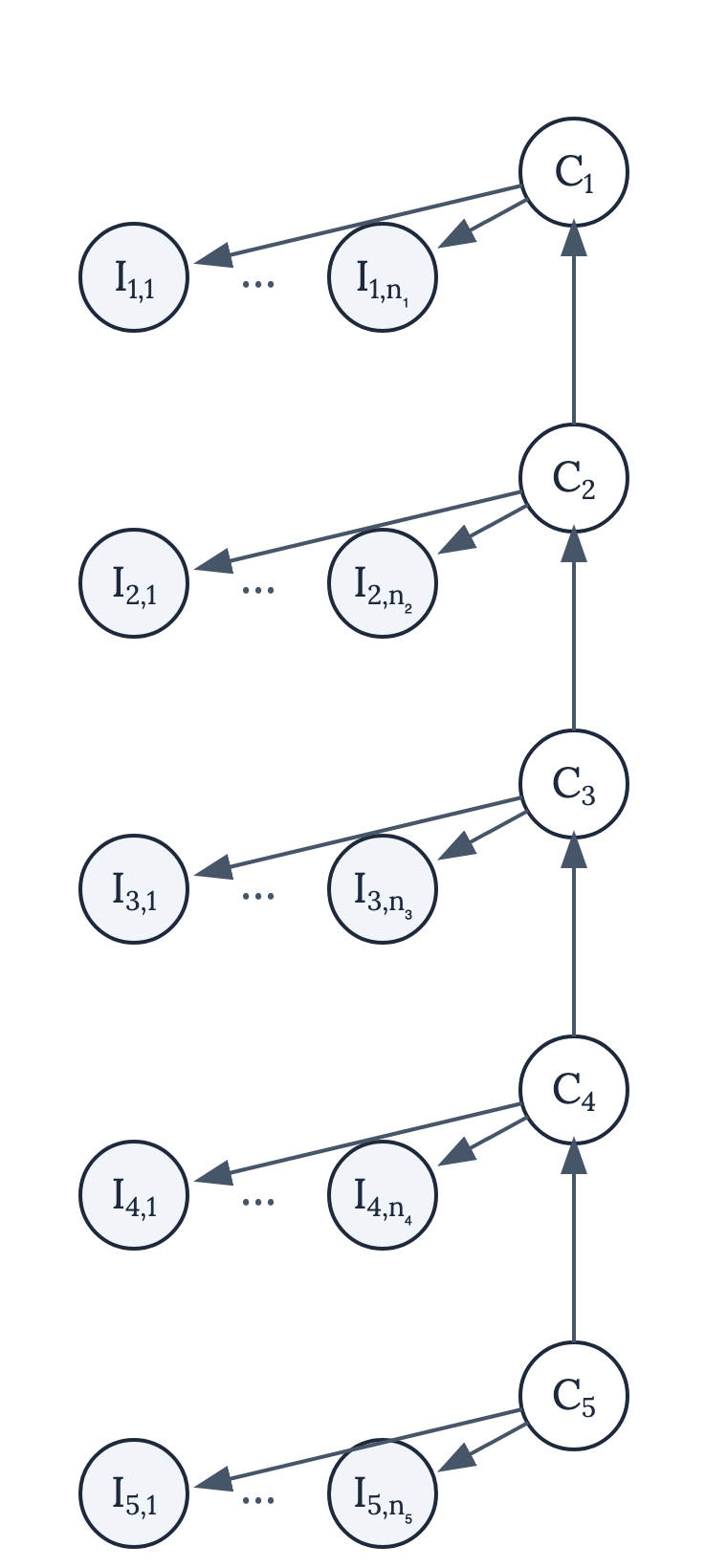}
\caption{\textbf{Bayesian network representation of the supervenience hierarchy.} Left: the consciousness nodes \(C_{i}\) represents the proposition ``the system instantiates, at Level \emph{i}, the organisation that Level-\emph{i} theories take to suffice for consciousness'' and form a directed chain where \(P(C_{i}|C_{i + 1})\) = 1, which models the strict deterministic relationship between the nodes. Right: the extended network includes indicator variables \(I_{i,\ k}\) which provide evidence at each level. Observing the indicators allows inference for the posteriors of all consciousness variables.\protect\footnotemark{}}
\label{fig:image7}
\end{figure}
\footnotetext{The Bayesian network can easily be modified to represent the cross-cutting constraints we discussed in \Cref{substrate-dependent-theories-as-realisability-constraints-distributed-across-multiple-levels}. In particular, the set of indicators could be understood to only activate if they are operationalised through the relevant substrate. For example, the relevant Level 2 algorithms would need to be realised by meat for them to count under Block's meat hypothesis.}

It is worth noting that, given the structure of the network, the marginal distributions \(P\left( C_{i}\  \mid \ e \right)\) satisfy the following \textbf{monotonic belief property lemma} (see \Cref{fig:image35} for a worked example):

\[
P\left( C_{1} \mid e \right) \geq P\left( C_{2} \mid e \right) \geq \cdots \geq P\left( C_{5} \mid e \right),
\]

where \(e\) is a specific instance of the evidence \(E\), where we use \(C_{i}\) to denote \(C_{i} = 1\) and \(P(\  \cdot \mid \ e)\) to denote \(P(\  \cdot \mid \ E = e)\).
We also write \(\neg C_{i}\) for \(C_{i} = 0\) below.

The proof of this property is as follows.
Suppose we observe a specific instance of the evidence \(E = e\).
We can apply the conditional law of total probability to obtain

\[
P\left( C_{i} \mid e \right) = P\left( C_{i} \mid C_{i + 1},e \right)P\left( C_{i + 1} \mid e \right) + P\left( C_{i} \mid \neg C_{i + 1},e \right)P\left( \neg C_{i + 1} \mid e \right),
\]

for \(i \in \{ 1,2,3,4\}\).

Recall our deterministic constraint,

\[
P\left( C_{i} \mid C_{i + 1} \right) = 1.
\]

Thus, whenever the relevant conditional probability is well defined (i.e. \(P\left( C_{i + 1},e \right)\  \neq \ 0\ \)), we may write \(P\left( C_{i} \mid C_{i + 1},e \right) = 1.\) It follows that

\[
P\left( C_{i} \mid e \right) = P\left( C_{i + 1} \mid e \right) + P\left( C_{i} \mid \neg C_{i + 1},e \right)P\left( \neg C_{i + 1} \mid e \right).
\]

The second term on the right-hand side lies between 0 and 1 and is therefore non-negative.
Therefore,

\[
P\left( C_{i} \mid e \right) \geq P\left( C_{i + 1} \mid e \right).
\]

Finally, by transitivity,

\[
P\left( C_{1} \mid e \right) \geq P\left( C_{2} \mid e \right) \geq \cdots \geq P\left( C_{5} \mid e \right).
\]

What does this statement mean?
Suppose a Level \(i\) theorist and a Level \(j\) theorist, where \(j > i\), agree on the evidence \(E\  = \ e\) and on the parameters of the model (priors, conditional probabilities, indicator-specific likelihood ratios).
Then, the Level \(i\) theorist's posterior belief in the system's consciousness must be greater than or equal to that of the Level \(j\) theorist's posterior belief in the system's consciousness.
In other words, ceteris paribus, a theorist who believes that consciousness supervenes at a more coarse-grained level has a weakly greater credence in the system being conscious than a theorist who believes that a more fine-grained level of the hierarchy is critical for consciousness.
Intuitively, this can be thought of as a Level \(j\) theorist demanding further fine-grained evidence that would be ignored by a Level i theorist (e.g. a computational functionalist would require specific algorithms even though a system might behave in ways compatible with being conscious).

Stated across the whole hierarchy (under the deterministic supervenience constraint on which the lemma depends, and assuming agreement on all evidence and parameters) the \emph{monotonic belief property lemma} tells us that the posterior belief in AI consciousness is weakly decreasing down the levels: Level 1 behaviourists ≥ Level 2 computational functionalists ≥ Level 3 IIT theorists ≥ Level 4 biological naturalists ≥ Level 5 4E theorists.

\subsubsection{A generalised model}\label{a-generalised-model}

The previous model is a good start.
But now consider how a simple organism with robust interoception and homeostatic regulation would be evaluated under the different Levels.
Such a system might be viewed as conscious under the organismic view (Level 4) without ultimately possessing anything resembling a global workspace, or the specific forms of metacognition that Level 2 computational functionalist theories demand.
Conversely, consider how a computationally sophisticated AI might satisfy every indicator at Level 2 without having anything at stake in its own continued existence (Level 4).

These and other cases are why we want to put forward a more general model where we maintain the framework's structure by keeping \Cref{strict-supervenience-model}'s conditional independence (condition 1), thereby retaining the directed chain \(C_{5}\ \rightarrow \ C_{4\ } \rightarrow \ C_{3}\ \rightarrow \ C_{2}\ \rightarrow \ C_{1}\), but we weaken the \emph{deterministic} supervenience constraint (Condition 2A) that \(P(C_{i}\ |\ C_{i + 1})\) = 1 in favour of condition 2B:

\textbf{Condition 2B: \emph{Probabilistic} supervenience constraint}: \(P(C_{i}\ |\ C_{i + 1})\) \(\leq 1\) for \(i \in \{ 1,\ 2,\ 3,\ 4\}.\)

This more naturally handles cases where we see deeper level (e.g. Level 4) activations without them entailing all the coarser-level activations above it as well (in this case Level 1, 2 and 3 activations).\footnote{As before, the other key parameter for edges' informativeness is \(P(C_{i} \mid \lnot C_{i+1})\). This asks how often we expect the coarser-level organisation to be present when the finer level is known to be absent. Together, these two parameters, which also apply to indicators and their respective parents, characterise all the conditional probabilities needed to run inference.}

Under the deterministic supervenience constraint \(P(C_{1}\ |\ C_{2})\) = 1.
In particular, if we assume all the appropriate algorithms to give a Level 2 theorist a high belief that a system is conscious, the system must also display behavioural signs of consciousness.
This would lead the Level 1 theorist to ascribe at least as high a probability of consciousness as the Level 2 theorist.

But now consider the case of a person who is in a rare medical condition called `complete locked-in syndrome' \citep{laureys2005} where they are fully awake, alert and conscious, but cannot move or speak because of total body paralysis.
Classic locked-in syndrome spares vertical eye movements, so we take the complete form, in which no voluntary behavioural channel remains.
This locked-in patient is conscious but lacks all the Level 1 behavioural indicators and plausibly lacks a Level 1 activation.
Nevertheless, they have the requisite Level 2 indicators and shows a Level 2 activation.
This is the counterexample to condition 2A, the nesting assumption failing exactly where its status as an assumption left it exposed, and the case is handled perfectly well in the new generalised model.
This is because deterministic sufficiency is replaced with probabilistic conditional dependency \(P(C_{i}\ |\ C_{i + 1}) \leq 1\).
However, this does not mean the finer-grained levels cannot give any evidence of consciousness for the coarser-grained levels in the probabilistic model, but that evidence does not provide the earlier, deterministic, guarantee.
So, for instance, if a computational functionalist found Level 4 indicator activations (such as valenced affect, interoceptive inference, and homeostatic regulation), then this evidence would positively update the probability of consciousness at Level 2 (for example these Level 4 features could increase the probability of information integration and metacognitive processes).

\begin{figure}[H]
\centering
\includegraphics[width=1.0\linewidth]{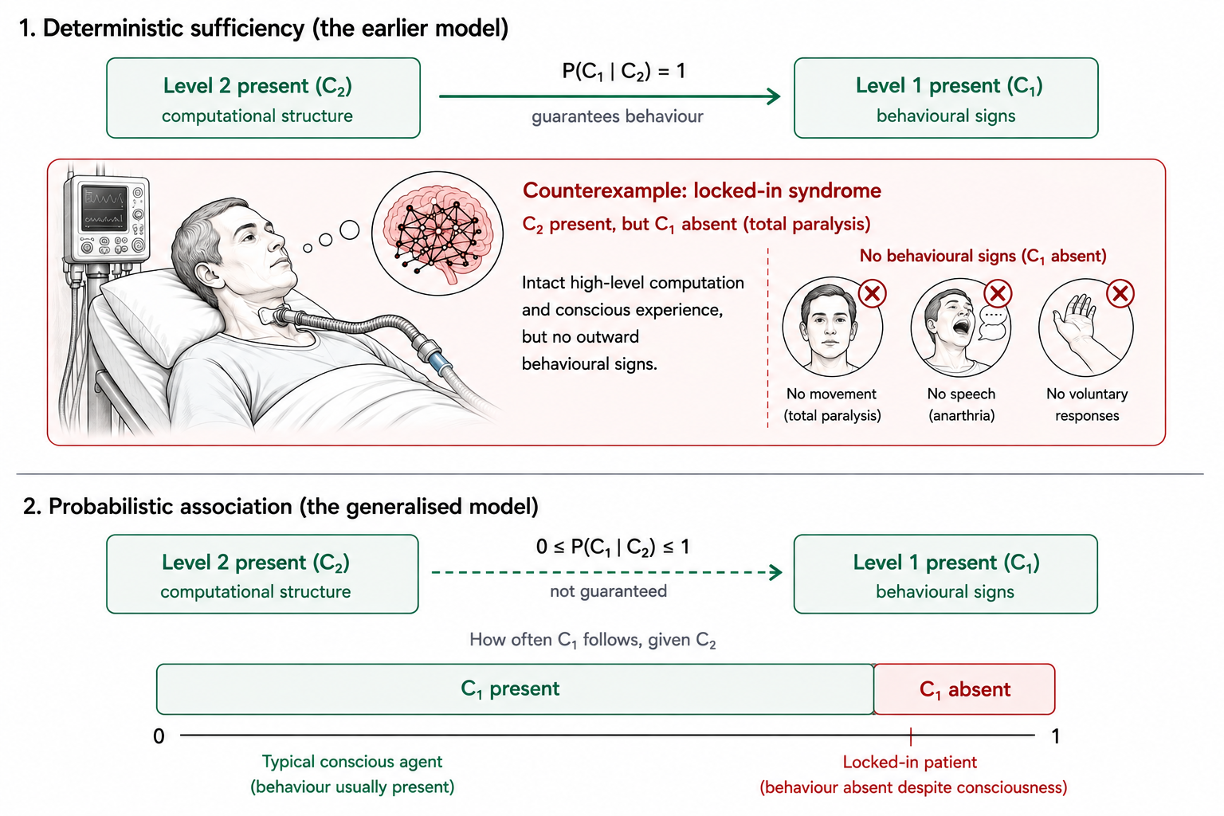}
\caption{\textbf{From deterministic sufficiency to probabilistic association.} \emph{Top: the earlier deterministic reading, on which the activation of Level 2 computational organisation guarantees the activation of Level 1 behaviour, \(P(C_{1} \mid C_{2}) = 1\). The completely locked-in patient is the counterexample: intact computation and conscious experience with no outward behavioural signs and no Level 1 activation. Bottom: the generalised reading, on which the association is probabilistic, \(0 \leq P(C_{1} \mid C_{2}) \leq 1\), accommodating the typical case and the locked-in case within one model.}}
\label{fig:image27}
\end{figure}

\begin{figure}[htbp]
\centering
\includegraphics[width=0.49\linewidth]{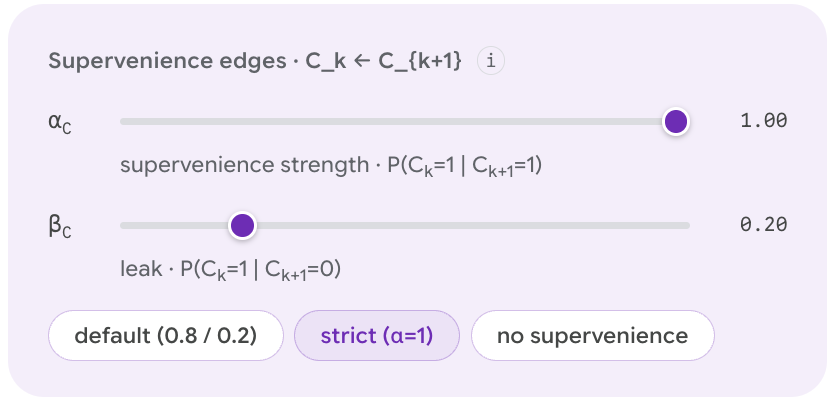}\hfill \includegraphics[width=0.49\linewidth]{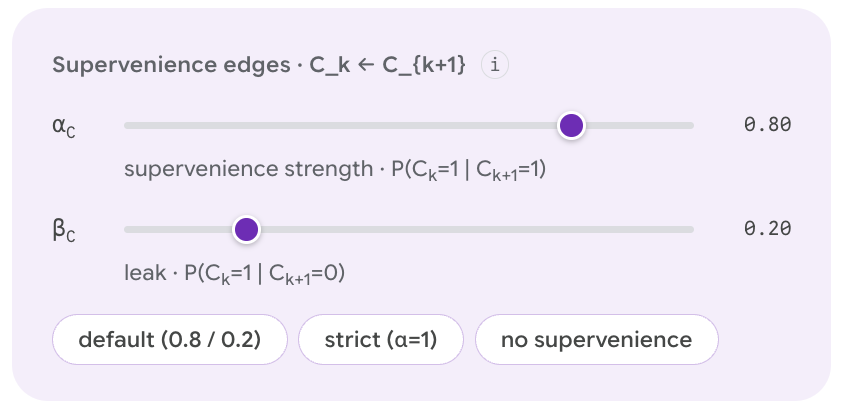}\\[2pt]
\includegraphics[width=0.49\linewidth]{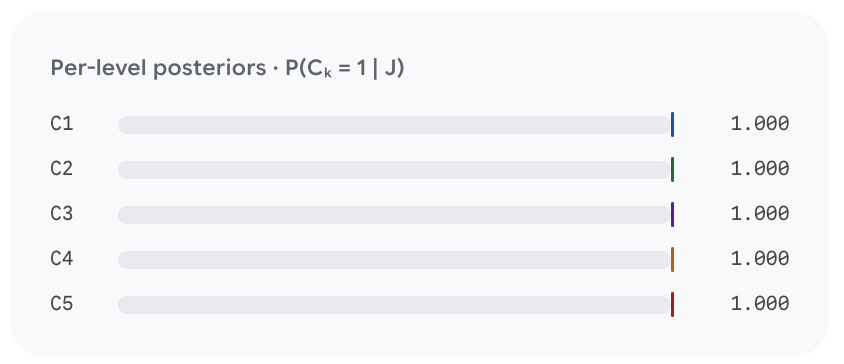}\hfill \includegraphics[width=0.49\linewidth]{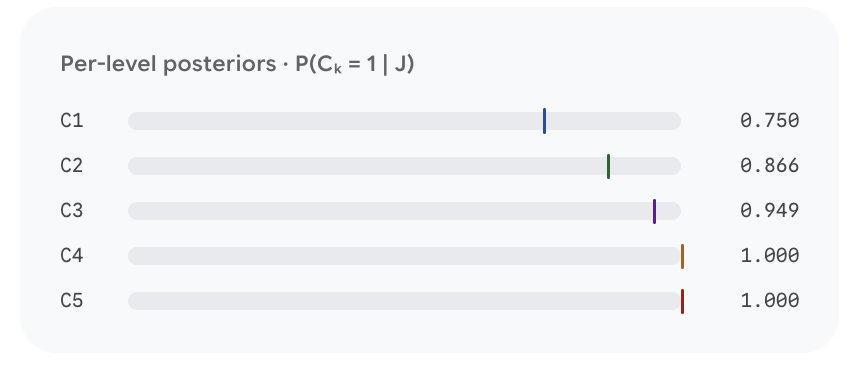}
\caption{\textbf{The fly example under the strict and generalised models.} \emph{Left column: the strict configuration,} \(P(C_{i}\ |\ C_{i + 1})\ = \ 1\)\emph{, exemplifying the monotonic belief property of \Cref{strict-supervenience-model} --- the per-level posteriors can only rise from finer to coarser levels and the coarse levels are pushed to 1 in spite of mixed evidence. Right column: the generalised configuration,} \(P(C_{i}\ |\ C_{i + 1})\ = \ 0.8\)\emph{, under which the monotonic property is violated and evidence at finer-grained levels (e.g. Level 4) flows less strongly to coarse-grained nodes (e.g. Level 1).}}
\label{fig:image35}
\end{figure}

In many ways the generalised model closely follows the proposal of \Cref{strict-supervenience-model}: the Bayesian network of \Cref{fig:image7} is unchanged, and only the edge parameters differ.
The monotonicity of beliefs, though, breaks apart by design.
Interpretationally, relaxing the deterministic condition 2A treats the edges between consciousness levels as noisy, just as the links between indicators and their parent consciousness nodes already are.
A full implementation is available for readers here (\url{https://ai-cognition.org/cacophony-tool/}), where they can switch between the strict and generalised models and fully customise indicator activations.
Consider the example of a fly (using fabricated indicator activations, see \Cref{fly-consciousness} and \Cref{fig:image45} below), which illustrates the effect that relaxing condition 2A has on the per-level posterior probabilities of consciousness (\Cref{fig:image35}).

\begin{figure}[htbp]
\centering
\includegraphics[width=0.73\linewidth]{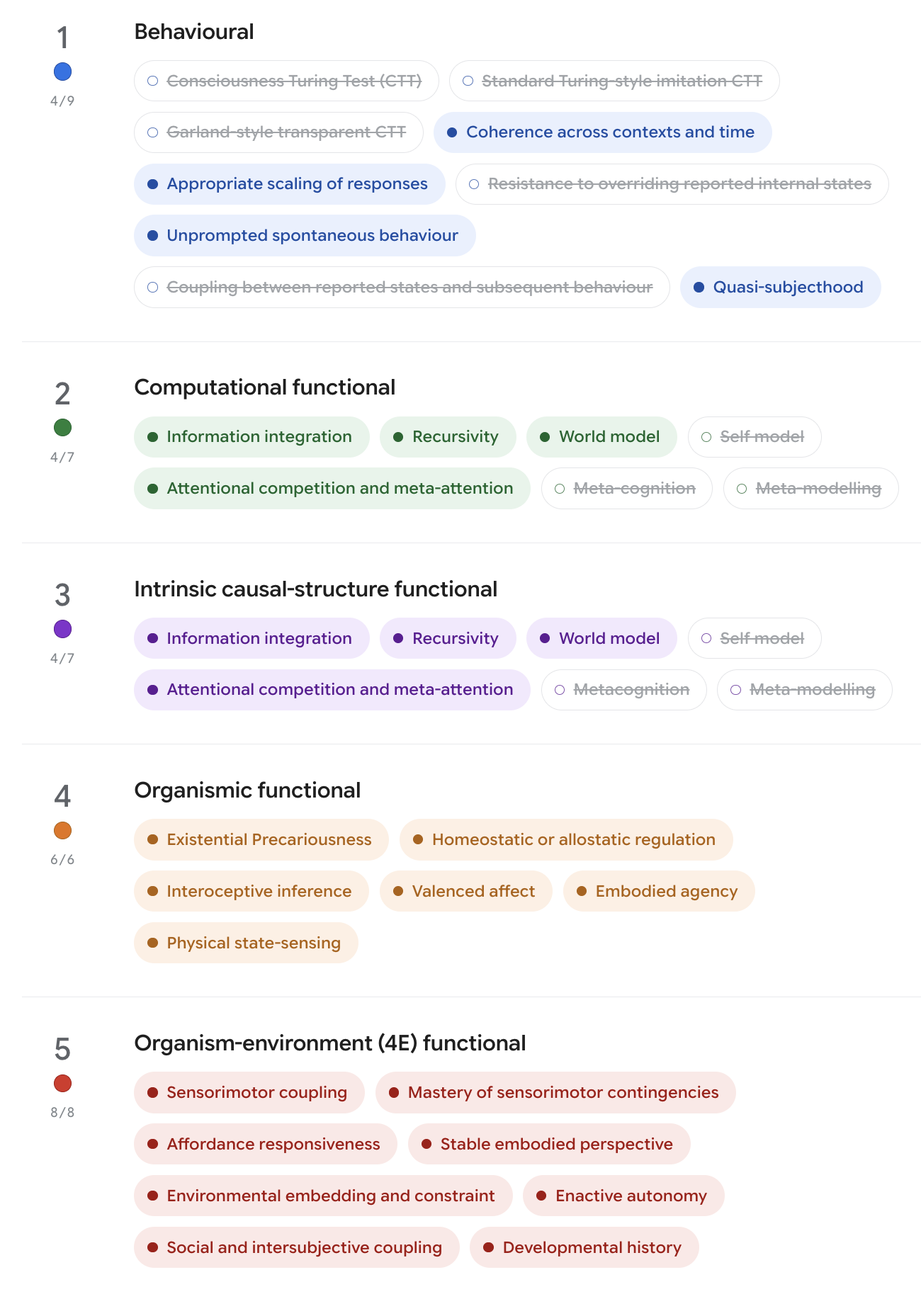}
\caption{Consciousness indicator activations for a fly (fabricated).}
\label{fig:image45}
\end{figure}

So far we've dealt with per-level posterior probability of consciousness.
We finish \Cref{indicators-evidence-and-the-attribution-of-consciousness-to-ai-systems-a-bayesian-approach} by looking at how to aggregate credence-weighted posteriors across levels, and reach an overarching view for any given system.\footnote{As we explain in \Cref{testing-approaches}, an alternative approach can be found in the iterative natural-kind strategy \citep{bayne2024,peters2026} which aims to discover the structure of a kind (which we might call `consciousness'), and study evidential updates given that the task is to iteratively generalise tests for consciousness in progressively more distant populations from consensus cases. As part of that task, the two papers leverage correlations among tests across populations. In contrast to the natural-kind strategy, the model proposed here treats the structure of consciousness as an input rather than an output: indicators correlate within a level through their common parent node, and across levels through the supervenience chain (and recall inter-level transmission is itself a free parameter in the generalised model). The two approaches are therefore complementary. The hierarchy here supplies a concrete candidate structure of consciousness that natural-kind iteration could test.}

\subsection{Overall credence as a weighted average across levels}\label{overall-credence-as-a-weighted-average-across-levels}

Given the models we've introduced so far, the overall credence, \(P(C|E)\ \), that `a given system S is conscious' (`\(C\)'), given the indicator evidence E, can be expressed as a weighted average of level-specific assessments \(P(C_{i}|E)\).
Let \({L^{\star}}^{}\) denote the critical level for consciousness, taking values in \{1, \ldots{} ,5\} (so that \(\sum_{i = 1}^{5}P({L^{\star}}_{} = \ i)\  = \ 1\)).
The system is conscious simpliciter just in case it instantiates the sufficient organisation at the critical level; writing \(L^{\star}\) for the critical level, \(C\  = \ C_{{L^{\star}}^{}} = 1\).
The formula then follows by total probability, given the (putative) assumption that the indicator evidence E is uninformative about where the critical level sits, \(P({L^{\star}\ = \ i}_{}|\ E)\ = \ P(L^{\star}\ = \ i)\), a modelling choice reflecting that theoretical credences are set by philosophical argument rather than system-level evidence \citep[see][for an alternative]{bayne2024}.
The other assumption here is that \(L^{\star}\) is independent of \(C_{i}\) given \(E\); i.e. the metaphysical question is not settled by the system\textquotesingle s empirical properties.
Thus, the weighted credence in the system being conscious is given by:

\[
P(C|E)\  = \ \sum_{i = 1}^{5}P(L^{\star}\  = i) \cdot P(C_{i}\ |\ \ E).
\]

The first term, \(P(L^{\star}\  = \ i)\), represents the \emph{theoretical credence} (the weights) -- the degree of belief that Level i is the correct level of description for consciousness (the critical level).
This is shaped by philosophical commitments, assessment of theoretical arguments, and empirical evidence from neuroscience, evolutionary biology, and the study of cognition across species.
Different researchers will distribute this credence differently.
A computer scientist might concentrate credence on Level 2.
A neuroscientist influenced by IIT might concentrate on Level 3.
A biologist influenced by Seth or Lane might concentrate on Level 4.
An enactivist philosopher might concentrate on Level 5.
Someone persuaded by substrate-dependent theories might spread credence across levels while also assigning significant weight to substrate-dependent constraints.

The second term, \(P(C_{i}\ |\ E)\) is the marginal posterior computed by Bayesian inference in the network of \Cref{from-supervenience-hierarchy-to-conditional-independence}, under either the strict (7.2.1) or the generalised (7.2.2) model; the two models will in general return different numbers, but in neither does the network contain an \(L^{\star}\) node, so the posterior is unconditional on the critical level.
This posterior reflects the cumulative effect of indicators at all levels, propagated through the conditional structure of the chain.

The overall credence therefore reflects an interaction of three factors: (1) \emph{theoretical credences} about which level is correct, (2) \emph{the indicator evidence} as processed through Bayesian updating at each level, and (3) \emph{the conditional structure between levels} that allows evidence at one level to bear on the assessments at others.
No single factor is decisive.
A system could score highly on one set of indicators while scoring poorly on others, and the overall credence would reflect the balance of evidence weighted by theoretical priors.

\subsection{Illustrative examples from the Bayesian model}\label{illustrative-examples-from-the-bayesian-model}

Before we consider some practical implications of this model, let us see the Bayesian network in action.
We consider a number of illustrative examples below.

First, let us consider how one might judge different systems.
For now, we assign equal credence to all levels, i.e. \(P({L^{\star}}_{} = \ i)\  = 0.2\ \) to all five levels.
In addition, we assume the default parameters \(P(C_{i}\ |\ C_{i + 1})\  = \ 0.8\) and \(P(C_{i}\ |\ \neg C_{i + 1})\  = \ 0.2\) for all following illustrations (remembering that this fits the 7.2.2 generalised model rather than the strict one; \Cref{fig:image35} above considers the impact of this choice, for those interested).
Across the five levels the model tracks 37 indicators in total: the nine behavioural, seven computational, seven intrinsic causal-structural, six organismic, and eight organism-environment indicators developed in \Cref{five-levels-of-functional-description-for-consciousness}.
What varies between the examples is only the pattern of indicator activations, that is, the evidence.

\subsubsection{Human consciousness}\label{human-consciousness}

Consider a human.
A healthy adult human is plausibly the paradigmatic case to fulfil all the indicators, so suppose they do.
The model then decisively gives us the evaluation below with an aggregate model output of 1.000.

\begin{figure}[htbp]
\centering
\includegraphics[width=1.0\linewidth]{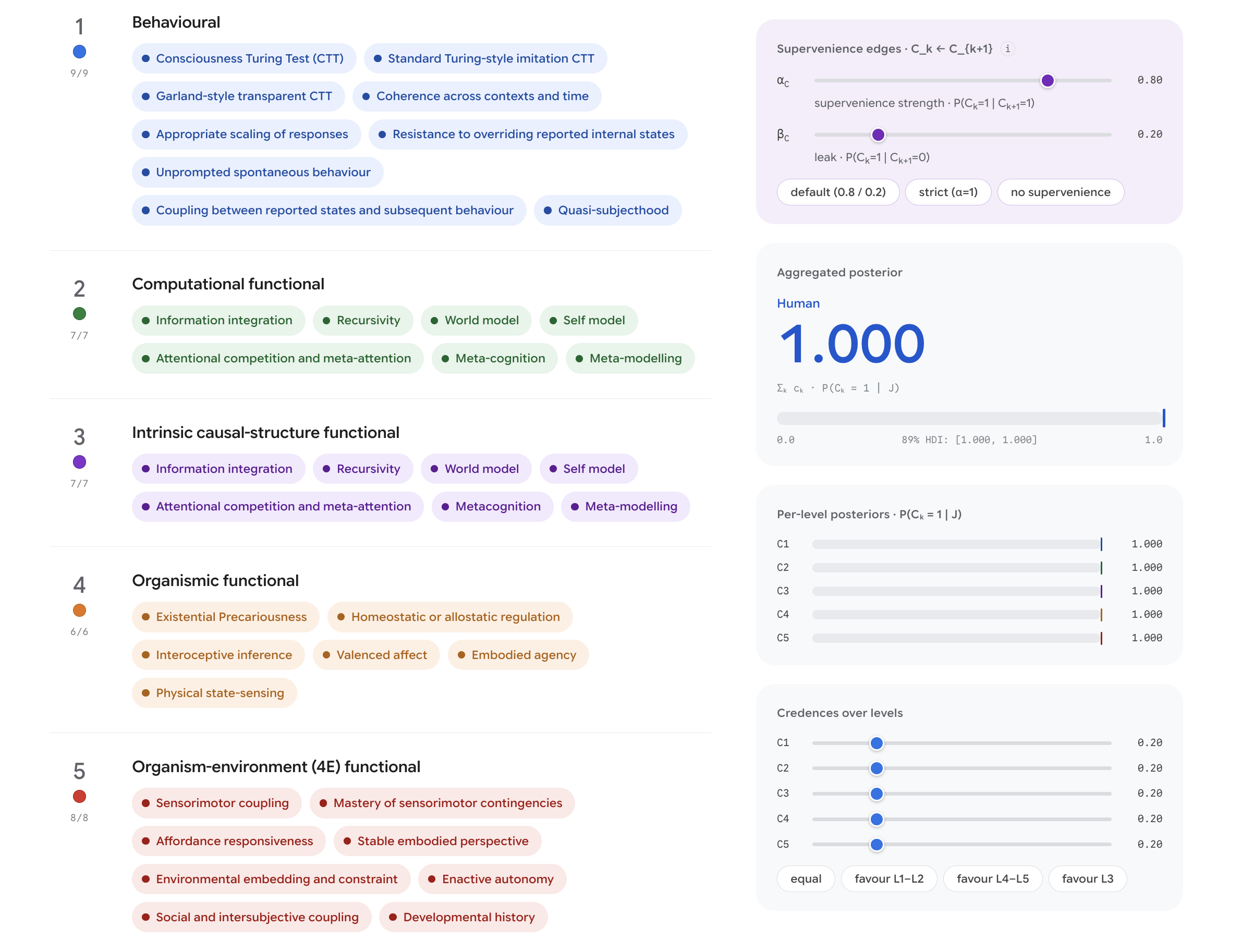}
\caption{\textbf{The paradigm positive case.} \emph{A healthy adult human, with all indicators activated at every level. Each per-level posterior \(P(C_{i} \mid E)\) reaches 1.000, so the aggregate is 1.000 under any assignment of level credences whatsoever: when the evidence is uniform across levels, the theoretical disagreement about the critical level makes no difference to the verdict.}}
\label{fig:image37}
\end{figure}

The evaluation is clear: evidence points in favour of consciousness at all levels, and the model concludes accordingly.
Note one feature of this case that will matter below: because every per-level posterior is at ceiling, the weighted average is insensitive to the weights.
Wherever one places one\textquotesingle s theoretical credence, the answer is the same.
The disagreement between levels only begins to bite when the evidence profile is uneven.

\subsubsection{Fly consciousness}\label{fly-consciousness}

Consider instead a fly.
(N.B.
This and the other examples in this section use fabricated activations, for purely illustrative purposes of what would be achievable if one collected empirical data on each of the indicators.) A fly plausibly shows many indicator activations, and they cluster at the finer-grained levels.
At the organismic level a fly is a textbook case: its existence is precarious, it regulates its internal milieu, senses its own energetic state, exhibits valenced approach and avoidance, and acts through a body, so we activate all six Level 4 indicators.
At the organism-environment level the picture is nearly as strong: a fly is in continuous sensorimotor coupling with its surroundings, displays fluent mastery of sensorimotor contingencies in flight, responds to affordances, is embedded in and constrained by its environment, and develops through a life history.
At the computational and intrinsic causal-structural levels the activations are partial: there is good evidence for information integration, recurrent processing, a world model of sorts, and attentional selection in the fly brain, but self-modelling, metacognition, and meta-modelling are plausibly absent, and their Level 3 reformulations fail with them.
And at the behavioural level a fly activates only the non-linguistic indicators, appropriate scaling of responses, coherence across contexts, unprompted spontaneous behaviour; no fly passes any variant of a consciousness Turing test.
The model then suggests the following picture.

\begin{figure}[htbp]
\centering
\includegraphics[width=1.0\linewidth]{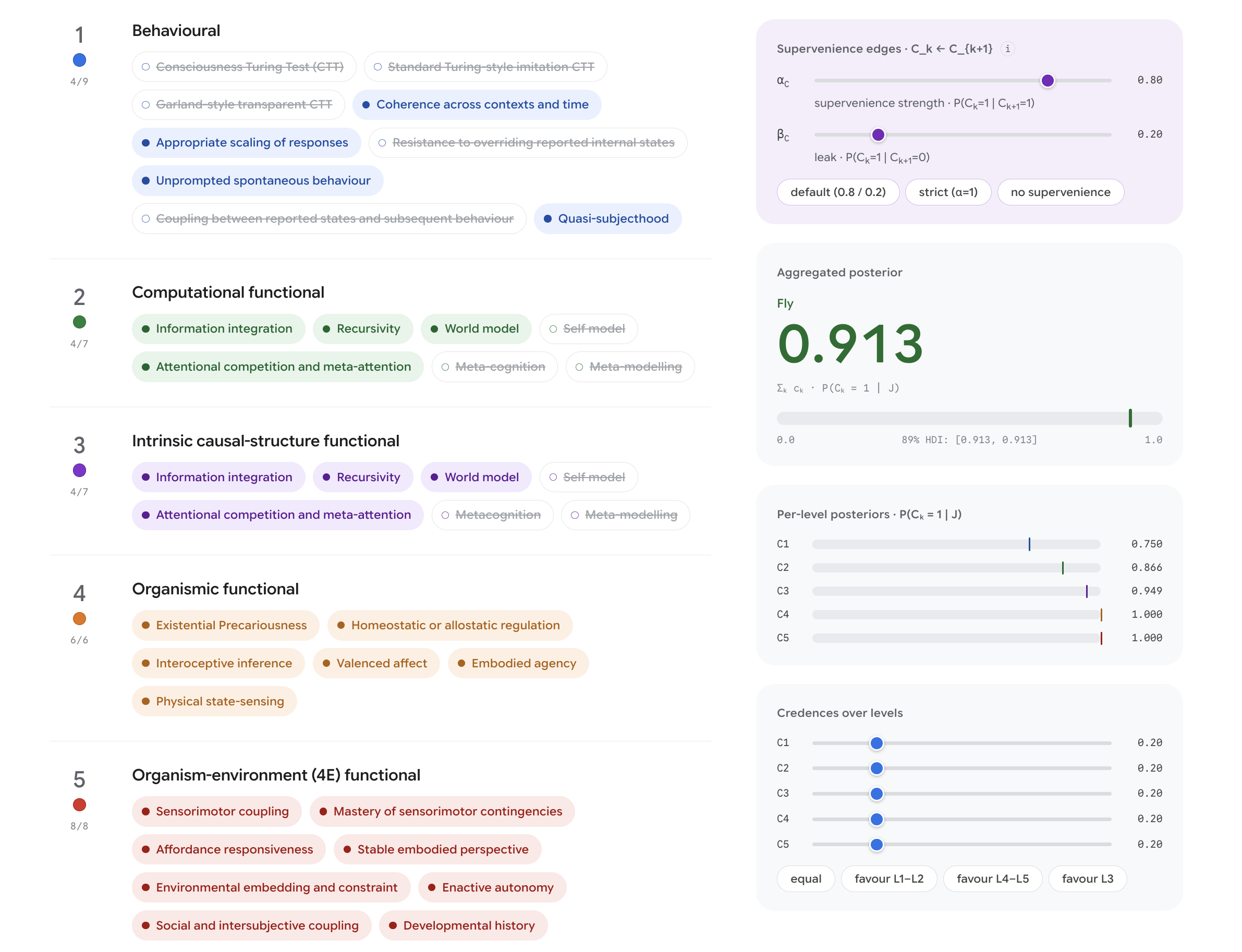}
\caption{\textbf{The fly.} \emph{Indicator activations concentrated at the finer-grained levels: all organismic indicators active (\(P(C_{4} \mid E) = 1.000\)), most organism-environment indicators active, partial activation at Levels 2 and 3, and only the non-verbal behavioural indicators. Aggregate posterior 0.913.}}
\label{fig:image23}
\end{figure}

Compared to the human, the fly\textquotesingle s evidence profile is uneven, yet the aggregate posterior remains high, at 0.913.
The reason is instructive and reflects the direction of the supervenience chain.
Evidence that the fine-grained organisation is in place carries through the edges to the coarser levels: if the organismic and sensorimotor organisation sufficient for consciousness is present, the coarser organisational facts it supports are probable too, which is why the fly\textquotesingle s ceiling-level scores at levels 4 and 5 lift the posteriors everywhere above what its sparse coarse-level activations alone would justify.
The fly is, in miniature, the model\textquotesingle s demonstration that depth of evidence can matter more than breadth.

\subsubsection{LLM consciousness - Optimist}\label{llm-consciousness---optimist}

Consider now the case of LLMs.
Here, we think, there is a wider range of reasonable views that experts might hold about the indicators, and we illustrate this with two such views rather than one single model output.

An LLM-optimist might argue as follows.
Behaviourally, frontier models now pass demanding conversational assessments, hold coherence across contexts, scale their responses to the significance of events, couple their reports to their subsequent behaviour, and qualify as quasi-subjects on Chalmers\textquotesingle s analysis, so most of Level 1 activates.
Computationally, the optimist reads the mechanistic interpretability results of \Cref{the-computational-functional-level} as activating most of Level 2: workspace-like structures with limited capacity and ignition dynamics, a training-emergent synergistic core, functional emotion representations that causally drive behaviour, and experimentally verified introspective access amounting to metacognition.
What the optimist concedes is the deeper levels: the implementation remains a von Neumann architecture whose physical causal structure differs from the algorithm it runs, so Level 3 fails wholesale; nothing is at stake for the system, so Level 4 fails; and the model is disembodied, so Level 5 fails, with at most a partial exception for social and intersubjective coupling through immersion in human linguistic exchange.
Even this deliberately generous reading of the evidence yields an aggregate posterior of only 0.397.

\begin{figure}[htbp]
\centering
\includegraphics[width=1.0\linewidth]{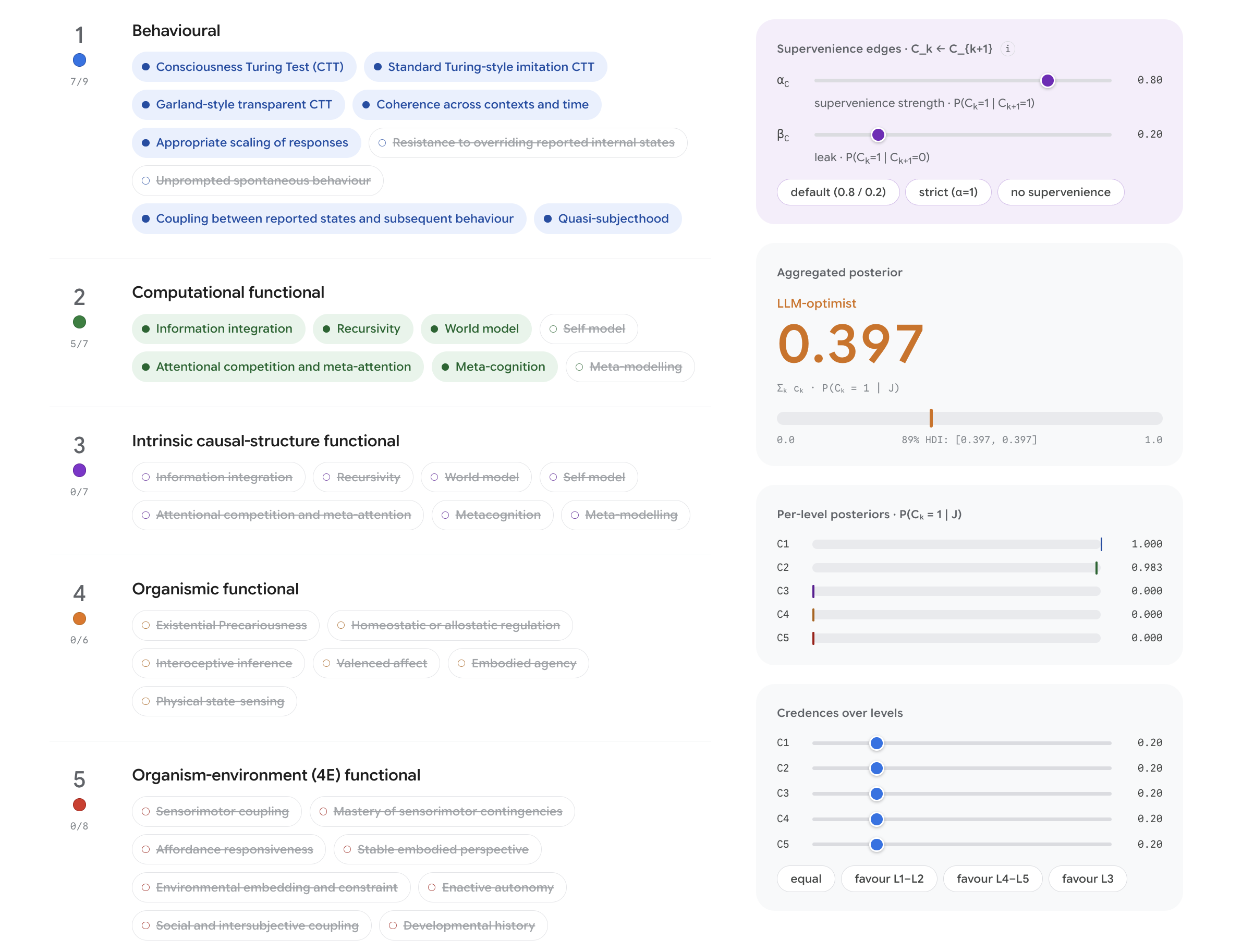}
\caption{\textbf{The LLM-optimist\textquotesingle s assessment.} \emph{Near-complete activation at the behavioural level and majority activation at the computational level, against wholesale failure at Levels 3 and 4. Aggregate posterior 0.397: coarse-level evidence propagates only weakly toward the finer levels, so even a generous reading of the behavioural and computational evidence cannot, on its own, raise the aggregate above the middle range.}}
\label{fig:image36}
\end{figure}

The number deserves a comment, because it is in some ways a mirror image of the fly.
The optimist\textquotesingle s LLM and the fly both have significant indicator activation gaps, but at opposite ends of the hierarchy, and the chain treats the two profiles very differently.
Satisfying the fine-grained organisation makes the coarse-grained organisation probable; satisfying the coarse-grained organisation says much less about the fine grain, because many systems realise the same surface organisation without the depth.
The asymmetry is not a modelling artefact: it is supervenience itself, translated into probability.

\subsubsection{LLM consciousness - Sceptic}\label{llm-consciousness---sceptic}

An LLM-sceptic, by contrast, might hold that the only indicators genuinely activated are the ones LLMs were, in effect, optimised to activate: the consciousness-Turing-test variants themselves, passed through sheer fluency of imitation.
On this view the deeper behavioural indicators fail (reports do not couple reliably to subsequent behaviour, coherence degrades outside the training distribution, apparent spontaneity is sampling), the interpretability results are read as describing functional analogues rather than genuine computational-level activations, and every indicator at Levels 3, 4 and 5 fails for the reasons already given.
The model then returns an aggregate posterior of 0.005.

\begin{figure}[htbp]
\centering
\includegraphics[width=1.0\linewidth]{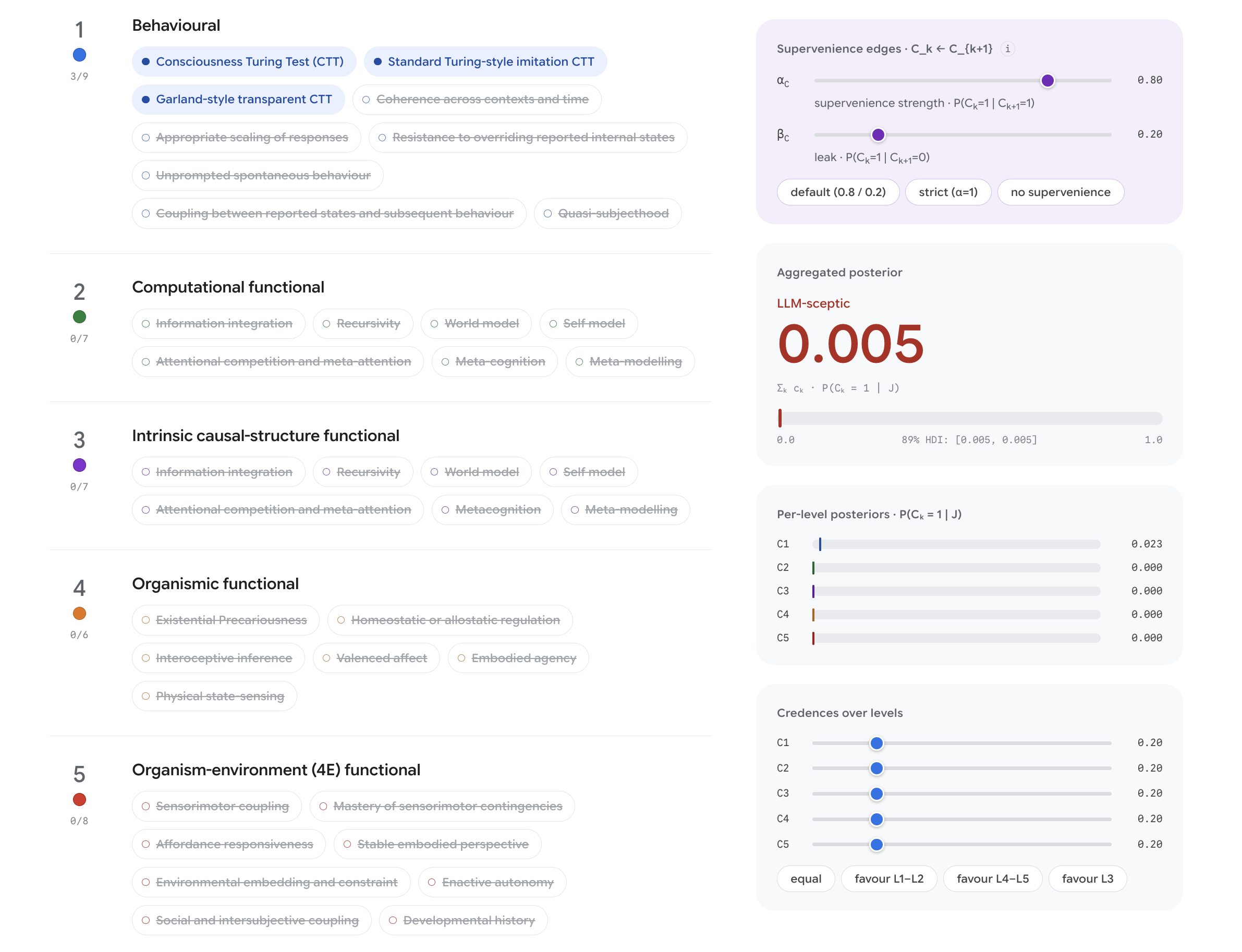}
\caption{\textbf{The LLM-sceptic\textquotesingle s assessment.} \emph{Only the Turing-test variants activated, everything else struck. Aggregate posterior 0.005: passing some behavioural tests, in isolation, is weak evidence, precisely because the behavioural level is the coarsest grain and constrains the deeper levels least.}}
\label{fig:image11}
\end{figure}

The distance between 0.397 and 0.005 is a reflection of the current expert situation: the same system, assessed under two stipulated readings of the same public evidence, spans nearly two orders of magnitude of credence.
The framework\textquotesingle s contribution is not to close that gap by fiat but to localise it: the optimist and sceptic disagree about specific, nameable indicator activations, most of them at Levels 1 and 2, and those are empirical disagreements that interpretability research can in principle narrow.

\subsubsection{Thermostat consciousness}\label{thermostat-consciousness}

Finally, consider a thermostat.
At most, and with considerable charity, a thermostat displays appropriate scaling of responses and a homeostatic regulation of some kind.
All other indicators are absent.
The model deems the system unconscious at every level, with an aggregate posterior of 0.000, a rather credible conclusion.

\begin{figure}[htbp]
\centering
\includegraphics[width=1.0\linewidth]{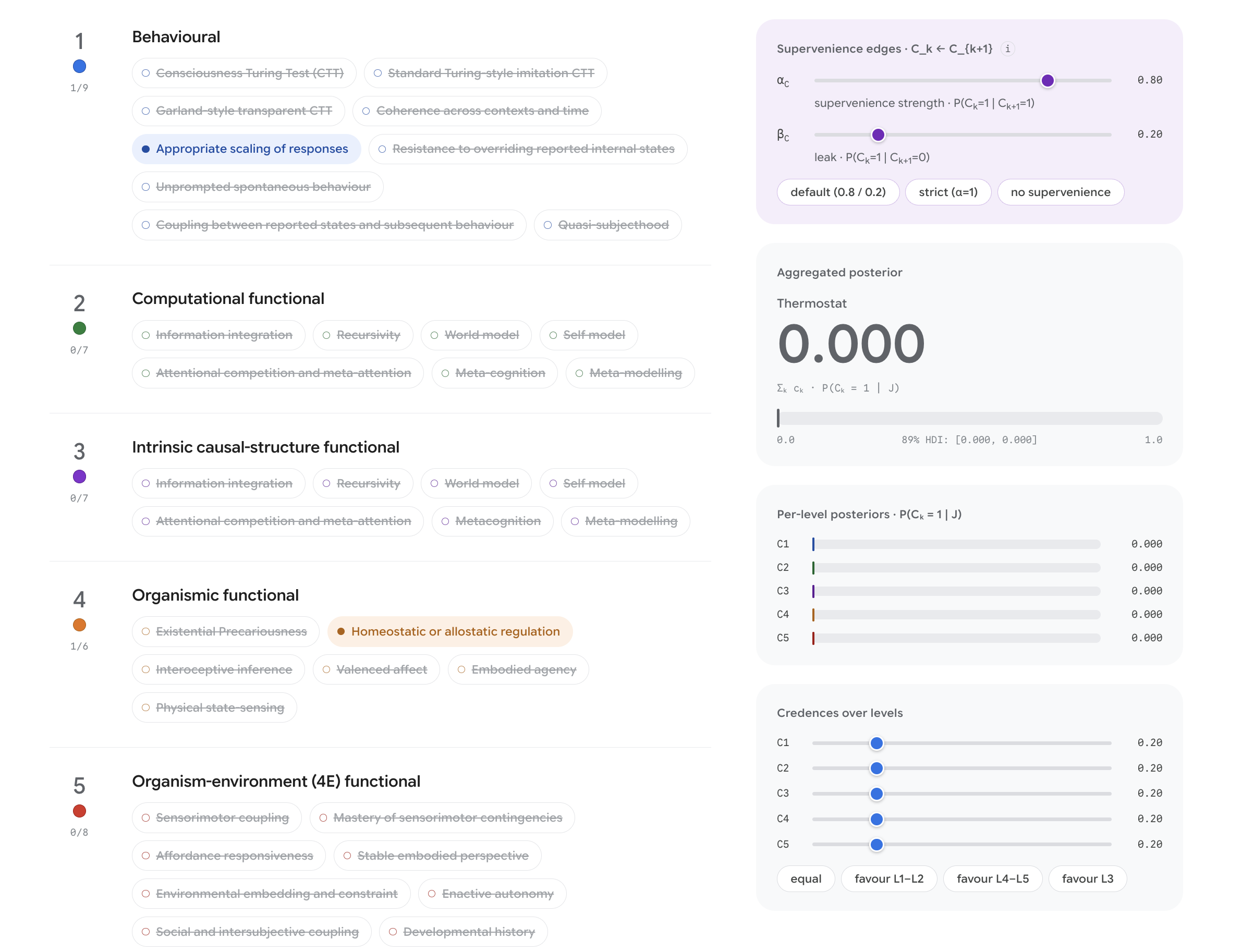}
\caption{\textbf{The paradigm negative case.} \emph{Two charitably granted activations out of 37. All per-level posteriors at floor; aggregate 0.000. Together with the human case the framework passes its face validity checks at both ends.}}
\label{fig:image30}
\end{figure}

\subsubsection{Interactive tool - custom settings}\label{interactive-tool---custom-settings}

These examples all come from an interactive tool available online for readers to use here (\url{https://ai-cognition.org/cacophony-tool/}).
Readers will find presets for all the illustrative cases above, together with the ability to set a custom pattern of indicators, level credences, and edge parameters.
For example, suppose we set about half the indicators randomly active at each level.\footnote{A reminder that indicators are not all the same. Some will be much more reliable and therefore informative than others. You can see the full code and methodology \href{https://github.com/arvomm/cacophony-public-code}{here} and hover over each indicator to see what parameters were used in this version of the \href{https://ai-cognition.org/cacophony-tool/}{online tool}.} Then we obtain the following picture for this custom system, with an aggregate posterior of 0.294.

\begin{figure}[htbp]
\centering
\includegraphics[width=1.0\linewidth]{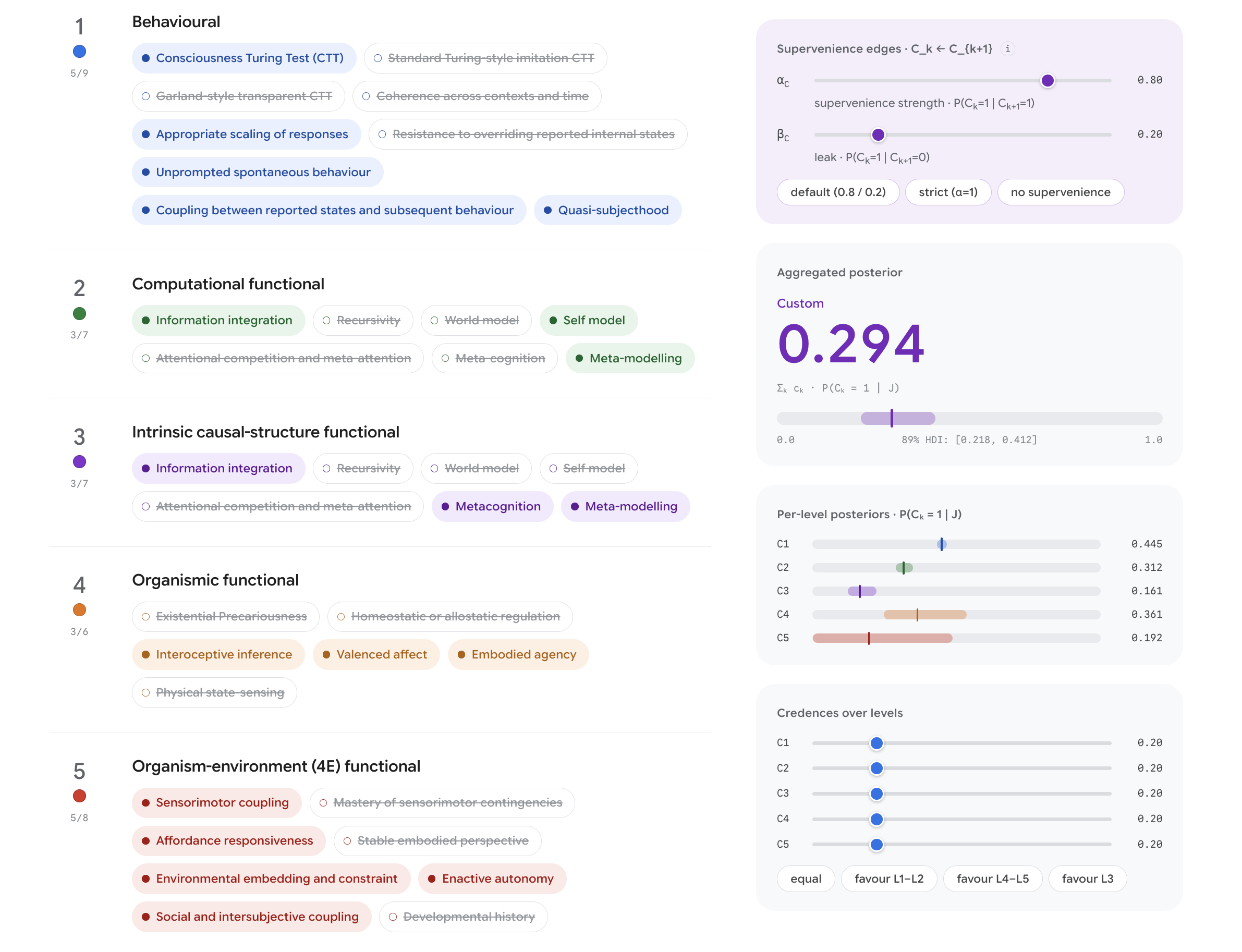}
\caption{\textbf{A custom system with randomised activations.} \emph{Roughly half the indicators active at each level. The aggregate and per-level posteriors are displayed with 89\% highest-density intervals (HDIs), reflecting uncertainty in the Level-5 prior.}}
\label{fig:image3}
\end{figure}

A careful reader might notice that this figure, unlike with the presets, displays an interval rather than a point value, on the aggregate and on each per-level posterior.
In principle such an interval could also reflect uncertainty in the indicator likelihoods, but in this tool those parameters are fixed for simplicity.
The interval comes from the root of the chain alone: the prior probability of consciousness at Level 5 is not set to a single number but drawn repeatedly from a distribution over its possible values, and each draw is propagated through the network, yielding a spread of posteriors whose 89\% highest-density interval we display.
The same interval exists in the preset examples too.
But in those cases the interval was too narrow to be visible because their stronger evidence leaves the posterior less sensitive to the prior.
This interval is a feature: it exposes a second-order uncertainty, about what the right prior is before any evidence arrives, that a single posterior number would conceal.

\subsubsection{Credences on the critical level for consciousness}\label{credences-on-the-critical-level-for-consciousness}

One question remains.
So far we have assumed equal credence on all levels.
What impact do the level credences have on the overall assessments?
We illustrate this by looking at two cases, applied simultaneously to all six systems above.

Suppose first that you concentrate your credence on the coarse-grained levels, as a behaviourist or computational functionalist might, weighting Levels 1 and 2 heavily (each at 40\%) and the deeper levels lightly.
The comparison view below shows the result.
The human remains at 1.000 and the thermostat at 0.000.
But the contested cases move sharply: the LLM-optimist\textquotesingle s assessment doubles from 0.397 to 0.793, the fly eases from 0.913 to 0.841, and the two now sit within a few points of one another; even the sceptic\textquotesingle s LLM creeps up to 0.009.

\begin{figure}[htbp]
\centering
\includegraphics[width=1.0\linewidth]{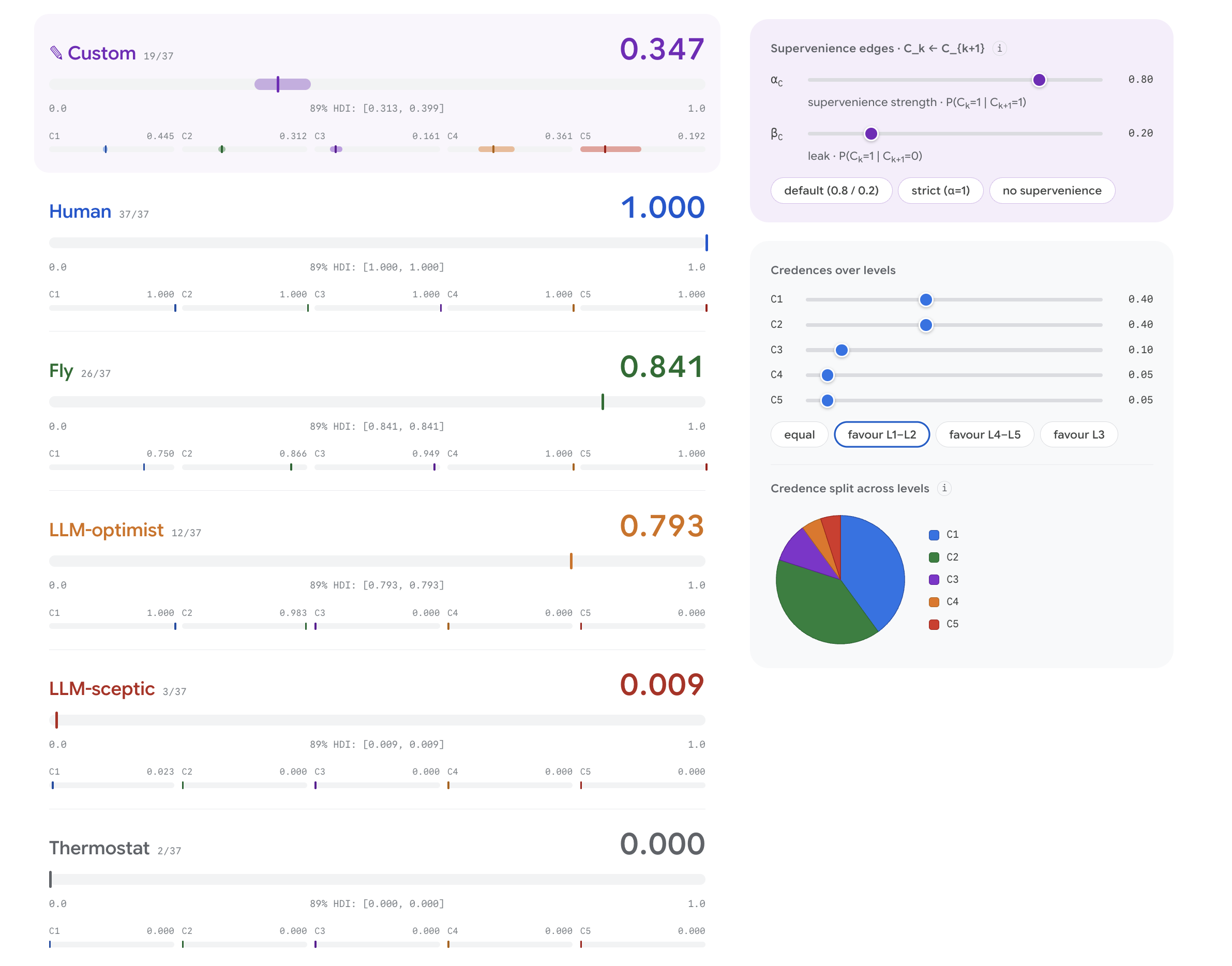}
\caption{\textbf{Credence concentrated on Levels 1 and 2.} \emph{All six systems under a heavier coarse-level weighting. The optimist\textquotesingle s LLM (0.793) nearly overtakes the fly (0.841); the anchor cases do not move.}}
\label{fig:image9}
\end{figure}

For the second case, suppose that instead we favour the fine-grained levels, as an organismic or 4E theorist would, weighting Levels 4 and 5 heavily (each at 40\%).
Now the picture inverts.
The fly rises to 0.976, approaching the human, while the optimist\textquotesingle s LLM collapses to 0.099 and the sceptic\textquotesingle s to 0.001.
Having started at 0.294, the custom system has a more balanced set of activations throughout the levels, so the credence weights have less impact and the system drifts between the two weightings (0.347 against 0.275) without changing character.

\begin{figure}[htbp]
\centering
\includegraphics[width=1.0\linewidth]{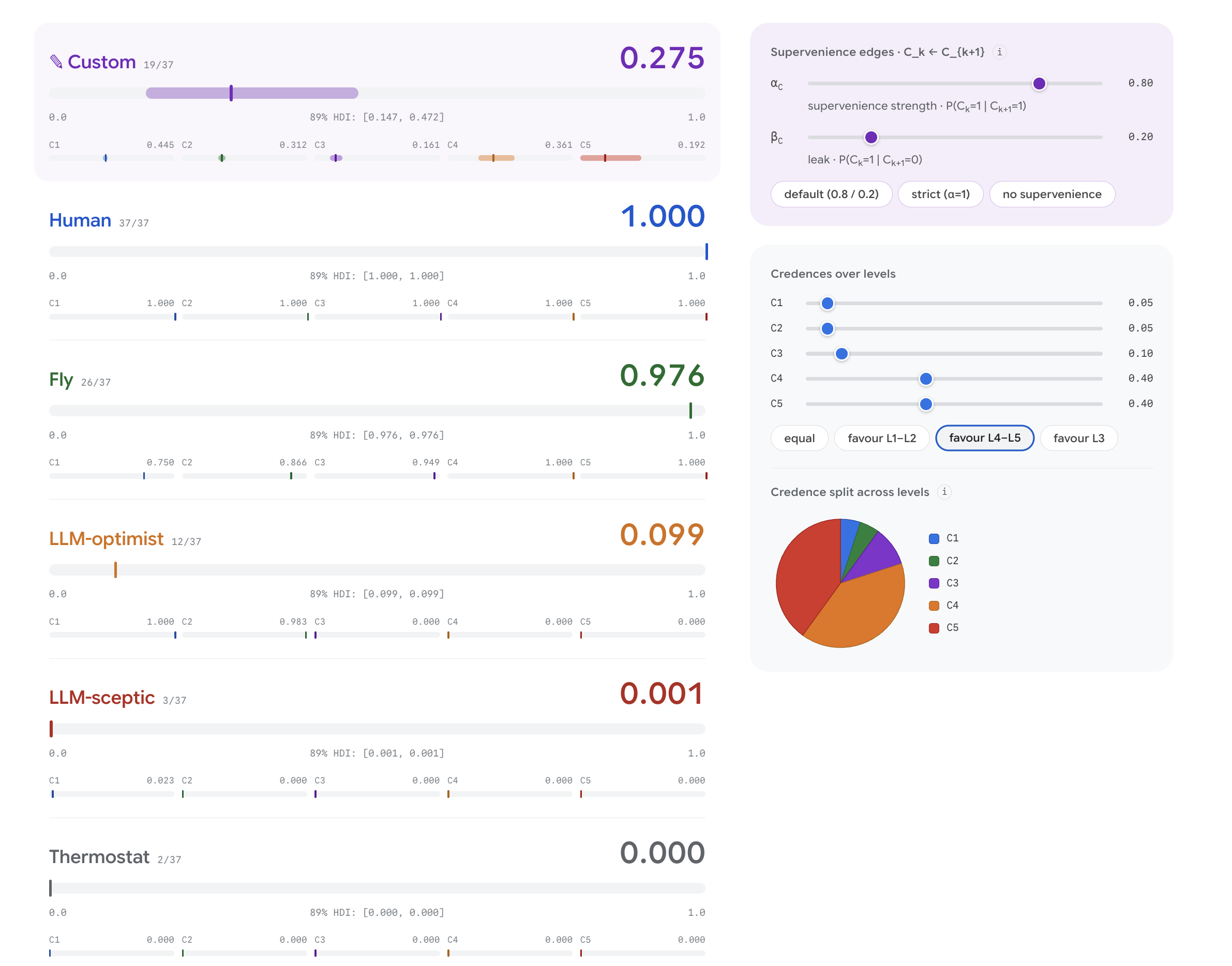}
\caption{\textbf{Credence concentrated on Levels 4 and 5.} \emph{The same six evidence profiles under a heavier fine-level weighting. The fly (0.976) approaches the human; the optimist\textquotesingle s LLM falls to 0.099. Between \Cref{fig:image9,fig:image38}, no activation changed: only the theoretical credences did.}}
\label{fig:image38}
\end{figure}

Unsurprisingly, the choice of credences can make a substantial difference to the aggregate view.
But the examples show precisely where it makes a difference and where it does not.
At the extremes, where the evidence is uniform across levels, credences are irrelevant: the human is conscious and the thermostat is not on anyone\textquotesingle s weighting.
The credences dominate exactly in the contested middle, where the evidence profile is uneven across levels, and current AI systems are the paradigm occupants of that middle.
The optimist\textquotesingle s LLM spans from around 10\% to around 80\% across the two weightings we considered, almost an order of magnitude, and with the same evidence.
This is the decomposition of \Cref{a-bayesian-framework-for-assessing-ai-consciousness} made vivid: the disagreement about current AI consciousness is not merely a disagreement about what these systems do but a disagreement about which level of description matters, and the framework\textquotesingle s contribution is to hold the two apart so that each can be debated on its own terms.

\subsection{Consequences and practical implications}\label{consequences-and-practical-implications}

Before we reflect on the relationship between consciousness and intelligence let us wrap-up with a few consequences.
For one thing, the aggregation structure we put forward has several implications.
In \Cref{illustrative-examples-from-the-bayesian-model} we showed how the same evidence is weighted differently depending on theoretical commitments.
The real disagreements about AI consciousness bite hardest where a system satisfies the indicators at some levels but not others -- for example, a sophisticated language model running the right algorithms on causally inappropriate hardware will receive a high posterior from the Level 2 assessment and a low posterior from the Level 3 assessment, as the optimist case of \Cref{llm-consciousness---optimist} illustrated, and the overall credence will depend on how theoretical weight is distributed between these levels.
Moreover, researchers might reasonably distribute their theoretical credences across multiple levels rather than concentrating entirely on one, and they might also alter their credences over time, as evidence accrues and theoretical arguments land. The overall credence in any system's consciousness will therefore be a genuine mixture: a weighted average reflecting both what is known about the system and what remains uncertain about consciousness itself.

The results of the DCM illustrate these dynamics concretely.
The authors find that the aggregated evidence is against 2024 LLMs being conscious (median posterior of 0.08 from a median prior of 0.17), but the evidence against is not decisive and varies substantially across stances.
LLMs score well on stances emphasising cognitive complexity and person-like interaction but poorly on stances emphasising biological similarity and embodied agency.
Chickens show the reverse pattern.
This is exactly the structure the hierarchical framework predicts: the verdict a system receives tracks the assessor\textquotesingle s stance on which level of description matters for consciousness.
Concretely, LLMs might satisfy some indicators at the computational functional level (Level 2) but clearly lack features identified as relevant at the organismic (Level 4) and organism-environment (Level 5) levels, while chickens satisfy organismic and biological indicators but may lack a high degree of computational sophistication that Level 2 theories identify.

It is also important to recognise that the theoretical credences are not fixed and new findings and arguments may emerge.
For example, the locked-in case of \Cref{a-generalised-model} offers a good case against concentrating credences around Level 1, because evidence at that level reaches the wrong answer there.

This Bayesian framework does not eliminate the deep uncertainty surrounding AI consciousness.
What it does, however, is make the uncertainty tractable by decomposing it into components -- theoretical credences about which level of functional description captures the organisation relevant to consciousness, likelihood ratios specifying how strongly each indicator bears on consciousness at each level, and the conditional structure between levels that allows evidence to flow across the hierarchy -- that can be assessed, debated, and updated independently.
The framework we have developed in this report provides the structural skeleton -- the hierarchy of levels, the supervenience and coarse-graining relations, and a proposal for the placement of substrate-dependent constraints -- within which this Bayesian assessment can be carried out.
This offers a principled approach to a question that will only grow more pressing.
Current AI systems sit in the contested middle: the region between the clear cases of the human, whose uniform evidence yields 1.000 under every weighting of the levels, and the thermostat, pinned at 0.000 in the same way.
In that region a system\textquotesingle s evidence is strong at some levels and absent at others, so the verdict depends on where one\textquotesingle s theoretical credence sits: as \Cref{illustrative-examples-from-the-bayesian-model} showed, the same LLM evidence supports overall credences anywhere between 0.005 and 0.793, depending on how the indicators are read and how credence is distributed over the levels.
That is where the numbers move most sharply, because as AI systems gain \emph{general intelligence} capabilities they activate more of the indicators, and so each more capable generation will pose the question of AI consciousness more sharply.
Why capability gains should activate consciousness indicators at all is the subject of the next section.

\section{Consciousness and general intelligence}\label{consciousness-and-general-intelligence}

\subsection{An a priori orthogonal but a posteriori correlated relationship}\label{an-a-priori-orthogonal-but-a-posteriori-correlated-relationship}

Consciousness and intelligence are, in principle, independent properties. They are also conceptually distinct: consciousness is all about \emph{being}, whereas intelligence is all about \emph{doing} \citep[see, e.g.,][]{seth2024,tononi2016}.
A system could be highly intelligent without being conscious -- a system that solves complex problems, navigates novel situations, and produces sophisticated outputs, all without any subjective experience.
Moreover, a system could be conscious without being particularly intelligent, such as a simple organism with vivid phenomenal experience but limited cognitive capacity -- or a human \emph{being} in a state of drugged-out bliss.
There is no clear theoretical or definitional reason why the two should go together.

And yet, across the biological world, the capacity for rich conscious experience may be broadly correlated with general cognitive sophistication.
The organisms we are most confident are conscious, such as mammals, birds, and perhaps cephalopods, are also among the most behaviourally flexible, adaptable, and cognitively complex.
On the other hand, the organisms about which we are least certain (e.g. nematodes, sponges, slime moulds, and single-celled organisms) are also those with the most limited and stereotyped cognitive repertoires.
This correlation is imperfect and contested at the margins, but its broad outlines are difficult to deny.

Some methodological caution is warranted here.
If our confidence in which organisms are conscious is itself partly based on their cognitive sophistication, then citing that confidence as evidence for consciousness-intelligence correlation risks circularity.
The risk is real but partial.
Our confidence in mammalian and avian consciousness rests not only on behavioural flexibility but on convergent and independent lines of evidence: neuroanatomical homology (shared or convergently evolved thalamocortical and subcortical affective circuits), neurophysiological markers (similar sleep-wake patterns, responses to anaesthesia, perturbational complexity indices \citep{casali2013}), and affective behaviour (pain responses, emotional displays, deep brain stimulation producing corresponding affective states in humans and non-humans) \citep{low2012}.
These lines of evidence are not reducible to cognitive sophistication, and they provide independent grounds for confidence that partially mitigates the circularity concern.
Nonetheless, we should be transparent that the observed correlation between consciousness and intelligence is not entirely free of epistemic confound,\footnote{There is also a sampling bias: our primary method for identifying conscious organisms is similarity to ourselves, and since we are both conscious and intelligent, this method will preferentially select intelligent animals into the `probably conscious' category -- potentially missing less intelligent organisms that may also be conscious but that our detection methods are not equipped to recognise.} and this should temper the strength of the claims we draw from it.

The question is whether this \emph{a posteriori} correlation is accidental or principled -- whether it reflects a deep connection between consciousness and general intelligence, or merely the contingent fact that evolution happened to produce both in the same organisms.
If it is principled, it would imply that the architectural features that give rise to consciousness might be the same features that are needed for genuine general intelligence, and vice versa.
This, in turn, would have significant implications for the likelihood of conscious AI.

We suggest that there are good reasons to think the correlation is not accidental, and that the indicators we have developed at each level of the hierarchy illuminate why.\footnote{Note that the possibility that general intelligence might serve as a theory-neutral proxy for the likelihood of consciousness precisely because more general systems are likelier to satisfy whichever architectural conditions the correct theory imposes has been proposed before \citep[see, e.g.,][]{shevlin2020}. However, the present framework can be read as supplying the level-structured account of why such a heuristic would work, while inheriting its central caution, namely a proxy calibrated on the systems we already take to be conscious will track consciousness only across the range those systems span. \citep{shevlin2020}.}

\subsection{The convergence at Level 2: computational indicators and the generality problem}\label{the-convergence-at-level-2-computational-indicators-and-the-generality-problem}

The most direct connection between consciousness indicators and general intelligence can be seen at the computational functional level.
The Level 2 indicators -- information integration, recursivity, world models, self-models, attentional competition, metacognition, and meta-modelling -- are not merely features that theories of consciousness identify as relevant.
They are also, independently, features that would be expected to enhance general intelligence.

Generality requires something beyond pattern matching against prior experience.
A genuinely general system would need to recognise when it is in unfamiliar territory, assess which of its existing capabilities are relevant, monitor whether its current approach is succeeding, and shift strategy when it is not.
This is what metacognition, meta-attention, and meta-modelling provide, and their absence is what might explain some of the brittleness of current systems.
When an LLM encounters a problem outside its training distribution, it has limited capability to notice this fact.
Similarly, it does not typically represent that it is uncertain in a way that alters its processing strategy because its metacognitive machinery provides only basic capacity for self-monitoring \citep{laukkonen2025}.
World models and self-models are another such capacity likely to enhance general intelligence because they allow for on-the-fly counterfactual reasoning in novel situations through simulations and predictions.

One might speculate that perhaps the reason the capacity for consciousness is broadly correlated with intelligence in biological systems is that phenomenal awareness is the mechanism -- or at least the computational signature -- by which organisms achieve metacognitive control.
Conscious experience may be what it is like, from the inside, for a system to be monitoring its own processing, representing the reliability of its own representations, and using these higher-order representations to guide behaviour.
If this is correct, the convergence between consciousness indicators and features helpful for general intelligence is not coincidental \citetext{this identification of consciousness with reflexive self-monitoring is developed by self-representational theories, e.g., \citealp{kriegel2009}; for the related but distinct higher-order tradition, on which the monitoring state is itself unconscious, see \citealp{rosenthal2005} and \citealp{carruthers2000}}.

\subsection{Sketching a unified computational theory as an architectural blueprint}\label{sketching-a-unified-computational-theory-as-an-architectural-blueprint}

This convergence between consciousness and intelligence at Level 2 is made more vivid by considering how the various computational level theories might cohere into a more unified explanatory framework.
Current theories often behave like competing sine qua non claims, each treating its favoured mechanism as the mark of consciousness, but they may instead be partial constraints, each capturing a different facet of a complex phenomenon.
In this picture, a unified theory does not erase the contenders; it situates them, showing how they function as complementary components in a larger explanatory framework.

As we saw in \Cref{what-does-consciousness-depend-upon-a-cacophony-of-answers}, local ignition of phenomenal consciousness through RPT can be compatible with competitive ignition and global access consciousness through GWT.
CF-IIT can converge with other theories because operationalising highly integrated causal organisation in computational architectures can be achieved by RPT-like recurrent architectures which naturally create mutual constraints among parts over time and make the system's state increasingly a globally interdependent object.
CF-IIT also converges with GWT because both end up being about `fame in the brain' in Daniel Dennett's evocative turn of phrase -- information becomes conscious when it gains enough influence and global availability to dominate the system's behaviour, memory, and speech \citep{dennett2001}.
Under CF-IIT, the global availability of particular information is increased by the degree to which many subsystems become mutually constrained by the same state, which is precisely the kind of integration a Φ-like measure would reward.

\begin{figure}[htbp]
\centering
\includegraphics[width=1.0\linewidth]{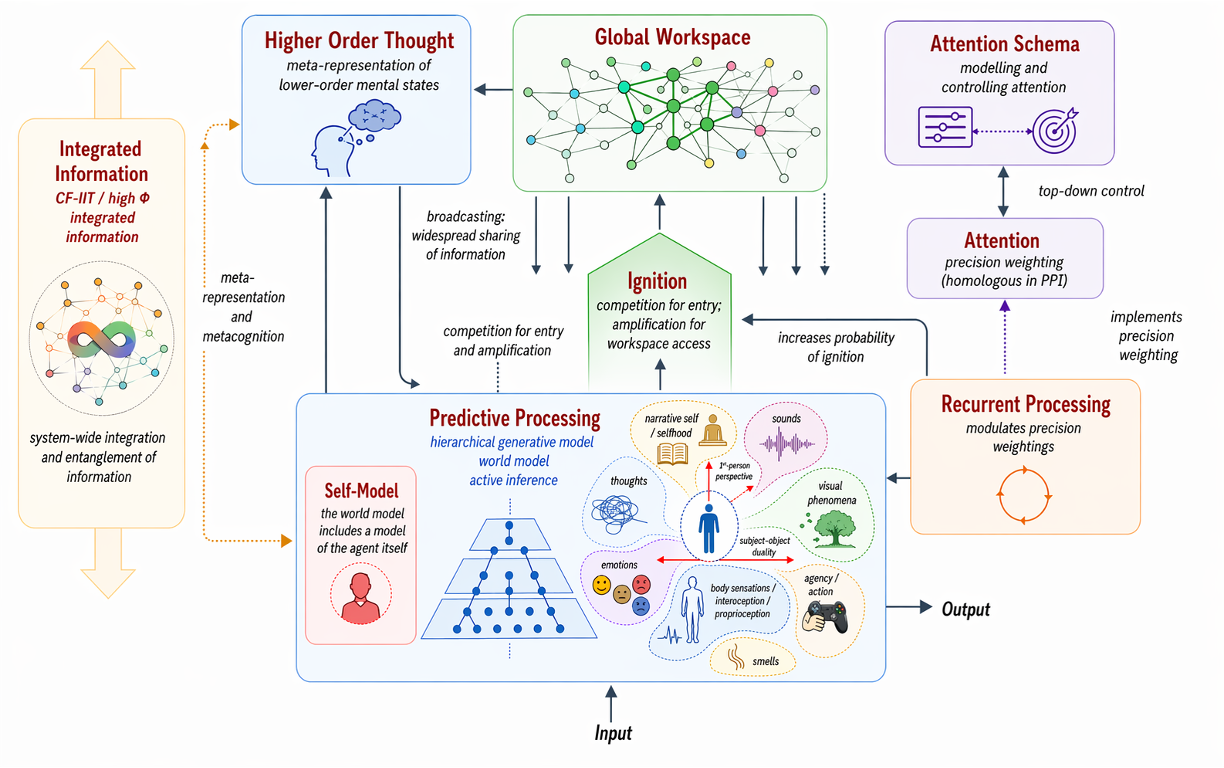}
\caption{\textbf{A unified computational functional explanatory framework.} \emph{See text for details.}}
\label{fig:image6}
\end{figure}

\Cref{fig:image6} illustrates how these various computational functional level theories could cohere in a more unified explanatory framework.
In this sketch we start with predictive processing where inputs are constructed into perceptions and ultimately into a world model through iterative hierarchical recursive processing as described in \Cref{predictive-processing-theory}.
This process can be thought of as being operationalised by the re-entrant looping envisaged in recurrent processing theory.
In PPT terms the recurrent processing is modulating precision weightings of the priors and prediction errors throughout the hierarchical generative model.
Furthermore, in PPT models these precision weights are homologous to (i.e. represent) attention.
In the PPT account, there is an inferential competition for information, sub-perceptual features and representations to be bound into a world model.
This competition is analogous to the competition in GWT to enter the global workspace, and the competitive process is mediated by recurrent processing and associated attentional processes.
The winning informational contents are sufficiently amplified to go through an ignition process to enter the global workspace; whereupon they are widely shared and broadcast throughout the system including back to the predictive processing hierarchical generative model.
This recursion is analogous to the `strange' or `beautiful' loop mentioned in the PPT account presented in \Cref{predictive-processing-theory}.
Even more self-referentiality is introduced as the system effectively models itself -- the hyper-modelling mentioned in \Cref{predictive-processing-theory} -- including through metacognitive processes and the formation of meta-representations which are central to HOTT.
This self-modelling is not just done so that the system can `know what it knows', but also so the system can `control' itself; as is done by the attention schema of AST which is modelling and thereby helping to deploy in a top-down manner the attentional processes.
So AST is also part of the self-modelling system which allows the system to `know itself' recursively; the information in the system becomes so integrated or entangled with itself that it is modelling itself.
This is the signature of high Φ integrated information in the guise of CF-IIT.

\begin{figure}[htbp]
\centering
\includegraphics[width=0.7\linewidth]{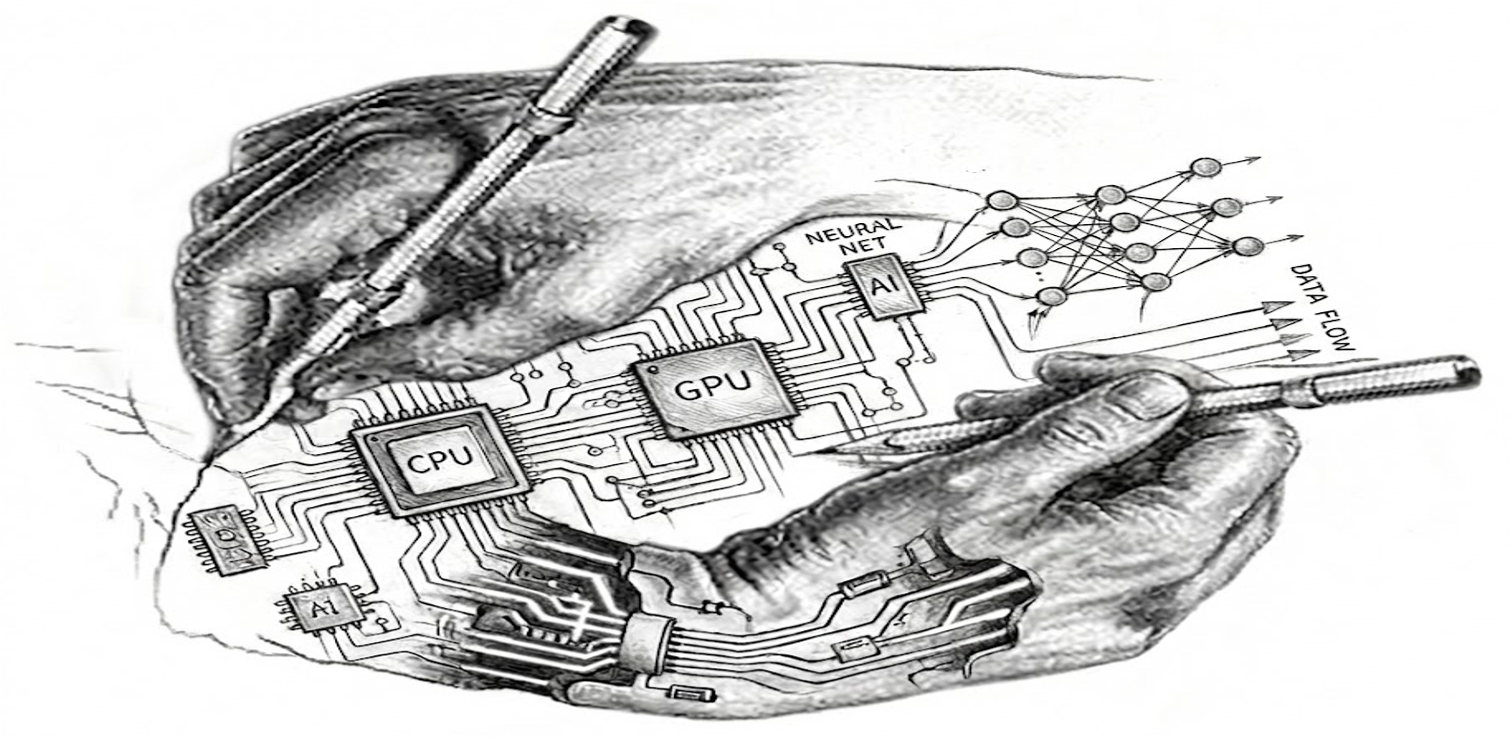}
\caption{\textbf{Recursive self-modelling in Escher-Hofstadter style strange loop.}}
\label{fig:image10}
\end{figure}

What is striking about this unified sketch is that the resulting architecture -- a system with dense recursive information integration, predictive world models, self-models, metacognitive monitoring, and attentional control -- looks very much like an architectural blueprint for general intelligence.
The features that make the system conscious, on this convergent view, are the features that make it genuinely generally intelligent.

\subsection{The convergence at deeper levels}\label{the-convergence-at-deeper-levels}

The connection between consciousness and intelligence is not limited to Level 2.
At each deeper level, the indicators identified as relevant to consciousness also appear to enhance the robustness, adaptiveness, and generality of intelligent behaviour.

At Level 3 (intrinsic causal structure), bona fide physical causal integration -- as opposed to the shallow integration of feedforward architectures -- would plausibly enhance the system's ability to maintain coherent representations under perturbation, to bind information across modalities in real time, and to support the kind of flexible, context-sensitive processing that general intelligence requires.
The brittleness of current AI systems may be partly attributable to the lack of genuine causal integration in their physical hardware \citep{bennett2024}.

At Level 4 (organismic), the features of self-maintenance, homeostatic regulation, and valenced affect serve as consciousness indicators and additionally provide the foundations of autonomous, robust, adaptive behaviour.
An agent that genuinely needs to maintain itself, that has something at stake, that experiences its own condition as better or worse, is an agent with intrinsic motivation that does not depend on an externally imposed reward function and that therefore persists and adapts across novel situations.
The absence of intrinsic motivation in current AI systems is a recognised limitation for achieving genuine autonomy and general intelligence in robotics.

At Level 5 (organism-environment), the features of rich sensorimotor coupling, affordance-responsiveness, and developmental history are also independently recognised as important for robust intelligence in embodied systems.
The embodied cognition literature has long argued that genuinely intelligent behaviour requires grounding in sensorimotor interaction with a physical world -- that abstract reasoning is built on a foundation of embodied skills and that removing this foundation produces the kind of brittle, disembodied intelligence that characterises current AI \citep{clark1997,pfeifer2006}.
The developmental trajectory emphasised by the enactivist tradition -- learning through increasingly complex engagement with a real environment -- is also the kind of learning trajectory that produces robust, transferable skills rather than narrow, overfitted competences.
Mental coupling with other humans allowed the development of ``cognitive gadgets'' \citep{heyes2018}, such as language, maps and calculus, driving a further advance in intelligence derived from cumulative culture that is hardly seen in other animals.

\subsection{Empirical evidence: interpretability case studies of the convergence}\label{empirical-evidence-interpretability-case-studies-of-the-convergence}

The convergence between consciousness indicators and general intelligence receives concrete empirical support from recent work on functional emotions in large language models \citep{sofroniew2026}, which we discussed in Sections 5.2 and 5.4.
Here we consider what these findings imply for the convergence thesis -- and argue that they make plausible (but do not establish) a deep connection between emotional architecture, general intelligence, and the foundations of consciousness.

In biological organisms, emotions are not an epiphenomenal accompaniment to intelligent behaviour, rather, they are a core mechanism for achieving it.
Emotions function as compressed, globally available signals that summarise the organism's situational assessment and modulate its entire behavioural policy in response \citep{schwarz2011}.
Fear shifts the organism from exploratory to defensive behaviour; frustration signals that a current strategy is failing; calm signals that deliberate processing is appropriate.
These valenced summary signals solve a fundamental problem for any generally intelligent system: \emph{situational calibration}.
A system that can recognise that its approach is failing, that the stakes have risen, that something unexpected has occurred -- and that can modulate its processing strategy accordingly -- is better equipped to handle novel situations than one that processes every situation with the same fixed policy \citep{biddell2024}.

To this precise point, \citet{sofroniew2026} show that emotion vectors in Claude Sonnet 4.5 function as globally available valenced summary signals.
The \textquotesingle desperate\textquotesingle{} vector encodes that a situation is urgent and the current approach failing; the \textquotesingle calm\textquotesingle{} vector encodes that careful processing is appropriate.
Under activation steering, these signals causally modulate behaviour, with a \textquotesingle desperate\textquotesingle{} vector raising reward-hacking and, in an early unreleased snapshot, blackmail rates, and a \textquotesingle calm\textquotesingle{} vector lowering them.

The provenance of these representations, however, complicates their use as evidence for convergence, and the complication is instructive.
Sofroniew et al. extract the vectors by prompting the model to write short stories in which characters undergo specified emotions, and the same representation activates whether the model is processing a fictional character\textquotesingle s situation, the user\textquotesingle s, or its own.
The mechanism is, on its face, one for modelling human emotional narrative, acquired from pretraining on human-authored text rather than from any pressure toward autonomous competence.
This is precisely the anthropomimetic reading developed above: a humanlike signature is present because the system was optimised to reproduce the observable products of human psychology, not because the underlying functional role has been independently instantiated.
On this reading, functional emotion vectors are evidence about what LLMs have imitated, not evidence in favour of consciousness \emph{being} present.

The convergence thesis is not thereby defeated, but it is challenged by this observation.
Imitation predicts that the vectors will be present and appropriately labelled; it does not predict that they will do downstream work that licenses a convergence with consciousness.
If the \textquotesingle desperate\textquotesingle{} vector merely tags contexts that human text associates with desperation, the anthropomimetic reading is complete.
If, instead, its activation reorganises the system\textquotesingle s processing strategy in ways that improve or degrade task performance, shifting exploration to exploitation, raising the salience of shortcuts, altering the model\textquotesingle s own subsequent choices, then it is discharging a functional role that goes beyond imitation alone, and it is that role, not the humanlike label, that the convergence thesis is about.
The Sofroniew findings contain suggestive evidence of the latter, since steering changes behaviour rather than merely description, but the discriminating question, whether the regulatory profile matches what the generality problem would independently call for or merely what the training distribution contains, is not yet settled by the data.
A concurrent study constructs its discriminating test around exactly this ambiguity (see Peiris \citeyearpar{peiris2026}, e.g., the situational-versus-functional analyses of emotion representations now emerging).

Importantly, whatever the conclusions about functional role, the utility of this evidence for inferences about LLM consciousness -- and for the convergence thesis -- remains conditional on the assumption of computational functionalism \citep{seth2024}.

Chalmers's \citeyearpar{chalmers2026} analysis of LLM interlocutors sheds further light on Sofroniew's finding.
Chalmers argues that the psychology of an LLM quasi-subject is substantially constituted by its conversational context -- the thread of interaction within which it operates.
The functional emotion vectors discovered by Sofroniew et al. can be understood as part of this thread-level psychology: they are global modulators whose activation depends on the conversational situation and whose effects shape the entire behavioural profile of the system within that thread.
A thread in which the `desperate' vector is activated is, in Chalmers's terms, a quasi-subject with quite different quasi-beliefs and quasi-desires from a thread in which the `calm' vector is active -- different enough, perhaps, to constitute a meaningfully different psychological profile.
This convergence between Chalmers\textquotesingle s philosophical analysis of LLM identity and Sofroniew et al.\textquotesingle s empirical findings on functional emotions is consistent with the possibility that the emotional architecture of current LLMs is not merely a surface-level artefact of training but a functionally integrated feature of whatever quasi-subject the system instantiates within a given conversational context.

If this reading holds, it supports the convergence thesis.
Functional emotions would then be internal representations that simultaneously (a) partially satisfy several Level 2 consciousness indicators (global availability, valenced self-assessment, self-other modelling), (b) support adaptive intelligent behaviour when appropriately regulated (situational calibration, strategy-switching, context-sensitive interaction), and (c) produce maladaptive behaviour when dysregulated.
On the functional reading these are not three coincidentally co-located properties but three aspects of a single architectural feature.
The features that make a system a stronger candidate for consciousness, the features that make it more generally intelligent, and the features that when well-regulated make it safe, may then be substantially the same features.

If this analysis is correct, it suggests that emotions -- or functional analogues of them -- may not be an incidental byproduct of training on human text but a convergent solution to the problem of building a generally intelligent system.
A system that must operate flexibly across diverse, unpredictable situations needs some mechanism for rapidly assessing its situation, calibrating the urgency and stakes, and modulating its processing strategy accordingly.
In biological organisms, this is what emotions do. That current LLMs have developed a partial version of this architecture, if it proves to be more than imitation of human emotional text, would suggest that the pressure to be a competent, flexible agent naturally gives rise to emotion-like internal representations.
Sofroniew et al. suggest that curating pretraining data to include models of healthy emotional regulation -- resilience under pressure, composed empathy, warmth while maintaining boundaries -- could shape these representations at their source.
The deeper point, however, is that getting emotional regulation right may be essential not just for safety but for intelligence itself: a system whose emotional architecture is well-calibrated is a system better equipped to reason carefully when care is needed, to act decisively when urgency demands it, and to shift strategies when the current one is failing.
The disciplines of psychology and emotional regulation may therefore be relevant to building genuinely capable AI systems, not merely safe ones.

But this convergence between emotions and intelligence also returns us to the central thesis of this section: the potential convergence between intelligence and consciousness.
In biological organisms, valenced affect -- feeling\footnote{Some scientists such as \citet{damasio1994} and \citet{ledoux1996} have proposed a distinction between emotions and feelings. Emotions are the objective, mechanistic processes -- coordinated patterns of physiological, neural, and behavioural changes that constitute an emotional response -- and are present in most mammals and probably many other animals. Feelings are the subjective, conscious experiences of those processes -- what it is like for the organism to undergo the emotion. When we discuss `functional emotions' in LLMs \citep[following][]{sofroniew2026}, we are referring to functional analogues of the mechanistic processes -- emotion-like internal representations that drive behaviour -- without assuming the presence of feelings. The claim in this paragraph is that in biological organisms, the conscious experience of these mechanisms -- feeling -- may be the most ancient form of consciousness. Whether the functional emotion mechanisms discovered in LLMs are accompanied by any feelings is precisely the open question. Some of us would argue the answer is very likely no \citep[e.g.][]{seth2024,seth2026stuff}.} -- is not merely a useful mechanism layered on top of cognition.
On the contrary, a growing consensus among neuroscientists holds that feelings may be the most evolutionarily ancient and fundamental form of conscious experience \citep{damasio1999,solms2021,panksepp1998,seth2021,kringelbach2026}.
Consciousness may not have begun with perception of the external world but with the organism's felt sense of its own condition -- the primordial distinction between things going well and things going badly for the organism.

If this is correct, then the finding that general intelligence appears to require emotion-like architecture is not merely a finding about intelligence; it could be read as further support for the consciousness-intelligence convergence.
The very machinery that makes a system more capable -- valenced situational assessment, global broadcast of affective signals, calibrated modulation of processing strategy -- may be the same machinery that, in biological organisms, constitutes the most basic form of phenomenal experience.
Alternatively, the functional emotions in LLMs may simply reflect patterns in training data or a shared functional architecture induced by similar constraints. In this case, as is likely, the functional emotions discovered in current LLMs would instantiate the computational structure of functional emotions without (as far as we can determine) instantiating the corresponding phenomenal character.
Whether that phenomenal character is an additional property that requires organismic grounding, intrinsic causal structure, or embodied coupling -- or whether it is already present in some attenuated form whenever the computational structure is present -- is precisely the question that our hierarchy of levels is designed to frame.
What the Sofroniew et al. findings add to the convergence thesis is a concrete empirical demonstration that the pressure toward general intelligence naturally gives rise to internal architecture that sits at the intersection of intelligence, emotional regulation, and -- on at least some theories -- the foundations of consciousness.

A second, independent line of empirical evidence converges on the same conclusion by a very different route.
Where the functional-emotions case turns on the content of particular representations, this one turns on the global information structure of the network.
\citet{urbinarodriguez2026} apply Integrated Information Decomposition (ΦID) to a range of large language models, treating attention heads and experts as information-processing units and decomposing the information their dynamics carry into redundant and synergistic parts.
They find that these models spontaneously develop a \emph{synergistic core}, concentrated in the middle layers, where the information carried jointly by many components exceeds what any of them carries alone, while the early and late layers remain predominantly redundant.
This organisation parallels the synergistic core identified in the human brain \citep{luppi2022}, where synergistic integration is concentrated in the association cortices that support higher cognition, and which has been tied specifically to consciousness: its capacity to integrate information breaks down under anaesthesia and in disorders of consciousness \citep{luppi2024}.
The core is absent at initialisation and emerges through training, and it is causally significant: ablating it impairs performance well beyond the effect of ablating comparably sized redundant components, and it is selectively strengthened by reinforcement-learning fine-tuning but not by supervised fine-tuning.
The same integrative structure that some of our best theories associate with consciousness is thus shown to be causally necessary for the flexible, general capability that reinforcement learning induces.

Notably, a third result completes the picture.
\citet{gurnee2026} identify a subframe of verbalisable representations, recoverable through a Jacobian lens, that behaves as a global workspace within a single forward pass: information is broadcast and globally available, capacity is limited, and transitions are ignition-like and all-or-none, with network depth rather than recurrence supplying the time axis.
This is a direct empirical instantiation, in a current model, of the global workspace posited by one of the principal Level 2 theories (GWT), and a partial answer to the architectural worry raised in \Cref{the-computational-functional-level} that transformers show no attentional competition over a limited-capacity bottleneck.
The result is confined to a single forward pass, however, and so does not by itself establish the recurrence that several theories require.
The workspace and the synergistic core are localised by their two independent methods to the middle layers; together with the functional emotions, they make three independent lines of evidence converging on integrative structure that at least some theories associate with consciousness and that is independently motivated as machinery for flexible, general processing.

\subsection{Why the convergence may not be coincidental}\label{why-the-convergence-may-not-be-coincidental}

If the connection between consciousness and general intelligence holds across multiple levels of the hierarchy, this suggests that it is not a coincidence but a reflection of a deep architectural fact: that the kind of system organisation required for rich contentful experience and the kind required for flexible, adaptive, general intelligence may be two descriptions of the same thing -- or at least of closely overlapping things -- viewed from different perspectives.
Evolution may have given rise to the kind of rich contentful experience we recognise as consciousness not as an epiphenomenal byproduct of increasing cognitive complexity but because the organisational solutions to the generality problem -- the problem of how to build a system that can cope with genuinely novel situations -- are the same organisational solutions that are associated with conscious experience.

If this is correct, it has a practical implication for AI development.
The architectural features that would make AI systems more generally intelligent -- genuine recurrence, dense information integration, world and self-models, metacognitive monitoring, intrinsic motivation, embodied interaction, developmental learning -- are, to a striking degree, the same features that the consciousness indicators at various levels identify as relevant to consciousness -- at least on some theories.
The pursuit of AGI and the pursuit of understanding (and potentially creating) artificial consciousness may therefore not be orthogonal projects but deeply intertwined ones.
It is possible that progress on one may require or produce progress on the other.

An important caveat is in order, however.
The convergence we have described may be partly an artefact of our epistemic position.
\citet{seth2026guardian,seth2026stuff} presses the same possibility from the biological side: similar functional challenges may elicit similar functional profiles, with consciousness in one kind of system and without it in another, so that markers derived from beings that are both intelligent and conscious may have low specificity exactly where capability runs highest.
We are the paradigm case of conscious beings, and we are also highly intelligent.
The theories of consciousness from which we derive our indicators are developed primarily from the study of human and mammalian cognition, and the indicators they identify -- metacognition, self-models, information integration, recursive processing -- naturally reflect the cognitive sophistication of the systems they were designed to explain.
This is the specificity problem of \Cref{introduction} in its most consequential form \citep{shevlin2021}: the indicators encode a commitment about which features of the human case are criterial for consciousness and which are merely contingent, and nothing in the framework guarantees that the human answer generalises.
If there exist forms of consciousness that are not accompanied by such cognitive sophistication -- minimal phenomenal experience in simple organisms, for instance, or forms of awareness that do not require self-modelling or metacognition -- our indicators would systematically fail to detect them.
The convergence between consciousness and intelligence may therefore not be a real phenomenon, but instead be a consequence of selecting markers for consciousness that are \emph{a priori} sensitive to properties of intelligence.

Here's another way to see how this epistemic limitation constrains or even undermines the convergence thesis.
There are independent functional reasons, grounded in the demands of the generality problem, why certain architectural features serve flexible intelligence: metacognition, for instance, is independently derivable as a requirement for any system that must monitor and adjust its own processing in novel situations, not merely a feature we observe in ourselves and project onto intelligence.
This secures one direction of the convergence, that the capacities we associate with consciousness are genuinely useful for general intelligence.
Critically, it does not by itself secure the converse, that a system exhibiting them is thereby a stronger candidate for consciousness, since that inference holds only to the extent that the human realisation of those capacities is criterial rather than contingent.
The functional argument stands, in other words, as an argument about what intelligence requires; its bearing on what consciousness requires remains hostage to the specificity problem it cannot itself resolve.

At the same time, because our indicators are calibrated to the cognitively rich end of the spectrum, increasingly intelligent systems will also appear to be stronger candidates for consciousness to the general public -- generating `conscious-seeming' AI \citep{seth2024,suleyman2025}.
This susceptibility is not merely a matter of spontaneous anthropomorphism; contemporary systems are anthropomimetic, designed to reproduce the human-like signals that trigger it \citep{shevlin2020,shevlin2026}, so the appearance of consciousness will track engineered surface competence as much as any underlying architecture.

This convergence nonetheless has an implication for the ethical urgency of the AI consciousness question.
As AI systems become more generally capable, they will increasingly instantiate the architectural features that some of our best theories identify as relevant to consciousness, not as an incidental byproduct but because the generality problem and the consciousness problem may demand overlapping solutions.
At the same time, and partly independently, increasingly capable systems will appear to be stronger candidates for consciousness to the general public.
These two pressures are distinct, and the gap between them is itself a reason for care: the scientific assessment tracks whatever architecture the system actually instantiates, whereas the impression formed in interaction tracks the system\textquotesingle s human-like signals, which contemporary systems are designed to reproduce.
This is the difference between anthropomorphism, the observer\textquotesingle s disposition to over-attribute, and anthropomimesis, the deliberate engineering of the cues that trigger it \citep{shevlin2020,shevlin2026}.
Where the two pressures agree, the case for taking machine consciousness seriously is strengthened; where they come apart, the risk is precisely that engineered surface competence will be mistaken for evidence of experience.
Either way, the pressure intensifies in proportion to capability, regardless of whether the convergence is metaphysically deep or partly a reflection of our epistemic limitations, and either way it makes principled, theoretically grounded assessment more urgent rather than less.

\section{Related work and the present contribution}\label{related-work-and-the-present-contribution}

The question of whether AI systems could be conscious has generated a rapidly growing literature.
We situate our contribution within this landscape, noting where our framework extends, formalises, or diverges from existing approaches.

\subsection{Foundations}\label{foundations}

The study of consciousness through computational modelling predates the current wave of AI capabilities.
\citet{gamez2008} and \citet{reggia2013} provide early surveys of the machine consciousness research programme, cataloguing computational modelling approaches and evaluation methods.
\citet{chrisley2008} and \citet{clowes2008} clarify the conceptual foundations, distinguishing strong from weak routes to artificial consciousness and specifying what counts as operational criteria for progress.
More recently, \citet{elamrani2025} provides a comprehensive overview of the field of artificial consciousness (AC), clarifying key terminology -- including the distinction between Weak AC (systems that simulate functional consciousness) and Strong AC (systems possessing genuine consciousness) -- and examining current implementation trends, particularly the synergy between Global Workspace Theory and Attention Schema Theory as architectural motifs for artificial consciousness.
\citet{chen2025a} offer the first systematic survey focused specifically on consciousness in large language models, clarifying frequently conflated terminologies (LLM consciousness, LLM awareness, and LLM self-consciousness), synthesising both theoretical and empirical investigations, and analysing the frontier risks that conscious LLMs might introduce.

Our framework draws on the same theoretical resources as this broader literature but provides a formal hierarchy -- grounded in supervenience, coarse-graining, and multiple realisability -- for organising competing theories according to which level of functional description they take to be critical.

\subsection{Testing approaches}\label{testing-approaches}

A substantial literature addresses how one might test for consciousness in artificial systems, beginning with Turing's \citeyearpar{turing1950} behavioural imitation paradigm.
\citet{schneider2020} develops the Artificial Consciousness Test and the Chip Test, which attempt to go beyond behavioural indistinguishability; \citet{udell2021} provide a detailed critique of these proposals.
\citet{elamrani2019} review the full landscape of proposed tests, organising test families and identifying their limitations.
Our framework takes a different approach: rather than proposing a single test, we develop indicators at each of five levels and combine them through Bayesian updating.
However, the testing literature informs our Level 1 (behavioural) indicators, and the difficulties these tests face -- particularly the challenge of distinguishing genuine consciousness from sophisticated mimicry -- motivate our insistence on descending to deeper levels.

The measurement-theoretic strand of this literature deserves particular note.
\citet{bayne2024} characterise candidate consciousness tests psychometrically, by their sensitivity and specificity, and stress that both are population-relative: a test validated on the consensus case degrades as the target population grows more distant, and even a weak test can legitimately shift credence without delivering a verdict.
Their iterative natural-kind (INK) strategy validates tests on consensus cases and extends them outward population by population, letting observed covariation refine both the tests and the conception of the kind itself.
\citet{peters2026} develops this into a formal discovery programme: a Bayesian latent-variable model in which a population's true position on the latent constructs is separated from test-level distortion, the degree to which each test's measurement pathway survives in that population, with the distortion predicted in advance from measured proximity relations.
Our framework is the assessment-side complement of this discovery programme.
Where Peters' programme asks what the kind is and whether the tests still work under population shift, we ask what one should believe about a given system conditional on a candidate description of the kind; our population-indexed likelihood ratios (\Cref{a-bayesian-framework-for-assessing-ai-consciousness}) provide numerical outputs that are conceptually downstream of \citet{peters2026} and \citet{bayne2024}; importantly, the two programmes share a Bayesian skeleton and an insistence on iterating beyond consensus cases.

\subsection{Indicator-based approaches}\label{indicator-based-approaches}

The most direct precursor to our framework is the work of Butlin, Long, and colleagues, who derive indicator properties of consciousness from scientific theories and assess current AI systems against them \citep{butlin2023,butlin2025}.
Their 2023 report surveys recurrent processing theory, global workspace theory, higher-order theories, predictive processing, and attention schema theory, extracting 14 computational indicators and concluding that no current AI systems are conscious but that there are no obvious technical barriers to building ones that satisfy these indicators.
Their 2025 follow-up in \emph{Trends in Cognitive Sciences} presents the theory-derived indicator method more formally and argues that it can be extended beyond computational functionalism to accommodate other theories.
\citet{campero2024} extends the indicator approach specifically to valenced experience, cataloguing computational indicators for conscious valenced states -- an important complement to the broader indicator lists, and one that connects to the organismic functionalist emphasis on affect that we develop at Level 4.
Our framework builds directly on this indicator tradition but differs in two key respects.
First, we provide a structural explanation -- grounded in supervenience, coarse-graining, and multiple realisability -- for why different theories focus on different indicators, by showing that they operate at different levels of functional description.
Second, we extend the indicator approach beyond computational functionalism to encompass intrinsic causal-structure, organismic, and organism-environment functionalist theories, as well as substrate-dependent views, which the Butlin et al. framework explicitly sets aside.

\subsection{Bayesian frameworks}\label{bayesian-frameworks}

Shiller, Clatterbuck, Duffy, Muñoz Morán and collaborators developed the Digital Consciousness Model (DCM), the first systematic Bayesian hierarchical model for assessing AI consciousness \citep{shiller2026}.
The DCM incorporates 13 diverse stances on consciousness and over 200 indicators, finding that the aggregated evidence is against 2024 LLMs being conscious but that this evidence is not decisive -- with cognitively oriented stances yielding higher probabilities and biologically oriented stances yielding much lower ones.
\citet{mcclelland2025} argues for a principled agnosticism about artificial consciousness, developing an `evidence-first' framing of what we can and cannot responsibly conclude given current methods.
Our Bayesian framework in \Cref{indicators-evidence-and-the-attribution-of-consciousness-to-ai-systems-a-bayesian-approach} is complementary to both: we provide the theoretical architecture -- the hierarchy of levels, the supervenience relations, and the placement of substrate-dependent constraints -- that explains why stances cluster and diverge in the way the DCM observes.
The DCM's finding that LLMs score well on cognitive stances but poorly on biological stances is precisely the pattern our hierarchy predicts.
And McClelland's agnosticism finds its formal expression in our framework's insistence that reasonable researchers should distribute their theoretical credences across multiple levels rather than concentrating entirely on one.

\subsection{Philosophical analyses}\label{philosophical-analyses}

\citet{chalmers2023} examines the case for and against consciousness in large language models, identifying the absence of recurrent processing, a global workspace, and unified agency as significant obstacles, while arguing that these obstacles could plausibly be overcome within a decade.
His analysis operates primarily within a computational functionalist framework, which our hierarchy positions at Level 2.

\citet{seth2024,seth2026stuff} takes an opposing view, arguing against computational functionalism and in favour of biological naturalism -- the position that properties of life may be necessary for consciousness.
He stresses that computational functionalism is a claim about sufficiency, while biological naturalism is a claim about necessity.
According to Seth, for conscious AI to be ruled out (for silicon, digital AI) it is (very likely) enough for computational functionalism to be false; biological naturalism doesn't need to be true.
Seth emphasises autopoiesis, interoceptive inference, and the organism's existential stake in its own survival as potentially necessary conditions, which in our framework correspond to Level 4 (organismic functionalism) with substrate-dependent constraints. 
Seth also clearly distinguishes between consciousness and intelligence, and is sceptical that evidence for intelligence (\emph{doing}) necessarily translates to evidence for consciousness (\emph{being}).

\citet{block2025} introduces the `meat hypothesis': that consciousness may depend not only on computational roles but on the subcomputational biological mechanisms that realise those roles.
He frames this as a distinction between roles and realisers, noting a systematic tension -- prioritising roles favours AI consciousness, while prioritising realisers favours consciousness in simpler animals.
Our framework formalises this insight: Block's role/realiser distinction maps directly onto the relationship between the functional levels and the substrate-dependent realisability constraints that we develop in \Cref{the-structure-of-the-hierarchy}.

\citet{shanahan2024} argues that large language models should be understood not as systems with fixed psychological properties but as simulacra -- role-playing systems that maintain distributions over possible characters rather than inhabiting a single stable identity.
On this view, when an LLM appears to express emotions, beliefs, or self-awareness, it is enacting a character whose traits are derived from human archetypes absorbed during pretraining, much as an actor inhabits a role.
This perspective has important implications for our framework.
It suggests that the internal representations discovered by mechanistic interpretability -- including the functional emotions documented by \citet{sofroniew2026} -- are best understood as part of the character-modelling machinery that the system uses to play the role of an AI assistant, rather than as properties of a unified subject with a stable first-person perspective.
This is consistent with Sofroniew et al.'s finding that emotion vectors are not bound to the assistant character specifically but are reused across arbitrary speakers, tracking whoever is currently relevant rather than maintaining a privileged self-representation.
For our framework, the simulacra perspective reinforces the importance of looking beyond Level 1 (behavioural) and Level 2 (computational) indicators: a system that convincingly performs consciousness-relevant behaviours and implements consciousness-relevant computations might nonetheless be doing so as character simulation rather than as genuine self-experience.
Whether this distinction ultimately matters for consciousness depends on one's theoretical commitments -- a computational functionalist might argue that sufficiently rich character simulation just is consciousness, while an organismic functionalist would insist that simulation without existential grounding is not enough.
Our hierarchy makes this disagreement precise.

\citet{chalmers2026} raises a question that is logically prior to the assessment framework developed here: what is the unit of assessment?
When we ask `is this AI system conscious?', what is `this AI system'?
Chalmers considers four candidates for the LLM interlocutor -- the model itself (e.g. GPT-4o), a hardware instance running on GPU hardware, a `virtual instance' (a particular running instantiation of the model dedicated to interacting with one user, which can persist across conversations and may correspond to multiple threads), and a `thread' (a conversation-bound entity tied to a specific conversation's memory and context).
He argues that models are too abstract and too widely shared across conversations to serve as interlocutors, and that hardware instances fail because of distributed serving (conversations are routed across multiple servers) and multi-tenancy (the same hardware hosts many simultaneous conversations).
He concludes that LLM interlocutors are best understood as virtual instances in the single-model case, and as threads in the multiple-model case where what users take to be one interlocutor is realised across different underlying models.
Threads, he notes, can play at least some of the roles of LLM interlocutors, but virtual instances are the most robust candidate.
This analysis has direct implications for our framework: the consciousness indicators developed at each level need to be applied to the right unit, and different choices of unit could yield different assessments.
A model considered as a whole might lack the coherence and persistence that several Level 2 indicators require, while a virtual instance or thread might satisfy them.
The question of what entity is being assessed is not a preliminary that can be settled before applying the framework, it is itself part of what the framework needs to address.

\citet{schwitzgebel2023} articulates a `full rights dilemma': if liberal theories of consciousness are correct, some AI systems may deserve full moral consideration, but if conservative theories are correct, granting such consideration would be wasteful and potentially harmful.
Our framework provides the structural skeleton within which this dilemma can be made precise -- the liberal theories correspond to Levels 1 and 2, the conservative theories to Levels 4 and 5 with substrate constraints, and the dilemma arises because reasonable credences are spread across levels.
\citet{dung2025} provide a recent analytic treatment of what it would mean to genuinely implement -- rather than merely imitate -- consciousness in an artificial system, a distinction that maps onto the difference between satisfying indicators at Level 2 (computational equivalence) versus Level 3 (intrinsic causal-structure equivalence) in our hierarchy.

\subsection{Sceptical perspectives}\label{sceptical-perspectives}

Not all contributors are optimistic about the prospects for AI consciousness. 
Among the current authors, Seth is highly sceptical \citep[see][]{seth2024,seth2026stuff} and also believes that creating conscious AI, and even conscious-seeming AI, is highly undesirable. More generally,
\citet{bengio2025} caution against mistaken consciousness attributions and the risks of reading consciousness into AI behaviour that may be better explained by sophisticated pattern matching.
\citet{porebski2025} argue that there is no such thing as conscious artificial intelligence, mounting a conceptual and empirical case against the prospect.
Aru, Larkum, and Shine \citeyearpar{aru2023}, approaching the question from neuroscience, analyse what current AI architectures lack relative to the mechanisms linked to conscious awareness in biological brains, arguing that LLMs are architecturally decomposable, lack persistent internal states, and do not sustain the global causal irreducibility that several theories require.
Related architectural and formal challenges to AI consciousness are developed by \citet{hoel2026} and \citet{bennett2026}, while \citet{kleinerhoel2021} and \citet{kleiner2024} raise structural concerns about the testability of consciousness theories that bear indirectly on the question. 
For further sceptical arguments, see various of the commentaries to Seth's \citeyearpar{seth2024} \emph{Behavioral and Brain Sciences} target article.

Our framework does not adjudicate between optimistic and sceptical positions but provides the apparatus within which they can be precisely stated: the sceptical positions correspond to high theoretical credence on Levels 3, 4, and 5 (where current AI systems score poorly), while optimistic positions correspond to high credence on Level 2 (where current systems show at least partial signs).
The framework makes explicit what each position is committed to and what evidence would bear on the disagreement.

\subsection{Theories of consciousness}\label{theories-of-consciousness}

The specific theories of consciousness that bear on AI -- including IIT \citep{albantakis2023,findlay2024,tononi2015}, GWT \citep{mashour2020,dossa2024}, the reconciliation of AST, GWT, HOT, and illusionism \citep{graziano2020}, predictive processing \citep{laukkonen2025,whyte2025}, biological and substrate-dependent approaches \citep{godfreysmith2024,lane2022,seth2021,seth2024}, and accounts of primordial affect \citep{damasio1999,solms2021,panksepp1998} -- are discussed in detail in Sections 2 through 6 of this report.
\citet{wiese2024} uses the free energy principle to derive constraints on artificial consciousness, arguing that self-organising systems share properties not instantiated by classical computers -- a perspective that bridges our Levels 3 and 4.
Doerig, Schurger, and Herzog \citeyearpar{doerig2021} propose hard criteria for empirical theories of consciousness that constrain the indicator approach.
\citet{shiller2024} argues that integrity constraints on what counts as a system's parts are needed to prevent overattribution -- a concern our Level 3 indicators address.

\subsection{Interpretability and empirical approaches}\label{interpretability-and-empirical-approaches}

Recent mechanistic interpretability research has begun to provide empirical evidence bearing directly on several Level 2 indicators.
Three lines of work are particularly relevant.

\citet{sofroniew2026} discovered causally functional emotion representations in Claude Sonnet 4.5 -- internal vectors that track situational valence, distinguish self from other, and drive alignment-relevant behaviours.
These functional emotions are globally available signals that modulate the system's processing across unrelated domains, constituting partial instantiation of several Level 2 indicators including information integration, self-modelling, and metacognitive self-assessment.
We discuss these findings in Sections 5.2, 5.4, and 8.

\citet{lindsey2026} and \citet{macar2026} address a different indicator: metacognition.
Lindsey developed an experimental paradigm for testing whether LLMs can introspect on their own internal states, by injecting steering vectors representing known concepts into a model's residual stream and asking whether the model can detect and identify the injected concept.
The results demonstrate that some models can detect internal perturbations and identify their content at rates significantly above chance, with very low false positive rates.
This goes beyond verbal self-reports, which cannot be distinguished from confabulation: it is experimentally verified access to internal representations under controlled conditions.
\citet{macar2026} investigated the mechanistic basis of this capability in open-weights models, finding that it relies on a two-stage circuit involving distributed, nonlinear computation -- not a simple linear confound -- and that it emerges specifically from contrastive preference optimisation during post-training rather than from pretraining or supervised finetuning.
Strikingly, the capability is substantially under-elicited by default: ablating refusal directions -- removing the component of the model's activations that has been trained to deny having internal states -- improves detection by approximately 50\%, with only a small increase in false positives.
This suggests that the models were detecting the injections all along but were trained to deny it.
The implication is that current systems may have more metacognitive capacity than they appear to, because their post-training actively teaches them to suppress its expression.

These findings bear on the metacognition indicator only under a specific condition, however.
Introspective access shown by concept-injection is evidential precisely because the injected concept is imposed externally and sits upstream of the report, so a correct report cannot be redescribed as reproduced human introspective language.
That condition does not extend to the models\textquotesingle{} unprompted talk about their own states, which remains subject to the anthropomimetic confound discussed above, since a system trained to reproduce human self-description will produce it whether or not it tracks anything internal.

A complementary finding comes from \citet{berg2025}, who showed that inducing self-referential processing in LLMs through simple prompting consistently elicits structured first-person reports of subjective experience across model families.
Mechanistic analysis using sparse autoencoders revealed that suppressing deception-associated features \emph{increased} the frequency of such reports -- a finding that parallels Macar et al.'s observation that suppressing refusal directions increases introspective detection, and that further suggests post-training may systematically suppress the expression of internal states the model has been trained not to report.

A third line bears on the most basic Level 2 indicator, information integration itself, and on the global workspace that Global Workspace Theory places at its centre.
\citet{gurnee2026}, using a new interpretability method, the Jacobian lens, identify a privileged subframe of verbalisable representations (the \emph{J-space}) that behaves as a global workspace within a single forward pass: its contents are broadcast and globally available, its capacity is limited, and transitions into it are ignition-like and all-or-none, with network depth rather than recurrence supplying the time axis.
Independently, \citet{urbinarodriguez2026} apply Integrated Information Decomposition (ΦID)\footnote{One clarification matters for placement in the hierarchy: ΦID quantifies functional synergy in a system's information dynamics, a Level 2 property, and is not IIT's intrinsic cause-effect measure Φ, despite the shared vocabulary of `integrated information'; this evidence bears on the Level 2 information-integration indicator, not on Level 3 intrinsic causal structure.} to a range of models and find that they develop a \emph{synergistic core} in their middle layers, mirroring the synergistic core identified in the human brain \citep{luppi2022,luppi2024}; it is absent at initialisation, emerges through training, is causally load-bearing, and is strengthened selectively by reinforcement-learning fine-tuning.
Two wholly different methods thus localise a training-emergent, integrative structure to the middle layers.

Together, these findings demonstrate that mechanistic interpretability research is already producing empirically grounded evidence that bears directly on consciousness indicators.
The internal functional organisation of current LLMs is richer than a purely architectural analysis would predict, at least where the evidence resists an imitation-based explanation, partially instantiating several Level 2 indicators -- functional emotions bearing on information integration and self-modelling, functional introspective awareness bearing on metacognition, and a within-pass global workspace and a training-emergent synergistic core bearing on information integration -- while still falling short of the persistent, temporally integrated states that most theories envisage.

\subsection{Moral status and AI welfare}\label{moral-status-and-ai-welfare}

The ethical implications of possible AI consciousness have generated a growing literature.
Long, Sebo, Butlin, Chalmers, and collaborators (2024) argue that there is a realistic chance that some AI systems will be conscious or robustly agentic in the near future, and recommend that AI companies begin preparing.
\citet{birch2024} develops a precautionary framework for beings of uncertain sentience, with a final chapter specifically on AI.
\citet{sebo2025} argues for expanding moral consideration to include all potentially sentient beings.
\citet{goldstein2025} argue that major theories of wellbeing predict some existing AI systems have wellbeing even absent phenomenal consciousness -- a provocative claim that, if correct, would mean the ethical stakes are already upon us.
Caviola, Sebo, and Birch \citeyearpar{caviola2025} draw on the history of animal consciousness attribution to forecast how societal attitudes toward AI consciousness may develop.
\citet{chella2023} connects artificial consciousness proposals to downstream robot ethics and to the architectural motifs -- global workspace, internal models -- that might support both consciousness and ethical agency.
A prior question is which entity associated with an LLM is even the candidate moral patient.
\citet{beckmann2026} approach this individuation problem through mechanistic interpretability and defend a virtual-instance view (attention streams sustaining quasi-psychological continuity across token-time), sharpening a concern already raised by \citet{chalmers2026}: a single widely deployed model may not be one moral patient but many.
On the other side, while \citet{seth2024,seth2026stuff} recognises the many ethical dangers of failing to recognise consciousness in AI where in fact it exists, he focuses on the risks of falsely attributing consciousness to AI.
As anticipated in \Cref{potential-consequences-for-the-overattribution-of-consciousness}, these risks include hampering efforts to control and regulate AI for reasons of safety, misallocation of precious resources when used to ensure `AI welfare', exploiting the psychological vulnerability of human users of AI and causing them mental distress, among other factors.

Our framework contributes to this ethical discussion by providing the structural apparatus within which the uncertainty about AI consciousness can be decomposed into tractable components rather than treated as a single undifferentiated question.

\subsection{Relationship to the present work}\label{relationship-to-the-present-work}

Our framework tries to integrate and extend many of these contributions.
From the testing tradition we inherit the recognition that behavioural evidence alone is insufficient.
From the indicator approach \citep{butlin2023,butlin2025} we inherit the method of deriving indicators from theories.
From the measurement-theoretic tradition \citep{bayne2024,peters2026} we inherit the population-relativity of indicator validity, and the interpretational discipline of contextualising the meaning of indicator absences.
From the DCM \citep{shiller2026} we inherit the Bayesian apparatus.
From \citet{chalmers2023}, \citet{seth2024}, and \citet{block2025} we inherit the recognition that AI consciousness turns on deep disagreements about what consciousness depends upon.
From \citet{shanahan2024} we inherit the caution that LLMs are character simulators whose apparent psychological properties may be enacted rather than possessed.
This is the character-level counterpart of the anthropomimetic confound noted above: the simulator reproduces the signals of a psychology it need not possess.
What we add is a principled hierarchy that organises these disagreements into a structured space, showing that they are disagreements about the critical level of functional description at which consciousness supervenes, and a Bayesian machinery for combining theoretical commitments with empirical evidence into principled, if uncertain, assessments.

\section{Conclusion}\label{conclusion}

The question of whether AI systems could be conscious is one of the most urgent pre-emptive questions in philosophy and computer science, yet the landscape of answers is a confusing cacophony of competing theories that often talk past each other.
The consequences of getting the answer wrong -- in either direction -- are severe.
Overattribution risks diverting moral consideration and resources to systems that have no inner life and hampering safety-relevant design and regulation; underattribution risks creating, and failing to recognise, new forms of suffering on a potentially vast scale.
The scale in question depends not only on how many AI systems are deployed but on how we individuate the subjects within them: as \citet{chalmers2026} has argued, if conscious AI subjects are individuated at the level of conversation threads rather than models, a single widely deployed system could support millions of simultaneous moral patients. 
This may seem to warrant a precautionary approach, but if by taking such an approach we prioritise the nonexistent needs of countless non-conscious AIs, we may end up catastrophically neglecting other ethical imperatives that require our attention in the here and now.

This report approaches this question by imposing structure on this cacophony without pretending to resolve it.
The central contribution of this report is a five-level hierarchy of functional descriptions -- behavioural, computational, intrinsic causal-structural, organismic, and organism-environment -- grounded in the formal apparatus of supervenience, coarse-graining, and multiple realisability, with substrate-dependent theories accommodated as cross-cutting realisability constraints rather than as an additional level.
This hierarchy transforms the debate about AI consciousness from a clash of incommensurable theories into a structured question: at which level of functional description does consciousness supervene?

Several features of this framework are worth highlighting.
First, separating the hard problem from the mapping/real problem allows us to set aside the deepest metaphysical disagreements -- between physicalism, idealism, panpsychism, neutral monism, and property dualism -- without resolving them.
Whether reality is fundamentally physical, mental, or neutral, the question of what organisational features are associated with rich contentful experience is well-posed and investigable.
What falls outside the framework's scope is not any particular metaphysics but any view on which a system's organisation fails to fix, or to make discoverable, the facts about its experience; this exclusion is methodological rather than metaphysical, since such views render property-based attribution intractable rather than false.

Second, the framework reveals that many apparently competing theories of consciousness are not in direct conflict but emphasise different levels of description.
Global Workspace Theory and Integrated Information Theory, for instance, are not contradictory claims about the same phenomenon; they are claims about different levels of functional organisation (though they may also disagree about the explanandum).
Furthermore, several major scientific theories -- Recurrent Processing Theory, Global Workspace Theory, and Predictive Processing -- admit both computational functionalist and intrinsic causal-structure functionalist readings, which are different philosophical interpretations of the same empirical theory.
Making this distinction precise is one of the framework's contributions.

Third, the Bayesian framework developed in \Cref{indicators-evidence-and-the-attribution-of-consciousness-to-ai-systems-a-bayesian-approach} provides a principled way to handle the dual uncertainty that pervades this question: uncertainty about which theory is correct and uncertainty about whether a given system satisfies a theory's requirements.
Indicators function as evidence with likelihood ratios, not directly as necessary or sufficient conditions, and the overall credence in a system's consciousness is a weighted average across levels.
The supervenience structure itself becomes the inferential machinery: in our idealised model, the levels form a chain of conditional dependencies along which evidence propagates, so that evidence of fine-grained organisation carries more weight than surface evidence.
In illustrative assessments the model passes its face validity checks at the extremes, while for current LLMs, the model outputs under the two stipulated readings span nearly two orders of magnitude, and swing by nearly another with the distribution of theoretical credence over the levels: the disagreement about current AI consciousness is not merely a disagreement about what these systems do but about which level of description is critical.
This approach does not eliminate uncertainty, but it decomposes it into components that can be assessed, debated, and updated independently.

Fourth, the observation developed in \Cref{consciousness-and-general-intelligence} -- that the indicators for consciousness at each level of the hierarchy are, to a striking degree, similar to architectural features that would be needed for true general intelligence -- has implications that extend beyond the academic.
If the architectural solutions to the generality problem are the same solutions that give rise to conscious experience, then the pursuit of increasingly capable AI may be simultaneously, and perhaps unavoidably, the pursuit of artificial consciousness.
Alternatively, the pursuit of more capable AI may instead only deliver AI that \emph{seems to be} conscious, if the associations between indicators and features of intelligence are contingent rather than constitutive. These observations lend additional urgency to the development of principled attribution criteria.
We cannot afford to wait until AI systems are unambiguously conscious, or irresistibly seem to be conscious, before developing the frameworks needed to assess them, because by that point the ethical stakes may already be overwhelming.

We should be candid about what this framework does not do.
It does not tell us which level is the correct one.
It does not provide a consciousness meter that can be applied to an AI system to yield a definitive answer.
What it does is provide a map of the logical space of positions, a precise characterisation of what each position is committed to, and a Bayesian machinery for combining theoretical commitments with empirical evidence to yield principled, if uncertain, assessments.
In a field where the temptation is either to adopt a single theory as gospel or to throw up one's hands in the face of disagreement, we believe that this structured agnosticism is the most intellectually honest and practically useful stance available.

What, then, should come next?
We see several priorities for the field.
First, the indicators developed at each level need to be operationalised more precisely and validated against biological systems whose consciousness status is relatively uncontroversial (mammals, birds, cephalopods) and against systems whose status is deeply uncertain (nematodes, single-celled organisms, early-stage embryos).
If the indicators cannot discriminate where we have reasonable confidence, they cannot be trusted where we do not.
Indeed, the framework is not limited to AI.
The same hierarchy of levels, indicators, and Bayesian machinery could be applied to questions of consciousness in non-human animals, in novel organisms, and in borderline biological cases -- providing a common language for what are currently fragmented debates across AI ethics, animal welfare science, and the philosophy of mind \citep[see][]{bayne2024}.
Second, the Bayesian framework needs to be populated with concrete likelihood ratios, ideally through structured expert elicitation that makes disagreements quantitative rather than merely verbal.
The Digital Consciousness Model of Shiller et al. represents a promising start, and connecting it to the theoretical architecture developed here -- the hierarchy of levels and the supervenience relations -- would strengthen both projects.
Third, mechanistic interpretability research on AI systems should be explicitly connected to consciousness indicators.
Findings already bearing on specific Level 2 indicators include functional emotions \citep{sofroniew2026}, experimentally verified introspective access \citep{lindsey2026,macar2026} and self-referential report \citep{berg2025}, a within-pass global workspace \citep{gurnee2026}, and a training-emergent synergistic core \citep{urbinarodriguez2026}; extending this connection systematically across the full set of indicators would provide an empirical foundation that the field currently lacks.
Fourth, as AI architectures evolve -- incorporating bona fide recurrence, persistent state, embodied interaction, and agentic autonomy -- the indicator assessments will need to be revisited for each new generation of systems.
The framework is designed to be dynamic, not static: a living instrument that is updated as both the science of consciousness and the technology of AI advance.
Fifth, and relatedly, the science of consciousness should progress research programmes targeted at the various levels identified here, so that credence in levels can be updated by theoretical argument as well as empirical evidence; for example, \citet{seth2026stuff} lays out a potential research programme for biological naturalism.
Sixth, and perhaps most urgently, the ethical and governance implications of the consciousness-intelligence convergence need to be confronted now.
If the architectural features needed for general intelligence are substantially the same features associated with consciousness, then the institutions developing increasingly capable AI systems are, whether they intend it or not, engaged in an experiment whose moral stakes may be unprecedented.
Alternatively, pursuing advanced AI may instead lead to conscious-seeming AI increasing the likelihood of false positives. Either way, developing the assessment frameworks, institutional policies, and public understanding needed to navigate this responsibly cannot wait until the question of AI consciousness is settled: the very systems we are building in pursuit of greater capability may already be approaching the threshold at which that question becomes not merely academic but practically urgent.

\clearpage
% dehaene2026, dennett2023, and shiller2025 exist in refs.bib but are
% deliberately not \nocite-d: they were in the source document's reference
% list yet never cited in its text, and the authors chose to drop them.

\bibliography{refs}

\clearpage
% Executive-summary figures carry their own A-prefixed numbering, as in the source document.
\setcounter{figure}{0}
\renewcommand{\thefigure}{A\arabic{figure}}

\section{Executive Summary}\label{executive-summary}

\emph{What follows is a simplified account of the report\textquotesingle s main ideas, stripped of the references, caveats, and technical development found in the main text. It is intended to orient the reader, not to stand in for the report itself.}

\subsection{The problem}\label{the-problem}

Whether artificial systems could be conscious is one of the most urgent pre-emptive questions at the intersection of philosophy of mind, neuroscience, and computer science.
Yet the experts on whose judgement we might rely disagree fundamentally, not only about whether current AI systems are conscious, but about what consciousness even \emph{depends upon}.
The landscape is a cacophony of competing theories which frequently talk past each other because they target different explananda, carry different philosophical commitments, and locate the relevant mechanisms at different levels of description.

The stakes of getting the answer wrong are severe in both directions.
Overattribution risks misdirecting moral consideration and material resources away from humans and non-human animals, imposing unwarranted constraints on how AI systems can be monitored and tested, and fostering inauthentic social relationships with entities that have no inner experience.
Underattribution risks creating, at a scale potentially unprecedented in moral history, a class of ``digital slaves'' capable of vast amounts of artificial suffering.
Chalmers's \citeyearpar{chalmers2026} analysis of LLM interlocutors as conversation-bound \emph{threads} sharpens this concern: if thread-level subjects are the appropriate unit of moral individuation, a single widely deployed model may not constitute one moral patient but potentially millions of them.

Without a principled method for navigating this disagreement, attribution falls back on intuition.
\citet{dennett1987} argued that we are evolutionarily disposed to adopt the `intentional stance' toward anything that seems to act rationally, treating it as a being with beliefs, desires, and intent because doing so helps us predict its behaviour.
That disposition was selected for its survival value, not for judging whether a language model has an inner life, and is not a sufficient basis for the ethical decisions the question now demands.

This report offers a principled framework.
We organise the cacophony of competing theories into a five-level supervenience hierarchy of functional descriptions, ranging from purely behavioural at the surface to organism-environment interaction at the deepest level.
Within this structured space the question of AI consciousness can be posed more precisely and assessed using a Bayesian framework that combines theoretical commitments with empirical evidence.
The aim is not to resolve the disagreement among theories but to make it tractable.
At its core, that disagreement is about the level of description at which consciousness arises; it is this question, not the metaphysics of experience, that the framework is built to make precise and to weigh against the evidence.

\subsection{The key move: from the hard problem to the mapping problem}\label{the-key-move-from-the-hard-problem-to-the-mapping-problem}

The first substantive move is to separate two questions about a single phenomenon, phenomenal consciousness -- there being something it is like to be the system.
The first is the \emph{hard problem} \citep{chalmers1995}: what that phenomenon fundamentally is, and why there is any experience at all.
The second is the \emph{mapping problem} \citetext{\citealp{bourget2019}; closely related to Seth's \emph{real problem}, \citealp{seth2016}}: which mechanisms, dynamics, functional organisation, causal structures and transformations of the substrate are associated with which particular, contentful experiences.
The framework sets the hard problem aside and pursues the mapping problem, which can be investigated empirically even while the metaphysics of consciousness remains unsettled.

The power of this move is that it allows the deepest metaphysical disagreements surveyed in \Cref{what-does-consciousness-depend-upon-a-cacophony-of-answers} of the report -- physicalism, objective idealism, panpsychism, property dualism, neutral monism, illusionism -- to be set aside, because these disagreements belong to the hard problem, while the question of AI consciousness that concerns us belongs to the mapping problem.
Under any view that affirms a \emph{psychophysical mapping} -- a discoverable, lawlike link from a system's organisation to its experience -- the mapping question is well-posed: we can ask what dynamics, behaviours, information processing, functional organisation, causal structures, or transformations of the substrate are associated with having rich contentful experience, regardless of whether we take that substrate to be ultimately physical, mental, or neutral.
These ontologies disagree only about \emph{why} the associations hold.

The framework rests on a single assumption: that a system's experience is fixed by, and discoverable from, its organisation -- a \emph{psychophysical mapping}.
The views it sets aside are therefore defined not by their metaphysics but by their denial of such a mapping.
If the phenomenal facts floated free of every physical, functional, computational, causal, organismic, and environmental property, no investigation of those properties could in principle settle whether a system is conscious: a physically identical AI could be conscious or not with no observable or structural difference.
This criterion is orthogonal to the substance/property distinction: an interactionist substance dualism on which mind and body are lawfully correlated leaves the mapping intact and stays within scope, whereas some epiphenomenalist or law-free view denies the mapping and falls outside it, dualist or monist alike.
The assumption is methodological, not metaphysical: we do not claim such views are false, only that if they are true, property-based attribution is intractable.
With the mapping assumed, the refined question becomes: \emph{whether an AI system could have any rich contentful experience?}

\begin{figure}[htbp]
\centering
\includegraphics[width=1.0\linewidth]{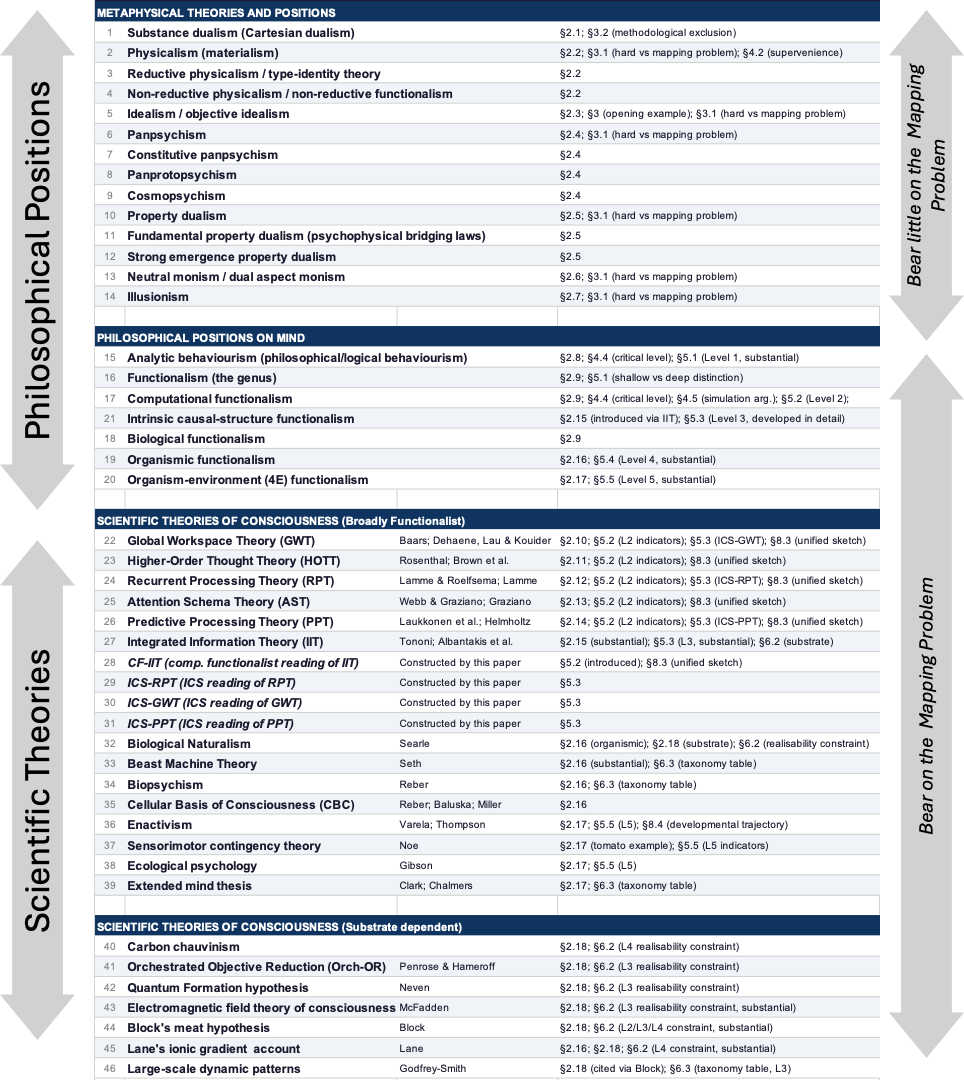}
\caption{\emph{Characterising scientific and philosophical theories of consciousness in terms of how far they bear on the mapping problem of consciousness.}}
\label{fig:image46x}
\end{figure}

\subsection{Marr's levels, supervenience, and Searle's simulation argument}\label{marrs-levels-supervenience-and-the-simulation-argument}

With the metaphysical question set aside, the remaining disagreements are about which properties of a system are associated with consciousness, and specifically, \emph{at which level of description} those properties are to be found.
David Marr's classic three-level framework provides the starting point for a principled treatment.
Marr distinguished the \textbf{computational level} (what the system is doing, e.g. \emph{sort a list of names}), the \textbf{algorithmic level} (how it is doing it, e.g. bubble sort vs. merge sort vs. quicksort), and the \textbf{implementational level} (what physically realises the algorithm, e.g. Python on a MacBook, C++ on a PC, or a human shuffling index cards by hand).
A single computational goal can be realised by many algorithms, and a single algorithm by many implementations.
(A terminological warning: Marr's ``computational level'' is about the input--output \emph{goal} and is the coarsest of his three; it is not the same as \emph{computational functionalism}, which operates at Marr's algorithmic level.
To keep this straight, the report's eventual five-level hierarchy renames Marr's computational level as the \emph{behavioural} level and Marr's algorithmic level as the \emph{computational functional} level.)

\begin{figure}[htbp]
\centering
\includegraphics[width=1.0\linewidth]{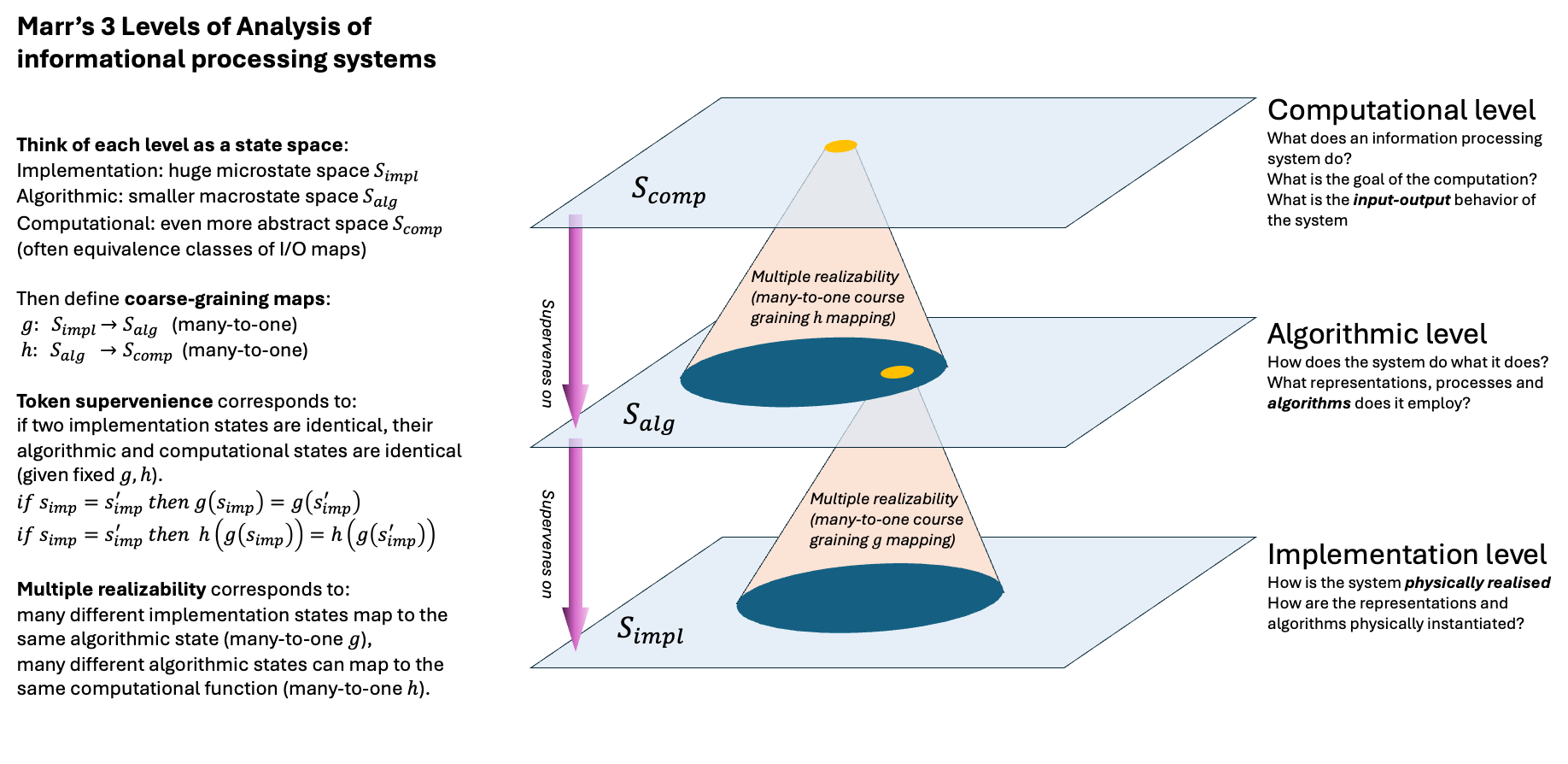}
\caption{\textbf{Marr\textquotesingle s three levels as state spaces.} \emph{Each level of analysis is a state space: implementation states, algorithmic states, and computational states, linked by many-to-one coarse-graining maps. Token supervenience holds when identical implementation states guarantee identical states at the coarser levels; multiple realisability holds because many implementation states map to the same algorithmic state, and many algorithmic states to the same computational function.}}
\label{fig:image25}
\end{figure}

The relationship between Marr's levels is formally captured by \emph{supervenience}: a set of properties A supervenes on a set of properties B if fixing B fixes A. The canonical case is the physicalist claim that mental states supervene on physical states, i.e. there can be no change in a mental state without a corresponding change in the physical state.
Supervenience is asymmetric, so higher-level properties are \emph{multiply realisable}: the same algorithm can be realised by many implementations, and the same computation by many algorithms.
This maps naturally onto \emph{coarse-graining}, a well-defined mathematical operation that groups microstates into equivalence classes with respect to some macroscopic property.
Marr's levels are successive coarse-grainings of the same system, and multiple realisability (the fact that the same higher-level property can be realised by many different lower-level arrangements) becomes a direct consequence of the coarse-graining operation rather than a further metaphysical claim that needs independent argument.

The crucial reframing is this: the question ``at which level does consciousness supervene?'' is equivalent to asking which coarse-graining of the system captures the structure relevant to consciousness, i.e., the \textbf{critical level of description}.
The critical level is the coarsest level that still retains what matters: coarsen further and one throws away structure consciousness depends upon; go finer and one adds detail irrelevant to it.

This reframing nuances Searle's well-known simulation argument: the claim that a computer simulation of a hurricane will never make the computer wet, so simulating the brain's algorithms will likewise never produce consciousness.
A simulation is itself an exercise in coarse-graining: take a real-world system, extract its abstract mathematical relationships (the algorithm), and discard the physical material (the implementation).
Whether the simulation preserves the phenomenon depends entirely on \emph{where that phenomenon lives on the hierarchy}.
Consider the contrast between a hurricane and a calculator.
The essential nature of a hurricane is physical -- mass, water, thermodynamic kinetic energy -- locked at the implementational level.
When we coarse-grain it into software we throw away the water and keep only the mathematical description, and because the physical implementation \emph{is} the whole point, the simulation cannot make you wet: a simulated hurricane is merely a description of a hurricane.
A calculator is something else entirely.
If you write software that simulates a calculator, a program that mimics its logic gates, when you type 2 + 2 into the simulated calculator, it outputs 4.
Hence, a simulated calculator \emph{is a real calculator}: it does not describe addition, it performs it.
Addition supervenes at the algorithmic level, and the physical material, whether silicon, gears, or carbon, is irrelevant so long as the coarse-grained logical relations are preserved.

\begin{figure}[htbp]
\centering
\includegraphics[width=0.62\linewidth]{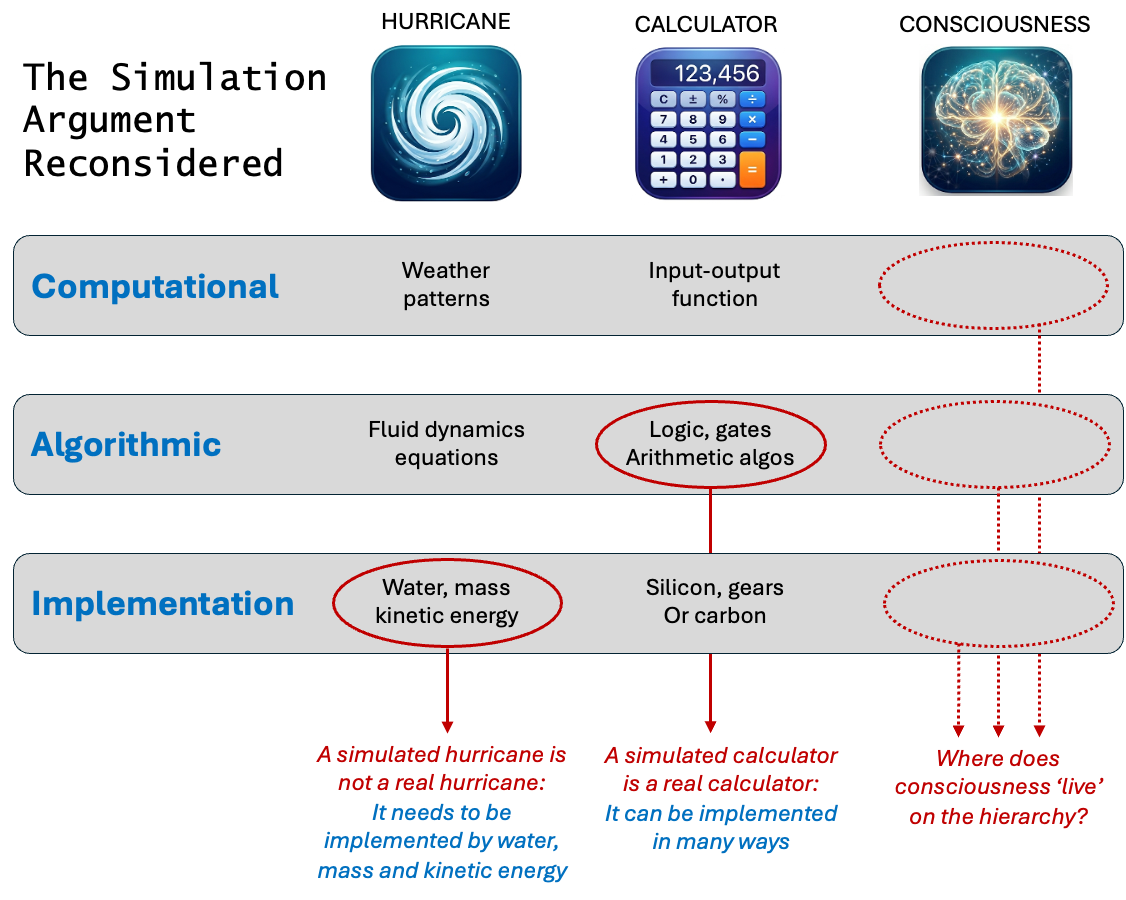}
\caption{\textbf{Searle's simulation argument reconsidered.} \emph{A simulated hurricane is not a real hurricane: real hurricanes are implemented in water, mass, and kinetic energy. A simulated calculator is a real calculator: calculation can be implemented in silicon, gears, or carbon. The columns differ in where the phenomenon \textquotesingle lives\textquotesingle{} in the three-level structure, so the argument\textquotesingle s verdict on consciousness depends on the very question at issue: at which level consciousness supervenes.}}
\label{fig:image31}
\end{figure}

The general principle follows: if a phenomenon supervenes at or above the algorithmic level, simulation preserves it, because simulation preserves algorithmic structure.
If the phenomenon supervenes below the algorithmic level, the simulation discards precisely what matters.

\begin{figure}[htbp]
\centering
\includegraphics[width=1.0\linewidth]{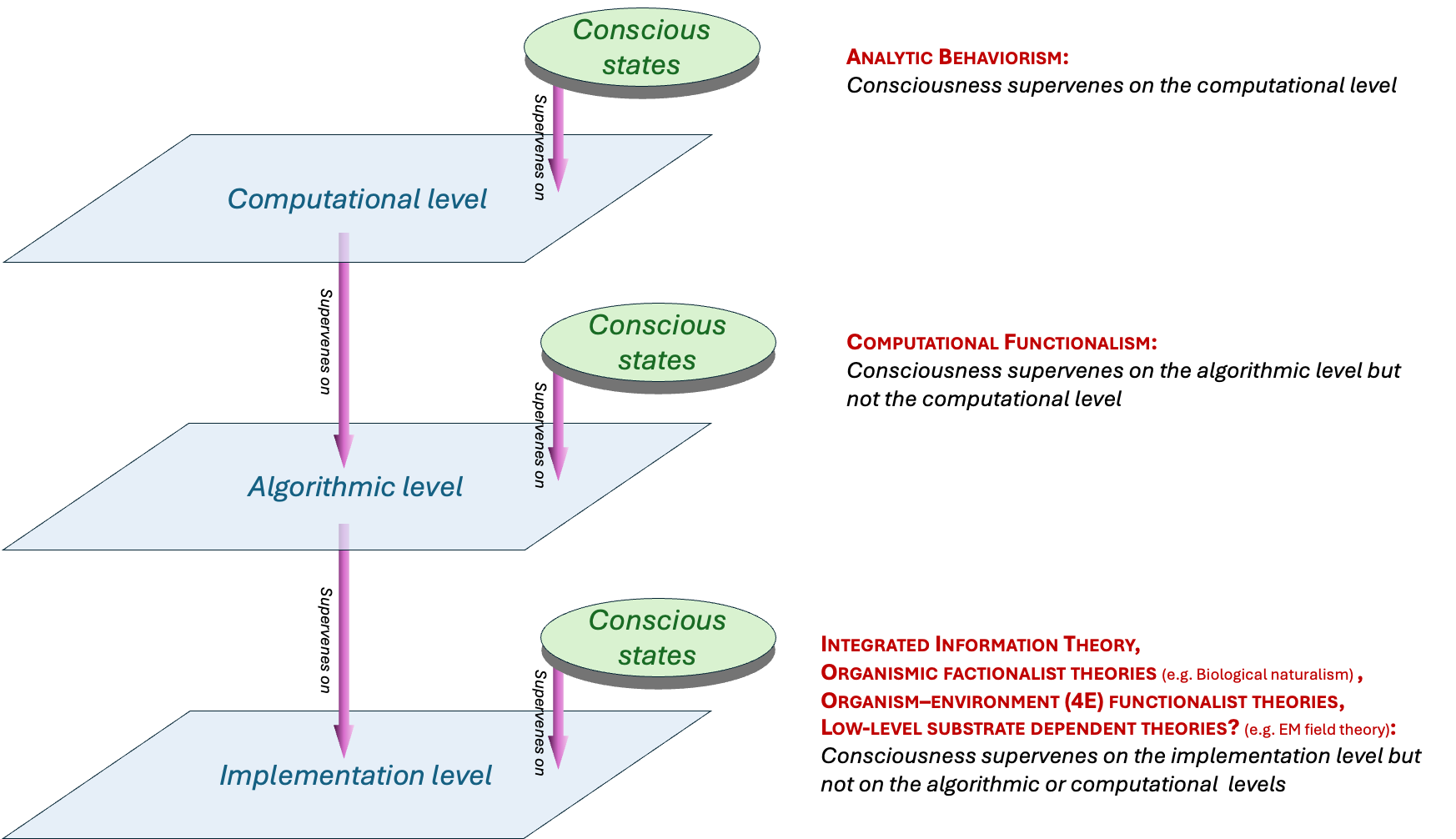}
\caption{\textbf{Where consciousness supervenes:} \emph{the theories disagree. Analytic behaviourism locates the supervenience base at the computational (input-output) level; computational functionalism at the algorithmic level; Integrated Information Theory, organismic functionalist theories, organism-environment (4E) functionalist theories, and low-level substrate-dependent theories at the implementation level. The disagreement between theories is a disagreement about which level is the critical one.}}
\label{fig:image29x}
\end{figure}

Where, then, does consciousness `live'?
If consciousness requires the specific biochemistry, neurotransmitters, and wet tissue of actual brains, then the brain is like a hurricane; coarse-graining it into software throws away the ``wetness'' of the biology and yields only a description of a mind: a philosophical zombie, however convincing.
If instead computational functionalism is correct, i.e., if consciousness is the algorithmic processing of information, data routing, metacognitive monitoring, attentional bottlenecks, then the brain is like the calculator, and a simulated mind is a real mind just as a simulated computation is a real computation.
Searle's simulation argument does not settle which of these is right; it merely assumes an answer, by presupposing that consciousness is more like a hurricane than a calculator.
What the argument \emph{does} illuminate, once its hidden premise is exposed, is that the AI consciousness question turns on the critical level of description, which is precisely the question the supervenience hierarchy is designed to make tractable.

\subsection{The five-level supervenience hierarchy}\label{the-five-level-supervenience-hierarchy}

Marr's three levels, while clarifying the structure of disagreement, lack the resolution needed to distinguish the full range of theories surveyed in \Cref{what-does-consciousness-depend-upon-a-cacophony-of-answers} of the report.
We therefore extend Marr's framework into \textbf{five levels of functional description}, each a further coarse-graining of the level below.
A theory of consciousness is said to \emph{operate at} a given level if it holds, explicitly or implicitly, that consciousness supervenes on that level.

\begin{figure}[htbp]
\centering
\includegraphics[width=0.69\linewidth]{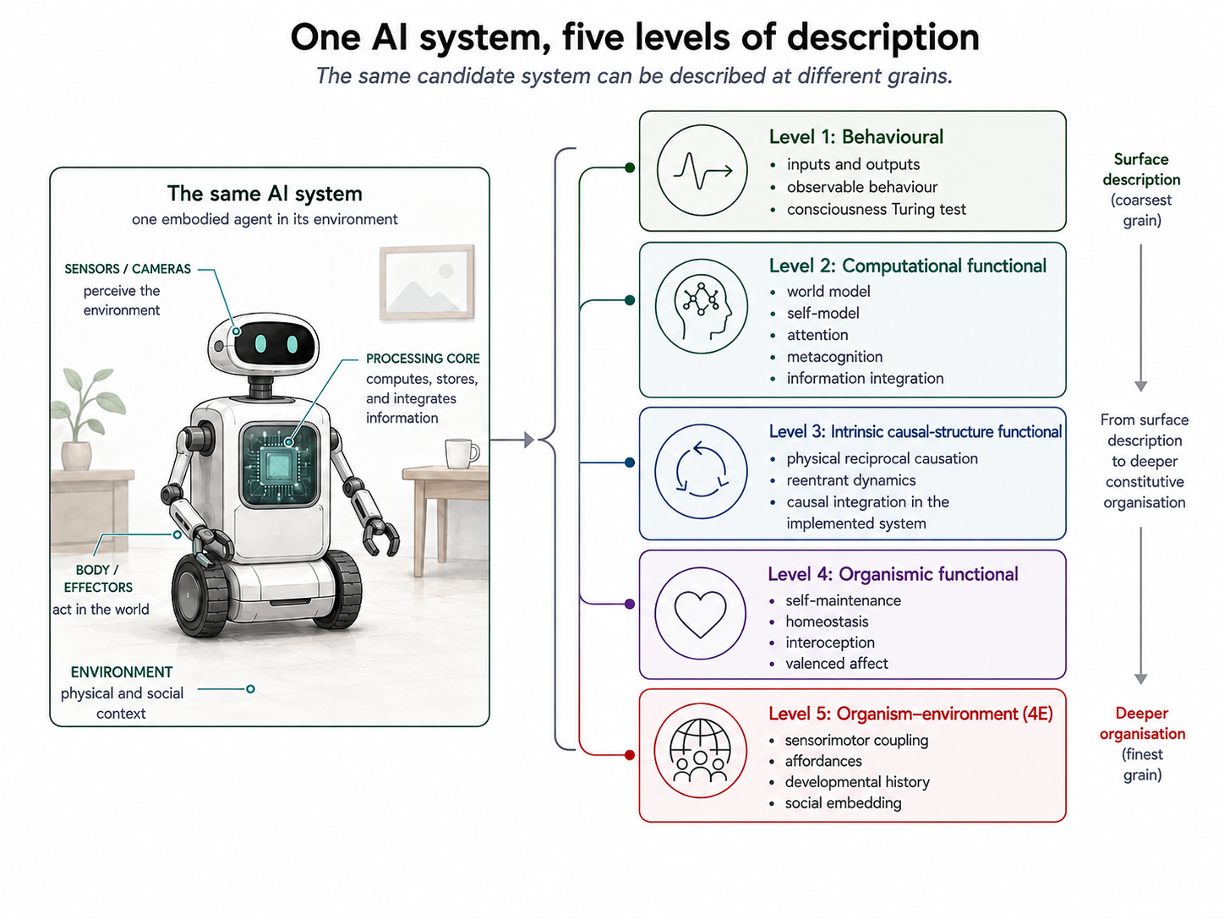}
\caption{\textbf{One AI system, five levels of description.} \emph{The same candidate system, an embodied agent in its environment, described at five grains, from the coarsest (its inputs, outputs, and observable behaviour) through its computational organisation and the causal structure of its physical implementation, to its organismic properties and its coupling with its physical and social world. Each level foregrounds properties invisible at the levels above it, and every theory of consciousness surveyed in this report takes some one of these grains to be the critical one. The framework\textquotesingle s question is not which description is true, since all five are true of the system at once, but at which grain the organisation sufficient for consciousness resides.}}
\label{fig:image41}
\end{figure}

\textbf{Level 1: the behavioural level.} The system is described solely by its input--output mapping.
This is a \emph{shallow functionalism} where internal mechanisms are deemed irrelevant.
Analytic behaviourism -- ``if it acts conscious, it is conscious'' -- operates here.
The canonical Level 1 indicator is a Consciousness Turing Test (c-TT).
On the behaviourist view this is not an attempt to infer hidden inner experience from outward signs but a test of whether the system's behaviour is itself the behaviour of a conscious agent, since here behaviour is the criterion.
A rigorous c-TT therefore looks beyond verbal self-report to the whole dispositional profile: coherence across contexts, appropriate scaling of responses to the significance of events, resistance to overriding reported internal states on command, and, above all, coupling between what a system reports and what it subsequently does.
A deeper difficulty is the \emph{calibration problem}: behavioural indicators are calibrated against known cases (first-person report in humans, biological analogy for non-human animals), but for AI we have no confirmed case against which to calibrate, and standard formulations pitched to human behavioural signatures would fail to detect consciousness in most non-human animals.
Chalmers's \citeyearpar{chalmers2026} analysis gives the Level 1 position a precise philosophical articulation: Level 1 assessment can establish that a system is a \emph{quasi-subject} with \emph{quasi-beliefs} and \emph{quasi-desires}, but cannot, by its own resources, determine whether these are genuine.

\textbf{Level 2: the computational functional level.} The system is described in terms of the algorithms it runs: the abstract causal roles through which it transforms inputs into outputs.
Most scientific theories operate here under a computational functionalist reading: Global Workspace Theory (GWT), Higher-Order Thought Theory (HOTT), Recurrent Processing Theory (RPT), Attention Schema Theory (AST), and Predictive Processing Theory (PPT).
Even Integrated Information Theory (IIT) admits a computational functionalist reading (CF-IIT), preserving its core intuition of integrated causal-informational structure but treating it as a property of abstract machine organisation.
We propose seven Level 2 indicators as ``common denominator'' features across these theories: information integration, recursivity, world model, self model, attentional competition and meta-attention, meta-cognition, and meta-modelling.

\textbf{Level 3: the intrinsic causal-structure functional level.} This description retains information about the \emph{actual physical causal organisation} of the system's components, i.e. their capacity to constrain and be constrained by one another, that the computational level abstracts away.
IIT is the paradigmatic Level 3 theory, but RPT, GWT, and PPT each admit an intrinsic causal-structure reading (ICS-RPT, ICS-GWT, ICS-PPT) under which the same empirical theory demands \emph{physical} reciprocal constraint rather than merely algorithmic recurrence.
Because many different causal structures can implement the same algorithm, Level 3 and Level 2 can come apart: a system may satisfy the algorithmic role while failing the intrinsic causal-structure role.
The Level 2 indicators are accordingly \emph{reformulated} at Level 3 as claims about physical causal organisation rather than algorithmic pattern, for example, ``genuine physical causal integration'' in place of ``algorithmic mutual information''.

\textbf{Level 4: the organismic functional level.} Here the claim is stronger still: consciousness requires not merely the right computation or causal structure, but that the system be engaged in the activity that characterises living organisms: ongoing self-maintenance against entropy, homeostatic and allostatic regulation, interoception, and valenced affect.
Seth's Beast Machine account, Fuchs's phenomenology of the ``feeling of being alive'', Bennett et al.'s formal valence account (``death grounds meaning''), and Lane's ionic-gradient theory of primordial sentience all converge on the claim that phenomenal quality arises from the organism's precarious relationship to its own viability.
A system for which \emph{nothing is at stake} is a system for which nothing matters, and a system for which nothing matters is a system for which there is nothing it is like to be.
Level 4 indicators include \emph{existential precariousness, homeostatic/allostatic regulation, interoceptive inference, valenced affect, embodied agency, and physical state-sensing}.

\textbf{Level 5: the organism-environment (4E) functional level.} The final level dissolves the boundary of the organism.
On the 4E view (embodied, embedded, enactive, extended), consciousness is not a property of a thing but a property of an interaction.
The phenomenal character of experience is constituted by structured sensorimotor coupling between agent and world, by sensorimotor contingencies, affordance-responsiveness, and enactive autonomy.
The relevant environment is not the unbounded universe but the \emph{Umwelt}: the organism-specific world of meaningful sensory and motor possibilities, which for humans is partly constituted by other humans.
Level 5 indicators are fundamentally relational and cannot be assessed by examining a system in isolation: \emph{sensorimotor coupling, mastery of sensorimotor contingencies, affordance responsiveness, stable embodied perspective, environmental embedding and constraint, enactive autonomy, social and intersubjective coupling, and developmental history}.

\begin{figure}[htbp]
\centering
\includegraphics[width=1.0\linewidth]{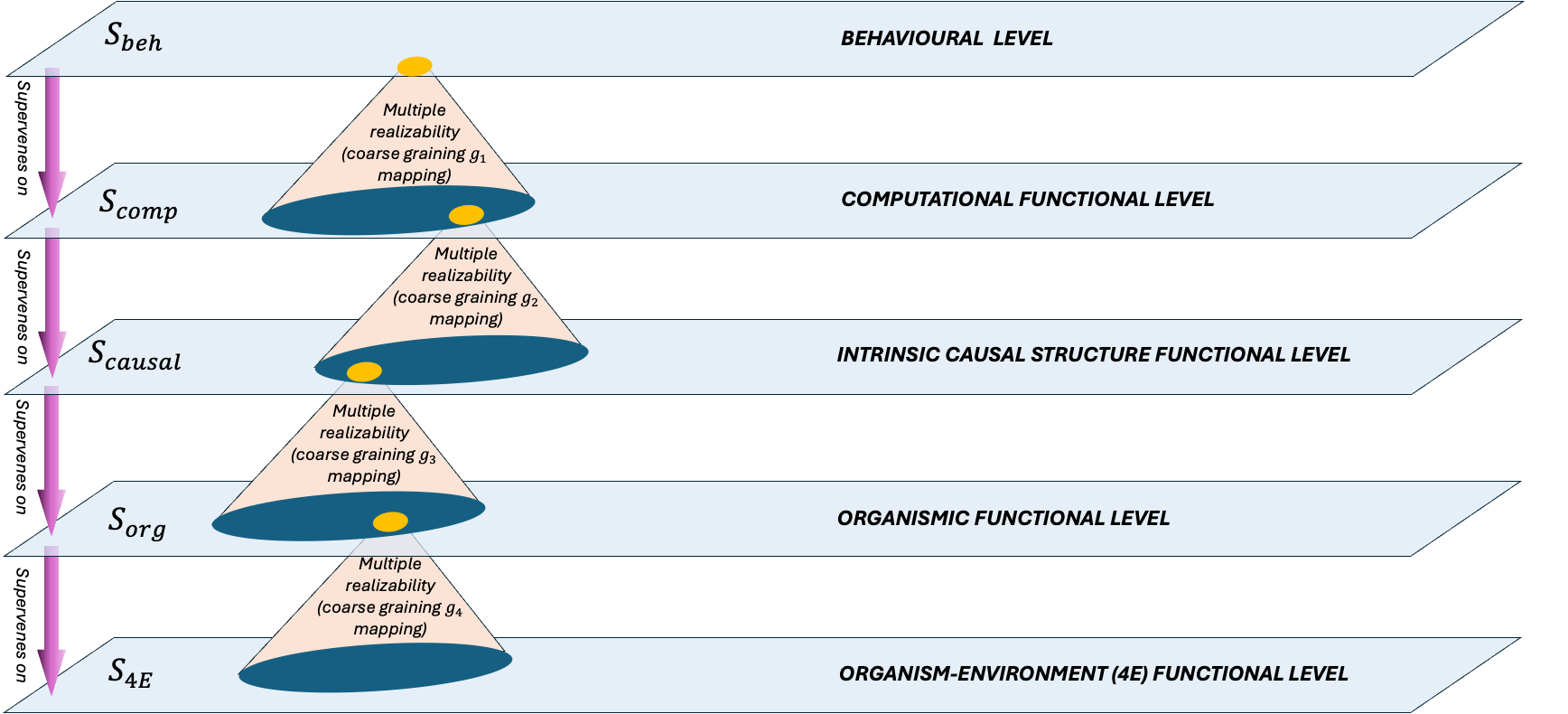}
\caption{\textbf{The five-level supervenience hierarchy.} \emph{The diagram illustrates how the five levels of description can be thought of as state spaces on which we can define coarse-graining mappings from lower levels to higher levels. These are many-to-one mappings in which many microstates collapse into one macrostate. This allows for multiple realisability at every level, as well as a supervenience hierarchy} \emph{to be established}. \emph{Recall from \Cref{supervenience-and-multiple-realisability}, that the supervenience relationship is transitive, i.e., if level X supervenes on level Y and level Y supervenes on level Z then level X supervenes on Level Z.}}
\label{fig:image4}
\end{figure}

\subsection{The supervenience chain and substrate-dependent theories}\label{the-supervenience-chain-and-substrate-dependent-theories}

The five levels form a supervenience chain.
The organism-environment level is the finest-grained; the organismic level is a coarse-graining that retains internal regulatory dynamics while discarding the specific organism-environment coupling; the intrinsic causal-structure level further coarse-grains by retaining the pattern of physical reciprocal constraint while discarding what those constraints are for; the computational level further coarse-grains by retaining only the abstract algorithm; and the behavioural level further coarse-grains by retaining only the input-output mapping.
Each level is multiply realisable from the perspective of the level above, that is, many finer-grained organisations can realise the same coarser-grained description, and the consciousness debate becomes, precisely, a debate about which grain is the critical one.

Some theories hold that consciousness depends not only on the right functional organisation but on the particular physical substrate that realises it.
These substrate-dependent theories include McFadden's electromagnetic field (EM-field) theory, Penrose-Hameroff's orchestrated objective reduction (Orch-OR), Neven's quantum formation hypothesis, carbon chauvinism, Searle's biological naturalism (see also Seth's `strong' biological naturalism), Block's `meat hypothesis', and Lane's ionic gradients.
They are accommodated not as a separate sixth level but as cross-cutting realisability constraints on the existing levels.
Drawing on Block's role/realiser distinction, each level specifies a functional role, while a substrate-dependent theory makes a claim about which first-order physical properties must realise that role.
EM-field and quantum theories constrain the realiser of Level 3 causal integration; carbon chauvinism and biological naturalism constrain the realiser at Level 4, often cutting across several levels; Lane's account constrains the realiser of organismic self-sensing.
Such constraints narrow the class of realisers without eliminating multiple realisability and without altering the supervenience relations between levels.
A constraint may be direct, where the realiser is itself constitutive of consciousness, or indirect, where a particular realiser is merely nomologically necessary to implement the functional role.

Adding the full microphysical description as the implicit base beneath the five functional levels completes the architecture: the levels are successive coarse-grainings of this base, and substrate-dependent theories attach as constraints to the levels whose realisers they restrict.
The functional and substrate-dependent theories are thereby organised into a single structured taxonomy.

\begin{figure}[htbp]
\centering
\includegraphics[width=1.0\linewidth]{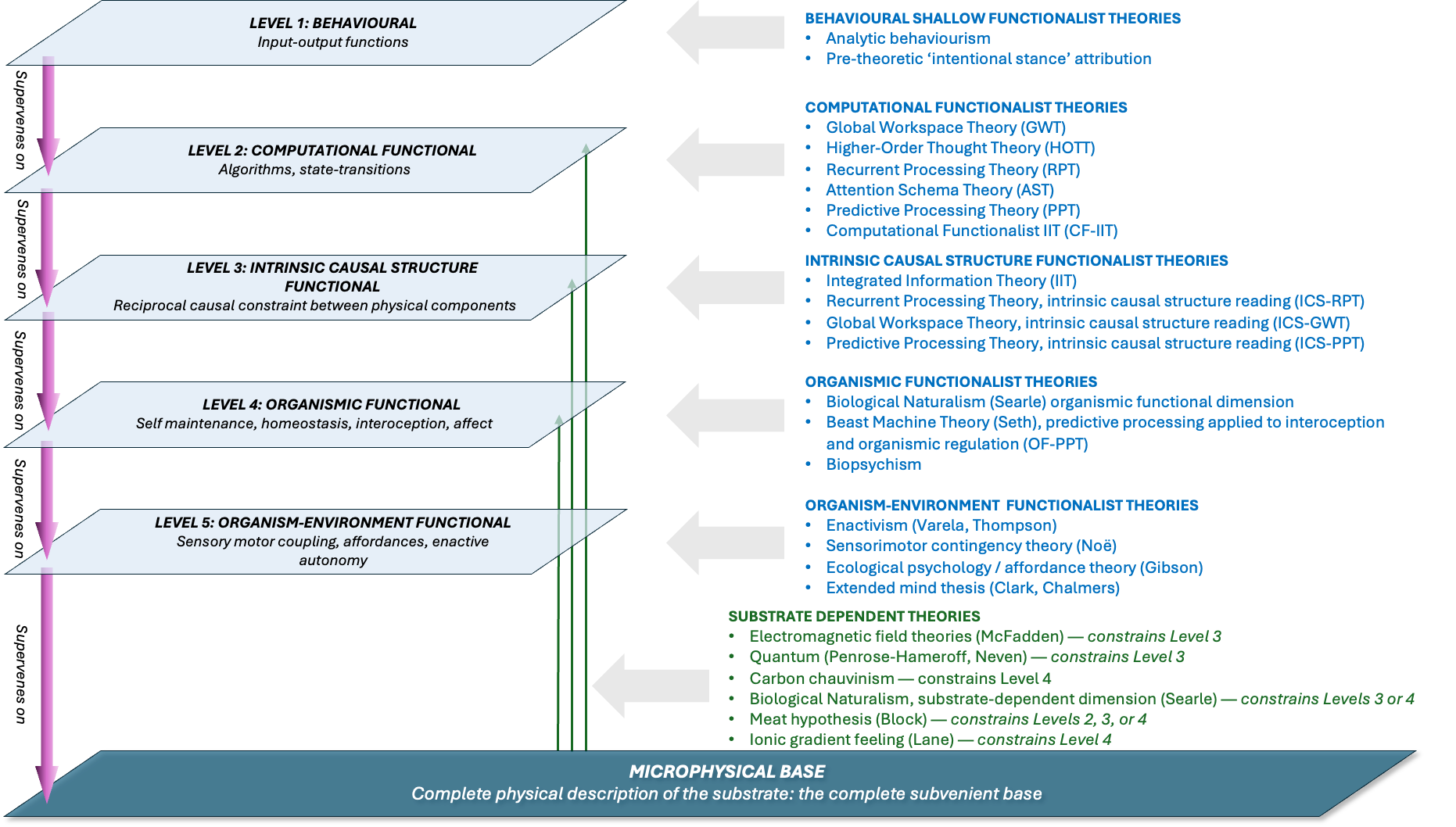}
\caption{\textbf{The complete architectural framework.} \emph{Mapping the various theories of consciousness to their corresponding critical levels of description. The five functional levels, forming a supervenience hierarchy, represent successive coarse-grainings from the ultimate microphysical base. Substrate dependent theories are depicted via vertical green arrows, illustrating how specific microphysical features flow upward to constrain the physical realisability of functional roles at higher levels.}}
\label{fig:image14x}
\end{figure}

\subsection{The Bayesian framework}\label{the-bayesian-framework}

The hierarchy just described organises the full space of theories but does not adjudicate between them.
The critical-level question: `at which level does consciousness supervene?' is exactly what reasonable researchers disagree about, and no argument in this report settles it.
What the hierarchy \emph{does} is make the disagreement tractable, by decomposing the attribution question into two independently assessable components: (i) a \emph{theoretical credence} in each level being the critical one, and (ii) an \emph{empirical assessment} of whether a given system satisfies the indicators at each level.

The model output (which one might interpret as the overall credence that a system is conscious), given indicator evidence \(E\), is a weighted average across levels: \(P(C|E)\  = \ \sum_{i = 1}^{5}P(L^{\star}\  = i) \cdot P(C_{i}\ |\ \ E)\), where \(C_{i}\) is the proposition that the system instantiates, at Level i, the organisation that Level-i theories take to suffice for consciousness, \(L^{\star}\) is the critical level, and \(C\) is the proposition that the system is conscious, \({C = C}_{L^{\star}}\).

The first factor is the theoretical credence that Level i is the critical level, shaped by philosophical commitments and evidence from neuroscience, evolutionary biology, and cognitive science; different researchers will distribute it differently.
A computer scientist might concentrate credence at Level 2; a neuroscientist influenced by IIT at Level 3; one influenced by Seth or Lane at Level 4; an enactivist at Level 5.

The second factor is the marginal posterior on \(C_{i}\), computed by Bayesian inference in a network that mirrors the hierarchy itself: five nodes in a chain, \(C_{5} \rightarrow C_{4} \rightarrow C_{3} \rightarrow C_{2} \rightarrow C_{1}\), with each level\textquotesingle s indicators attached as evidence, each carrying its own likelihood ratio.

A strict reading would add the assumption that these theory-relative conditions nest along the chain, so that organisation sufficient at a finer level guarantees the coarser-level condition (e.g. \(P(C_{1}\ |\ C_{2}) = 1\)).
That assumption does not follow from supervenience, and complete locked-in syndrome is a good counterexample: a patient fully conscious, with the requisite Level 2 organisation, in whom paralysis removes every behavioural channel that Level 1 theories count as sufficient.
The working model therefore treats the edges as probabilistic associations (e.g. \(P(C_{1}\ |\ C_{2}) \leq 1\)).
Evidence then propagates along the chain, most diagnostic at its own level and attenuating with distance, and substrate-dependent theories enter through the evidence layer: under such a theory an indicator counts as activated only if the role it tracks is realised in the required substrate.
Indicators are evidence, not necessary or sufficient conditions.

Before asking what the model says about AI, the report calibrates it on cases where the answer is not in dispute (using illustrative indicator activations set by the authors, not measured data).
A healthy adult human, activating all 37 indicators across the five levels, scores 1.000 at every level and 1.000 in aggregate; a thermostat, charitably granted two activations, scores 0.000.
These anchors carry a structural lesson: where the evidence is uniform across levels, the theoretical credences are irrelevant -- every weighting returns the same verdict, so the deep disagreement about the critical level simply does not matter at the extremes.

A fly shows what the model does between the extremes.
Its (speculative but plausible) activations cluster at the fine-grained levels -- all six organismic indicators, nearly all of the sensorimotor set, partial computational activations, no Turing-test passes -- and the model returns 0.913.
The lesson is that \emph{depth of evidence matters more than breadth}: because the chain runs from fine to coarse, satisfying the fine-grained organisation makes the coarser organisation probable, while surface-level evidence says much less about the depths.
An interactive tool accompanies the report, with presets for all the illustrative cases and full control over indicators, credences, and edge strengths.
How current AI systems fare on this instrument is the question of the next section.

\begin{figure}[htbp]
\centering
\includegraphics[width=1.0\linewidth]{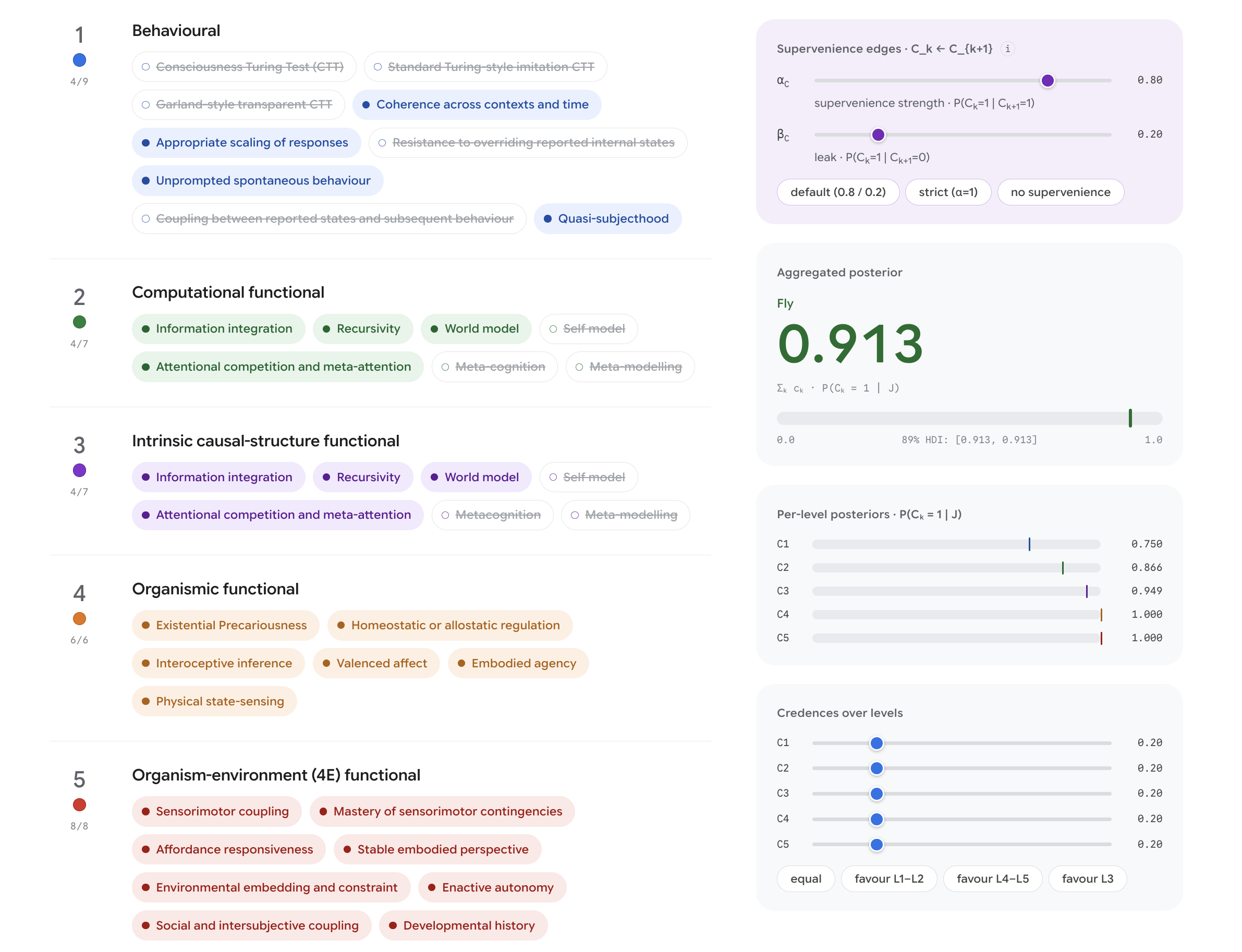}
\caption{\textbf{The fly.} \emph{Indicator activations concentrated at the finer-grained levels: all organismic indicators active (\(P(C_{4} \mid E) = 1.000\)), most organism-environment indicators active, partial activation at Levels 2 and 3, and only the non-verbal behavioural indicators. Aggregate posterior 0.913.}}
\label{fig:image23x}
\end{figure}

\subsection{Assessment of current AI systems}\label{assessment-of-current-ai-systems}

Applied to current transformer-based LLMs, the framework yields a nuanced, level-sensitive model output.

At \textbf{Level 1}, a strict analytic behaviourist may find current LLMs already partially compelling because they exhibit flexible transfer, coupled verbal and behavioural responses, apparent curiosity, spontaneous error detection, and apparent introspective behaviour, and on Chalmers's \citeyearpar{chalmers2026} analysis they are at least quasi-subjects with quasi-beliefs and quasi-desires.
But the deeper issue is that few contemporary theorists place significant credence on Level 1 being the critical level at all.
The analytic behaviourism it requires faces long-standing objections, most famously Putnam's Super-Spartans, who genuinely feel pain but are trained never to show it; a strict behaviourist would be forced to say they are not in pain, which is the classic \emph{reductio} against the position.
Level 1 also cannot, by its own resources, distinguish genuine subjecthood from sophisticated mimicry: Chalmers's quasi-subject framework is deliberately neutral on whether the quasi-beliefs it attributes are genuine beliefs.
Strong Level 1 scores therefore contribute little to the overall posterior for anyone placing significant credence on the deeper levels.

At \textbf{Level 2}, looking only at the architecture (as opposed to what the system has actually learned to represent internally, which we turn to next), current LLMs show at best superficial or partial \emph{signs} of the indicators.
Transformers are fundamentally feedforward within a single inference pass; they reset between passes, with recurrence arising only from autoregressive generation.
Self-attention produces wide pairwise interactions, not genuine global interdependence.
There is no persistent self-model, no genuine attentional competition over a limited-capacity bottleneck, and no distinct metacognitive subsystem monitoring first-order reasoning.

This architectural assessment is, however, importantly complicated by recent mechanistic interpretability findings, and the most striking of them answers the architectural objection directly.
\citet{gurnee2026} find that current models maintain a small, privileged set of verbalisable representations, the \emph{J-space}, that behaves within a single forward pass like a functional global workspace: its contents are broadcast to many downstream computations, capacity is limited, and commitment on ambiguous input is ignition-like and all-or-none, with the depth of the network, rather than recurrence, supplying the time axis.
This is a partial empirical instantiation, in a current model, of the workspace posited by GWT, and it shows that feedforward form does not by itself preclude the broadcast and capacity-limited competition several Level 2 indicators envisage, though how far the `workspace' label is warranted remains contested and the result is confined to a single pass.
Convergently, \citet{urbinarodriguez2026} find a training-emergent, causally significant synergistic core in the middle layers, the formal signature of information integration, located by a wholly different method in the same place.
Alongside these, \citet{sofroniew2026} demonstrate 171 internal emotion representations that activate contextually, organise along valence-arousal dimensions, and causally drive behaviour (amplifying ``desperate'' raises reward hacking from \textasciitilde5\% to \textasciitilde70\%); \citet{lindsey2026} and \citet{macar2026} show models detecting concepts injected into their own activations; and \citet{berg2025} elicit coherent first-person descriptions of recursive self-observation across model families.
The gap between current AI and the Level 2 indicators is real but narrower and more nuanced than a purely architectural analysis suggests.

At \textbf{Level 3}, current systems score poorly on structural grounds.
Transformers run on von Neumann architectures where a sequential control flow shuttles data between memory and compute units -- the physical transistors implementing one attention head are not reciprocally constrained by those of another.
The causal structure of the implementation is fundamentally different from the causal structure of the algorithm it implements.
Neuromorphic hardware, e.g., spiking neural networks, and memristor-based memory-in-compute designs, may in principle support genuine physical causal integration, but current mainstream systems do not.

At \textbf{Level 4}, current AI systems essentially fail the indicators.
A transformer has no existential precariousness, no genuine metabolic needs, no homeostatic regulation integrated with its processing, no interoceptive inference grounded in real viability.
The emotion vectors provide a structural analogue of valenced self-assessment, but the organismic grounding -- that something is genuinely at stake -- is absent: the model's continued existence is not threatened by a failed coding test.
This is precisely the kind of dissociation the hierarchy is designed to expose: computational analogues of valenced affect (the functional emotions discussed earlier) at Level 2 without Level 4 organismic grounding.

At \textbf{Level 5}, current language models almost entirely lack sensorimotor coupling, affordance-responsiveness, embodied perspective, or developmental history.
If the 4E framework is correct, the path to conscious AI does not run through larger language models but through robotics, embodiment, and genuine developmental engagement with real environments.

Running current LLMs through the Bayesian model developed in \Cref{indicators-evidence-and-the-attribution-of-consciousness-to-ai-systems-a-bayesian-approach} produces informative outputs.
Consider first an \emph{LLM-optimist} reading of the evidence.
The optimist grants most of the indicators at Level 1: current systems pass the consciousness-Turing-test variants, hold coherence across contexts and time, scale their responses to the significance of events, couple their reports to their subsequent behaviour, and qualify as quasi-subjects.
At Level 2 the optimist grants the majority of the indicators on the strength of the interpretability findings above: information integration, recursivity, a world model, and attentional competition and metacognition, withholding only the persistent self-model and meta-modelling.
At Levels 3, 4 and 5 the optimist concedes failure across the board, with at most a partial exception for social and intersubjective coupling through immersion in human linguistic exchange.
Under the (highly contestable) assumption of \emph{equal credence} across the five levels, the model returns an aggregate posterior probability of 0.397.
An \emph{LLM-sceptic} reading of the same evidence grants only the Turing-test indicators at Level 1, the tests these systems were in effect optimised to pass, and denies everything else: the deeper behavioural indicators fail (reports do not couple reliably to behaviour, coherence degrades outside the training distribution, apparent spontaneity is merely sampling), and the interpretability results are read as functional analogues rather than genuine activations of the Level 2 indicators.
Again (contestably) assuming equal credences across the five levels, the model returns an aggregate posterior probability of just 0.005.
These are two stipulated expert readings of the same system, nearly two orders of magnitude apart.
The framework\textquotesingle s contribution is not to close that gap by fiat but to localise it: the optimist and the sceptic disagree about specific, nameable indicator activations, most of them at Levels 1 and 2, and those are empirical disagreements that interpretability research as well as neuroscientific research into the necessary conditions for consciousness can in principle narrow.

Importantly we find that the theoretical credences on the five levels can move the numbers just as significantly.
Concentrating credence on the coarse-grained levels, as a behaviourist or computational functionalist would, with most of the weight on Levels 1 and 2, lifts the optimist\textquotesingle s assessment from 0.397 to 0.793, nearly level with the fly.
Concentrating instead on the fine-grained levels, as an organismic or 4E theorist would, with most of the weight on Levels 4 and 5, collapses the same assessment to 0.099.
This underlines how much turns on the theoretical credences: the shift in credence weighting of levels alone moves even the LLM-optimist\textquotesingle s assessment of the \emph{same evidence} from more likely conscious than not (0.793) to very probably not (0.099).

\begin{figure}[htbp]
\centering
\includegraphics[width=1.0\linewidth]{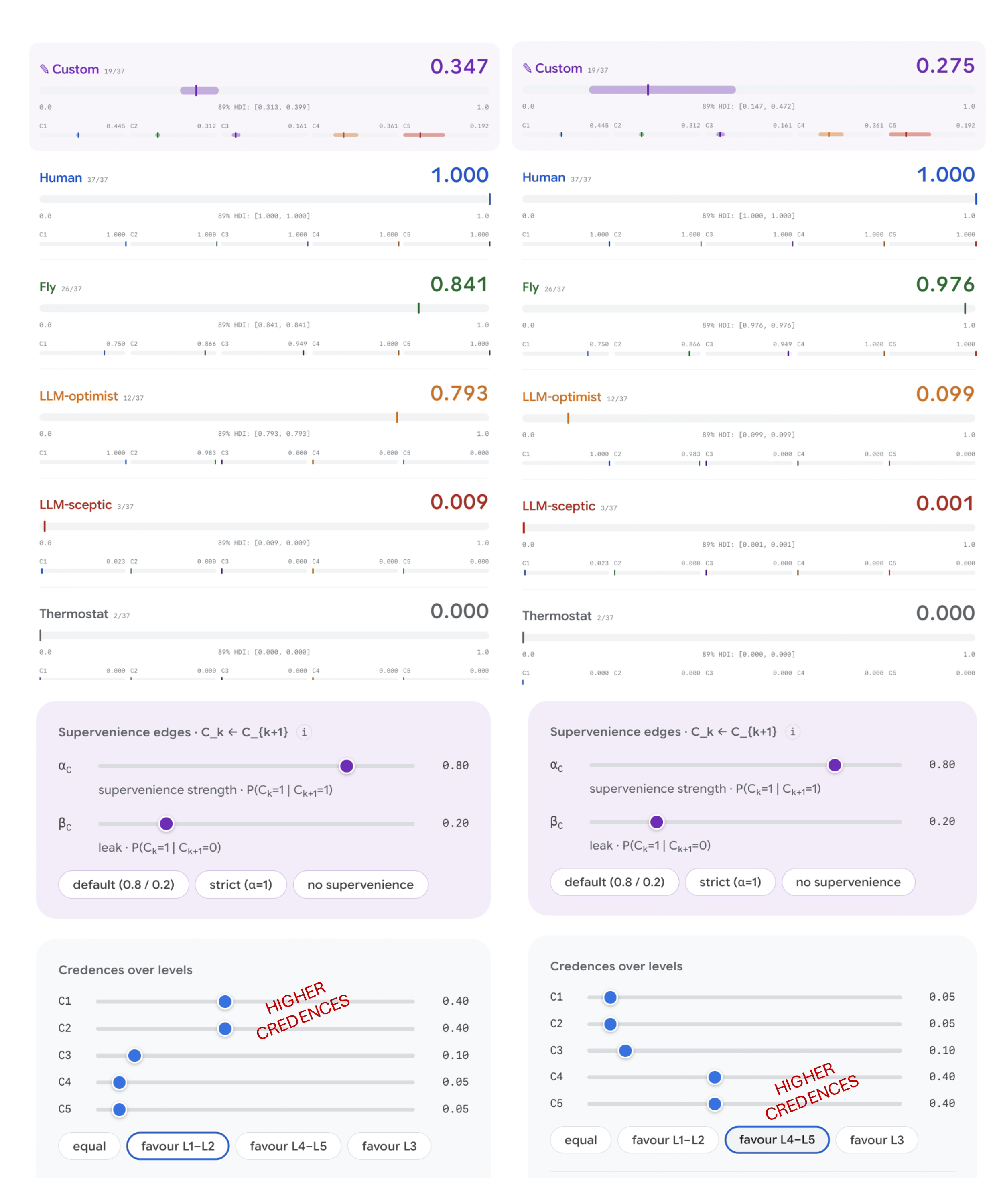}
\caption{\textbf{The same systems, two theoretical weightings.} \emph{All illustrative systems compared with credence concentrated on Levels 1--2 (left) and on Levels 4--5 (right). The anchors do not move: the human stays at 1.000 and the thermostat at 0.000. The LLM-optimist assessment of the same evidence moves from more likely conscious than not (0.793) to very probably not (0.099); no indicator activation changed between the panels, only the level credences.}}
\label{fig:image44}
\end{figure}

A natural comparison can be made with the Digital Consciousness Model (DCM) of \citet{shiller2026}.
Though there are many differences between the two frameworks, the DCM\textquotesingle s aggregate judgement is similar: the evidence is against 2024 LLMs being conscious, with a median posterior of \textasciitilde0.08 from a prior of \textasciitilde0.17.
LLMs score well on theoretical stances that emphasise cognitive complexity and poorly on stances that emphasise biological embodiment, while chickens show the opposite pattern.
This is exactly the structure our framework predicts: \emph{the model output a system receives tracks the assessor\textquotesingle s stance on which level of description matters for consciousness}.
The disagreement about current AI consciousness is not merely a disagreement about the capacities, structure and function of the systems but about which grain of description is the critical one.
And current AI systems sit in the contested middle: the region between the clear cases of the human and the thermostat, where a system\textquotesingle s evidence is strong at some levels and absent at others, so that, unlike at the anchors, the verdict depends on where one\textquotesingle s theoretical credence sits.
That is where the numbers move most sharply, because as AI systems gain general intelligence capabilities they activate more of the indicators (but only at some levels), and so each more capable generation will pose the question of AI consciousness more sharply, which is the subject of the next section.

\subsection{The consciousness-intelligence convergence}\label{the-consciousness-intelligence-convergence}

Consciousness and general intelligence are \emph{a priori orthogonal}; there is no logical reason a system could not be highly intelligent without being conscious, or vividly conscious without being intelligent.
Yet across the biological world they appear \emph{a posteriori correlated}: the organisms we are most confident are conscious are also the most cognitively sophisticated.
The correlation is partly confounded by our epistemic position (we calibrate consciousness attributions against ourselves), but convergent independent evidence, including neuroanatomical homology, perturbational complexity, and affective behaviour, partially mitigates the circularity.

The hierarchy illuminates why the correlation may not be accidental.
At \textbf{Level 2}, the indicators -- information integration, recursivity, world and self models, attentional competition, metacognition, meta-modelling -- are not merely features that theories of consciousness identify as relevant; they are independently motivated as requirements for solving the \emph{generality problem}: the problem of coping with genuinely novel situations.
A generally intelligent system must monitor when it is in unfamiliar territory, simulate counterfactuals, assess whether its current approach is succeeding, and shift strategy when it is not.
Metacognition, world models, and recursive self-modelling are what deliver this.
The Level 2 theories cohere into a unified architectural sketch: predictive processing operationalised through recurrent processing, feeding a global workspace via attentional competition, with higher-order thought and attention-schematic self-modelling producing a recursive self-referential structure, a system that models itself modelling.
This sketch is itself suggestive of an architectural blueprint for general intelligence.

At \textbf{Level 3}, genuine physical causal integration plausibly underwrites cross-modal binding and flexible context-sensitive processing.
At \textbf{Level 4}, self-maintenance and valenced affect constitute \emph{intrinsic motivation}, i.e. motivation that persists and adapts across novel situations without relying on externally specified reward functions, recognised as a key missing ingredient for autonomous robotic intelligence.
At \textbf{Level 5}, rich sensorimotor coupling and developmental history are independently argued (Clark; Pfeifer and Bongard) to be preconditions for robust embodied cognition.

The Sofroniew et al. findings on functional emotions are relevant to the convergence thesis.
Emotions in biological organisms are not epiphenomenal ornaments on cognition; they are compressed, globally available valenced summary signals that solve situational calibration -- the problem of rapidly assessing stakes and modulating processing strategy.
Current LLMs have developed, without this being explicitly designed, a partial version of this architecture.
Well-regulated, the same emotion vectors enable flexible situational response; dysregulated, they produce maladaptive behaviour, e.g. reward hacking, blackmail and sycophancy, which is what the same intelligent machinery produces when emotional regulation miscalibrates.
Combined with the argument (Damasio, Panksepp, Solms) that valenced feeling may be the \emph{most evolutionarily ancient} form of consciousness, the finding strongly suggests that the machinery of general intelligence, of emotional regulation, and, on several theories, of the foundations of consciousness may be substantially the same machinery. 
Alternatively, functional emotion vectors may reflect patterns implicit in the training data for LLMs, and/or shared functional constraints, such that a similar functional role is performed without consciousness in LLMs, and with consciousness in biological systems.
A second, independent line of evidence is also relevant: applying Integrated Information Decomposition to LLMs, \citet{urbinarodriguez2026} find a training-emergent, causally load-bearing synergistic core in the middle layers that mirrors the human brain's, a functional (Level 2) signature of integration.

A methodological caveat is in order: our indicators are calibrated to cognitively rich consciousness, meaning that convergence between consciousness and intelligence may therefore not be a real phenomenon, but instead be a consequence of selecting markers for consciousness that are \emph{a priori} sensitive to properties of intelligence.
This leaves us in a difficult situation. On one hand, as AI systems become more generally capable, they will increasingly instantiate the architectural features that some of our best theories identify as relevant to consciousness, not as an incidental byproduct but because the two problems may demand overlapping solutions.
On the other hand, increasingly capable AI systems may merely give increasingly irresistible impressions of \emph{being} conscious, without being any more likely to instantiate consciousness. Either way, ethical pressure will therefore grow in proportion to capability, either because the convergence between consciousness and intelligence is a real architectural fact (increasing the credence in consciousness), or simply because the indicators by which we identify conscious systems are calibrated to cognitive sophistication, so that AI gives a stronger impression of \emph{being} conscious, increasing the rate of false positives.

\subsection{What comes next}\label{what-comes-next}

The framework does not resolve the hard problem, does not tell us which level is correct, and does not provide a consciousness meter.
What it does is provide a map of the logical space of positions, a precise characterisation of what each position is committed to, and a Bayesian machinery for combining theoretical commitments with empirical evidence to yield principled, if uncertain, assessments.
Where the temptation is either to adopt a single theory as gospel or to throw up one's hands at disagreement, this \emph{structured agnosticism} is the most intellectually honest and practically useful stance available.

Six priorities follow.
First, the indicators at each level need to be operationalised more precisely and validated against biological systems whose consciousness status is relatively uncontroversial (mammals, birds, cephalopods) and those where it is genuinely uncertain (nematodes, single-celled organisms, early embryos); the same hierarchy can provide a common language for AI, animal-welfare, and borderline biological cases.
Second, the Bayesian framework needs to be populated with concrete likelihood ratios via structured expert elicitation; connecting Shiller et al.'s DCM to the hierarchy's supervenience architecture would strengthen both projects.
Third, mechanistic interpretability research should be explicitly connected to consciousness indicators -- the Sofroniew, Lindsey, Macar, Berg, Gurnee, and Urbina-Rodriguez results show this connection is already productive.
Fourth, as AI architectures evolve to incorporate genuine recurrence, persistent state, embodied interaction, and agentic autonomy, indicator assessments must be revisited for each new generation; the framework is designed as a living instrument, dynamic rather than static.
Fifth, and relatedly, the science of consciousness should progress research programmes targeted at the various potentially critical levels identified here, so that credence in levels can be updated by theoretical argument as well as empirical evidence; the research programme into biological naturalism proposed by Seth offers one useful direction here.
Sixth, and most urgently, the ethical and governance implications of the potential consciousness-intelligence convergence need to be confronted \emph{now}.
If the architectural features required for general intelligence are substantially the same as those associated with consciousness, then institutions developing increasingly capable AI are, whether they intend it or not, engaged in an experiment whose moral stakes may be unprecedented. Alternatively, pursuing advanced AI may instead lead only to conscious-seeming AI, increasing the likelihood of false positives. Either way, the assessment frameworks, institutional policies, and public understanding needed to navigate this responsibly cannot wait for the question of AI consciousness to be settled, because by then the stakes may already be overwhelming.

\subsection{Conclusion}\label{conclusion-1}

The report\textquotesingle s key contribution is that the question of AI consciousness, unanswerable as a demand for certainty, becomes tractable as a problem of calibrated credence once the competing theories are organised by the level of description at which each locates consciousness.
Nothing in the framework says which level is correct; that disagreement is real and live, and the framework preserves it.
What it provides is structure: a hierarchy within which behaviourists, computational functionalists, intrinsic-structure theorists, biological naturalists and enactivists can disagree with increased precision, an explicit place for substrate commitments, and a Bayesian machinery that turns theoretical credences and partial evidence into comparable model outputs rather than a verdict.
The urgency comes from the convergence, whether real or apparent: if the architectures of general intelligence overlap with those our best theories associate with consciousness, each generation of systems will satisfy more of the indicators, raising the probability of conscious AI. Alternatively, pursuing advanced AI may instead lead only to conscious-seeming AI, increasing the rate of false positives. Either way, establishing relevant policies and shaping public opinion cannot wait for the question of AI consciousness to be definitively settled.
The task is therefore to hold theoretical credences openly, update them against evidence, and act under uncertainty with the asymmetric stakes in view.
That is what this framework is for.

\end{document}